\documentclass{egpubl2}
\usepackage{eg2025}
 
\ConferencePaper        %

\CGFccby

\usepackage[T1]{fontenc}
\usepackage{dfadobe}  

\usepackage{cite}  %
\BibtexOrBiblatex
\electronicVersion
\PrintedOrElectronic
\ifpdf \usepackage[pdftex]{graphicx} \pdfcompresslevel=9
\else \usepackage[dvips]{graphicx} \fi

\usepackage{egweblnk} 
\usepackage{booktabs} %
\usepackage{amsmath}
\usepackage{amsfonts}
\usepackage{adjustbox}

\usepackage{soul}
\usepackage{multirow}
\usepackage{xcolor}
\usepackage{xspace}

\newcommand\features[0]{\phi_{4\text{D}}}

\newcommand{\methodname}{\emph{4-LEGS}\xspace}
\newcommand{\datasetname}{\emph{Grounding-PanopticSports}\xspace}

\newcommand{\whitetxt}[1]{{\color{white}#1}\normalfont}
\usepackage{color}
\definecolor{gray}{rgb}{0.6,0.6,0.6}
\definecolor{red}{rgb}{1,0,0}
\definecolor{green}{rgb}{0,1,0}
\definecolor{blue}{rgb}{0,0,1}
\definecolor{dark-green}{rgb}{0,0.4,0}
\definecolor{orange}{rgb}{1,0.55,0}
\definecolor{white}{rgb}{1,1,1}
\definecolor{black}{rgb}{0,0,0}
\definecolor{dark-brown}{rgb}{0.2,0.1,0}
\definecolor{light-blue}{rgb}{0.4,0.6,0.99}
\definecolor{dark-red}{rgb}{0.6,0,0}
\definecolor{light-red}{rgb}{1,0.2,0.6}
\definecolor{pink}{rgb}{1,0.2,0.6}
\definecolor{dark-pink}{rgb}{0.6,0,0.3}

\newbox\jsavebox
\newcommand{\jsubfig}[2]{%
	\sbox\jsavebox{#1}%
	\parbox[t]{\wd\jsavebox}{\centering\usebox\jsavebox\\#2}%
	}

\newcommand{\sayitt}[3]{{\small\protect\colorlet{col}{#2}\color{col}{} #3}}

    \newcommand{\archway}[1]{\sayitt{}{cyan}{#1}}
    \newcommand{\facade}[1]{\sayitt{}{orange}{#1}}
    \newcommand{\sundial}[1]{\sayitt{}{teal}{#1}}

\title{4-LEGS: 4D Language Embedded Gaussian Splatting}
\author[G. Fiebelman et al.]
{\parbox{\textwidth}{\centering Gal Fiebelman$^{1}$,\: Tamir Cohen$^{1}$,\: Ayellet Morgenstern$^{1}$,\: Peter Hedman$^{2}$,\: Hadar Averbuch-Elor$^{1}$ 
        }
\\
{\parbox{\textwidth}{\centering
$^1$Tel Aviv University \quad
        $^2$Google Research 
       }
}
\\
\\
{\parbox{\textwidth}{\centering
\small{\url{https://tau-vailab.github.io/4-LEGS/}}       }
}
}

\begin{document}

\teaser{\begin{center}

\includegraphics[width=0.99\textwidth]{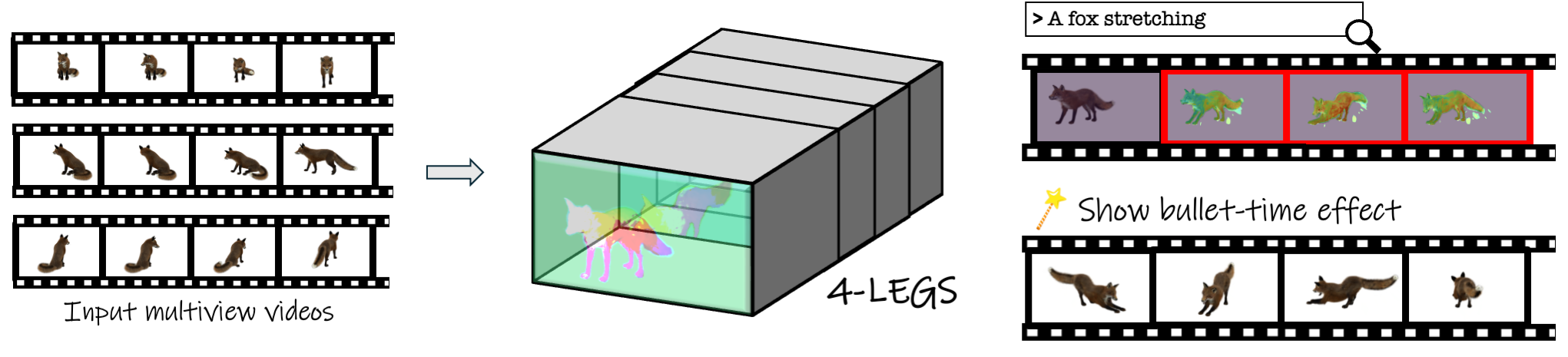}
\caption{\textbf{4D Language Embedded Gaussian Splatting}. Given multiview videos capturing a dynamic 3D scene (left), we augment a dynamic 3D Gaussian representation with a 4D language field (center). Our approach enables localizing queried texts in both space and time (top right, localized frames in red). The extracted spatio-temporal maps allow for creating various highlight effects, such as automatically visualizing a bullet-time display of the query (bottom right).}\label{fig:teaser}
\end{center}
}

\maketitle

\begin{abstract}
The emergence of neural representations has revolutionized our means for digitally viewing a wide range of 3D scenes, enabling the synthesis of photorealistic images rendered from novel views. 
Recently, several techniques have been proposed for connecting these low-level representations with the high-level semantics understanding embodied within the scene.
These methods elevate the rich semantic understanding from 2D imagery to 3D representations, distilling high-dimensional spatial features onto 3D space.
In our work, we are interested in connecting language with a dynamic modeling of the world.
We show how to lift spatio-temporal features to a 4D representation based on 3D Gaussian Splatting. %
This enables an interactive interface where the user can spatiotemporally localize events in the video from text prompts. We demonstrate our system on public 3D video datasets of people and animals performing various actions.

\end{abstract}

\section{Introduction}

Video editing is a cumbersome task that requires extensive knowledge of specialized commercial software.
The recent advancements in text-based generative modeling has shown great promise for streamlining this task. 
Given a single input video, current methods allow for modifying the appearance of the foreground or background regions~\cite{bar2022text2live,wu2023tune,ceylan2023pix2video,geyer2023tokenflow}. 
However, prior works operate directly on image pixels, which limit the representative power of these methods, particularly for handling occlusions and geometric modifications and for synthesizing novel view directions and relighting effects.

In parallel with these advancements, the emergence of volumetric neural fields has opened up new possibilities for modeling dynamic scenes, captured by monocular videos~\cite{gao2021Dynamic} as well as multi-camera setups~\cite{broxton20,li2022neural}. 
Prior work has demonstrated their potential for performing tasks such as novel-view synthesis~\cite{attal2023hyperreel}, compositing~\cite{duvall2019compositing,wu20234d}, and dense tracking of scene elements~\cite{luiten2023dynamic,wang2023tracking}. 
However, these neural techniques lack the high level semantics embodied within the scene, neglecting the key role of language in providing a semantic interface for performing volumetric text-driven edits.

In this work, we seek to connect volumetric representations capturing dynamic scenes with text describing spatio-temporal regions within
the scene. Our work, coined 4-LEGS (\textbf{{4}}D \textbf{L}anguage \textbf{E}mbedded \textbf{G}aussian \textbf{S}platting), extends language embedded volumetric representations modeling static scenes to dynamic ones, allowing to localize semantics not only in 3D space but also on the temporal domain. Technically, we ground spatio-temporal embeddings into a dynamic Gaussian Splatting representation~\cite{luiten2023dynamic}. To this end, we leverage a recent video-text model~\cite{wang2023internvid} for creating ground-truth rendered features. Once these features are grounded in the pretrained volumetric representation, we can extract volumetric probabilities, and render spatio-temporal maps from these probabilities -- finding both the temporal segment and the pixel-level probabilities for each given text query.

We show that our approach allows for creating various spatio-temporal highlight effects, such as automatically generating a bullet-time display over the desired content described using natural language (\emph{e.g.}, visualizing the moment the fox stretches from all viewpoints in Figure \ref{fig:teaser}).
Furthermore, to facilitate a quantitative evaluation, we densely annotate the six scenes from Panoptic Sports~\cite{luiten2023dynamic}, and compare with an existing 2D spatio-temporal grounding technique and a neural representation depicting static 3D environments. Our results show that \methodname{} allows for localizing textual descriptions depicting static and dynamic regions within dynamic environments, significantly surpassing the performance of alternative techniques.

\section{Related Work}
\noindent \textbf{Language Embedded 3D Neural Representations.} 
Connecting 3D representations to natural language descriptions is key to various capabilities and applications, from object localization~\cite{chen2020scanrefer, chen2022ham} and object recognition ~\cite{corona2022voxel, thomason2022language, huang2022multi} to semantic segmentation ~\cite{rozenberszki2022language} and visual question answering~\cite{gordon2018iqa, azuma2022scanqa, cascante2022simvqa}. Prior works localizing regions in 3D sparse given a textual description often operate over sparse point cloud representations~\cite{huang2022multi,chen2022ham,shafiullah2022clip,chen2023open}. %
Recently, with the rise of neural field-based 3D representations, research efforts shifted to connect these representations with textual data. LERF~\cite{kerr2023lerf} augment neural radiance fields (NeRFs)~\cite{mildenhall2021nerf} with CLIP~\cite{radford2021learning} embeddings, predicting a semantic feature field alongside the scene's geometry. HaLo-NeRF~\cite{dudai2024halo} optimize volumetric probabilties over a single text prompt at a time.

Several recent works embed spatial language features onto 3D Gaussians representations~\cite{qin2023langsplat,zhou2023feature,shi2023language, labe2024dgd}, which achieve real-time inference, significantly outperforming NeRF-based techniques. As inference speed is critical for enabling an interactive interface for spatiotemporally localizing events in videos, in our work, we also embed language onto Gaussians. However, unlike these prior works, we focus on the modeling of dynamic scenes, augmenting a dynamic 3D Gaussian representation~\cite{luiten2023dynamic} with high-dimensional spatio-temporal language features. Additionally, several work~\cite{kim2024garfield,ye2023gaussian,cen2023segment,hu2024semantic} segment 3D scenes via high-dimensional features optimized by leveraging 2D masks provided, for instance, by Segment Anything~\cite{kirillov2023segany}. However, these works do not directly utilize language, nor do they allow for text-based querying.

\smallskip
\noindent \textbf{Spatio-temporal Grounding in Image Sequences.} 
Spatial (or visual) grounding aims at localizing regions in an image given a textual description. It is a fundamental capability in visual semantic tasks, receiving ongoing attention in recent years~\cite{uijlings2013selective, zitnick2014edge, ren2015faster,jie2016scale,hu2017modeling,cui2021s}.
Considering image sequences, several works~\cite{gao2017tall, anne2017localizing} address the task of temporal grounding, which determine the temporal boundaries corresponding to input textual descriptions. 
Spatio-temporal grounding extends these two tasks, localizing textual descriptions both in space and time.
Earlier attempts assume additional provided cues, such as human gaze~\cite{vasudevan2018object} or the set of objects for grounding~\cite{yamaguchi2017spatio,zhou2018weakly}. Recent work~\cite{chen2019weakly,zhang2020does,yang2022tubedetr}  explore a more general 2D setting, outputting a sequence of bounding boxes, referred to as a spatio-temporal tube.
In our work, we perform spatio-temporal grounding in 4D space, bypassing challenges such as occluded scene elements. We render these volumetric probabilities to pixel-level probabilities, providing more accurate grounding results.

\smallskip
\noindent \textbf{Computational Video Editing.} Computational text-driven editing techniques have seen tremendous progress with the emergence of powerful large-scale vision-language models~\cite{radford2021learning} and diffusion models~\cite{ho2020denoising}. Although methods primary focus on editing single images~\cite{nichol2021glide,hertz2022prompt,patashnik2023localizing,ge2023expressive}, the problem of text-guided video editing is also gaining increasing interests.
Text2live~\cite{bar2022text2live} combines CLIP and layered neural atlas models~\cite{kasten2021layered} to generate consistent videos with new visual effects. Recent work leverage pretrained diffusion models, either with~\cite{wu2023tune,chai2023stablevideo,liu2023video} or without~\cite{ceylan2023pix2video,geyer2023tokenflow} additional training. 

Our volumetric spatio-temporal grounding technique could potentially allow for performing such text-driven video edits in 3D space. In our work, we demonstrate several applications benefiting from our approach, which are related to video editing tasks that were not previously tied to text guidance. For instance, prior work has tapped into the problem of automatic viewpoint selection \cite{leake2017computational,arev2014automatic}, editing the timing of different motions in a video \cite{lu2020layered}, and time slice video synthesis~\cite{cui2017time}, which aligns and stitches multiple videos into a single video. Unlike the applications we show, these tasks were previously addressed using a 2D representation, which allows for only restricted manipulation capabilities and small divergence from the captured camera viewpoints.

    \begin{figure*} %
\centering
\jsubfig{\includegraphics[width=0.99\textwidth]{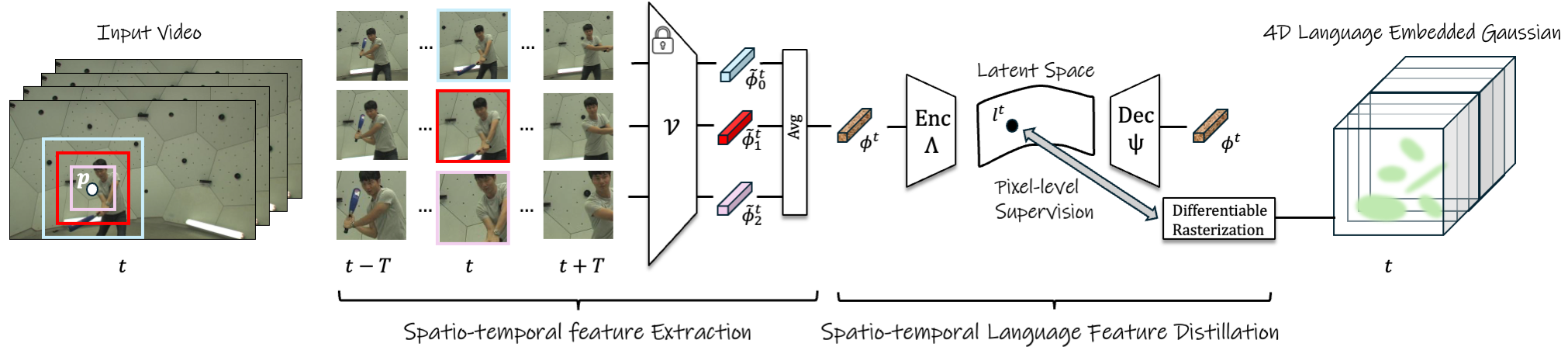}}{}
\vspace{-4pt}
\caption{\textbf{4D Language Field Optimization}. Given multiple videos capturing a dynamic 3D scene, we first extract pixel-aligned spatio-temporal language features at multiple scales using a video-text model. We average these features to produce spatio-temporal features, which are encoded into a more compact latent space that is used for supervising the optimization of a 4D language embedded Gaussian. 
}
\label{fig:overview}
\end{figure*}

    \section{Method}
    To embed language features onto a volumetric modeling of a dynamic scene, we propose an approach that augments a dynamic 3D Gaussian representation ~\cite{luiten2023dynamic,li2023spacetime,lu20243d,yang2023deformable,wu20234d, sun20243dgstream, lin2024gaussian}. 
    3D Gaussian Splatting~\cite{kerbl20233d} (3DGS) is an explicit representation of a 3D scene that uses a collection of anisotropic 3D Gaussians. 
    These 3D Gaussians are projected onto camera views through a differentiable  rasterization process, and their attributes are optimized to represent the 3D scene, as captured by a collection of input images. 
    Dynamic 3DGS extend this representation to model a dynamic 3D scene, optimizing additional attributes such as the Gaussian's center and 3D rotation across time. 
    
    In particular, we use the representation from Dynamic 3D Gaussians~\cite{luiten2023dynamic} which optimizes a set of 3DGS --- one for each timestep in the input videos. 
    Each 3DGS is optimized per timestep, with temporal consistency enforced over all parameters except the ones representing the Gaussian center and rotations.
    
    We show how to optimize a 4D language field onto a pretrained dynamic 3DGS representation via pixel-aligned spatio-temporal language features extracted from the input videos (Section \ref{sec:optimization}, Figure \ref{fig:overview}). We then describe how to query the trained 4D features using open-vocabulary prompts (Section \ref{sec:querying}). Our method outputs spatio-temporal relevancy maps, predicting both the temporal segment containing the queried description (\emph{i.e.}, exact time-frames) and spatial pixel-level probabilities for rendered views.

    \subsection{4D Language Field Optimization}
    \label{sec:optimization}
    In this section, we describe our approach for optimizing a 4D language field. 
    We first explain how we extract pixel-aligned features from the input videos depicting the dynamic scene (Figure \ref{fig:overview}, left). Afterwards, we describe how these features are used for supervising the optimization of a 4D language feature field (Figure \ref{fig:overview}, right).

    \medskip
    \noindent \textbf{Spatio-Temporal Feature Extraction}. We first generate ground-truth pixel-aligned features for each timestamp in all input videos, which we later distill into a dynamic 3DGS representation.
    Existing methods that optimize language embedded 3D representations~\cite{kerr2023lerf, qin2023langsplat} typically employ the CLIP~\cite{radford2021learning} image encoder for extracting image features. %
    However, CLIP is pretrained only over image--caption pairs, and consequently cannot easily capture temporal differences. 
    
    Therefore, we extract features using ViCLIP~\cite{wang2023internvid}, a video--text model based on CLIP, which contains a video encoder $\mathcal{V}$ that processes a sequence of input video frames. %
    To obtain pixel-aligned features (rather than image-aligned ones), we follow prior work targeting static 3D scenes and extract an image pyramid for each timestep $t$, containing image crops of different scales $s\in S$. Spatio-temporal features of scale $s$ are extracted by encoding a temporal sequence of image crops $c^{t}_{s}$. Written precisely, for each pixel $\textbf{p}$ and timestamp $t$, we extract spatio-temporal features of multiple scales $s$:
    \begin{equation}
    \label{formula: feature_concat}
         \tilde{\phi}^{t}_{s}(\textbf{p}) = \mathcal{V}(\text{Concat}(c^{k}_{s}(\textbf{p}) \; | \; k=t-L/2,...,t,...,L/2+t)),
    \end{equation}
    where $L$ is the length of the temporal tube, centered around $t$.

    To obtain the final spatio-temporal features used for optimization, we average the features across all scales:
    \begin{equation}
    \label{formula: feature_avg}
        \phi^{t}(\textbf{p}) =\frac{\sum_{s \in S}\tilde{\phi}^{t}_{s}(\textbf{p})}{|S|}.
    \end{equation} 
    In practice, we use a sliding window of crops as computing such a pyramid of crops centered around each pixel is prohibitively expensive. That is, to extract features for each pixel at a specific scale, we perform bilinear interpolation between the features of the four nearest crops and then average these bilinearly interpolated features to obtain our final spatio-temporal features used for supervising the optimization procedure.
    Averaging the features across scales allows for significantly reducing the inference speed, and yields a more feasible distillation pipeline, which we elaborate on next.

    \medskip
    \noindent \textbf{Spatio-Temporal Language Feature Distillation}.
    Now, we wish to distill the spatio-temporal features extracted in the previous section into a Dynamic 3D Gaussian Splatting representation, denoted by $\mathcal{D}_{T}$. 
    As we use an explicit representation for modeling our scene, each 3DGS in $\mathcal{D}_{T}$ is represented by 200-300k Gaussians. Directly learning the high-dimensional ViCLIP features significantly increases the memory requirements for storing 3D Gaussians, and could potentially lead to memory issues. %
    To reduce memory cost and improve efficiency, we follow \cite{qin2023langsplat} and pretrain a per-scene autoencoder, consisting of an encoder $\Lambda{}$ and a decoder $\Psi{}$. In our case, the autoencoder maps $D$-dimensional dense spatio-temporal features to a $d$-dimensional latent space (where $d\ll D$), thus reducing the memory requirements during distillation. The autoencoder is trained using an $L_{2}$ loss with a cosine distance loss as regularization. After training our scene-specific autoencoder, we encode all of the extracted spatio-temporal features into the scene-specific latent space using $\Lambda{}$.

    For spatio-temporal feature distillation, we equip each Gaussian with a $d$-dimensional trainable spatio-temporal feature $L^{t}_i$, denoting the spatio-temporal feature of the $i$-th Gaussian at timestep $t$ in $\mathcal{D}_{T}$.
    To encourage smoothness between the distilled features, we apply \emph{Self Attention} to between each Gaussian and its local neighborhood. In other words, we compute $\Bar{L^{t}_i} = \text{SelfAttention}(\text{kNN}(x^{t}_i))$, where $x^{t}_i$ is the mean location of the $i$-th Gaussian at timestep $t$ and $\text{kNN}$ returns the features of the $k$ spatially nearest Gaussians.

    We then render these smoothed spatio-temporal features at time $t$ using volumetric rendering and tile-based rasterization as in \cite{kerbl20233d}: 
    \begin{equation}
        l^{t}(\textbf{p}) = \sum_{i \in \mathcal{N}} \Bar{L^{t}_i} \alpha_i \prod_{k=1}^{i-1} (1 - \alpha_k),
        \label{eq:rendering_lang}
    \end{equation}
    where $\mathcal{N}$ denotes the Gaussians in the tile overlapping with the given pixel, $l^{t}(\textbf{p})$ is the rendered spatio-temporal feature at pixel $\textbf{p}$, $\alpha_i$ is the opacity of the $i$-th Gaussian and $\prod_{k=1}^{i-1} (1 - \alpha_k)$ is the product of opacity values of previous
    Gaussians in the tile overlapping the same pixel. This allows for lifting the spatio-temporal features to interact directly with the dynamic 3D representation.
    
    We optimize per timestep, thus given a time $t$, we compute the loss over all training views $j \in N$ corresponding to input videos $\{{V}_j | j =1,2,...N\}$:
    \begin{equation}
         \mathcal{L}^{t}_{feat} =  \sum_{j \in N} \left\lVert l^{t}_{j}(\textbf{p}) - \Lambda{}(\phi^{t}_{j}(\textbf{p})) \right\rVert_1.
        \label{eq:loss_feat_t}
    \end{equation}

    Our \methodname{} representation is a set of $T$ spatio-temporal language embedded 3DGS, each of which has a distilled latent spatio-temporal feature for each Gaussian. During inference, given a time $t$ we follow Equation (\ref{eq:rendering_lang}) and render the latent spatio-temporal features from 3D to 2D and then decode the rendered features using the pre-trained scene specific decoder, $\Psi{}$, resulting in a per-pixel dense spatio-temporal feature map $\Psi{}(l^{t}(\textbf{p})) \in \mathbb{R}^{H\times W\times D}$.

    \subsection{Spatio-Temporal Open-Vocabulary Querying}
    \label{sec:querying}
    We now leverage our 4D feature representation to localize a language query in both space and time. Essentially, we seek not only the \emph{object} described by the query, but also the \emph{action} it performs.
    The ViCLIP latent space is well-aligned between videos and text enabling our trained \methodname{} to easily support spatio-temporal open-vocabulary queries.
    Given a decoded spatio-temporal feature $\features{} = \Psi{}(l)$ extracted from our representation and an encoded text query $\phi_{\text{Q}}$, we compute compute a relevancy score $\text{R}(\features{}, \phi_{\text{Q}})$ following LERF~\cite{kerr2023lerf}.
    Specifically,
    \begin{equation}
       \text{R}(\features{}, \phi_{\text{Q}}) = \min_{\phi_{\text{C}}} 
       \frac
       {\exp(\features{} \cdot \phi_{\text{Q}})}
       {\exp(\features{} \cdot \phi_{\text{Q}}) + \exp(\features{} \cdot \phi_{\text{C}})},
        \label{formula: relevancy}
    \end{equation}
    where $\phi_{\text{C}}$ are the ViCLIP embeddings computed from a predefined canonical set of phrases 
    (\textit{``object"}, \textit{``things"}, \textit{``stuff"}, and \textit{``texture"}). This score expresses the probability of how well the feature $\features{}$ is described by text query Q compared to all the canonical phrases.
    
    \smallskip
    \noindent \textbf{Temporal Localization}.
    For temporal localization we compute this relevancy score over the \emph{volumetric} language embeddings at each Gaussian $G$ and time step $t$. In other words, for temporal localization we use $\features{} = \Psi{}(L^t)$. We chose to compute relevancy directly in 4D space instead of using rendered image-space features as this does not depend on any chosen viewpoint and thus is robust to occlusions.
    
    We seek features with a relevancy score $\text{R}(\features{}, \phi_{\text{Q}})$ greater than 0.5, meaning that the text query is more likely than any of the canonical phrases.
    To find the timestamps $t$ that match a  text query $\phi_{\text{Q}}$, we compute the fraction $s_t$ of how many Gaussian match the query at $t$ compared to the entire 4D sequence of Gaussians. To compute this fraction, we first extract the set of Gaussians with a relevancy score above 0.5, and compute the average relevancy score $\text{R}_{\text{avg}}$ among these \emph{relevant} Gaussians.
    We use this measure to compute the fraction $s_t$ as follows:
    \begin{equation}
        s_t = \frac
        {\#\{\text{R}(\phi^{t}_{\text{g}}, \phi_{\text{Q}}) > \text{R}_{\text{avg}} \,|\, g \in G\}}
        {\#\{\text{R}(\phi^{t'}_{\text{g}}, \phi_{\text{Q}}) > \text{R}_{\text{avg}} \,|\, g, t' \in G, T\}},
    \end{equation}
    where $G$ is the set of Gaussians and $T$ is the number of timesteps. 
    
    Then we predict that the query occurs in segment $(t_{1}, t_{2})$ if $s_{t_{i}} > k, \forall i \in \{t_{1},t_{2}\}$ for a threshold $k$.
    We set $k=\frac{1}{T}$ to select frames with more active Gaussians than the average over the sequence. Additional details are provided in the supplementary material.

    \smallskip
    \noindent \textbf{Spatial Localization}.
    Once the temporal segments associated with the text query are localized, we \emph{spatially} localize the action in the rendered output viewpoints. For all pixels $p$ we compute $\text{R}(\Psi{}(l^t(\mathbf{p})), \phi_{\text{Q}})$, where $\Psi{}(l^t(\mathbf{p}))$ is the \emph{rendered} language embedding at $\mathbf{p}$.
    
    \smallskip
    \noindent \textbf{Runtime}. Open-vocabulary querying is achieved at 10 FPS on a single RTX A5000 GPU. This includes rendering the language features, decoding them and computing the relevancy score.

\section{Evaluation}
In this section, we evaluate our method, comparing it to prior work addressing similar settings, including a 2D technique performing spatio-temporal grounding and a 3D method that embeds language features onto representations depicting static 3D environments. To perform a quantitative evaluation, we need multiple videos capturing dynamic environments, paired with ground truth spatio-temporal segmentation maps. To the best of our knowledge, existing datasets contain bounding box annotations for a single input video depicting the scene~\cite{tang2021human, Zhang_2020_CVPR}, as prior work only address a 2D setting. Therefore, we introduce the \datasetname{} bechmark in Section \ref{sec:benchmark}. We perform a comparison to prior work in Section \ref{sec:comparisons}, and ablations are provided in Section \ref{sec:ablations}. Finally, limitations are discussed in Section \ref{sec:limitations}.

\subsection{The \datasetname{} Benchmark}
\label{sec:benchmark}
We assemble the \datasetname{} benchmark by pairing the six multiview video sequences from the Panoptic Sports dataset used in ~\cite{luiten2023dynamic} (originally from the Panoptic Studio dataset~\cite{joo2015panoptic}) %
with ground-truth spatio-temporal segmentation maps, corresponding to fifteen different textual queries. We annotate five camera views per sequence, yielding a set of 90 annotated video--caption pairs, comprised of 13500 frames overall. Additional details are provided in the
supplementary material.

\smallskip
\noindent \textbf{Metrics.} We adopt a set of evaluation metrics that consider both the spatial and temporal dimensions, as well as the 3D nature of our multi-view data. 
We follow 2D techniques (\emph{e.g.} ~\cite{Zhang_2020_CVPR}) and define $vIoU$ as

\begin{equation}
    vIoU = \frac{1}{|S_u|}\sum_{t \in S_i}IoU(\hat{b}_t, b_t),
    \label{formula: viou}
\end{equation}
where $S_u$ and $S_i$ are the sets of frames in the union and intersection between the ground-truth and the prediction, and $\hat{b}_t$ and $b_t$ are the predicted and ground truth bounding boxes at time t. We extend this metric for spatio-temporal grounding at the pixel level by defining $vAP$. We calculate $vAP$ by measuring Average Precision (AP) for each annotated frame, evaluating segmentation quality on the pixel level. We then calculate:
\begin{equation}
    vAP = \frac{1}{|S_u|} \sum_{t\in S_i}{AP(\hat{m}_{t},m_{t})},
    \label{formula: vap}
\end{equation}
where $\hat{m}$ and m are the predicted and ground-truth segmentations.

To separately evaluate the quality of the temporal localization, we also measure Recall (tRec), Precision (tPrec), AP (tAP) and temporal IoU (tIoU) for each video, and average these across all annotated videos.
Additionally, to provide a measure for consistency across views, we also report the standard deviation among the different views over all metrics, averaging over the different scenes.

\subsection{Baselines}
\label{sec:baselines}
Prior work do not directly address the task of spatio-temporal grounding over dynamic volumetric representations. Hence, we compare our approach to a 2D spatio-temporal grounding method, to illustrate the benefits of operating over a volumetric representation. %
We also compare against a method embedding language features over a static 3D representation. %

For the 2D spatio-temporal grounding baseline methods, we consider TubeDeTR~\cite{yang2022tubedetr} and Context-Guided Spatio-Temporal Video Grounding (CGSTVG) \cite{gu2024context}. TubeDETR uses visual and text encoders to create video-text features based on short clips extracted from the video. A space-time decoder is then used to predict both bounding boxes for each frame and the start and end frame by using temporal self-attention to model the long-range temporal interactions in the video and masked cross-attention. CGSTVG uses a multimodal encoder for feature extraction and a context-guided decoder by cascading
a set of decoding stages for grounding. In each decoding stage, instance context is mined to guide query learning for better localization. The output of these methods is a sequence of bounding boxes matching to each frame of the estimated temporal tube.

For the comparison to a 3D language embedding technique, we compare with LangSplat~\cite{qin2023langsplat}, which optimizes a per-scene 3D static language field on a 3D Gaussian Splatting~\cite{kerbl20233d} representation using hierarchical SAM~\cite{kirillov2023segany} masks and CLIP~\cite{radford2021learning} features. 
To adapt this method for our temporal scenes, we optimize a LangSplat per timestep (as LangSplat only operates over static scenes). We optimize LangSplats for a small number of timesteps, as exhaustively optimizing LangSplat models across all timestamps and scenes would be computationally intractable.
As one LangSplat is made up of three 3DGS language fields, each for a different semantic scale, we average the relevancy map computed from each scale to create a single output for comparison.

\subsection{Comparisons} 
\label{sec:comparisons}

Temporally localizing actions presents unique challenges as actions span multiple timesteps and can occur multiple times. Our 4D representation addresses these challenges in ways that 2D and 3D baselines cannot.

\begin{figure*}
\rotatebox{90}{\whitetxt{Tubeg}}
  \jsubfig{\includegraphics[height=1.21cm]{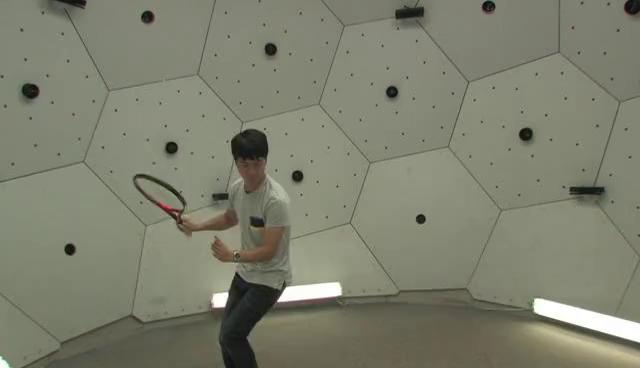} \hfill \includegraphics[height=1.21cm]{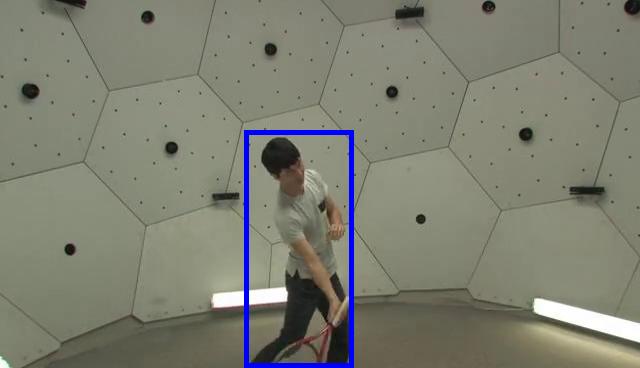}
  \includegraphics[height=1.21cm]{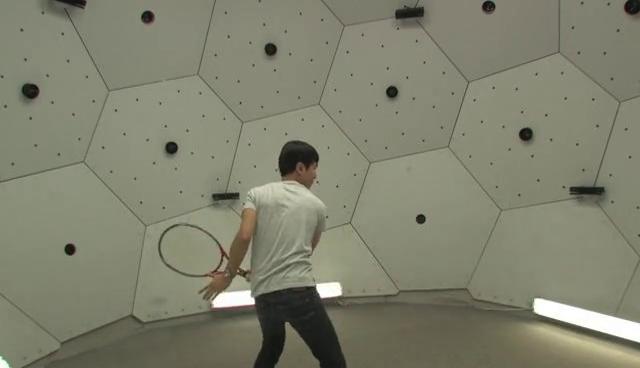}
    \includegraphics[height=1.21cm]{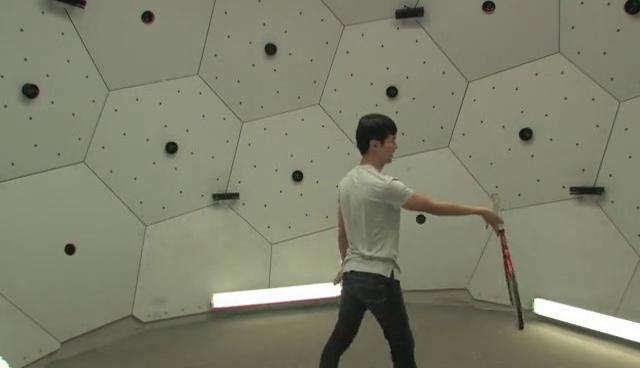}
  }{ }
  \hfill
   \jsubfig{\includegraphics[height=1.21cm]{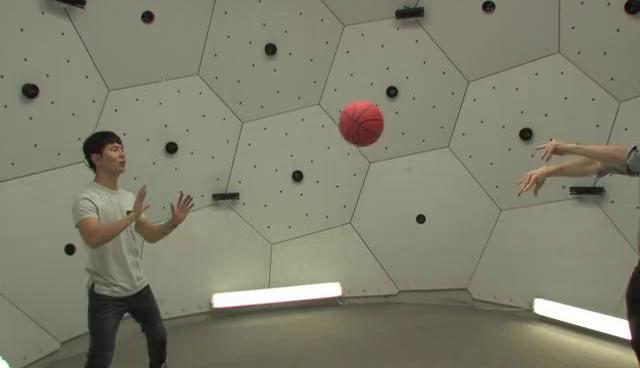} \hfill \includegraphics[height=1.21cm]{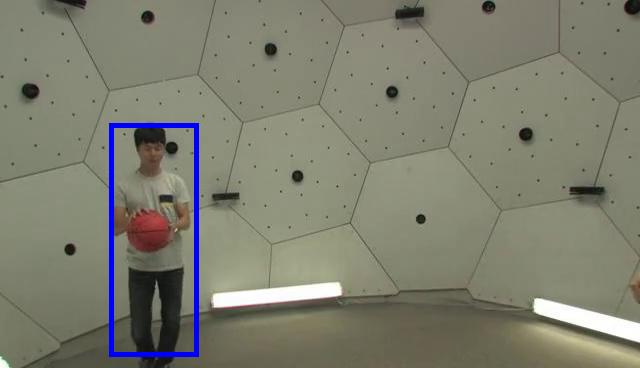}
  \includegraphics[height=1.21cm]{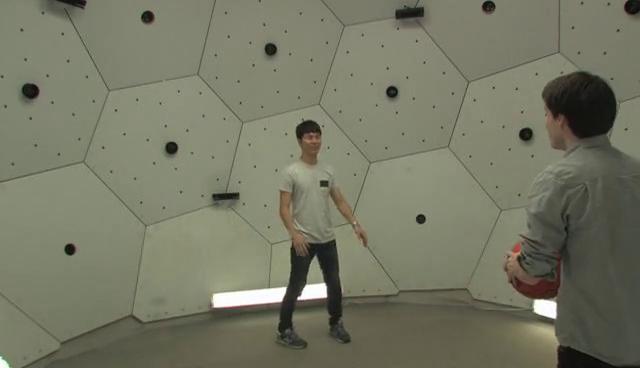}
    \includegraphics[height=1.21cm]{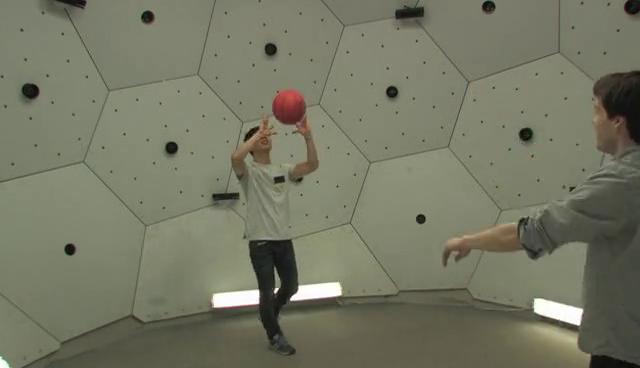}
  }{ }  \\ 
   \rotatebox{90}{\hspace{-10pt}TubeDETR\whitetxt{g}}
    \jsubfig{\includegraphics[height=1.21cm]{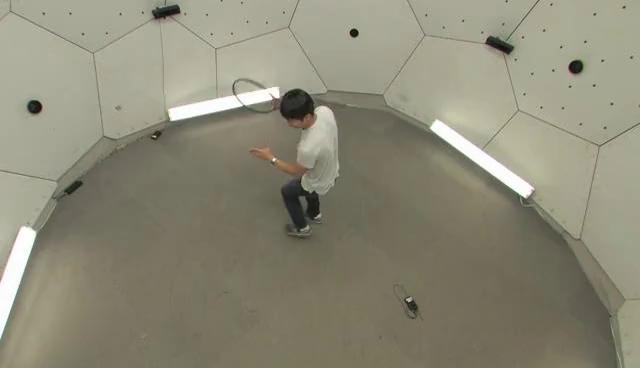} \hfill \includegraphics[height=1.21cm]{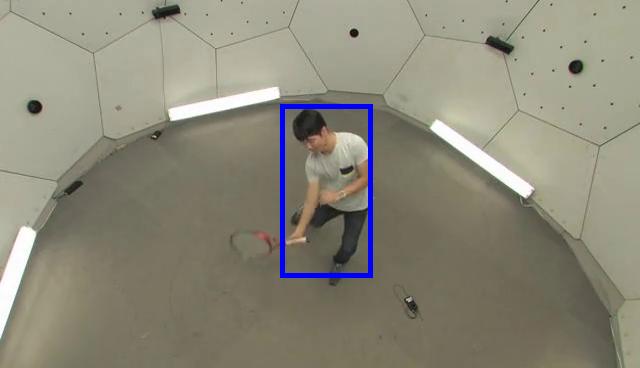}
  \includegraphics[height=1.21cm]{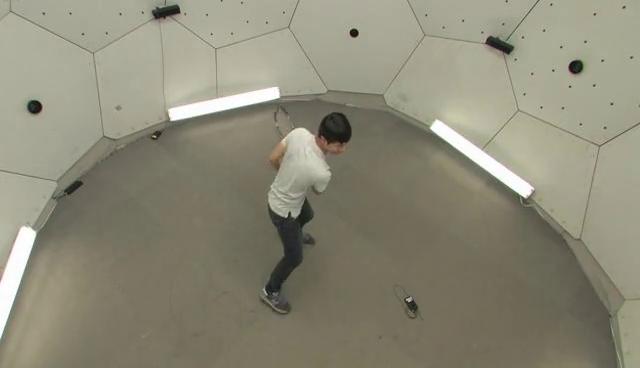}
    \includegraphics[height=1.21cm]{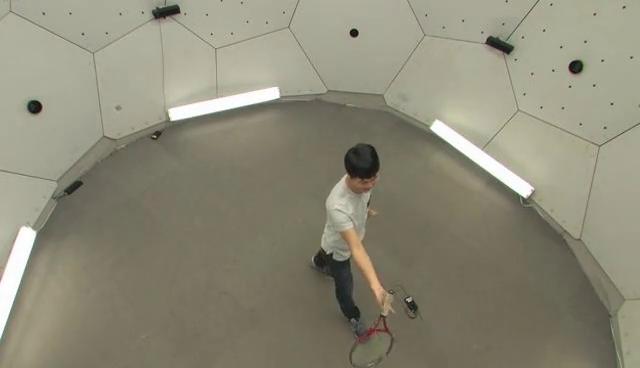}
  }{ }
    \hfill
  \jsubfig{\includegraphics[height=1.21cm]{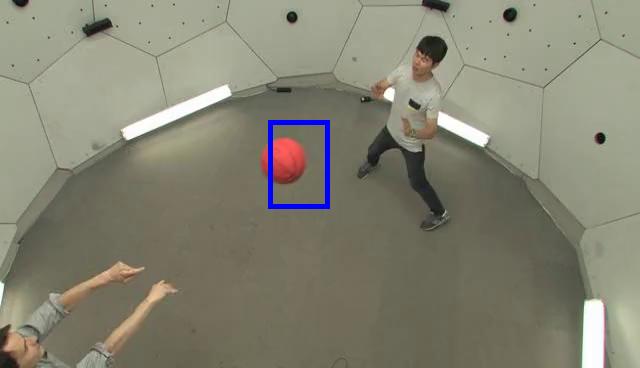} \hfill \includegraphics[height=1.21cm]{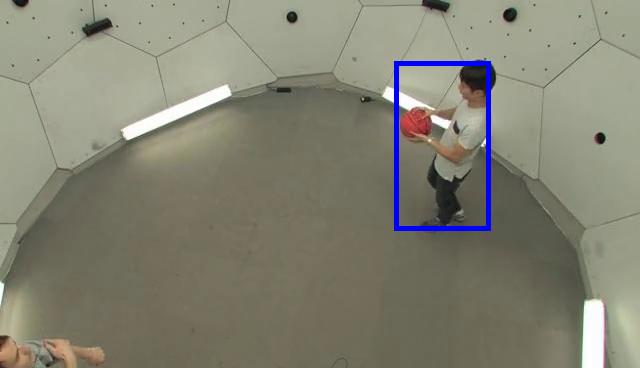}
  \includegraphics[height=1.21cm]{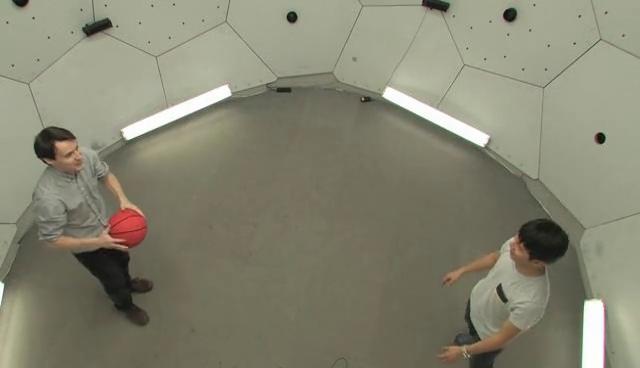}
    \includegraphics[height=1.21cm]{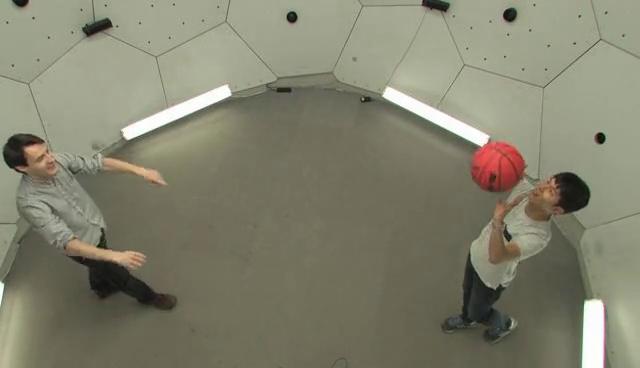}
  }{ }  \\  
    \rotatebox{90}{\whitetxt{Tubeg}}
  \jsubfig{\includegraphics[height=1.21cm]{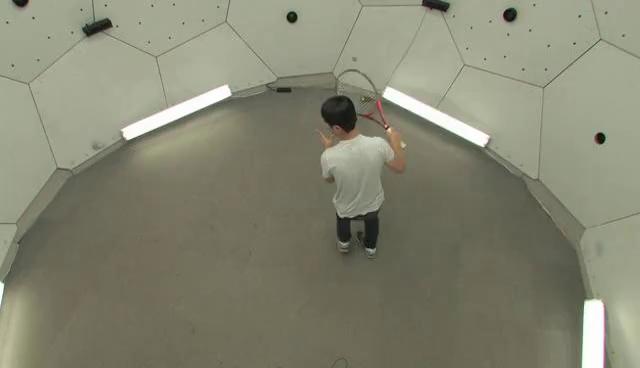} \hfill \includegraphics[height=1.21cm]{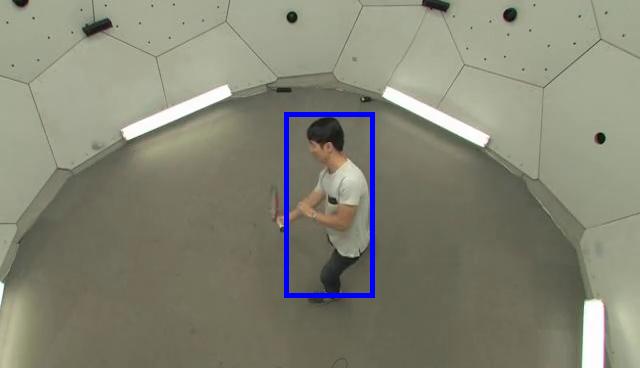}
  \includegraphics[height=1.21cm]{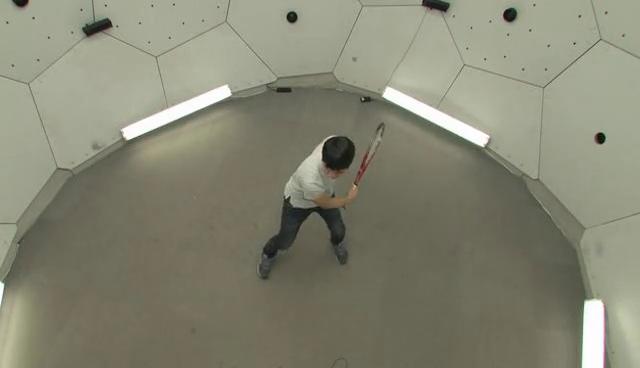}
    \includegraphics[height=1.21cm]{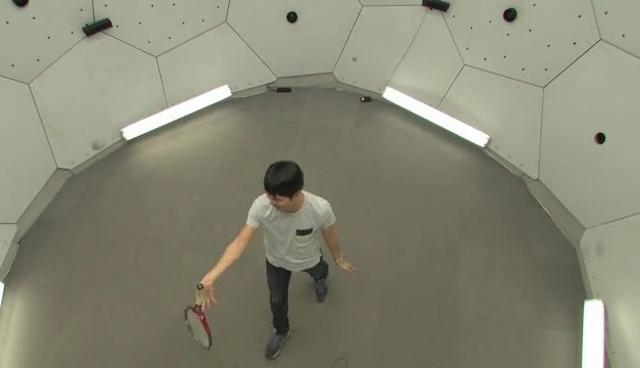}
  }{ }
  \hfill
  \jsubfig{\includegraphics[height=1.21cm]{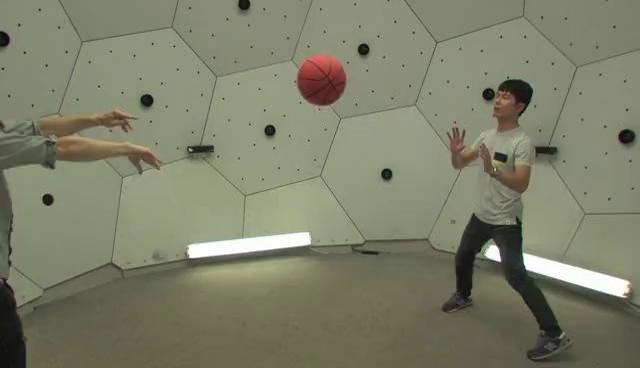} \hfill \includegraphics[height=1.21cm]{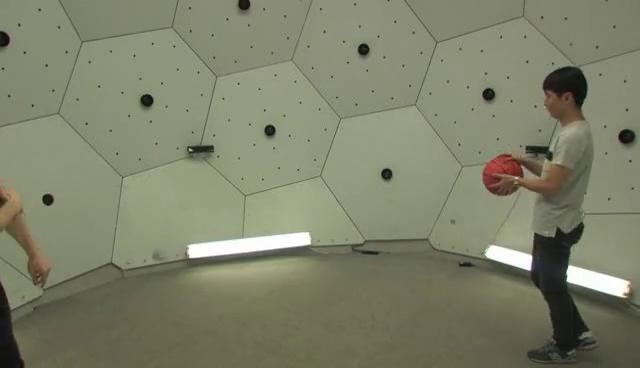}
  \includegraphics[height=1.21cm]{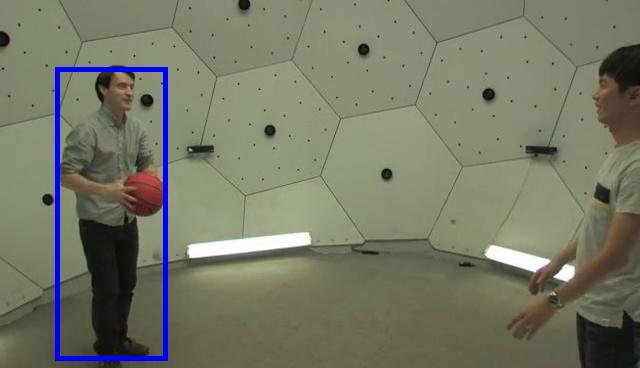}
    \includegraphics[height=1.21cm]{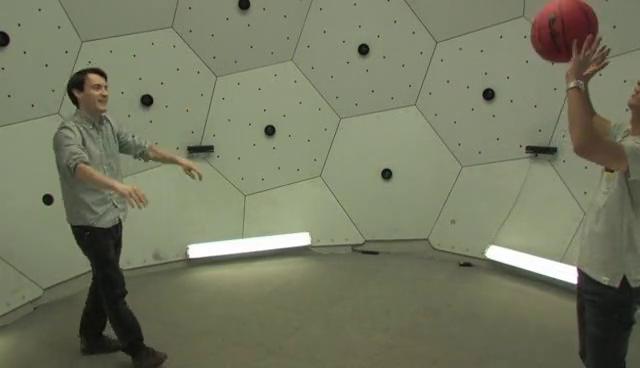}
  }{ } 
  \rotatebox{90}{\whitetxt{Tubeg}}
  \jsubfig{\includegraphics[height=1.21cm]{images/comparisons/tubedetr/0/a_person_playing_tennis_cam0_frame0_no.jpg} \hfill \includegraphics[height=1.21cm]{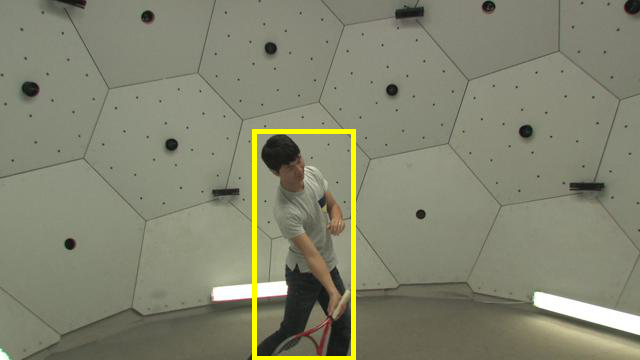}
  \includegraphics[height=1.21cm]{images/comparisons/tubedetr/0/a_person_playing_tennis_cam0_frame89_no.jpg}
    \includegraphics[height=1.21cm]{images/comparisons/tubedetr/0/a_person_playing_tennis_cam0_frame149_no.jpg}
  }{ }
  \hfill
   \jsubfig{\includegraphics[height=1.182cm]{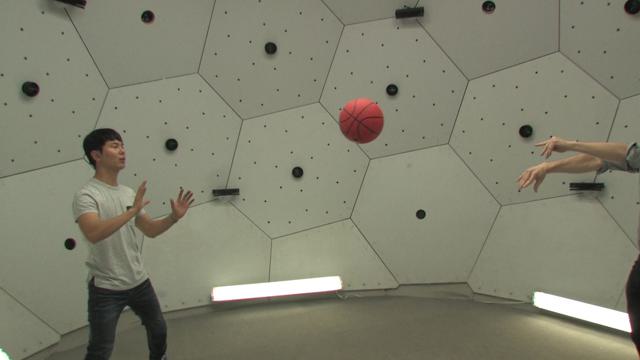} \hfill \includegraphics[height=1.182cm]{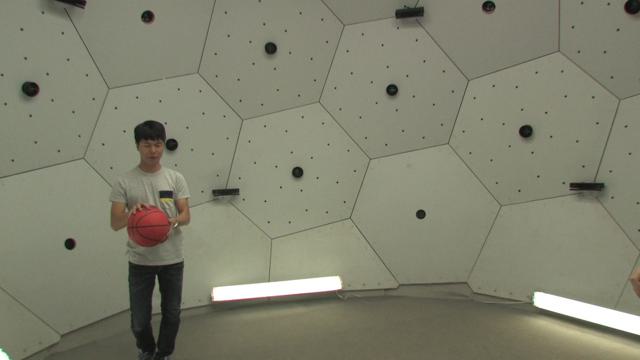}
  \includegraphics[height=1.182cm]{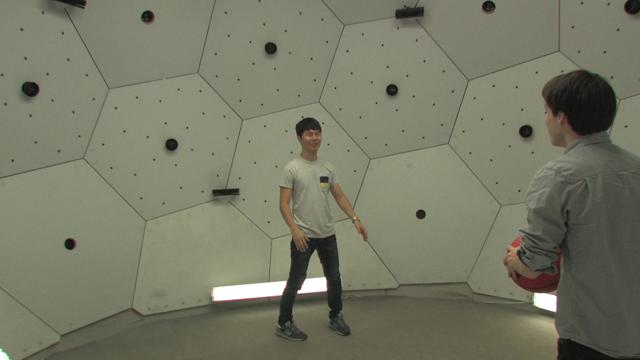}
    \includegraphics[height=1.182cm]{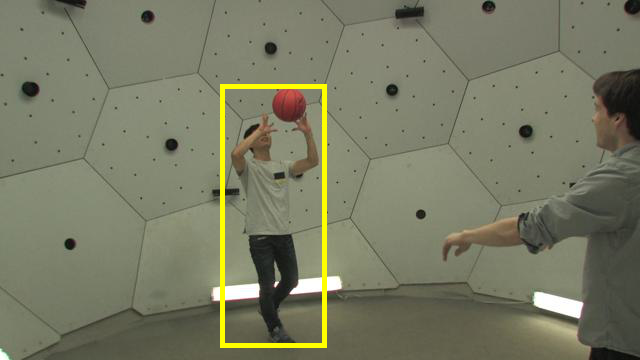}
  }{ } \\
  \rotatebox{90}{\hspace{-7pt}CGSTVG\whitetxt{g}}  \jsubfig{\includegraphics[height=1.21cm]{images/comparisons/tubedetr/6/a_person_playing_tennis_cam6_frame0_no.jpg} \hfill \includegraphics[height=1.21cm]{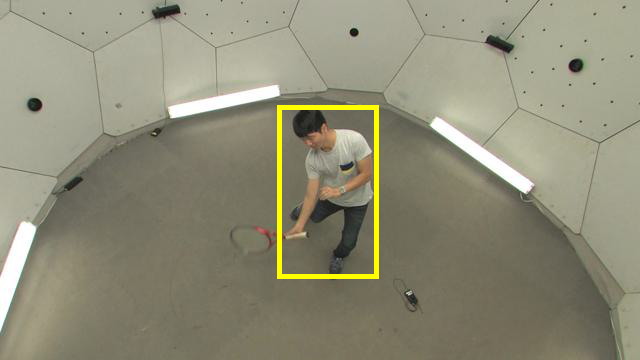}
  \includegraphics[height=1.21cm]{images/comparisons/tubedetr/6/a_person_playing_tennis_cam6_frame89_no.jpg}
    \includegraphics[height=1.21cm]{images/comparisons/tubedetr/6/a_person_playing_tennis_cam6_frame149_no.jpg}
  }{ }
  \hfill
\jsubfig{\includegraphics[height=1.182cm]{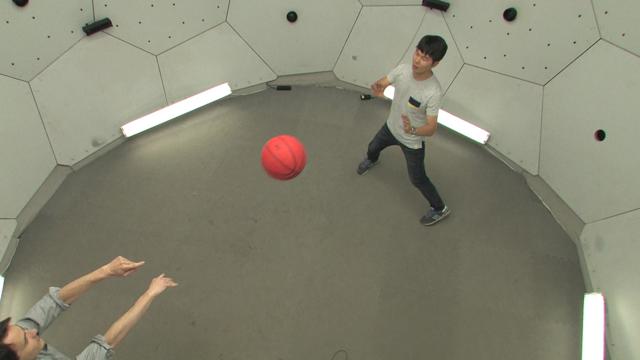} \hfill \includegraphics[height=1.182cm]{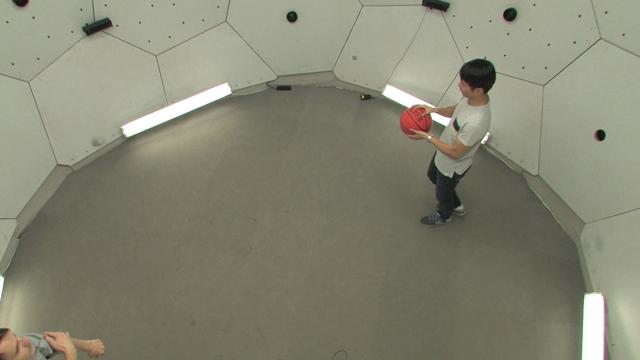}
  \includegraphics[height=1.182cm]{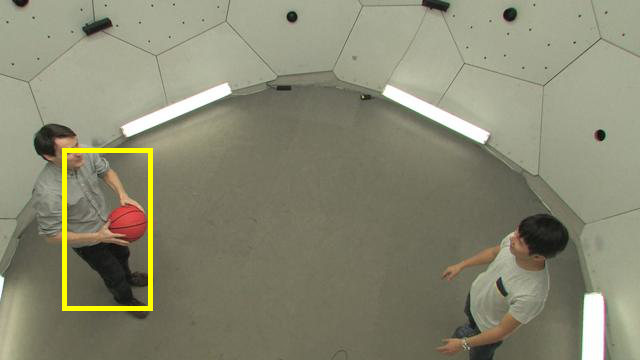}
    \includegraphics[height=1.182cm]{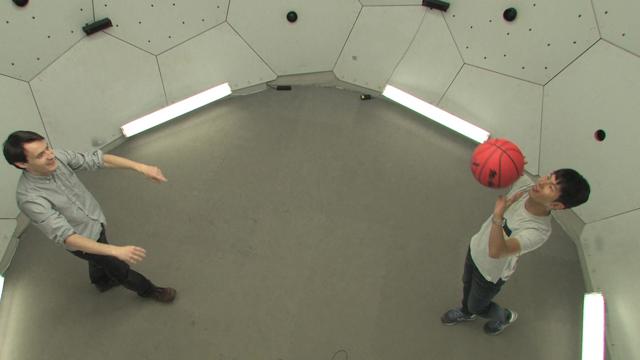}
  }{ } \\
  \rotatebox{90}{\whitetxt{Tubeg}}
   \jsubfig{\includegraphics[height=1.21cm]{images/comparisons/tubedetr/19/a_person_playing_tennis_cam19_frame0_no.jpg} \hfill \includegraphics[height=1.21cm]{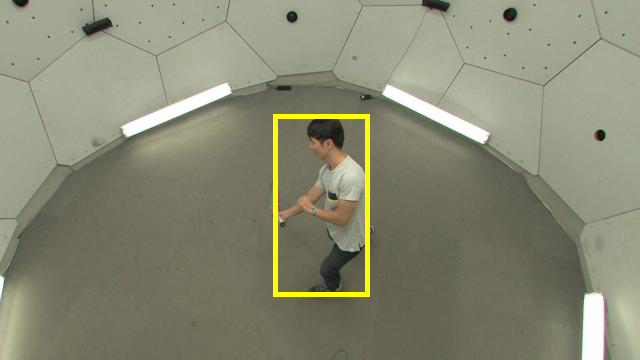}
  \includegraphics[height=1.21cm]{images/comparisons/tubedetr/19/a_person_playing_tennis_cam19_frame89_no.jpg}
    \includegraphics[height=1.21cm]{images/comparisons/tubedetr/19/a_person_playing_tennis_cam19_frame149_no.jpg}
  }{ }
  \hfill
  \jsubfig{\includegraphics[height=1.182cm]{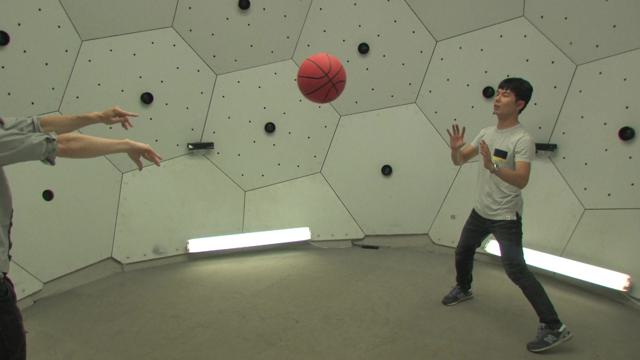} \hfill \includegraphics[height=1.182cm]{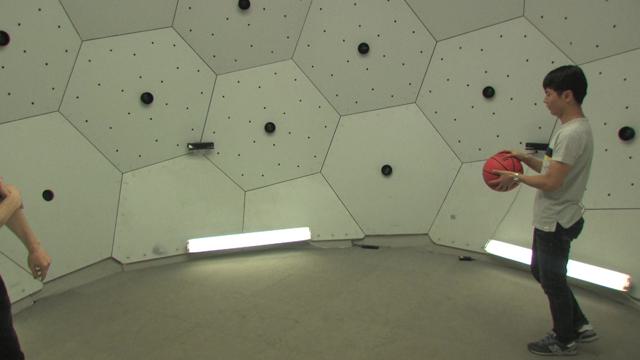}
  \includegraphics[height=1.182cm]{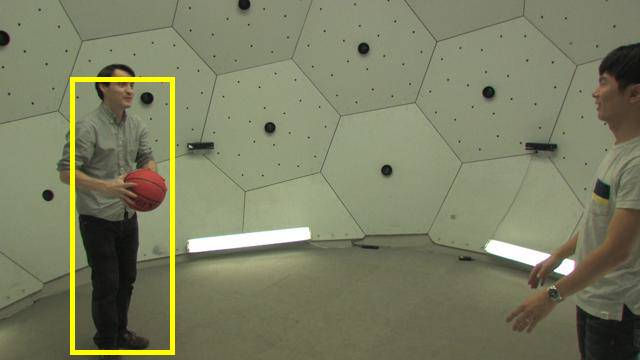}
    \includegraphics[height=1.182cm]{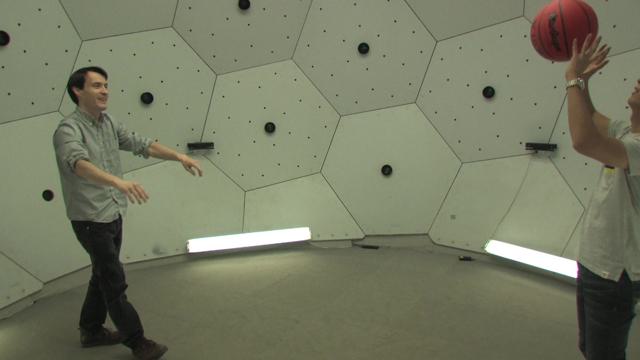}
  }{ } 

  \rotatebox{90}{\whitetxt{Tubeg}}
  \jsubfig{\includegraphics[height=1.182cm]{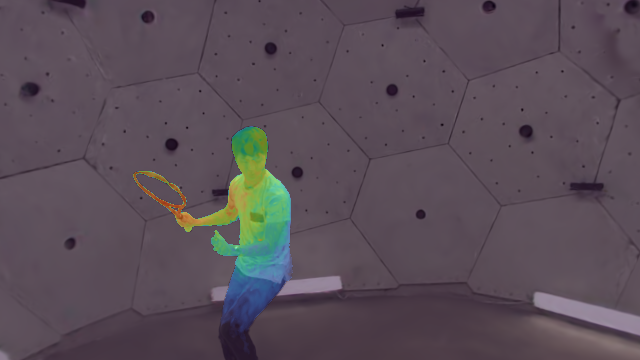} \hfill \includegraphics[height=1.182cm]{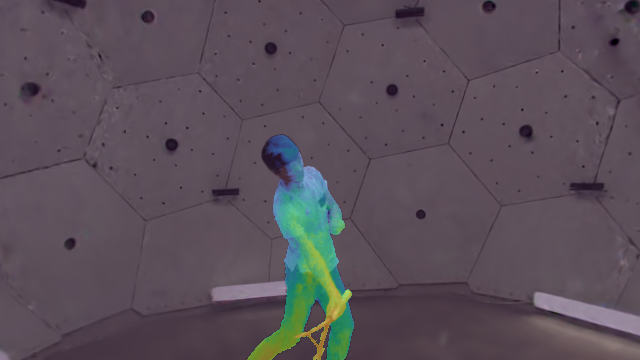}
  \includegraphics[height=1.182cm]{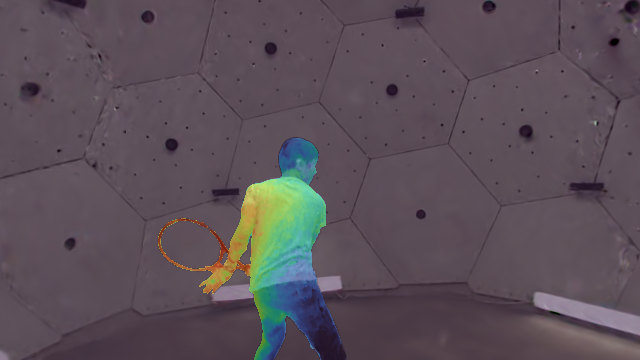}
    \includegraphics[height=1.182cm]{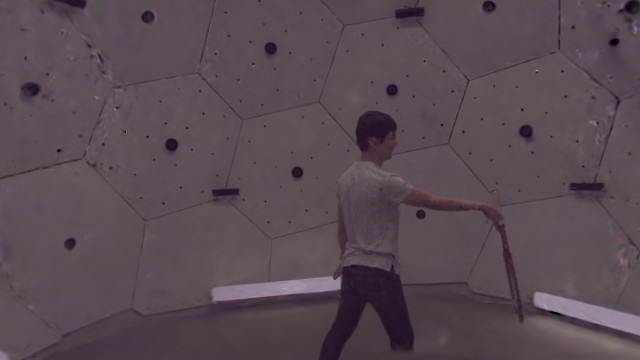}
  }{ }
  \hfill
   \jsubfig{\includegraphics[height=1.182cm]{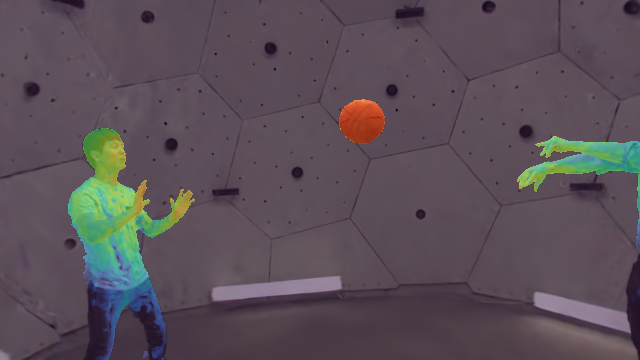} \hfill \includegraphics[height=1.182cm]{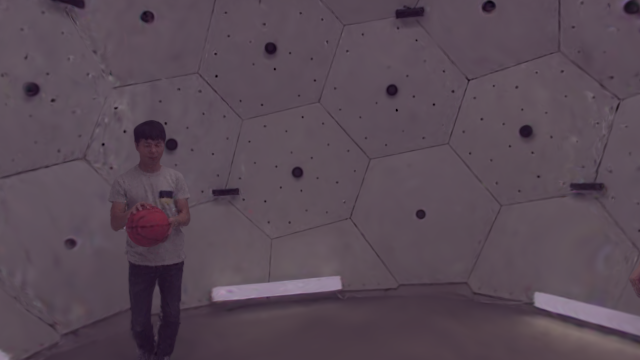}
  \includegraphics[height=1.182cm]{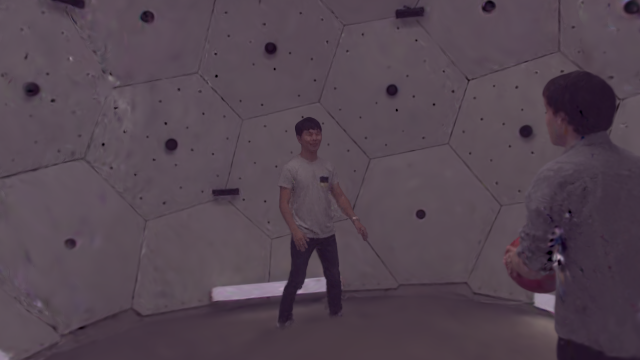}
    \includegraphics[height=1.182cm]{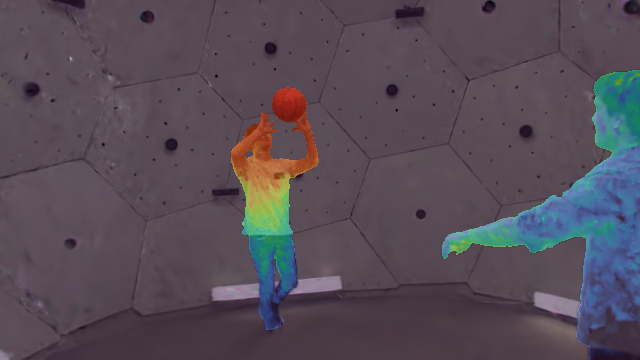}
  }{ } \\
  \rotatebox{90}{\whitetxt{gg}Ours}  \jsubfig{\includegraphics[height=1.182cm]{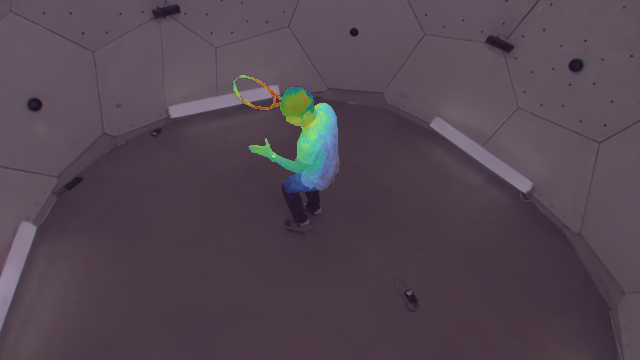} \hfill \includegraphics[height=1.182cm]{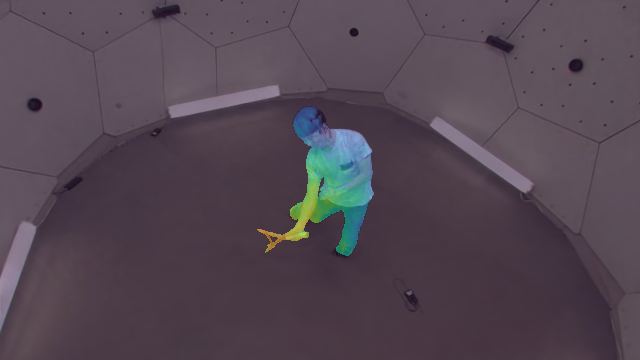}
  \includegraphics[height=1.182cm]{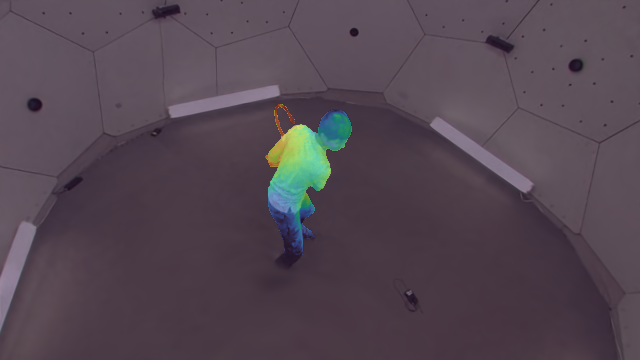}
    \includegraphics[height=1.182cm]{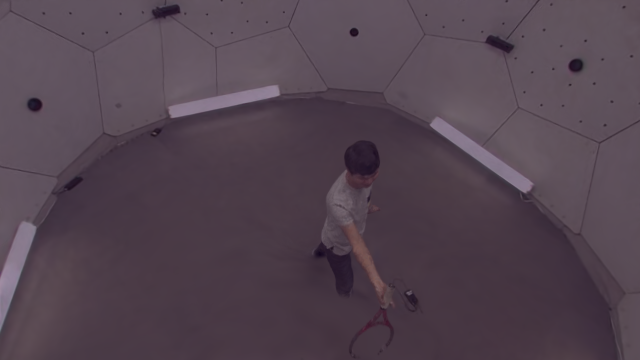}
  }{ }
  \hfill
\jsubfig{\includegraphics[height=1.182cm]{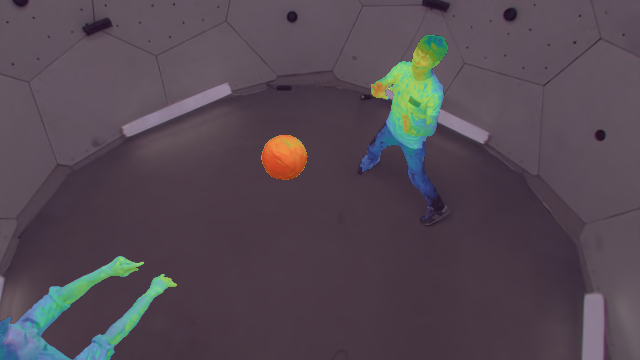} \hfill \includegraphics[height=1.182cm]{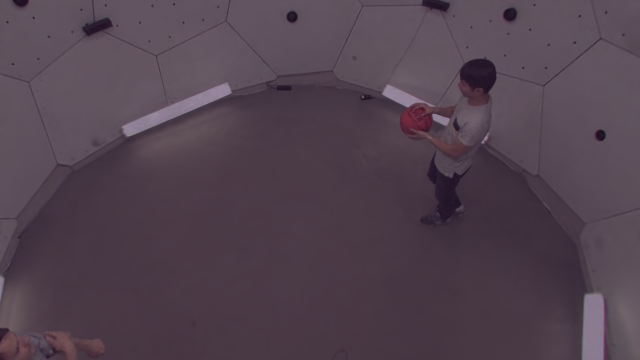}
  \includegraphics[height=1.182cm]{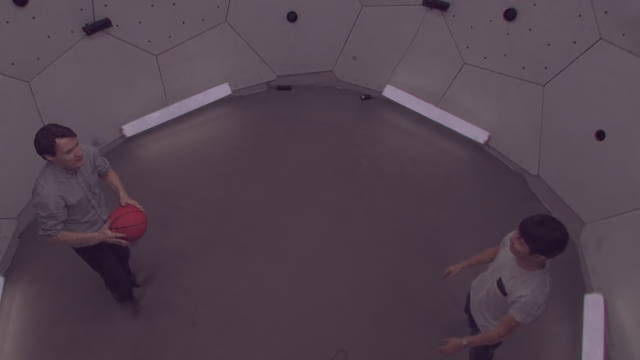}
    \includegraphics[height=1.182cm]{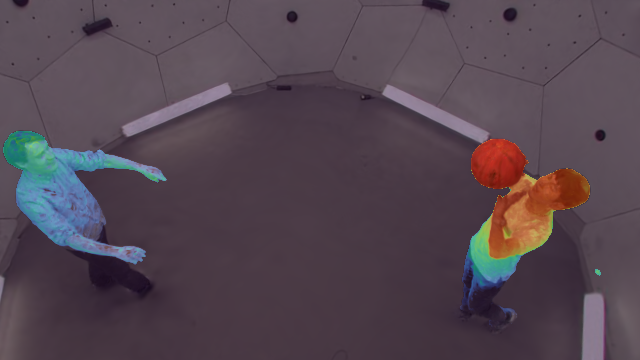}
  }{ } \\
  \rotatebox{90}{\whitetxt{Tubeg}}
  \jsubfig{\includegraphics[height=1.182cm]{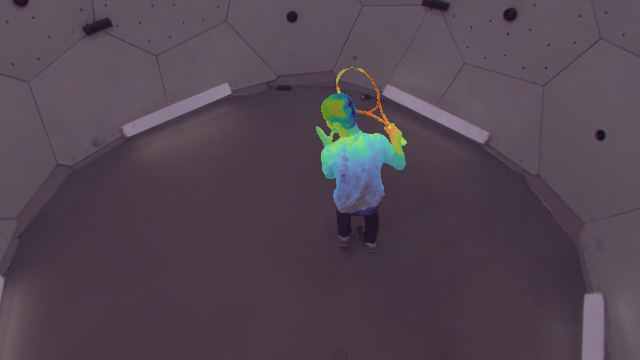} \hfill \includegraphics[height=1.182cm]{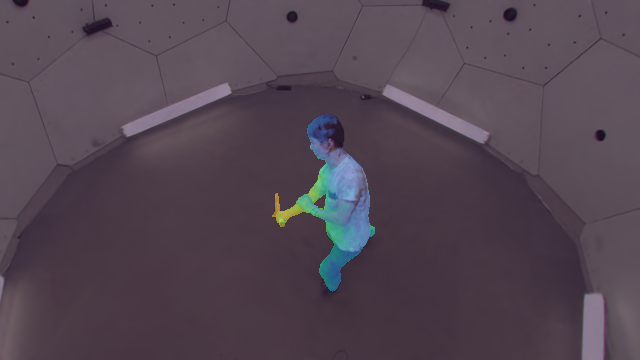}
  \includegraphics[height=1.182cm]{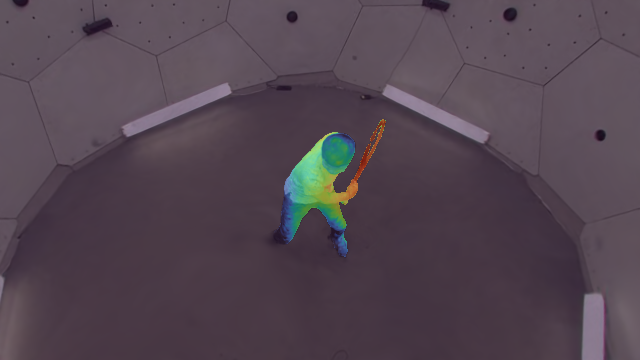}
    \includegraphics[height=1.182cm]{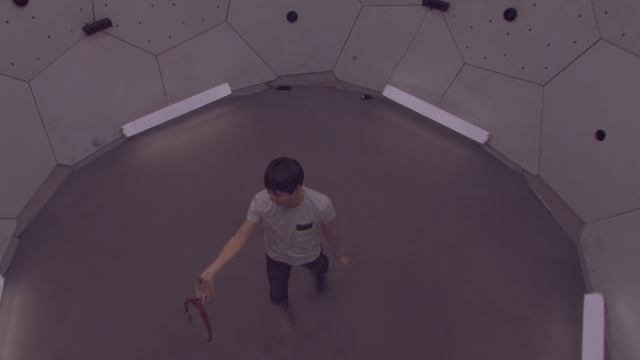}
  }{ }
  \hfill
  \jsubfig{\includegraphics[height=1.182cm]{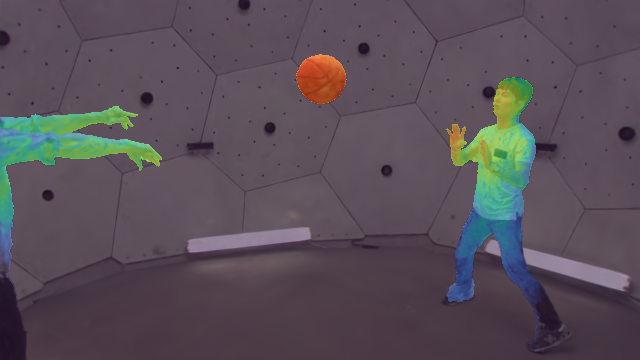} \hfill \includegraphics[height=1.182cm]{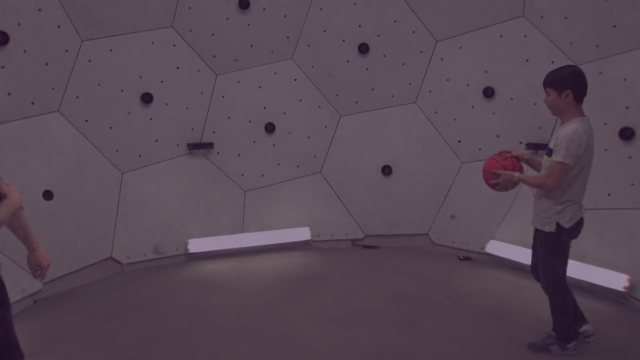}
  \includegraphics[height=1.182cm]{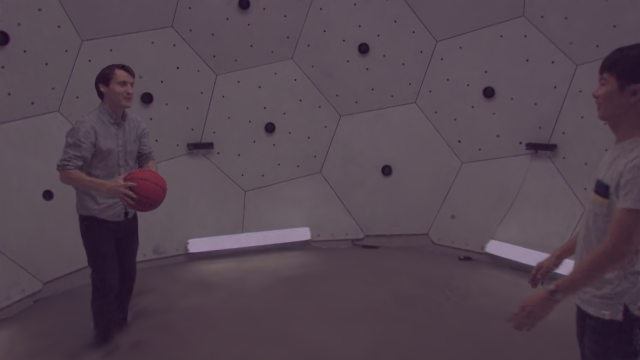}
    \includegraphics[height=1.182cm]{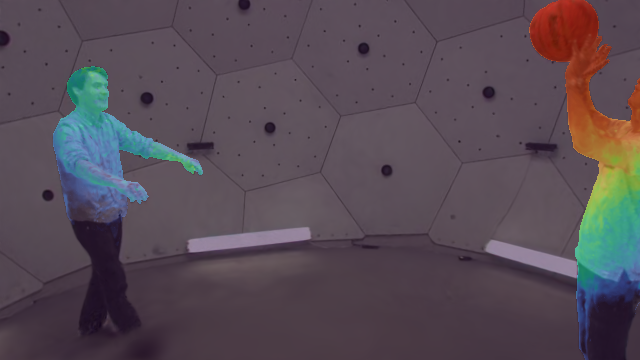}
  }{ } 
        \\
  \rotatebox{90}{\whitetxt{Tubeg}}
  \jsubfig{\includegraphics[height=1.182cm]{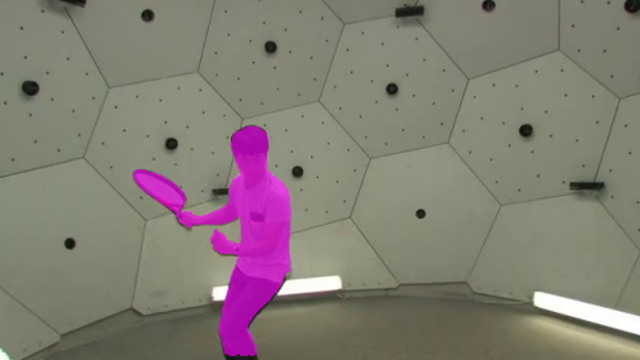} \hfill \includegraphics[height=1.182cm]{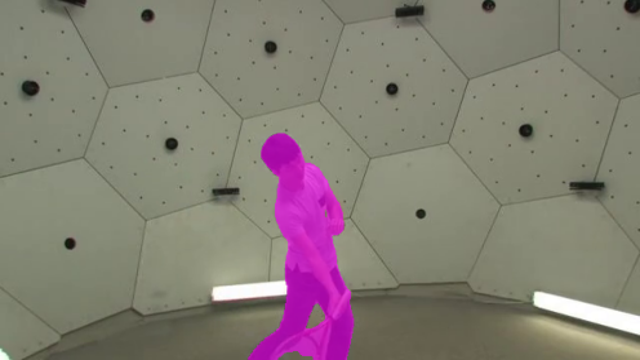}
  \includegraphics[height=1.182cm]{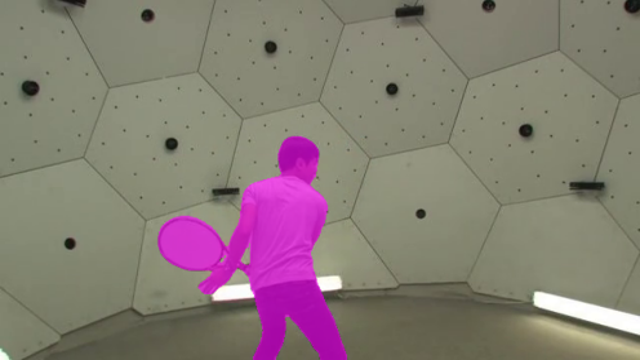}
    \includegraphics[height=1.182cm]{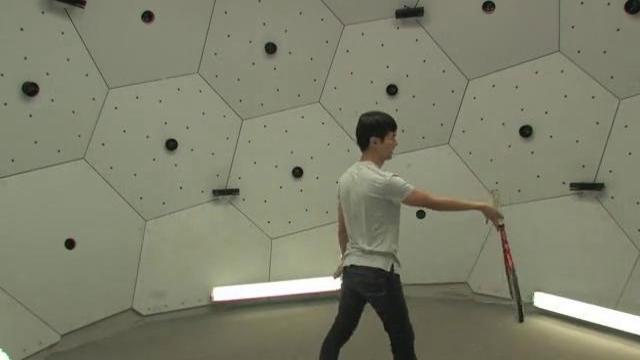}
  }{ }
  \hfill
  \jsubfig{\includegraphics[height=1.182cm]{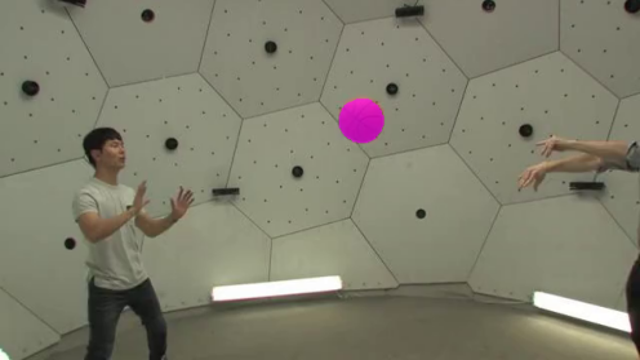} \hfill \includegraphics[height=1.182cm]{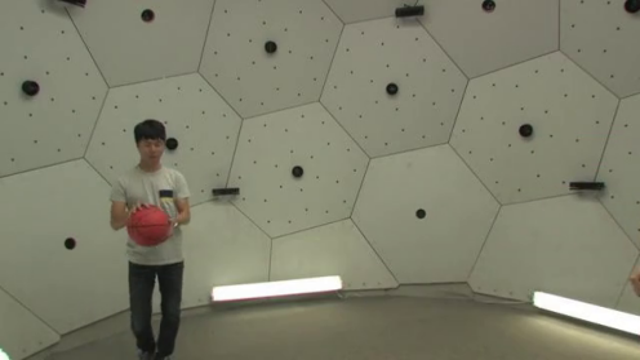}
  \includegraphics[height=1.182cm]{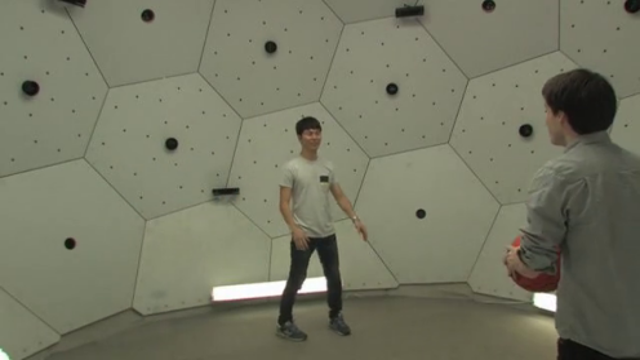}
    \includegraphics[height=1.182cm]{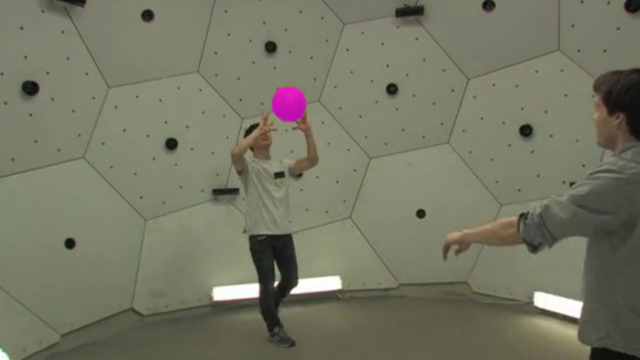}
  }{ } \\
  \rotatebox{90}{\whitetxt{ggg}GT}
    \jsubfig{\includegraphics[height=1.182cm]{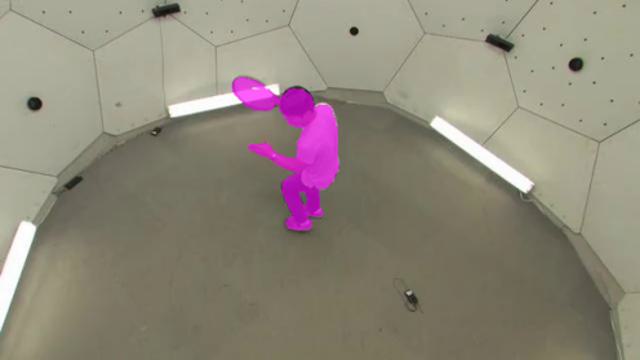} \hfill \includegraphics[height=1.182cm]{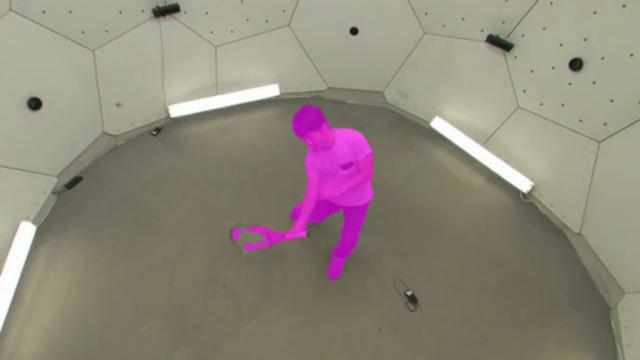}
  \includegraphics[height=1.182cm]{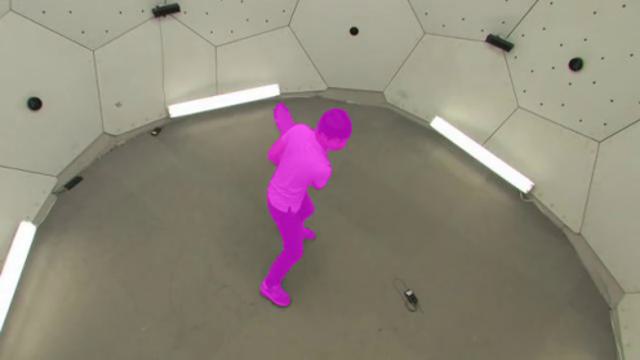}
    \includegraphics[height=1.182cm]{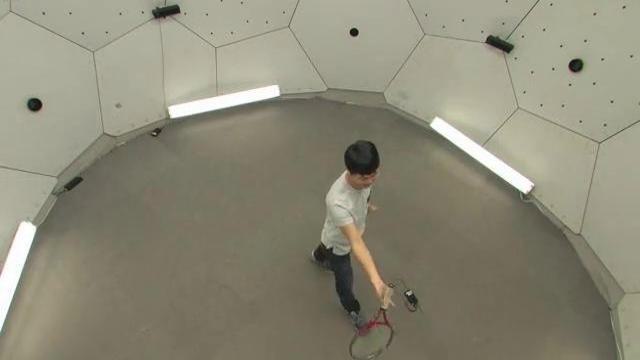}
  }{ }
  \hfill
  \jsubfig{\includegraphics[height=1.182cm]{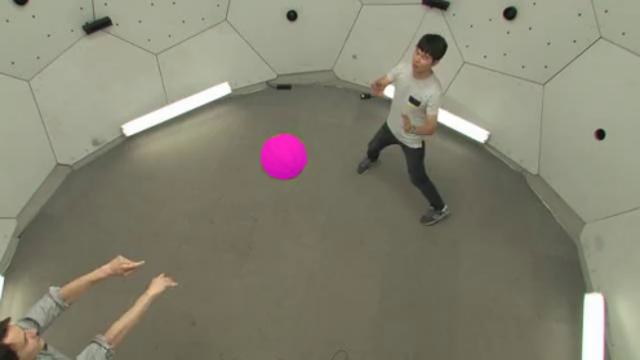} \hfill \includegraphics[height=1.182cm]{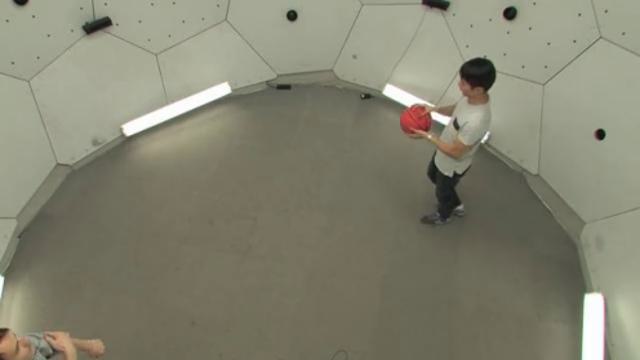}
  \includegraphics[height=1.182cm]{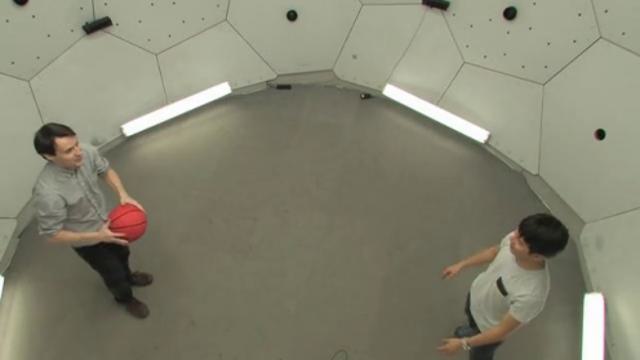}
    \includegraphics[height=1.182cm]{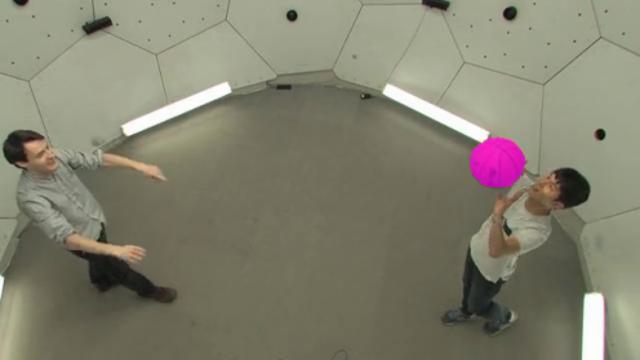}
  }{ } \\
  \rotatebox{90}{\whitetxt{Tubeg}}
  \jsubfig{\includegraphics[height=1.182cm]{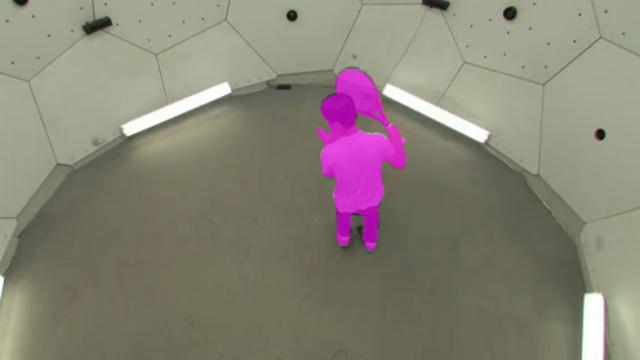} \hfill \includegraphics[height=1.182cm]{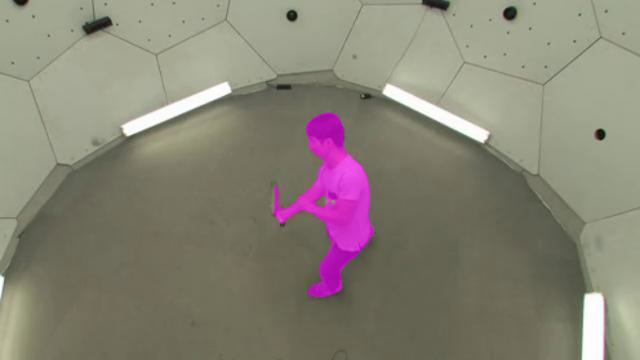}
  \includegraphics[height=1.182cm]{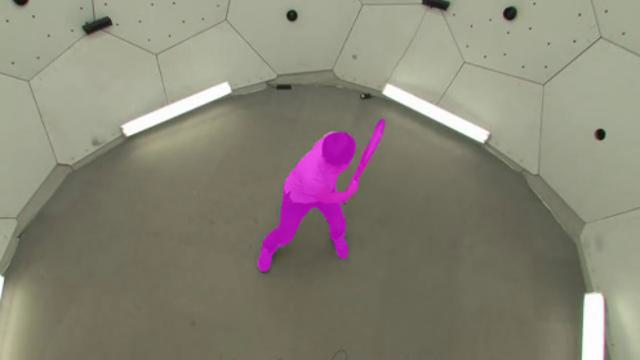}
    \includegraphics[height=1.182cm]{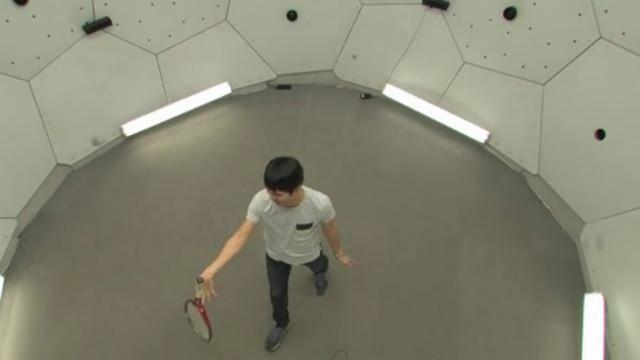}
  }{ Input query: \emph{A person playing tennis }}
  \hfill
  \jsubfig{\includegraphics[height=1.182cm]{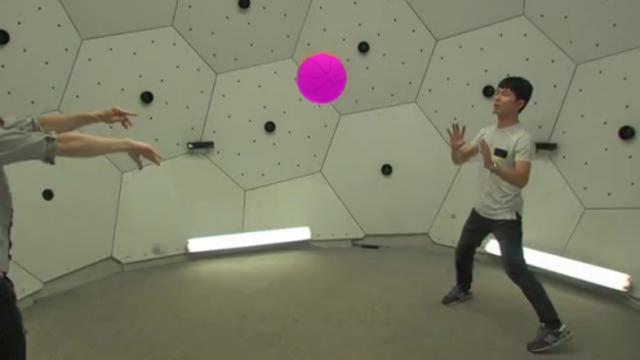} \hfill \includegraphics[height=1.182cm]{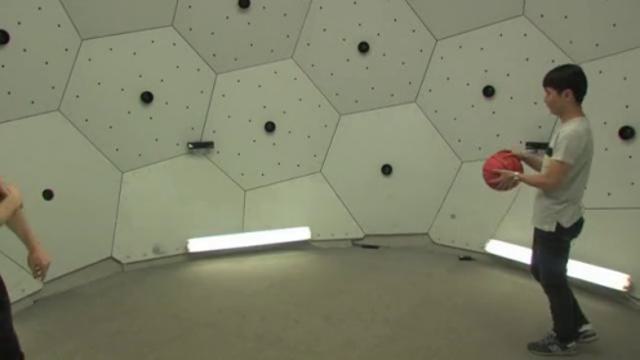}
  \includegraphics[height=1.182cm]{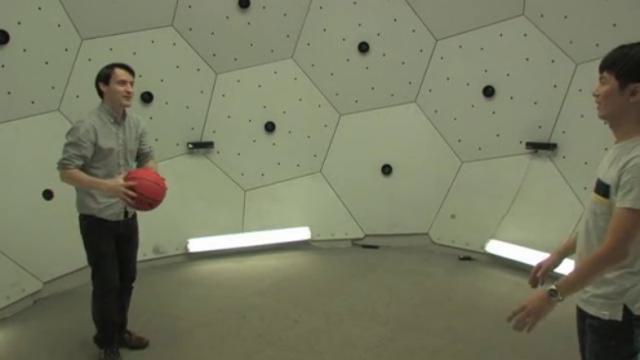}
    \includegraphics[height=1.182cm]{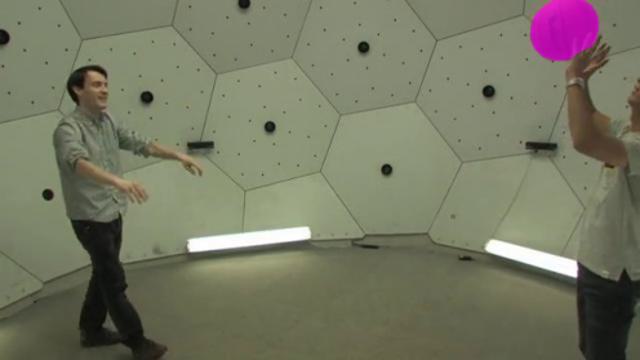}
  }{Input query: \emph{A ball flying in the air } } 
    \vspace{-5pt}
    \caption{
  \textbf{Comparison to 2D spatio-temporal grounding}. We show spatio-temporal localization results for TubeDETR~\cite{yang2022tubedetr} and CGSTVG \cite{gu2024context}, 2D baseline methods, along with our results. The textual queries and ground-truth segmentation maps are taken from our \datasetname{} benchmark. Results are illustrated over three different camera viewpoints (shown on different rows) and four different timestamps (shown on different columns). As illustrated above, these 2D methods cannot generate view consistent results, and often fails to correctly localize the queried text. Our approach, on the other hand, allows for localizing the regions in both space and time.  
  }\label{fig:comparison}
\end{figure*}

A quantitative evaluation on the \datasetname{} benchmark is provided in Table \ref{tab:comparisons}.
As illustrated in the table (and also qualitatively in Figure \ref{fig:comparison}), our method outperforms the baseline 2D methods significantly, yielding better results across all metrics.
The 2D methods suffer from view inconsistency as they infer over a single view, failing to utilize information from other viewpoints. In particular, their results highly vary (high standard deviation value). 
This is further illustrated in Figure \ref{fig:comparison} on the right, where different predictions are extracted for differing views during the same timestep. Our method achieves consistency between different views as it optimizes a volumetric representation in 4D space, thus utilizing data from all views. %

\begin{figure}
  \rotatebox{90}{\hspace{-6pt}LS}
    \jsubfig{\includegraphics[scale=0.11]{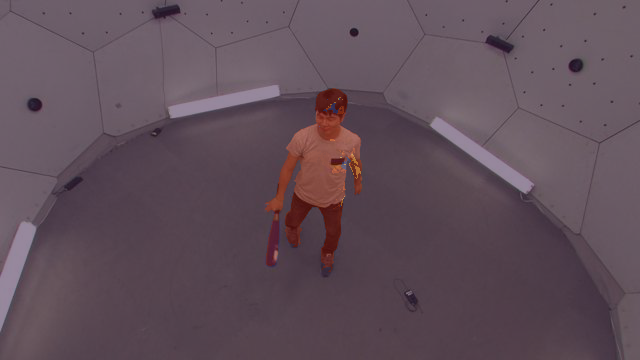}  \includegraphics[scale=0.11]{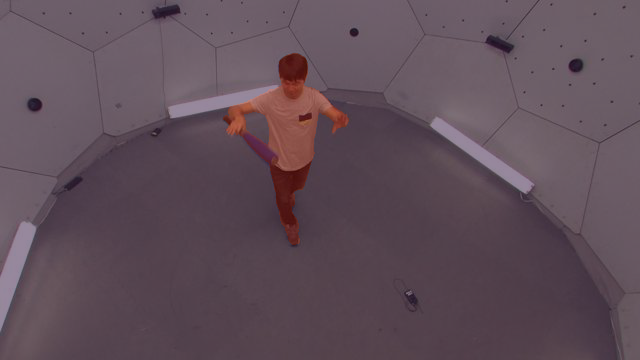}
    \includegraphics[scale=0.11]{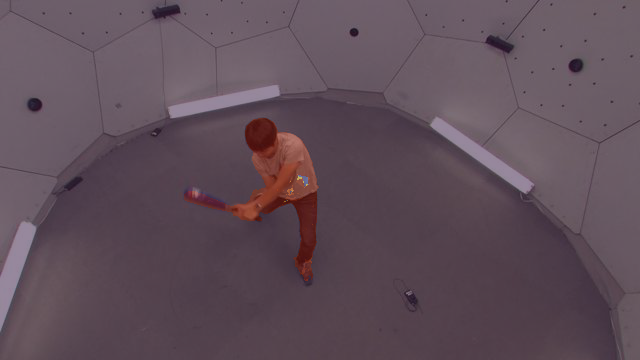}
  }{ }\\
  \rotatebox{90}{\whitetxt{LS}}
    \jsubfig{\includegraphics[scale=0.11]{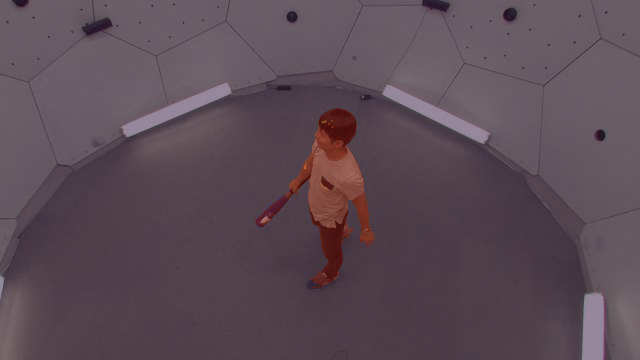}  \includegraphics[scale=0.11]{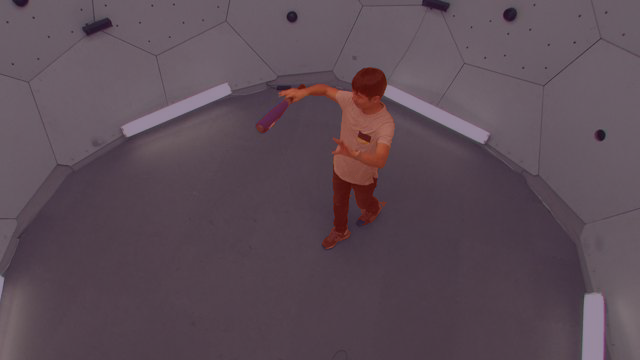}
    \includegraphics[scale=0.11]{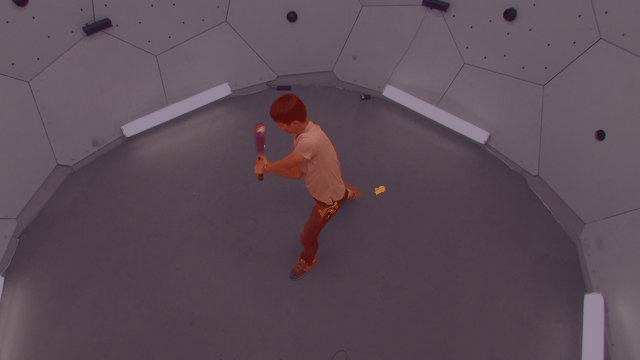}
  }{ }\\
      \rotatebox{90}{\hspace{-8pt}Ours}
    \jsubfig{\includegraphics[scale=0.11]{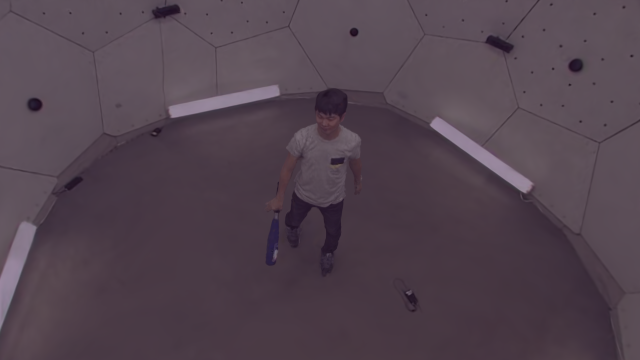}  \includegraphics[scale=0.11]{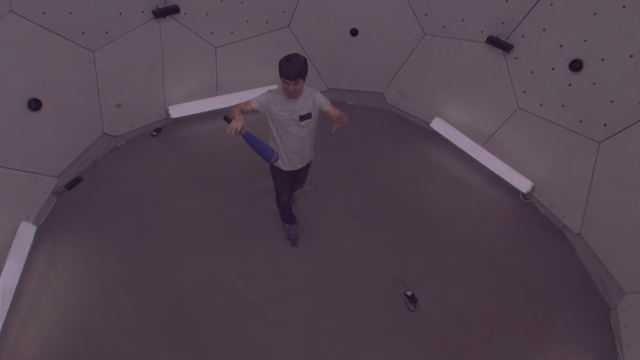}
    \includegraphics[scale=0.11]{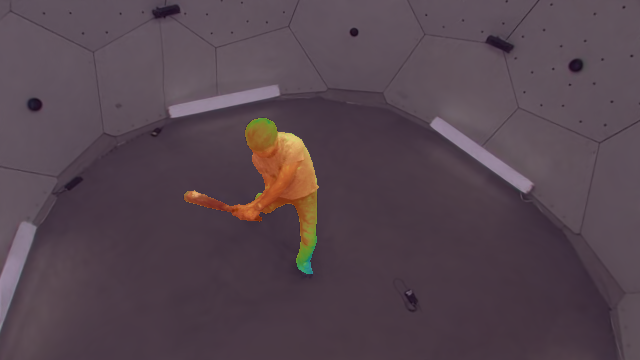}
  }{ }\\
  \rotatebox{90}{\whitetxt{Ours}}
    \jsubfig{\includegraphics[scale=0.11]{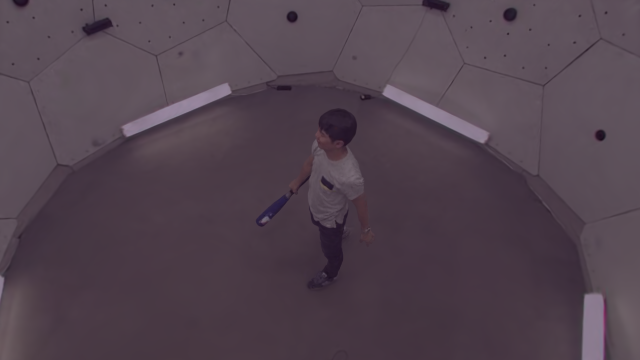}  \includegraphics[scale=0.11]{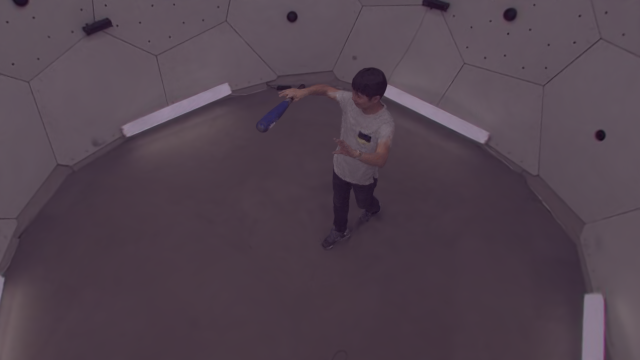}
    \includegraphics[scale=0.11]{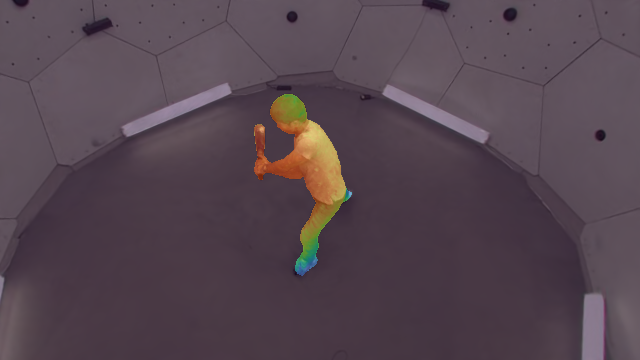}
  }{ }\\
    \rotatebox{90}{\hspace{-6pt}GT}
    \jsubfig{\includegraphics[scale=0.11]{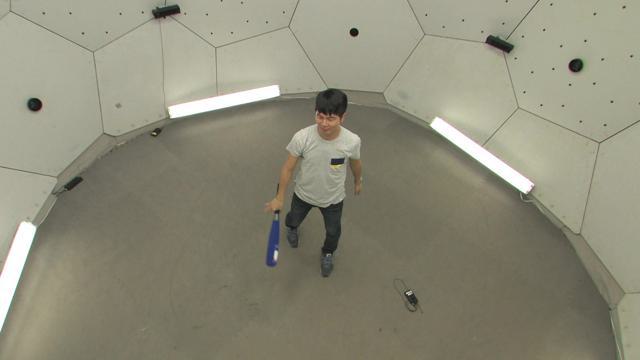}  \includegraphics[scale=0.11]{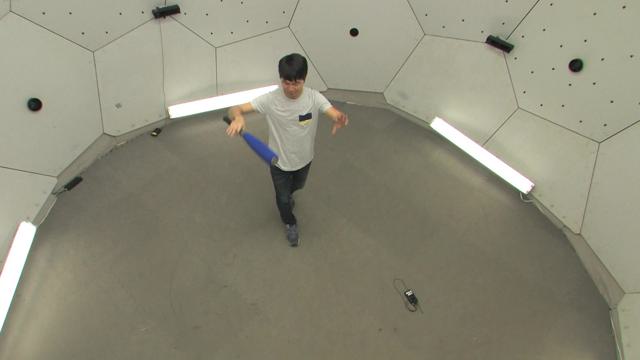}
    \includegraphics[scale=0.11]{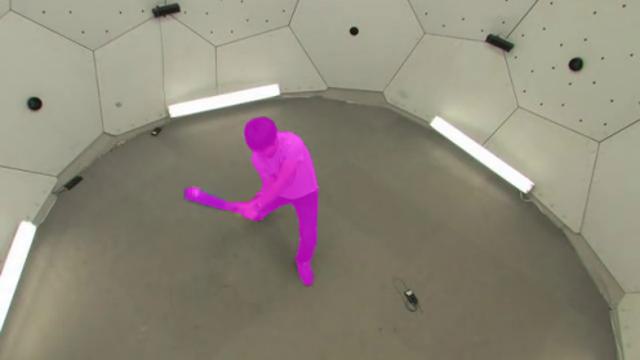}
  }{}\\
   \rotatebox{90}{\whitetxt{LS}}
    \jsubfig{\includegraphics[scale=0.11]{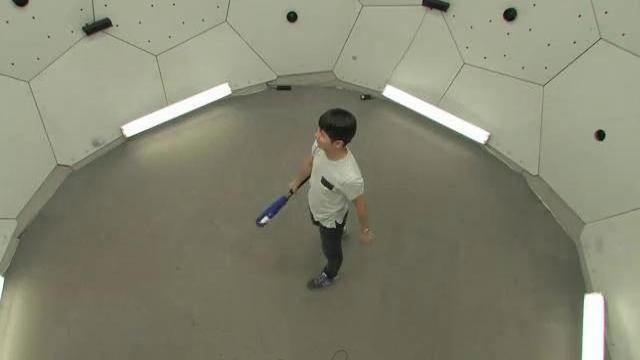} 
    \includegraphics[scale=0.11]{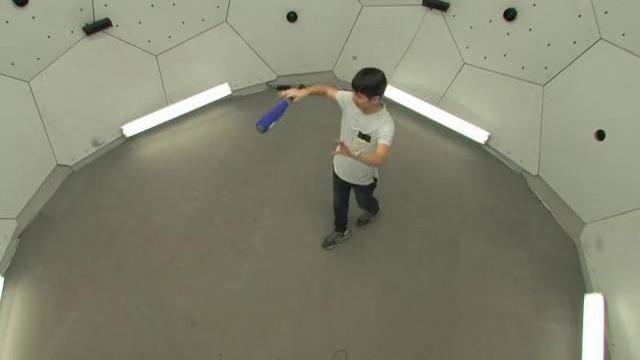}
    \includegraphics[scale=0.11]{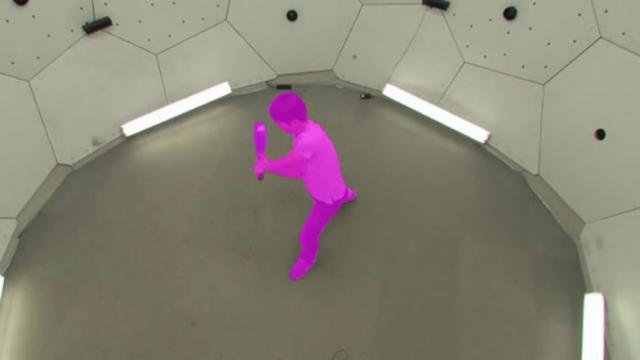}
  }{Input query: \emph{A person swinging the bat} }\\
    \vspace{-5pt}
    \caption{
  We show spatio-temporal localization results for LangSplat (LS) which embeds language features onto representations depicting static 3D environments, along with our results, over three different timesteps (shown on different columns) and two different camera viewpoints (shown in different rows). LS outputs high probabilities for all timesteps, showing difficulty in capturing the temporal segment of the action. As demonstrated above, methods targeting static 3D environments are not intended to operate over such dynamic settings.
  }\label{fig:comparison-langsplat}
\end{figure}

Figure \ref{fig:comparison-langsplat} shows a qualitative evaluation to LangSplat. As illustrated in Table \ref{tab:comparisons} and in the figure, our method outperforms LangSplat, emphasizing the importance of distilling temporal features in addition to spatial features. LangSplat suffers from lack of temporal awareness as it is optimized independently for each timestep. Figure \ref{fig:comparison-langsplat} shows that LangSplat cannot localize the input queries temporally, and that it outputs high probabilities for all timesteps. It also demonstrates that LangSplat yields high predicted probabilities even for input images do not depict the queried text. \methodname{} achieves temporal awareness and proper temporal localization of actions, by optimizing over all timesteps.

This is further observed in our ablations, where we can see that CLIP-based features struggle in localizing actions and instead focus on objects.
Our method, by contrast, is capable of temporally localizing an input query, due to the use of temporal features, as illustrated both numerically and in the figure. In the supplementary, we show results over additional views.

\begin{table}[t]
\centering
\resizebox{1.0\linewidth}{!}{
\begin{tabular}{lcccccc}
\toprule
 Method & $\text{vAP}\uparrow$ &  $\text{vIOU}\uparrow$ & $\text{tIOU}\uparrow$ & $\text{tRec}\uparrow$ & $\text{tPrec}\uparrow$ & $\text{tAP}\uparrow$\\
\midrule
TubeDETR & $21.3\pm3.7$ & $19.9\pm7.3$ & $29.3\pm9.7$ & $44.6\pm16.4$ & $63.7\pm16.3$ & $41.4\pm13.7$ \\
CGSTVG & $25.1\pm8.3$ & $24.3\pm7.7$ & $34.9\pm10.8$ & $56.7\pm17.3$ & $51.6\pm11.3$ & $48.1\pm14.4$ \\
LangSplat & $38.4\pm0.3$$^\star$& $13.9\pm4.1$$^\star$ & --- & --- & --- & --\\
Ours & $\mathbf{58.7\pm0.6}$ & $\mathbf{25.6\pm4.6}$ & $\mathbf{60.8\pm0.0}$ & $\mathbf{83.0\pm0.0}$ & $\mathbf{68.5\pm0.0}$ & $\mathbf{72.6\pm0.0}$\\  
\bottomrule
\end{tabular}
}
\caption{\textbf{Quantitative Evaluation over the \datasetname{} Benchmark}. We report performance over the spatio-temporal vAP and vIOU metrics, and over four additional metrics quantifying the quality of the temporal localization (tIOU,tRec, tPrec, tAP). %
$^\star$Note that LangSplat metrics are calculated over a small number of timesteps, as further detailed in Section \ref{sec:baselines}.}

\label{tab:comparisons}
\end{table}

\begin{figure*}
\jsubfig{\fcolorbox{cyan}{cyan}{\includegraphics[height=1.5cm]{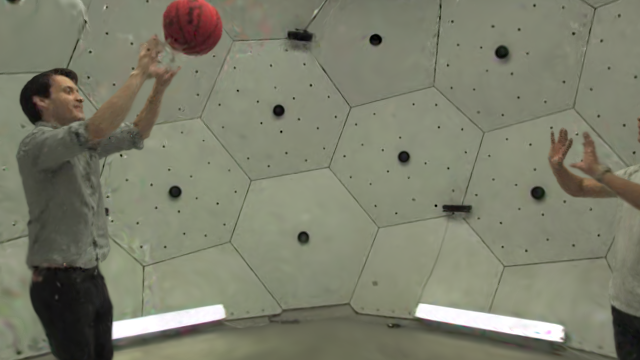}}}{}
\jsubfig{\fcolorbox{orange}{orange}{\includegraphics[height=1.5cm]{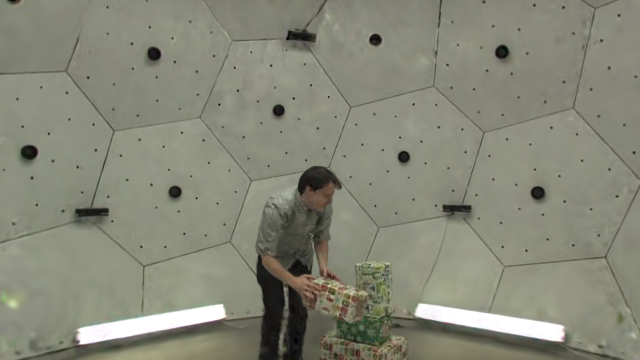}}}{} 
\jsubfig{{\includegraphics[height=1.5cm]{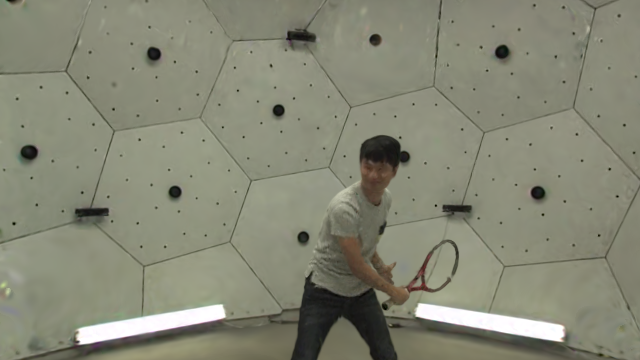}}}{} 
\jsubfig{{\includegraphics[height=1.5cm]{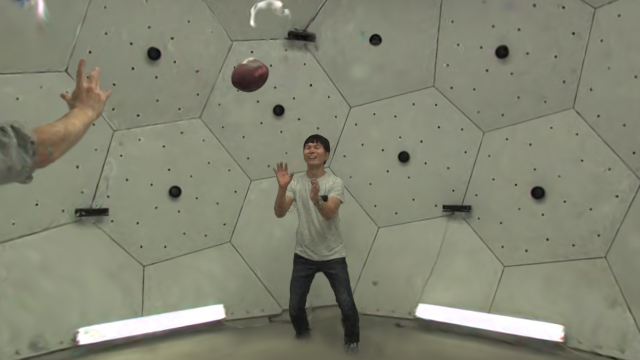}}}{} 
\jsubfig{\fcolorbox{teal}{teal}{\includegraphics[height=1.5cm]{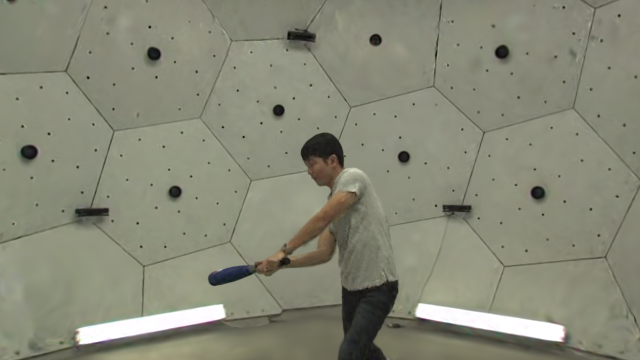}}}{} 
\jsubfig{{\includegraphics[height=1.5cm]{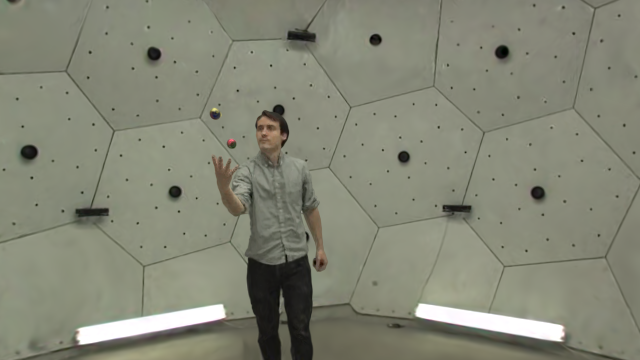}}}{} 
\\ 
  {\begin{flushleft} \leftskip=0.1pt   
  \textbf{Input queries}:
  \archway{
   \emph{A person throwing the basketball}
  },
  \facade{
   \emph{A person picking up the box}
  },
  \sundial{
  \emph{A person swinging the softball bat}
}
  \end{flushleft}} %
  \caption{
  \textbf{Spatio-temporal localization given a set of 4-LEGS}, representing multiple dynamic 3D environments. We query all six scenes from the Panoptic Sports dataset, each illustrated by a single frame above, over different queries shown in unique colors. As illustrated by the matching colors, our approach allows for retrieving the correct dynamic environment.
  }\label{fig:scene-selection}
\end{figure*}

\begin{figure*} %
\centering
\rotatebox{90}{\whitetxt{.}$\text{2D Features}$\whitetxt{,}}
\jsubfig{\includegraphics[scale=0.12]{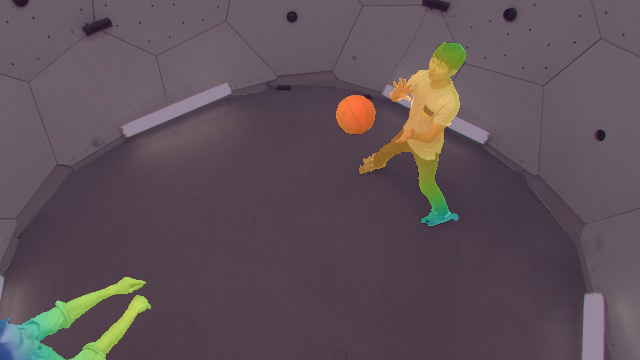}
\includegraphics[scale=0.12]{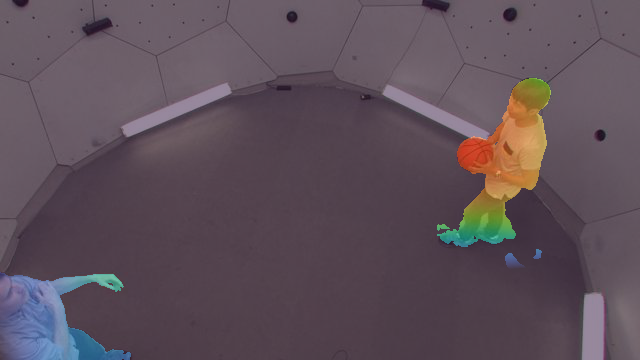}
\includegraphics[scale=0.12]{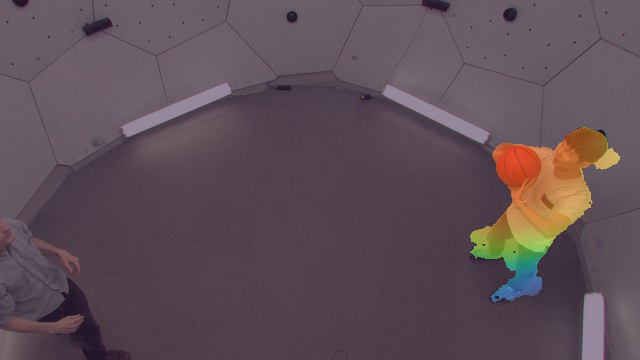}
\includegraphics[scale=0.12]{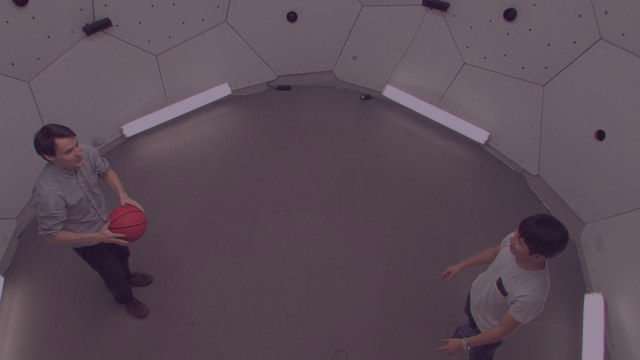}
\includegraphics[scale=0.12]{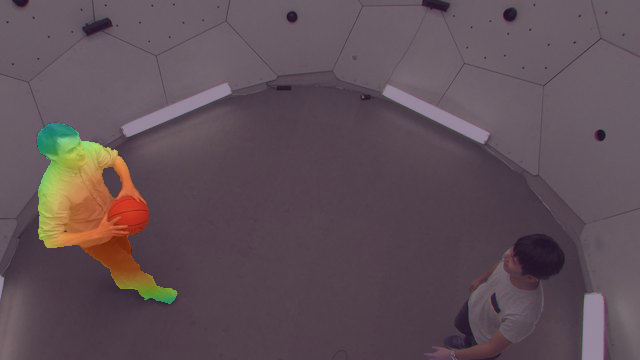}
\includegraphics[scale=0.12]{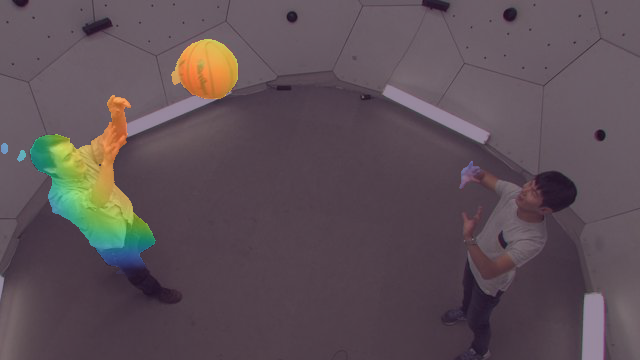}}{}%
\\
\rotatebox{90}{\whitetxt{p}$\text{Static}_{\text{CLIP}}$}
\jsubfig{\includegraphics[scale=0.12]{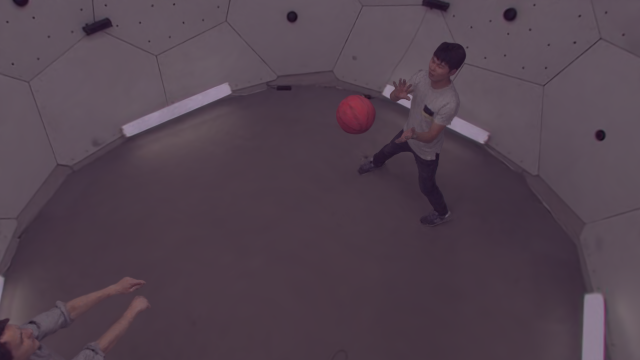}
\includegraphics[scale=0.12]{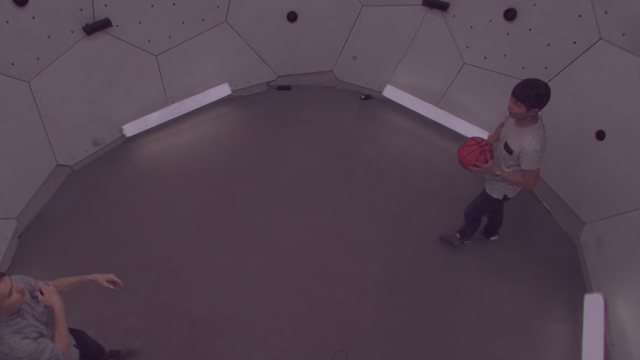}
\includegraphics[scale=0.12]{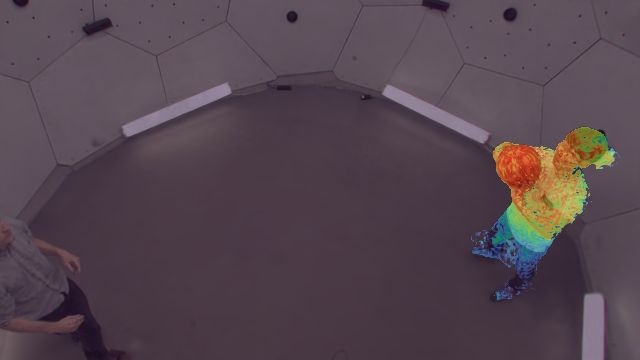}
\includegraphics[scale=0.12]{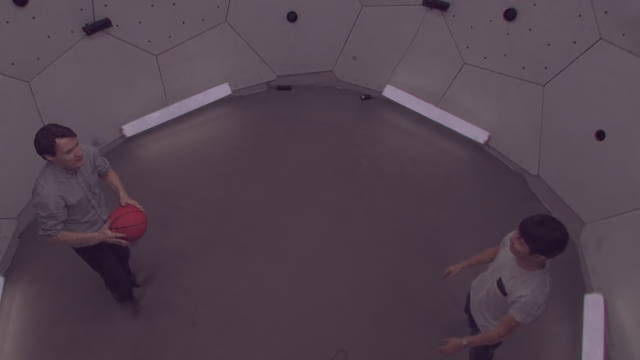}
\includegraphics[scale=0.12]{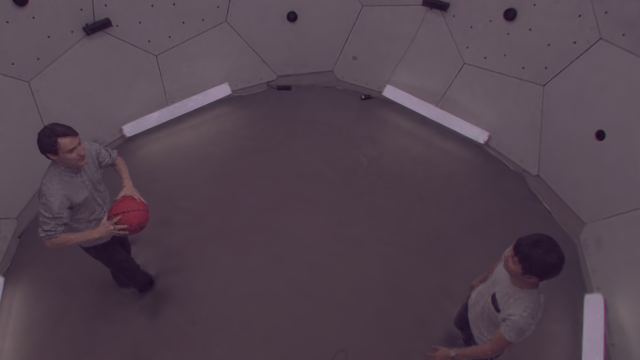}
\includegraphics[scale=0.12]{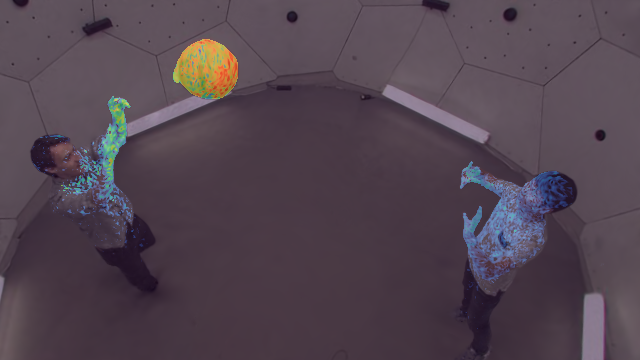}}{}%
\\
\rotatebox{90}{\whitetxt{p}$\text{AVG}_{\text{CLIP}}$}
\jsubfig{\includegraphics[scale=0.12]{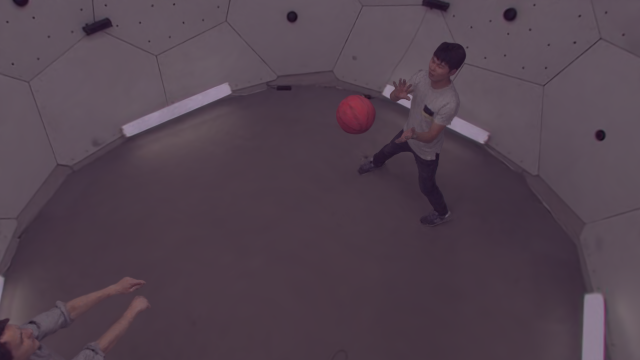}}{}
\jsubfig{{\includegraphics[scale=0.12]{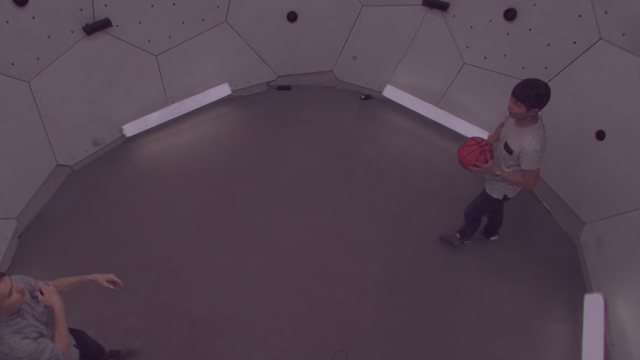}}}{}
\jsubfig{{\includegraphics[scale=0.12]{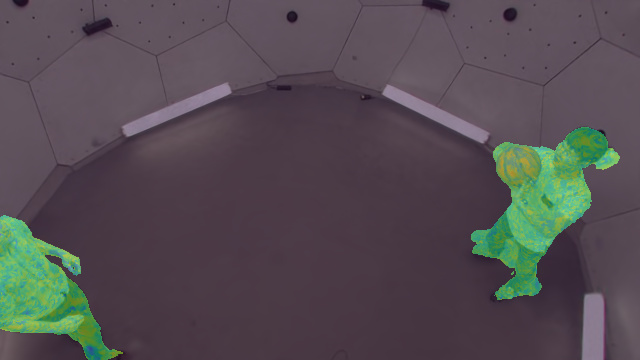}}}{}
\jsubfig{\includegraphics[scale=0.12]{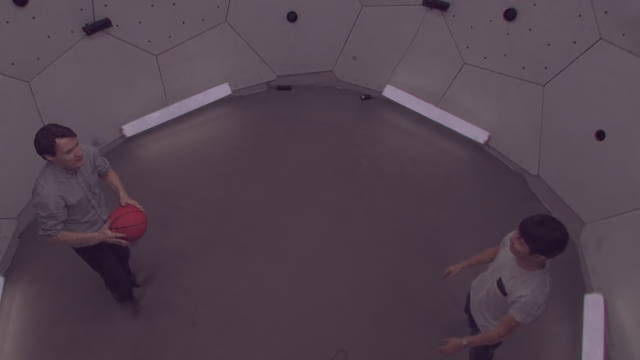}}{}
\jsubfig{{\includegraphics[scale=0.12]{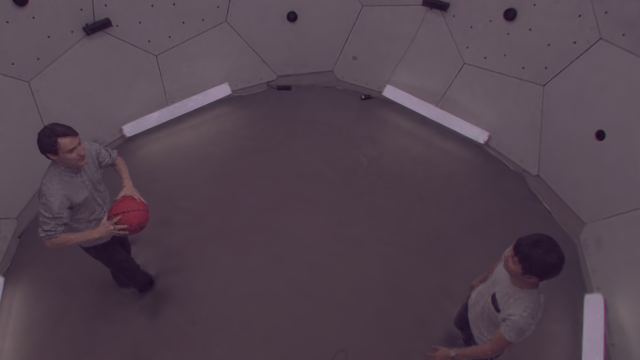}}}{}
\jsubfig{\includegraphics[scale=0.12]{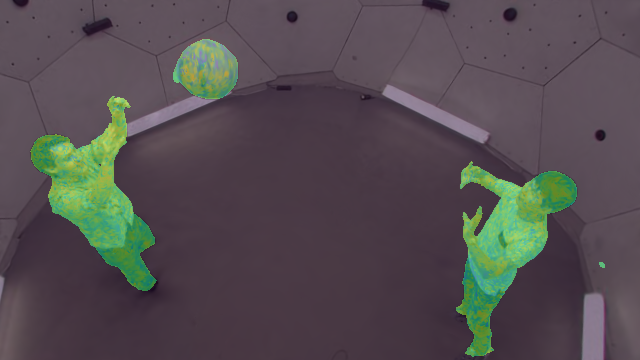}}{}
\\
\rotatebox{90}{\whitetxt{5pt}$\text{Ours}$}
\jsubfig{\includegraphics[scale=0.12]{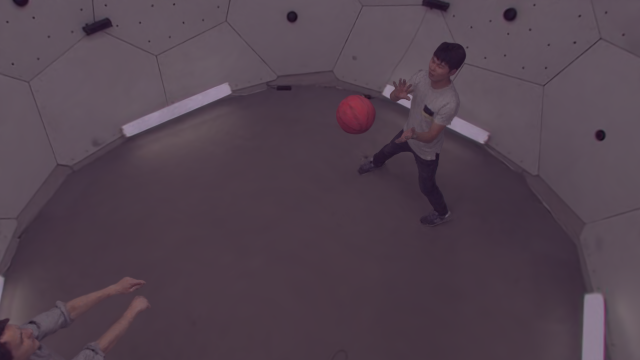}
\includegraphics[scale=0.12]{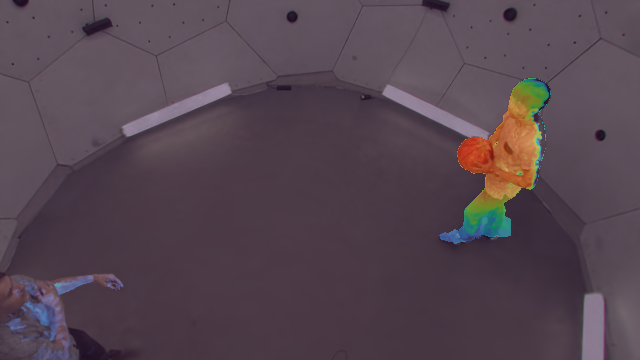}
\includegraphics[scale=0.12]{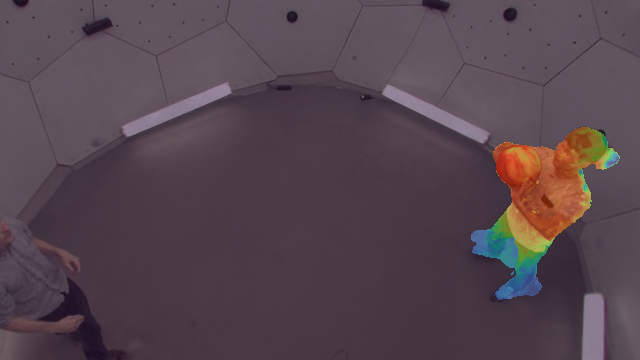}
\includegraphics[scale=0.12]{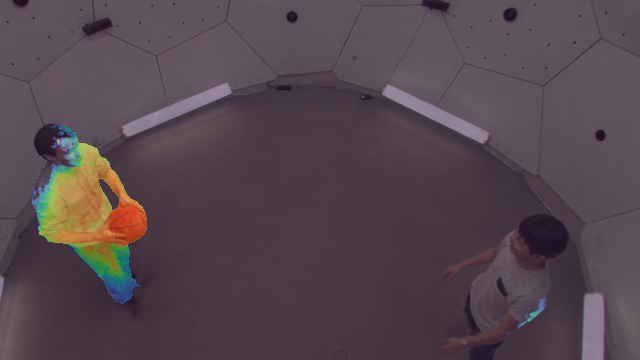}
\includegraphics[scale=0.12]{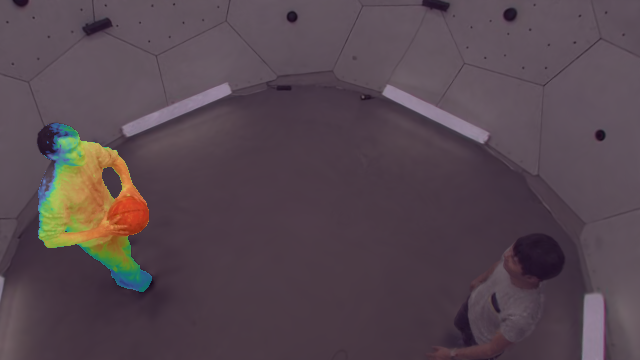}
\includegraphics[scale=0.12]{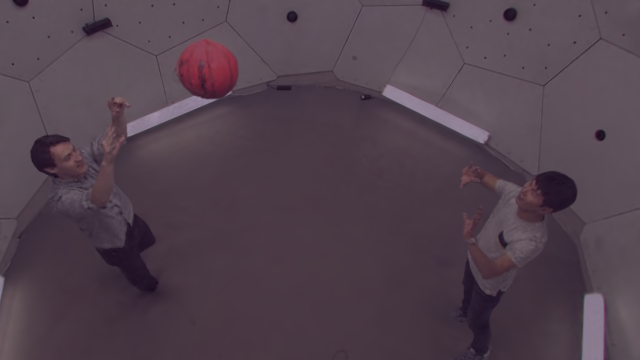}}{}%
\\
\rotatebox{90}{\whitetxt{10pt}$\text{GT}$}
\jsubfig{\includegraphics[scale=0.12]{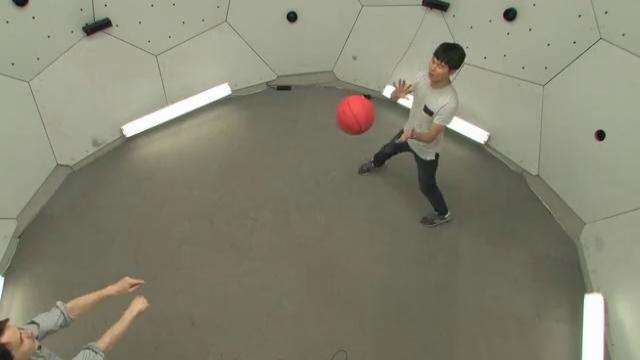}}{}
\jsubfig{\includegraphics[scale=0.12]{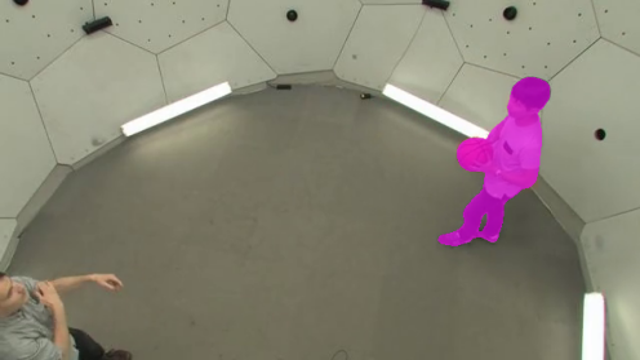}}{}
\jsubfig{{\includegraphics[scale=0.12]{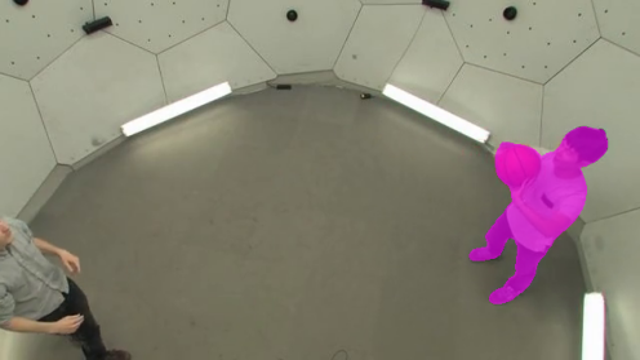}}}{}
\jsubfig{{\includegraphics[scale=0.12]{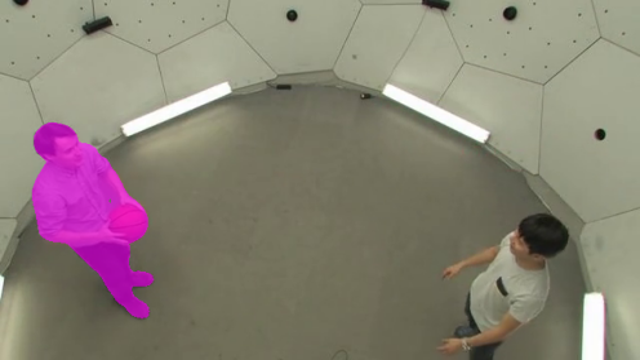}}}{}
\jsubfig{{\includegraphics[scale=0.12]{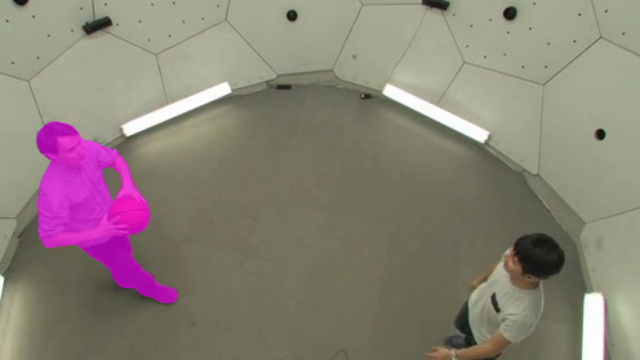}}}{}
\jsubfig{\includegraphics[scale=0.12]{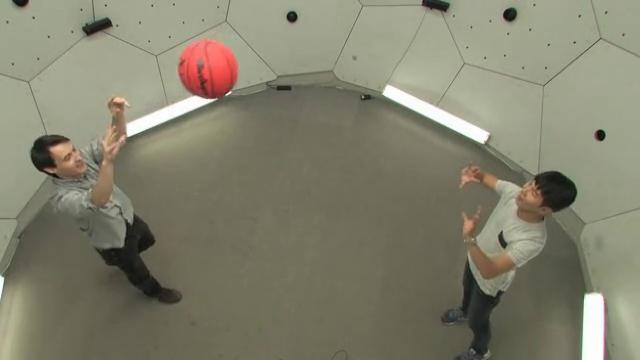}}{ }\\%\hspace{0.001pt}
{}
\caption{\textbf{Qualitative ablation results} for the input query \emph{A person holding the ball}. We ablate the use of a volumetric representation (2D features), the use of the ViCLIP video encoder ($\text{Static}_{\text{CLIP}}$, $\text{AVG}_{\text{CLIP}}$). As illustrated above, our approach outperforms these ablations -- both spatially and temporally, yielding more accurate results as our extracted features can better capture temporal different, and also because we learn 3D-consistent representations. } 
\label{fig:ablations_temporal}
\end{figure*}

\subsection{Ablations} 
\label{sec:ablations}

We ablate our use of the ViCLIP \emph{video} encoder, demonstrating the importance of extracting features with a video encoder, our use of a \emph{volumetric} representation, emphasizing the benefits of lifting features from 2D. When ablating our use of a video encoder we compare against two different methods for extracting pixel-aligned 2D spatio-temporal features: (i) static CLIP features (denoted as $\text{Static}_{\text{CLIP}}$), which uses a CLIP image encoder and does not look at a temporal tube centered at the current frame, and (ii) averaged CLIP features (denoted as $\text{Avg}_{\text{CLIP}}$), which averages the extracted CLIP image encoder features in a temporal window centered at our current frame. When ablating the use of a volumetric representation we compare against  
the extracted ViCLIP features from the input 2D videos (denoted as 2D Features). 
The CLIP-based methods suffer from poor temporal localization since they distill only static features.
This is illustrated in Figure \ref{fig:ablations_temporal}, 
{as these} methods struggle to localize the predictions temporally, even when using averaged features. In contrast \methodname{} distills temporally-aware features thus enabling correct temporal localization. This is further illustrated quantitatively 
in Table \ref{tab:ablations}, as our method achieves significantly higher scores across all metrics. The 2D Features baseline (the extracted ViCLIP features from the input 2D videos) is also outperformed by our method. Additionally, results highly vary across different viewpoints. This is in contrast to consistency achieved by our method, due to the use a volumetric representation.

Additionally, we ablate the use of self-attention and its effectiveness in enforcing local smoothness in 3D space between the disttilled features. We compare our use of self-attention against a 4D language feature field distilled without attention (denoted as w/o Attn) and its effectiveness in enforcing smoothness within a local spatial neighborhood with three methods of local smoothing, a neighborhood blur (denoted as $\text{Local}_{\text{Blur}}$), a neighborhood smoothness regularization loss (denoted as $\text{Local}_{\text{Reg}}$) and a MLP for the local neighborhood (denoted as $\text{Local}_{\text{MLP}}$. As illustrated in Table \ref{tab:ablations-smooth}, the addition of self-attention improves performance and it's ability to model local relationships is outperforms all other local smoothness operators. We provide additional details and a qualitative comparison in the supplementary material.

In the supplementary material, we conduct additional experiments, ablating our feature extraction aggregation
, our temporal localization score weighting method and the querying space usef for open-vocabulary querying. These ablations demonstrate that our averaging mechanism outperforms other aggregation techniques %
, that weighting the temporal localization scores by cardinality outperforms weighting by size and that querying in the feature space outperforms latent space querying.

\subsection{Limitations} 
\label{sec:limitations}

\begin{figure*} %
\centering
\jsubfig{\includegraphics[scale=0.12]{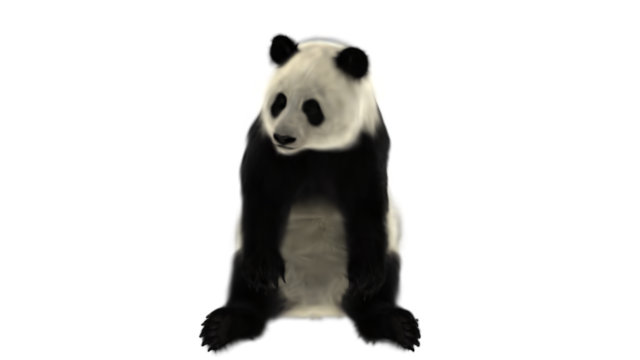}}{}
\jsubfig{\fcolorbox{white}{white}{\includegraphics[scale=0.12]{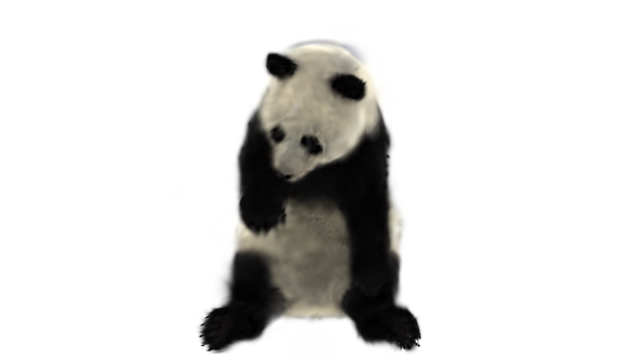}}}{}
\jsubfig{\fcolorbox{white}{white}{\includegraphics[scale=0.12]{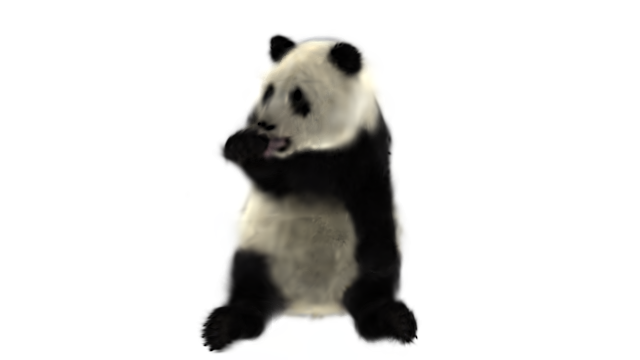}}}{}
\jsubfig{\includegraphics[scale=0.12]{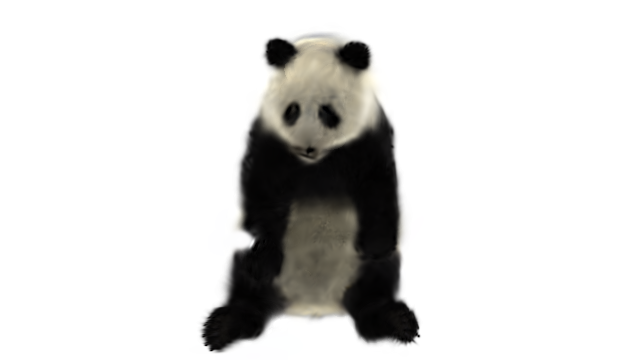}}{}
\jsubfig{\fcolorbox{white}{white}{\includegraphics[scale=0.12]{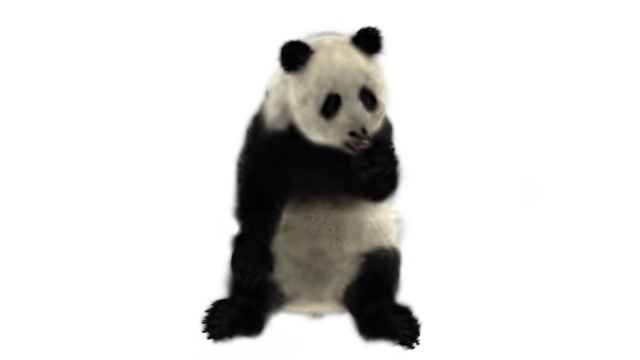}}}{}
\jsubfig{\includegraphics[scale=0.12]{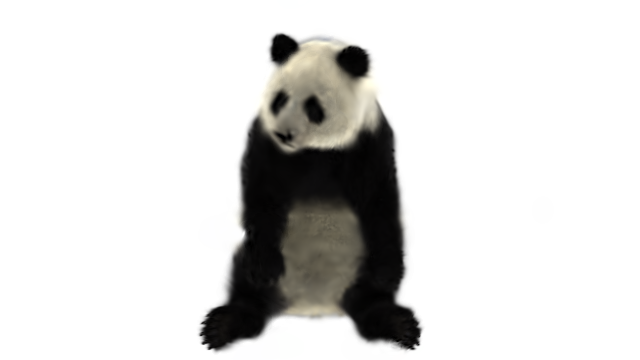}}{}%
\\
\jsubfig{\includegraphics[scale=0.12]{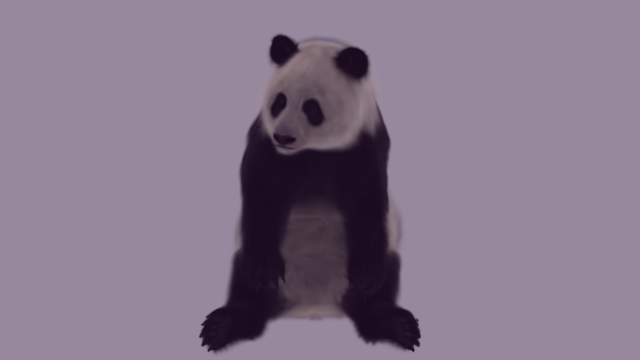}}{}
\jsubfig{\fcolorbox{red}{red}{\includegraphics[scale=0.12]{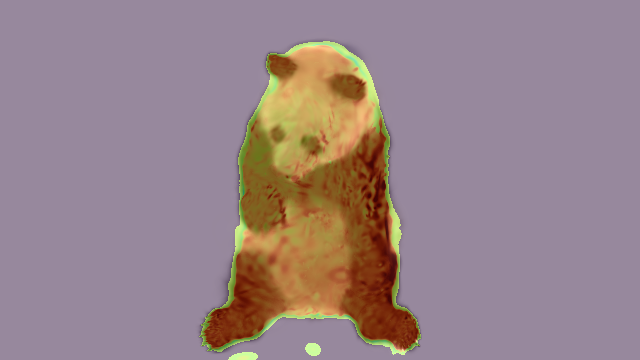}}}{}
\jsubfig{\fcolorbox{red}{red}{\includegraphics[scale=0.12]{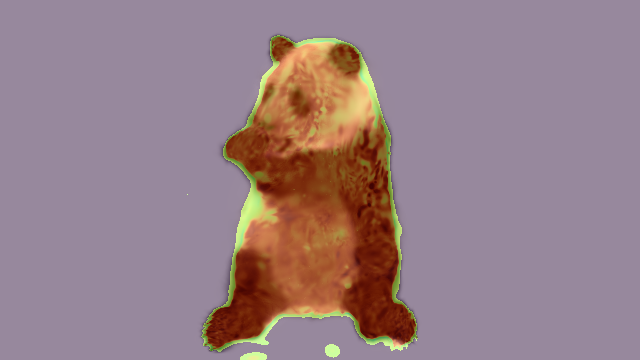}}}{}
\jsubfig{\includegraphics[scale=0.12]{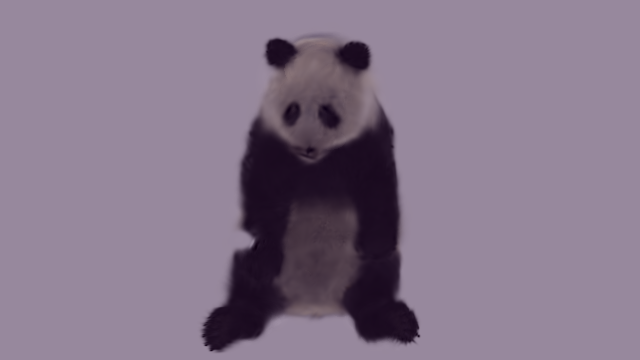}}{}
\jsubfig{\fcolorbox{red}{red}{\includegraphics[scale=0.12]{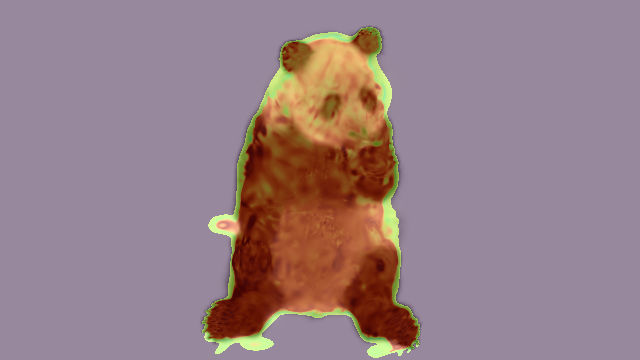}}}{}
\jsubfig{\includegraphics[scale=0.12]{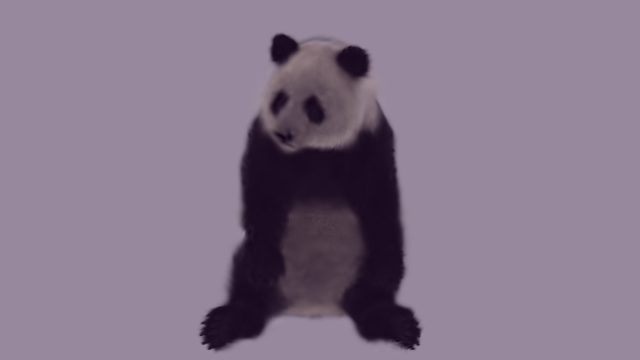}}{}
\\
\vspace{4pt}
{ Input query: \emph{A panda licking its paws}}
\vspace{4pt}
\\
\jsubfig{\includegraphics[scale=0.12]{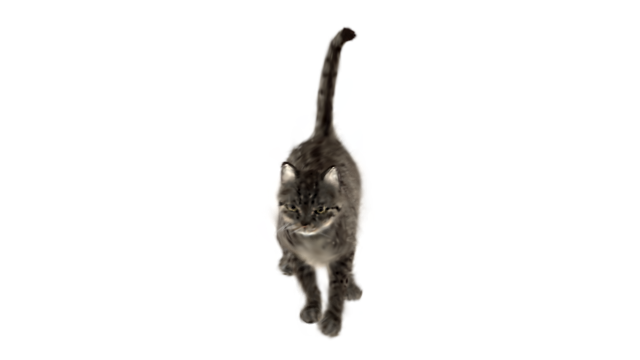}}{}
\jsubfig{{\includegraphics[scale=0.12]{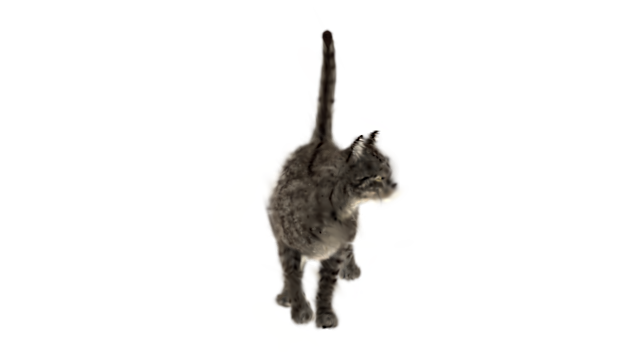}}}{}
\jsubfig{\fcolorbox{white}{white}{\includegraphics[scale=0.12]{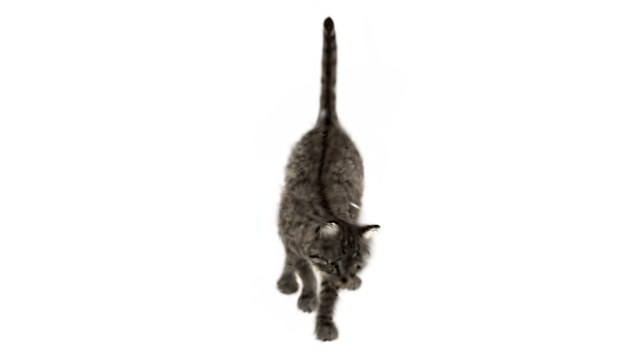}}}{}
\jsubfig{\fcolorbox{white}{white}{\includegraphics[scale=0.12]{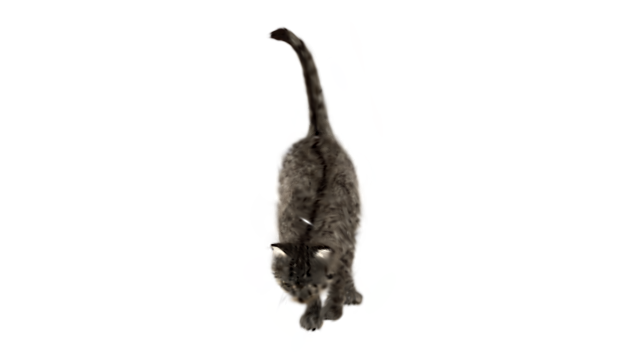}}}{}
\jsubfig{\fcolorbox{white}{white}{\includegraphics[scale=0.12]{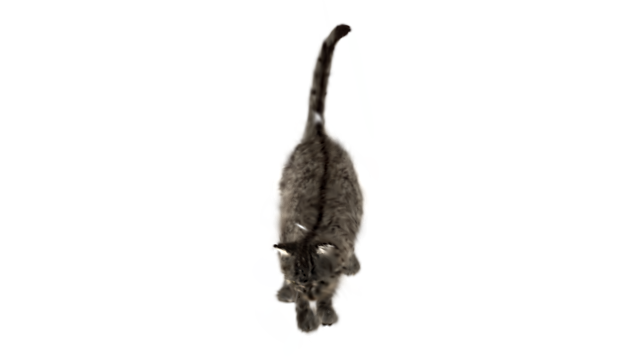}}}{}
\jsubfig{\includegraphics[scale=0.12]{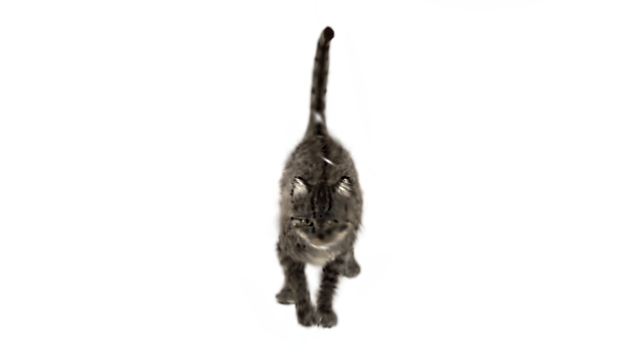}}{}%
\\
\jsubfig{\includegraphics[scale=0.12]{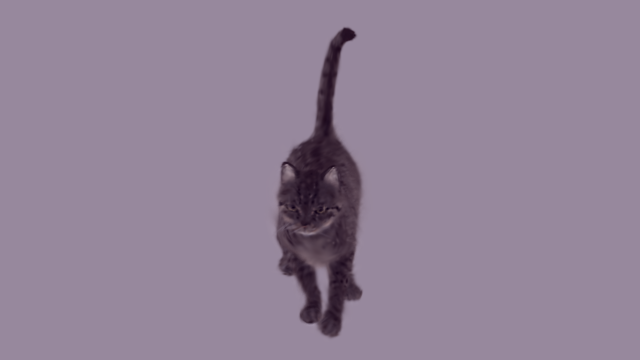}}{}
\jsubfig{\includegraphics[scale=0.12]{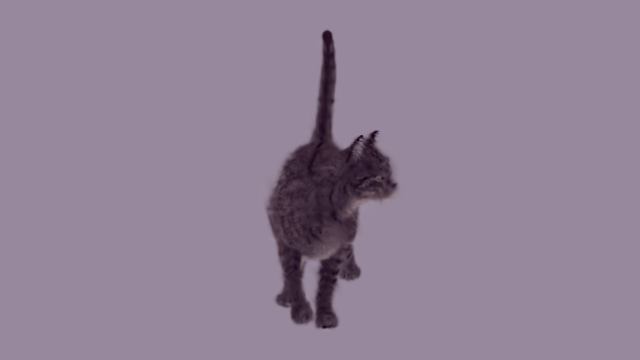}}{}
\jsubfig{\fcolorbox{red}{red}{\includegraphics[scale=0.12]{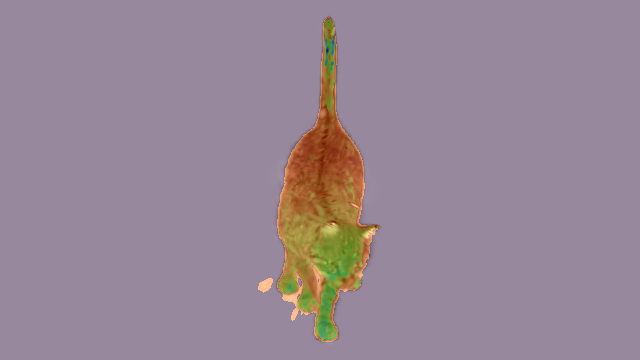}}}{}
\jsubfig{\fcolorbox{red}{red}{\includegraphics[scale=0.12]{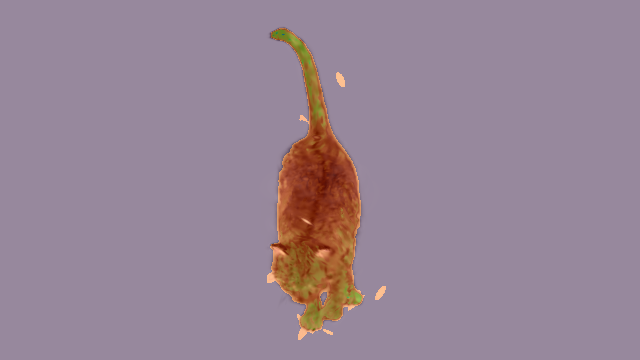}}}{}
\jsubfig{\fcolorbox{red}{red}{\includegraphics[scale=0.12]{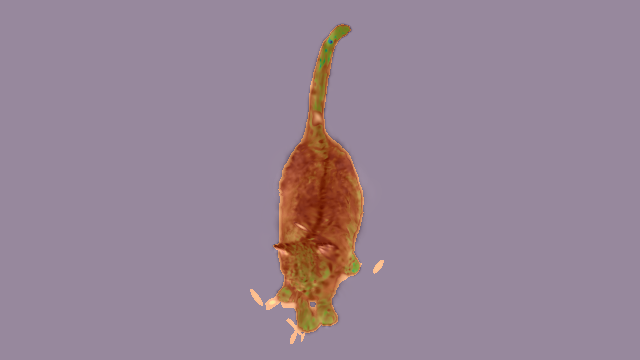}}}{}
\jsubfig{{\includegraphics[scale=0.12]{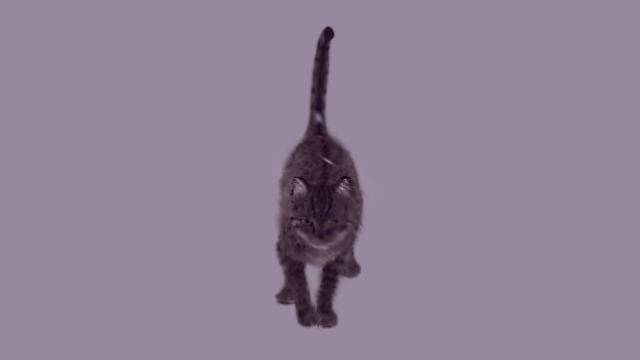}}}{}
\\
\vspace{4pt}
{ Input query: \emph{A cat sniffing the ground}}
\vspace{4pt}
\\
\jsubfig{\includegraphics[scale=0.12]{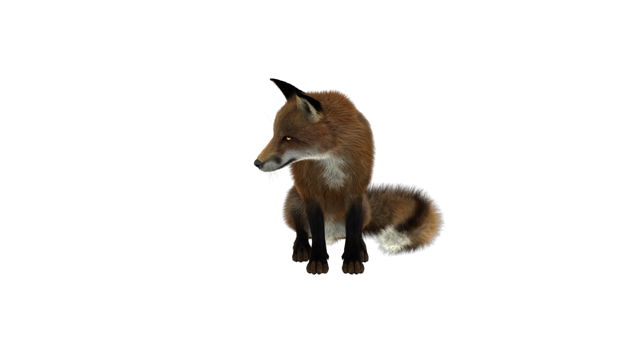}}{}
\jsubfig{{\includegraphics[scale=0.12]{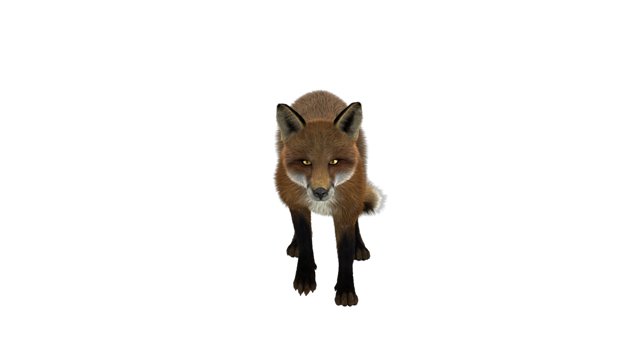}}}{}
\jsubfig{\fcolorbox{white}{white}{\includegraphics[scale=0.12]{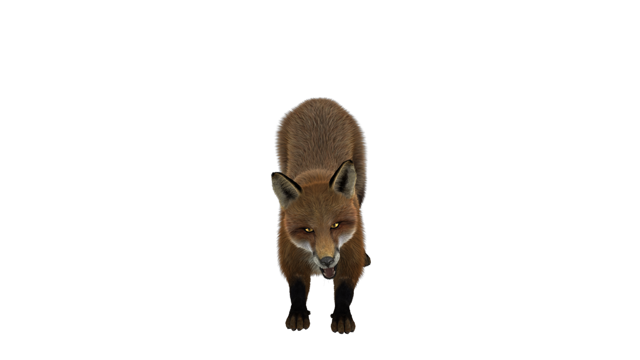}}}{}
\jsubfig{\fcolorbox{white}{white}{\includegraphics[scale=0.12]{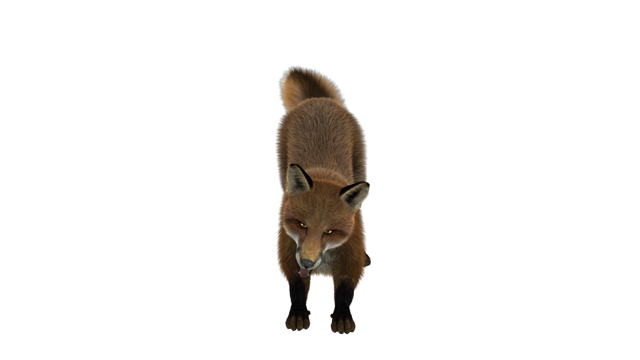}}}{}
\jsubfig{\fcolorbox{white}{white}{\includegraphics[scale=0.12]{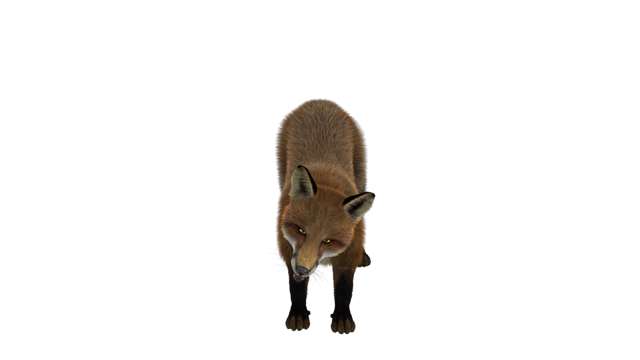}}}{}
\jsubfig{\includegraphics[scale=0.12]{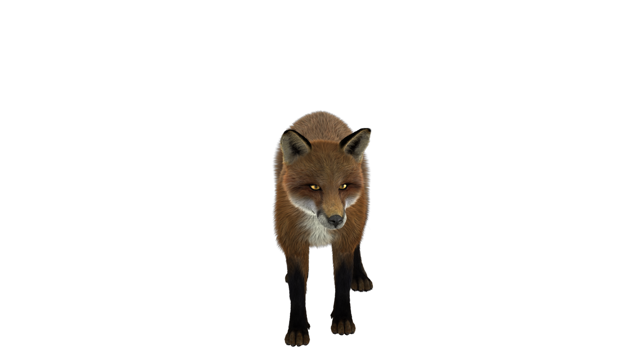}}{}%
\\
\jsubfig{\includegraphics[scale=0.12]{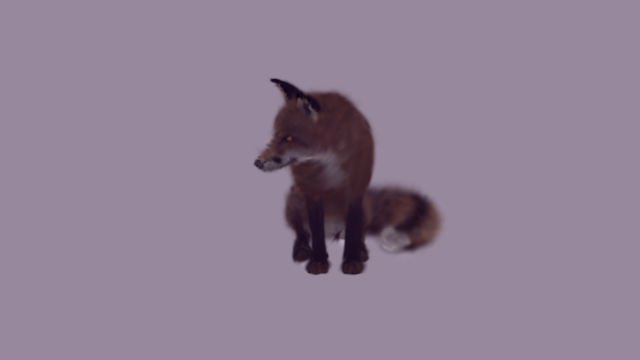}}{}
\jsubfig{\includegraphics[scale=0.12]{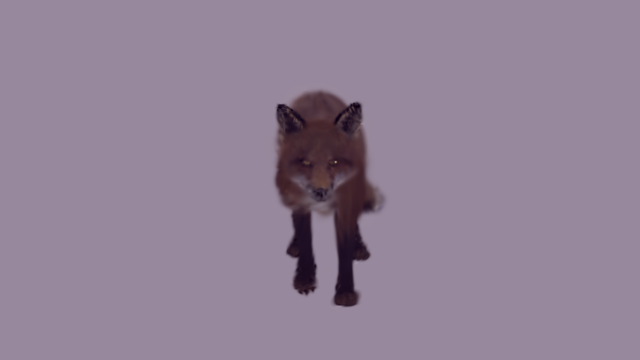}}{}
\jsubfig{\fcolorbox{red}{red}{\includegraphics[scale=0.12]{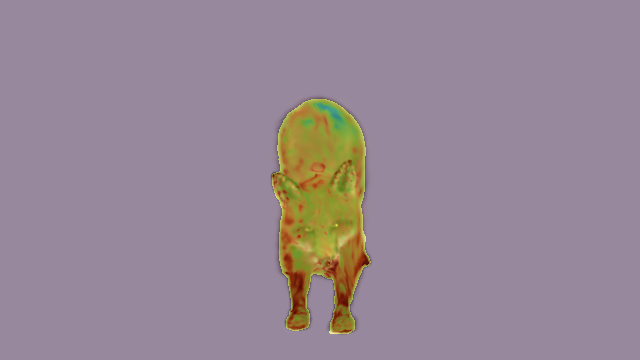}}}{}
\jsubfig{\fcolorbox{red}{red}{\includegraphics[scale=0.12]{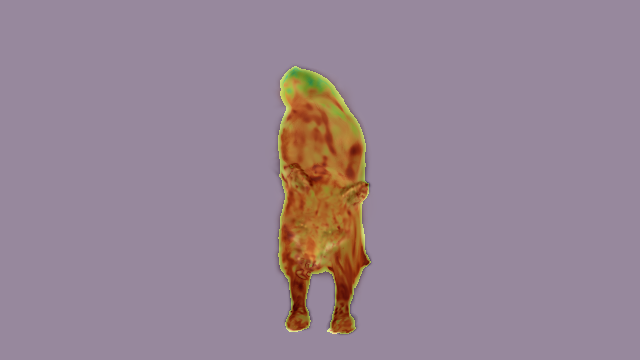}}}{}
\jsubfig{\fcolorbox{red}{red}{\includegraphics[scale=0.12]{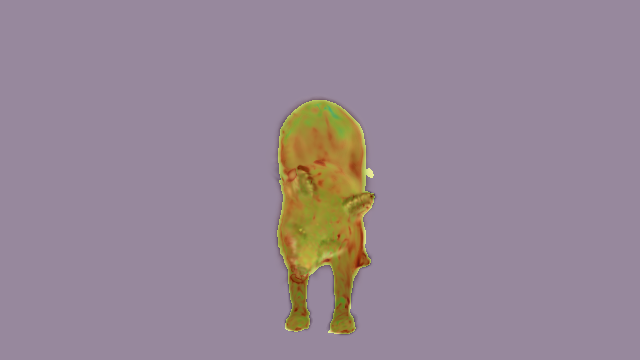}}}{}
\jsubfig{{\includegraphics[scale=0.12]{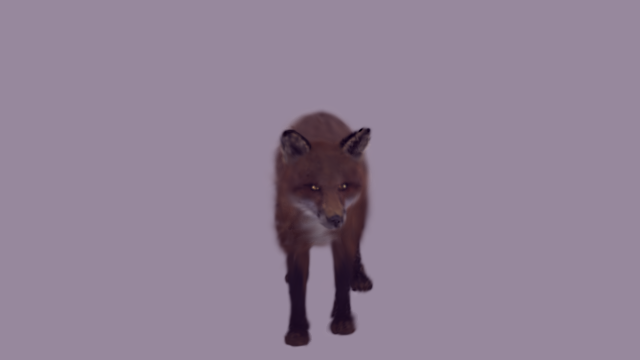}}}{}%
\\
\vspace{4pt}
{ Input query: \emph{A fox stretching}}
\caption{\textbf{Qualitative results on the Dynamic Furry Animals dataset}~\cite{luo2022artemis}, with the input frames depicted on top and the rendered probabilities directly below (frames localized temporally are in red). As illustrated above, our approach succeeds in temporally localizing subtle object-centric events.  %
}
\label{fig:results}
\end{figure*}

Our method can localize a given text query spatially and temporally. However, there are several limitations to consider. First, as the 3D Gaussian Splatting representation we build upon is optimized per timestep, runtime may potentially become an issue for longer videos. We also inherit its quality for rendering the scene from novel views. In the spatio-temporal highlights we created, we did observe some visual artifacts, particularly when the rendered Gaussians are close to the camera view.
Second, our localization accuracy is tied to our video encoder backbone. We found that ViCLIP struggled with fine grained localization of some actions (\emph{e.g.}, highlighting the entire Panda rather than its paws in Figure~\ref{fig:results}).
Given the high volume of research focusing on video-to-text models as well as field-based representations, we expect both localization accuracy and video quality to improve with future backbones.
Third, training both dynamic 3D Gaussians and an autoencoder per scene is time-consuming and computationally intense. Speeding up preprocessing is promising future work.

\begin{table}[!t]
\vspace{-8pt}
\centering
\resizebox{\linewidth}{!}{
\begin{tabular}{lcccccc}
\toprule
 Method & $\text{vAP}\uparrow$ & $\text{vIOU}\uparrow$&  $\text{tIOU}\uparrow$ & $\text{tRec}\uparrow$ & $\text{tPrec}\uparrow$ & $\text{tAP}\uparrow$\\
\midrule
2D Features & $44.9\pm11.7$ & $18.6\pm10.3$& $46.3\pm12.0$ & $73.3\pm14.2$ & $53.7\pm11.8$ & $60.3\pm12.5$ \\
$\text{Static}_{\text{CLIP}}$ & $33.3\pm1.0$ & $8.2\pm3.9$ &  $34.9\pm0.0$ & $62.2\pm0.0$ &  $43.4\pm00.0$ & $47.2\pm0.0$\\ 
 $\text{AVG}_{\text{CLIP}}$ & $32.5\pm0.7$ & $7.4\pm3.7$ & $35.2\pm0.0$ &$62.5\pm0.0$ & $43.6\pm0.0$ & $48.4\pm0.0$ \\
Ours & $\mathbf{58.7\pm0.6}$ &$\mathbf{25.6\pm4.6}$ & $\mathbf{60.8\pm0.0}$ & $\mathbf{83.0\pm0.0}$ & $\mathbf{68.5\pm0.0}$ & $\mathbf{72.6\pm0.0}$ \\  
\bottomrule
\end{tabular}
}
\caption{\textbf{Ablation study}, ablating the use of a 4D learned feature field (2D Features, the extracted ViCLIP features from the input 2D videos) and the video--text model ($\text{Static}_{\text{CLIP}}$, $\text{AVG}_{\text{CLIP}}$). Best results are highlighted in bold.}
\label{tab:ablations}
\end{table}

\begin{table}[!t]
\vspace{-8pt}
\centering
\resizebox{\linewidth}{!}{
\begin{tabular}{lcccccc}
\toprule
 Method & $\text{vAP}\uparrow$ & $\text{vIOU}\uparrow$&  $\text{tIOU}\uparrow$ & $\text{tRec}\uparrow$ & $\text{tPrec}\uparrow$ & $\text{tAP}\uparrow$\\
\midrule
$\text{Local}_{\text{Blur}}$ & $55.3\pm1.3$ & $19.8\pm5.1$ &  $57.9\pm0.0$ & $83.1\pm0.0$ &  $65.2\pm0.0$ &  $70.3\pm0.0$\\
$\text{Local}_{\text{Reg}}$ & $56.7\pm1.4$ & $21.1\pm4.7$& $59.3\pm0.0$ &$82.7\pm0.0$ & $66.4\pm0.0$ &  $71.1\pm0.0$\\ 
$\text{Local}_{\text{MLP}}$ & $51.7\pm1.2$ & $18.2\pm3.3$& $54.1\pm0.0$ & $\mathbf{85.4\pm0.0}$ & $59.6\pm0.0$ &  $67.7\pm0.0$\\
w/o Attn & $57.9\pm1.6$ & $20.1\pm4.2$&$60.6\pm0.0$ & $83.6\pm0.0$ & $67.7\pm0.0$ &  $67.9\pm0.0$\\  
Ours & $\mathbf{58.7\pm0.6}$ & $\mathbf{25.6\pm4.6}$ & $\mathbf{60.8\pm0.0}$ & $83.0\pm0.0$ & $\mathbf{68.5\pm0.0}$ & $\mathbf{72.6\pm0.0}$ \\
\bottomrule
\end{tabular}
}
\caption{\textbf{Self-Attention Ablations}, ablating the use of self-attention (w/o Attn) and the effectiveness of self-attention in enforcing local smoothness in the 3D space ($\text{Local}_{\text{Blur}}$, $\text{Local}_{\text{Reg}}$ and $\text{Local}_{\text{MLP}}$). Best results are highlighted in bold. See the supplementary material for additional details.}
\label{tab:ablations-smooth}
\end{table}

\begin{figure} %
\centering
\jsubfig{\includegraphics[scale=0.12]{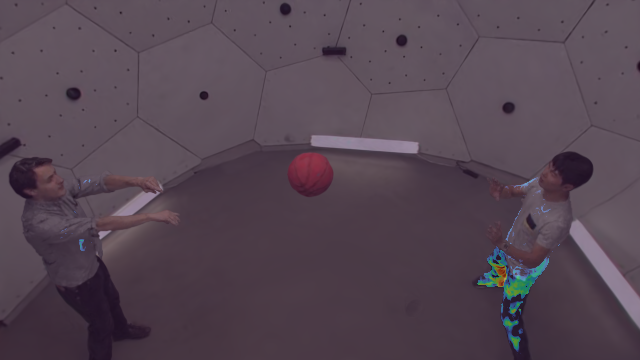}}{}
\jsubfig{\includegraphics[scale=0.12]{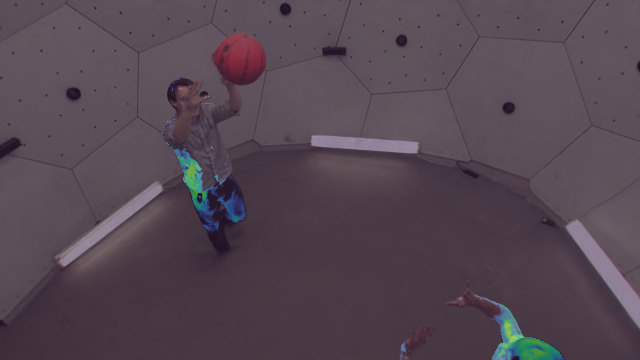}}{}
\jsubfig{\includegraphics[scale=0.12]{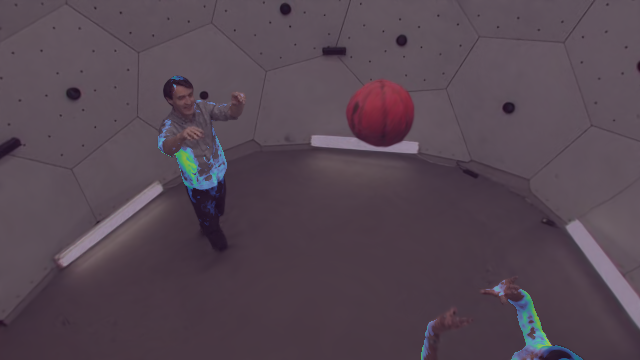}}{}
 \vspace{-2pt}
\caption{\textbf{Limitations}. If a query that does not appear in the scene is given as input our method might localize regions in the scene. For instance, above we show heatmaps obtained over the Basketball scene for the input query ``A person swinging the bat". Although the query does not occur in the scene, our method outputs non-negligible probabilities over certain timesteps.
}
\label{fig:limitations}
\end{figure}

Additionally, if the user inputs a query that does not appear in the scene, our approach may localize related regions. For instance, the query \emph{A person swinging the bat} in the Basketball scene yields localized timesteps where the ball is in the air (see Figure \ref{fig:limitations}). 
This happens because our approach assumes that the existence of 3D Gaussians with a relevancy score above 0.5 signify that the queried action indeed appears in the 3D scene. This issue could be alleviated by computing a more scene-specific relevancy score, which considers additional scene-relevant phrases as negative prompts.

\section{Applications}

We demonstrate the utility of \methodname{} for text-conditioned video editing tasks. Our primary application is  creating spatio-temporal highlights. %
We further show that our approach allows for spatio-temporal localization given multiple dynamic 3D environments. %

\begin{figure*}
\centering
\jsubfig{{\includegraphics[width=1\textwidth]{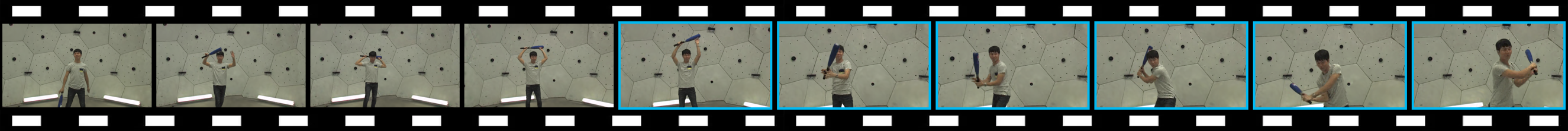}}}{} \\
{\emph{A person swinging the softball bat} (with \textcolor{cyan}{zoomed-in} frames)}
\\
\jsubfig{{\includegraphics[width=1\textwidth]{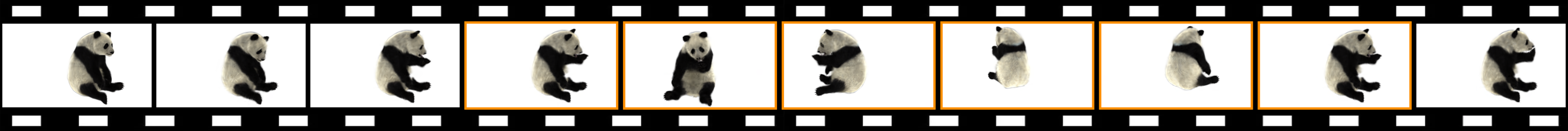}}}
{\emph{A panda licking its paws} (with \textcolor{orange}{bullet-time} frames)}
\\
\jsubfig{{\includegraphics[width=1\textwidth]{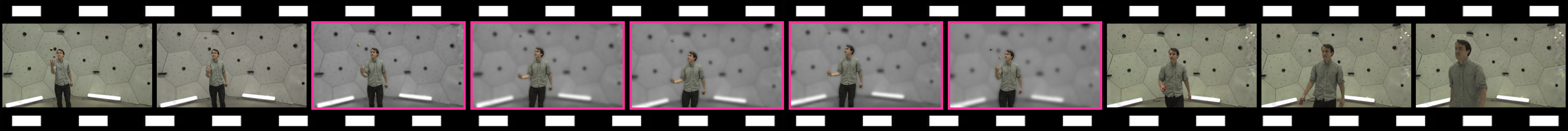}}}{} \\
{\emph{A person juggles} (with \textcolor{magenta}{desaturated background} frames)}
\vspace{-1pt}
  \caption{
 \textbf{Spatio-temporal highlights created with 4-LEGS}. Above we show three different highlights created automatically with our approach:  \textcolor{cyan}{zooming in} to the queried action, generating a \textcolor{orange}{bullet-time} display of the action, and \textcolor{magenta}{desaturating background} regions. We show sampled generated frames, visualized in these unique colors. Please refer to the supplementary material for the full video sequences.
  }\label{fig:apps}
\end{figure*}

\begin{figure*} %
\centering
\jsubfig{\includegraphics[scale=0.063]{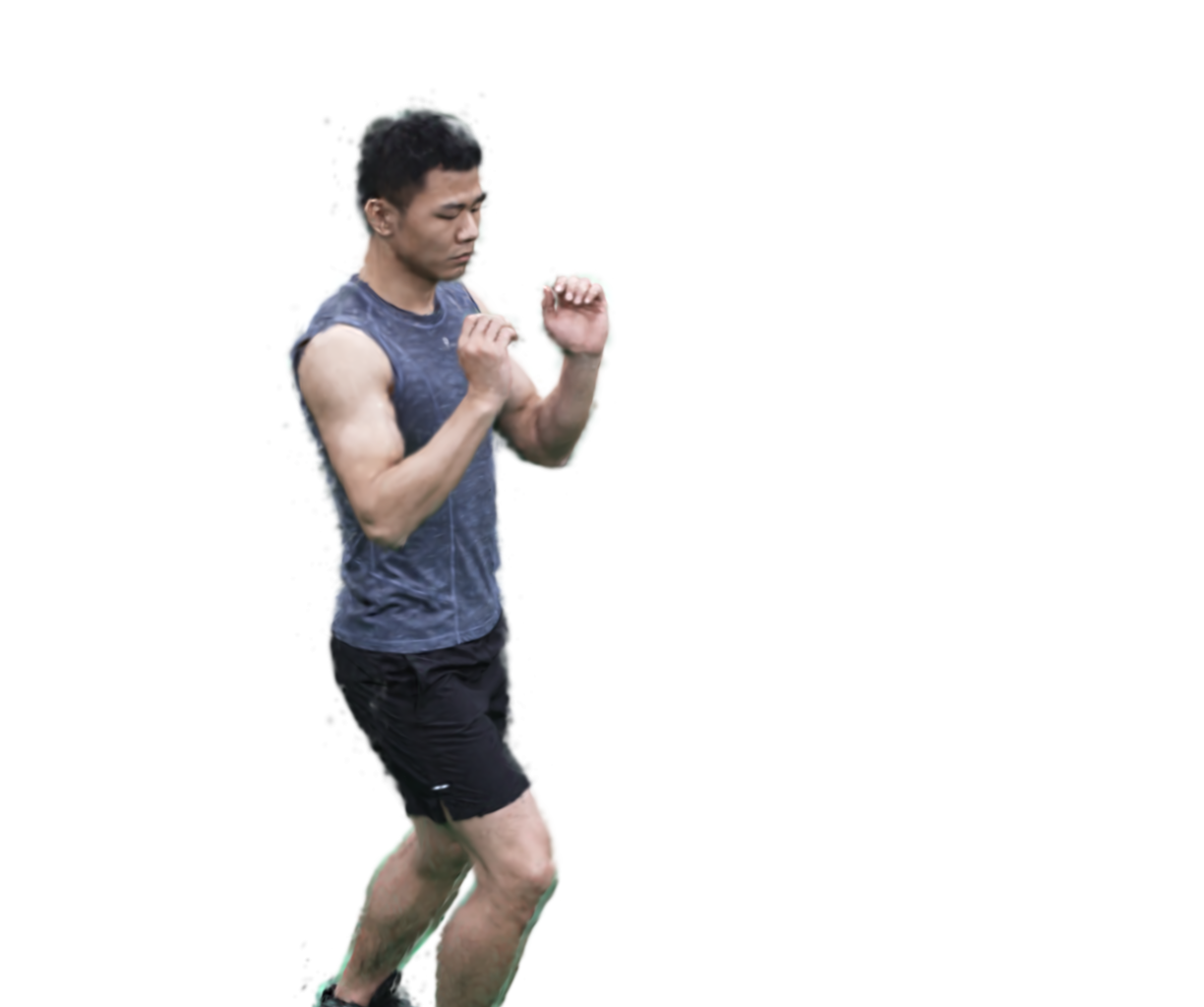}}{}
\jsubfig{{\includegraphics[scale=0.063]{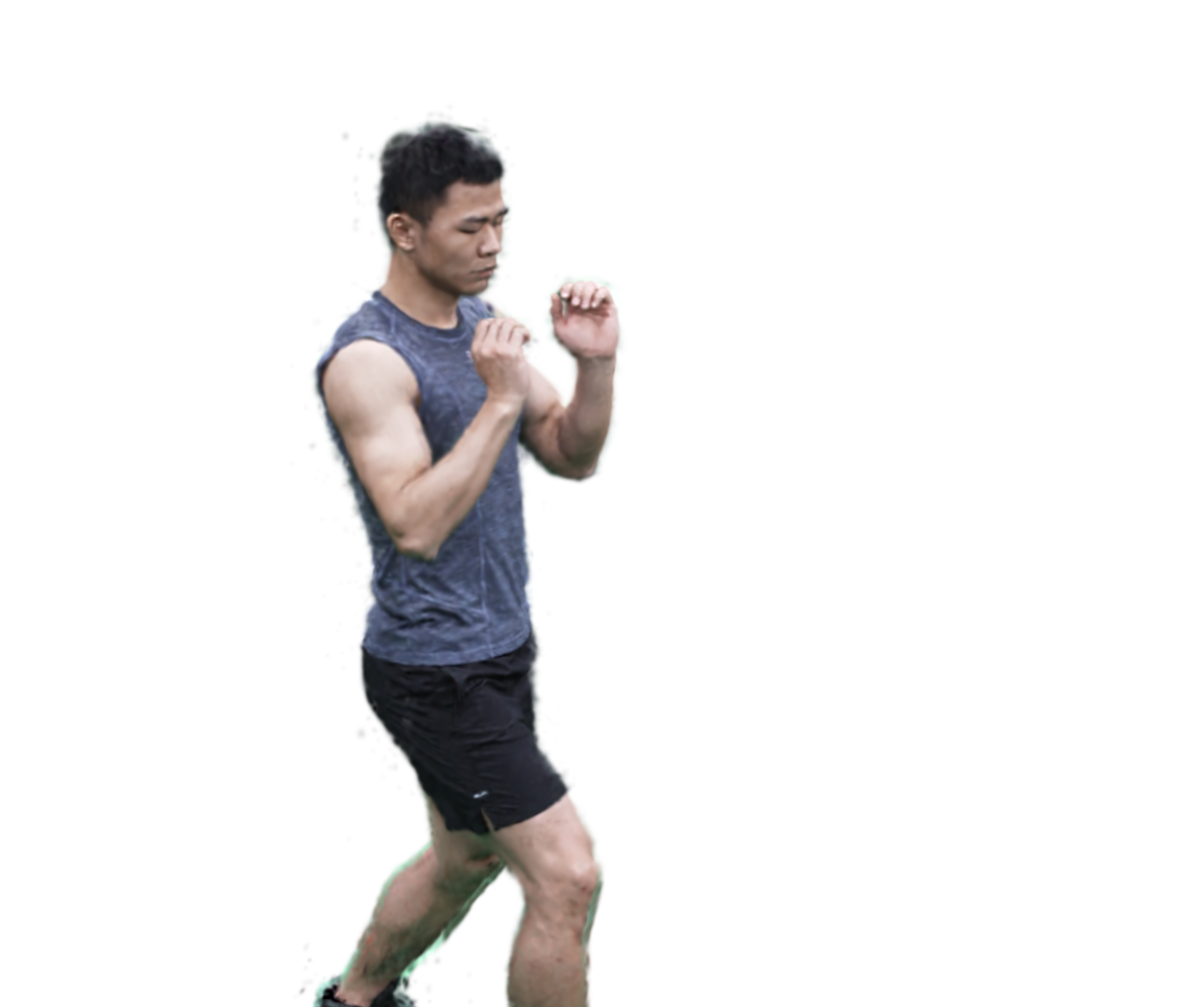}}}{}
\jsubfig{\fcolorbox{white}{white}{\includegraphics[scale=0.063]{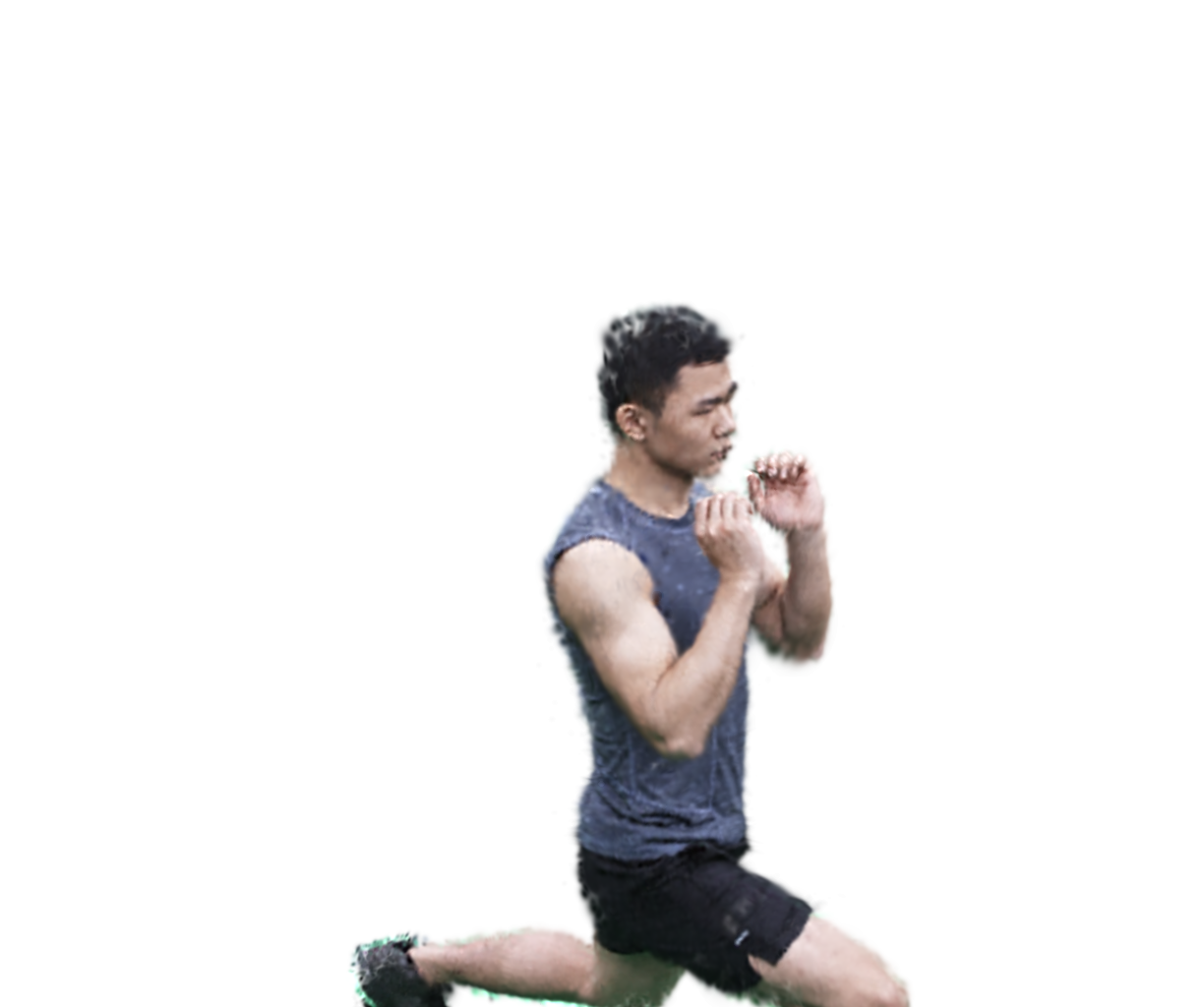}}}{}
\jsubfig{\fcolorbox{white}{white}{\includegraphics[scale=0.063]{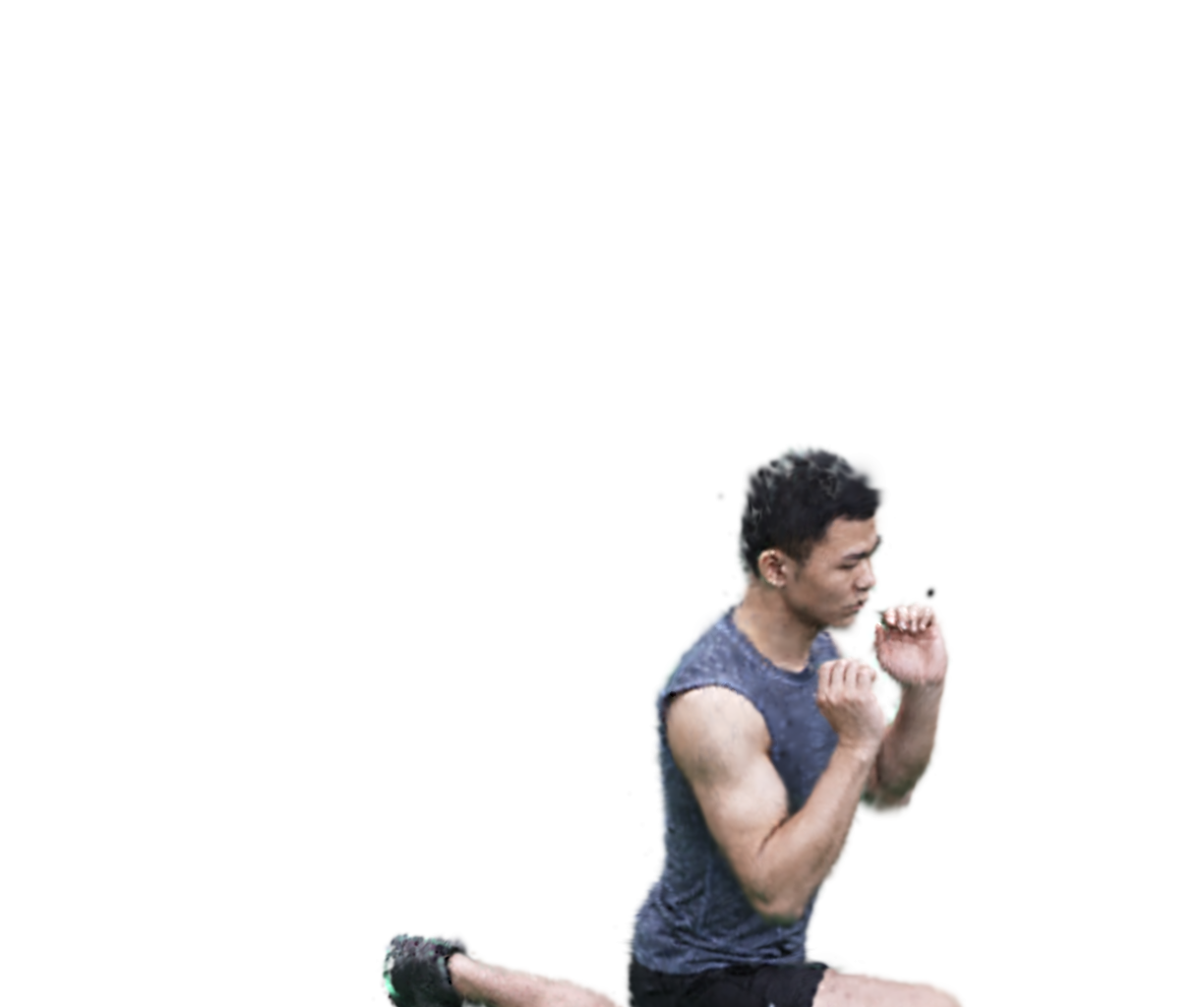}}}{}
\jsubfig{\fcolorbox{white}{white}{\includegraphics[scale=0.063]{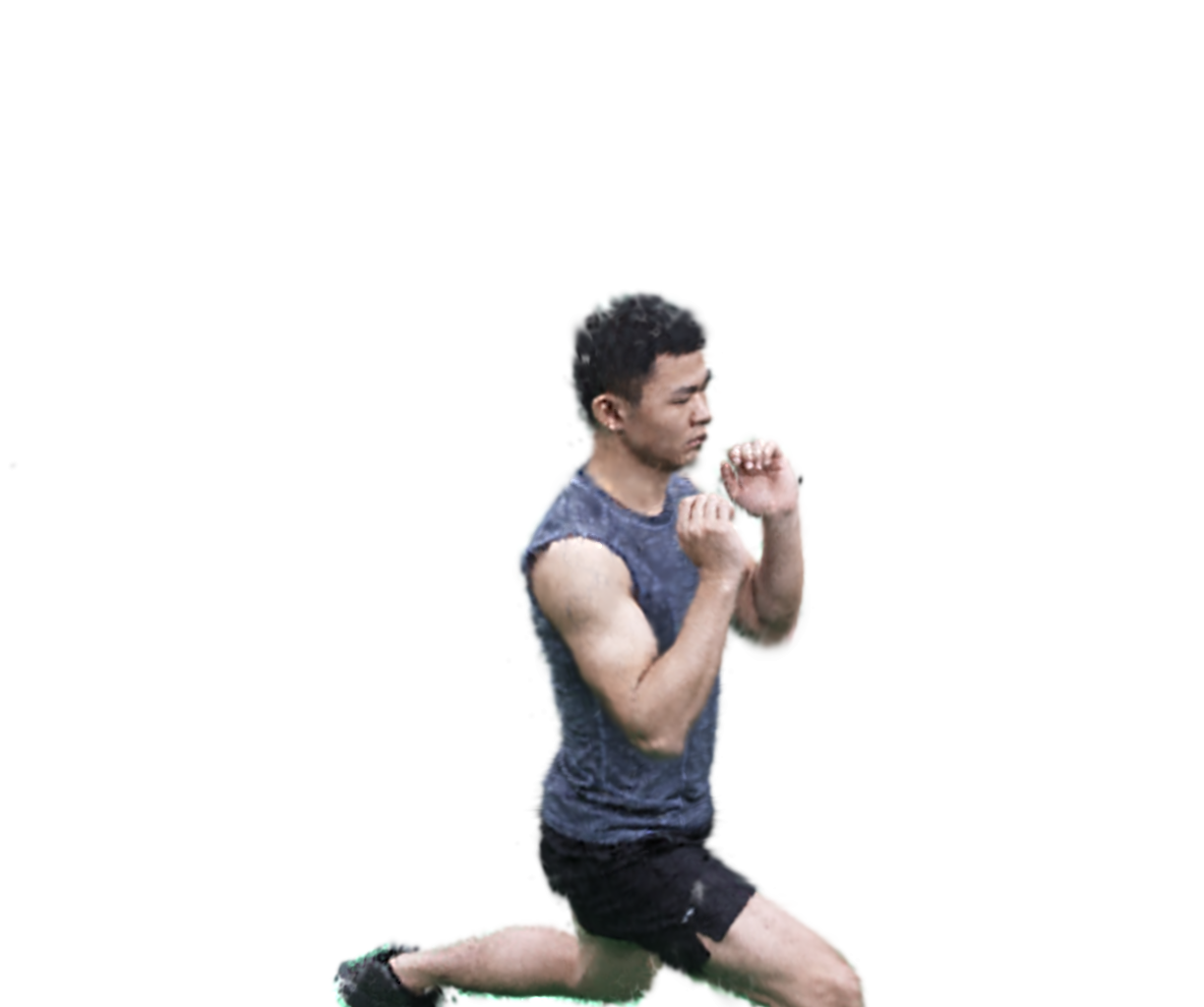}}}{}
\jsubfig{\includegraphics[scale=0.063]{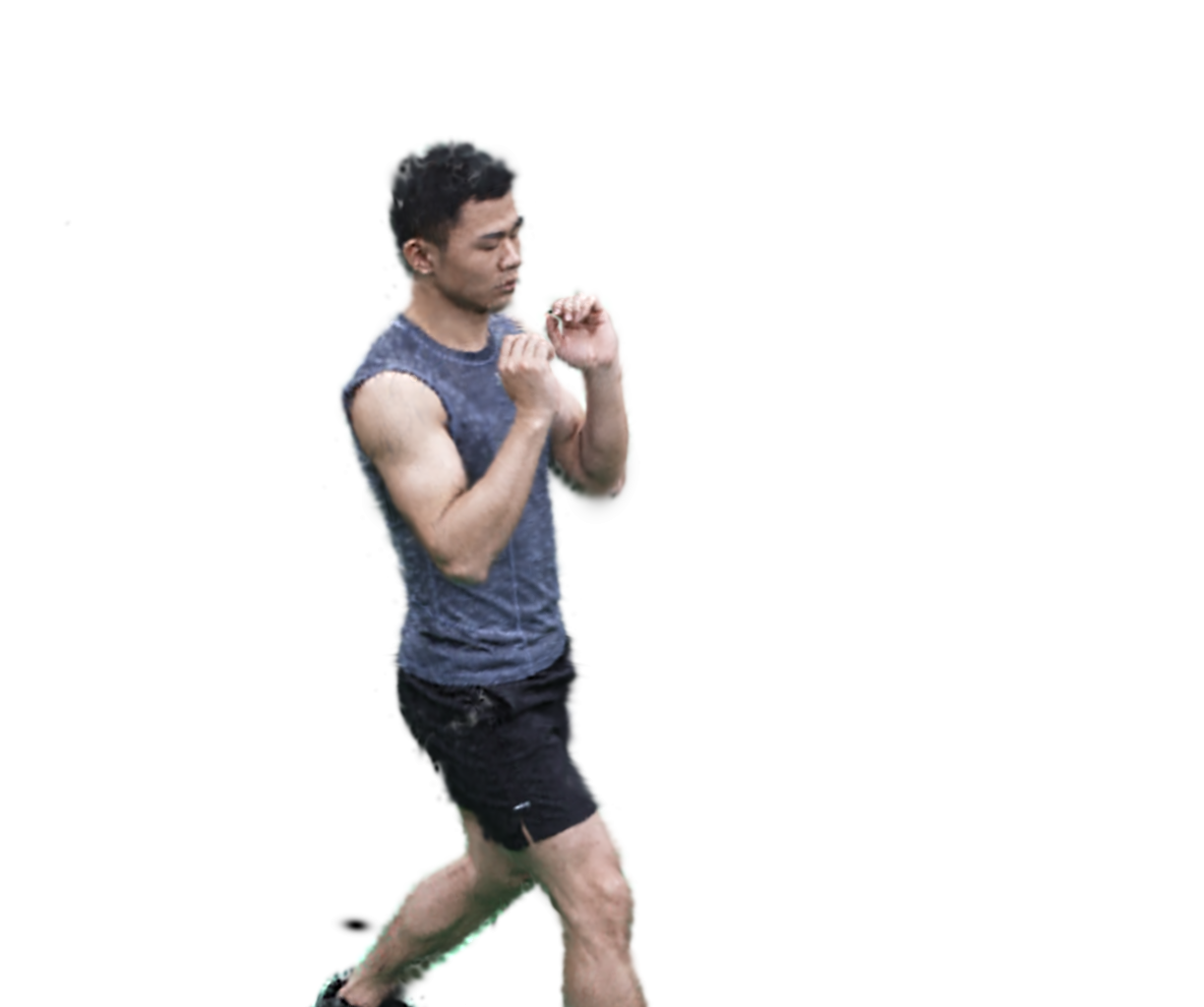}}{}%
\\
\jsubfig{\includegraphics[scale=0.063]{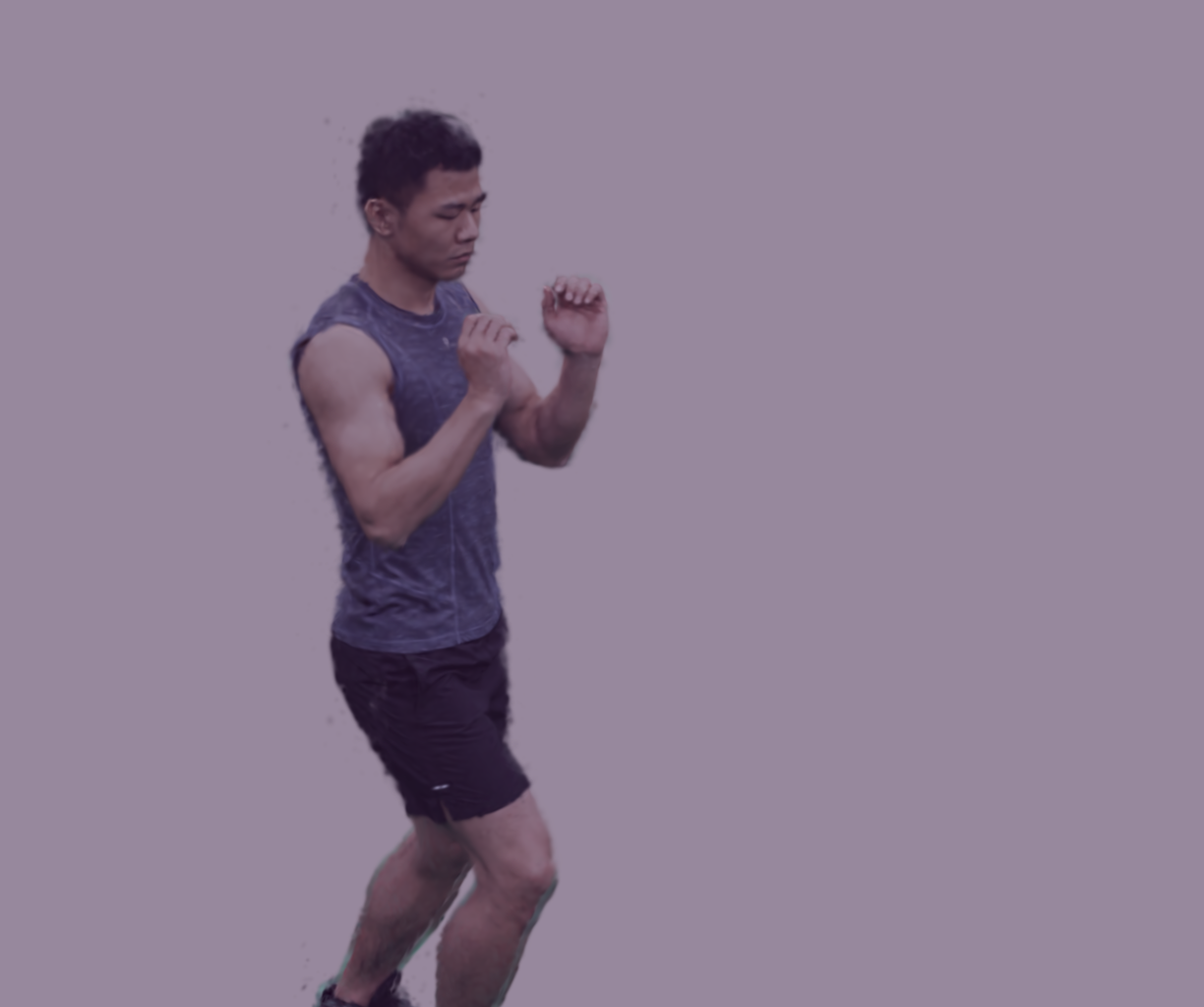}}{}
\jsubfig{{\includegraphics[scale=0.063]{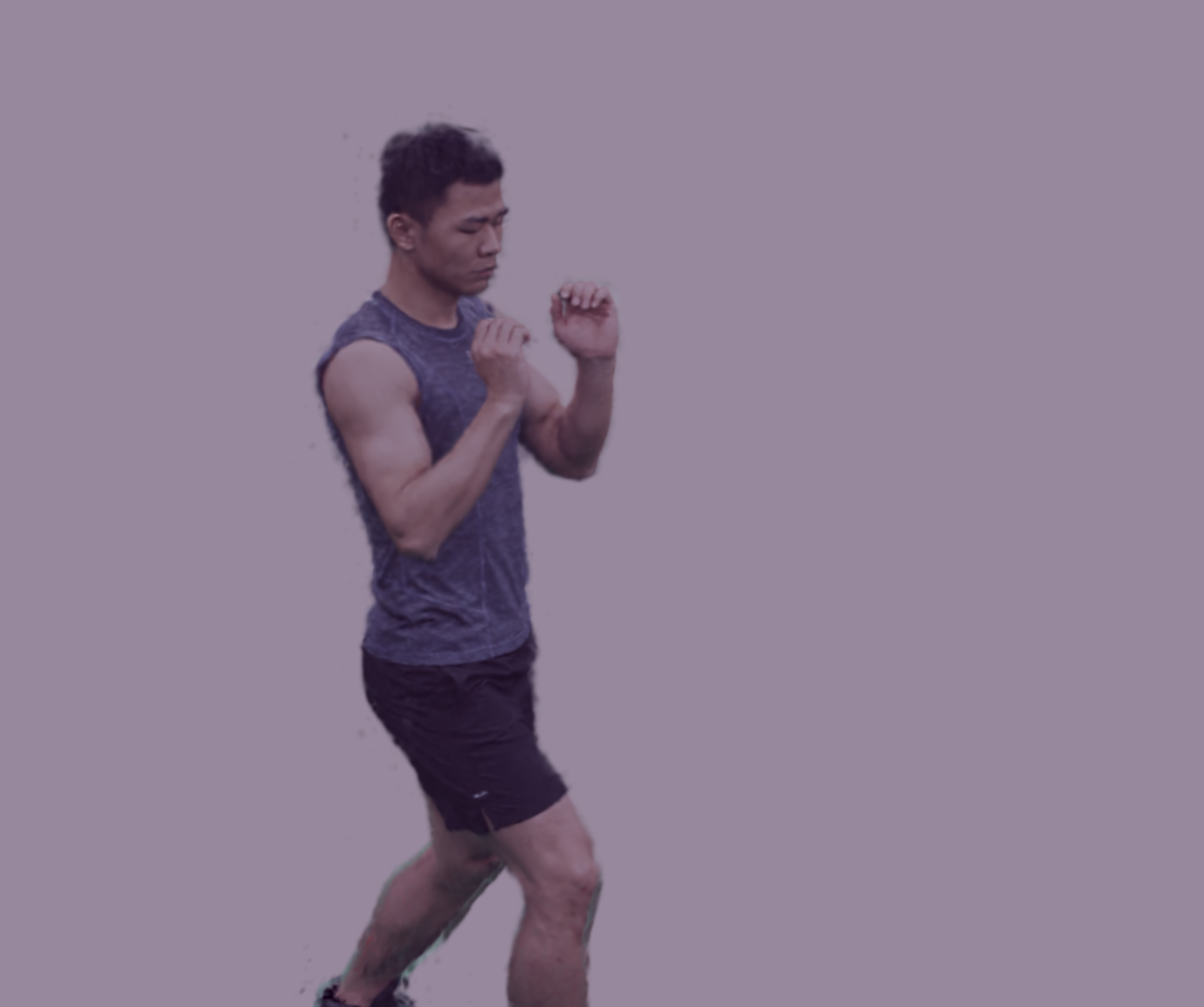}}}{}
\jsubfig{\fcolorbox{red}{red}{\includegraphics[scale=0.063]{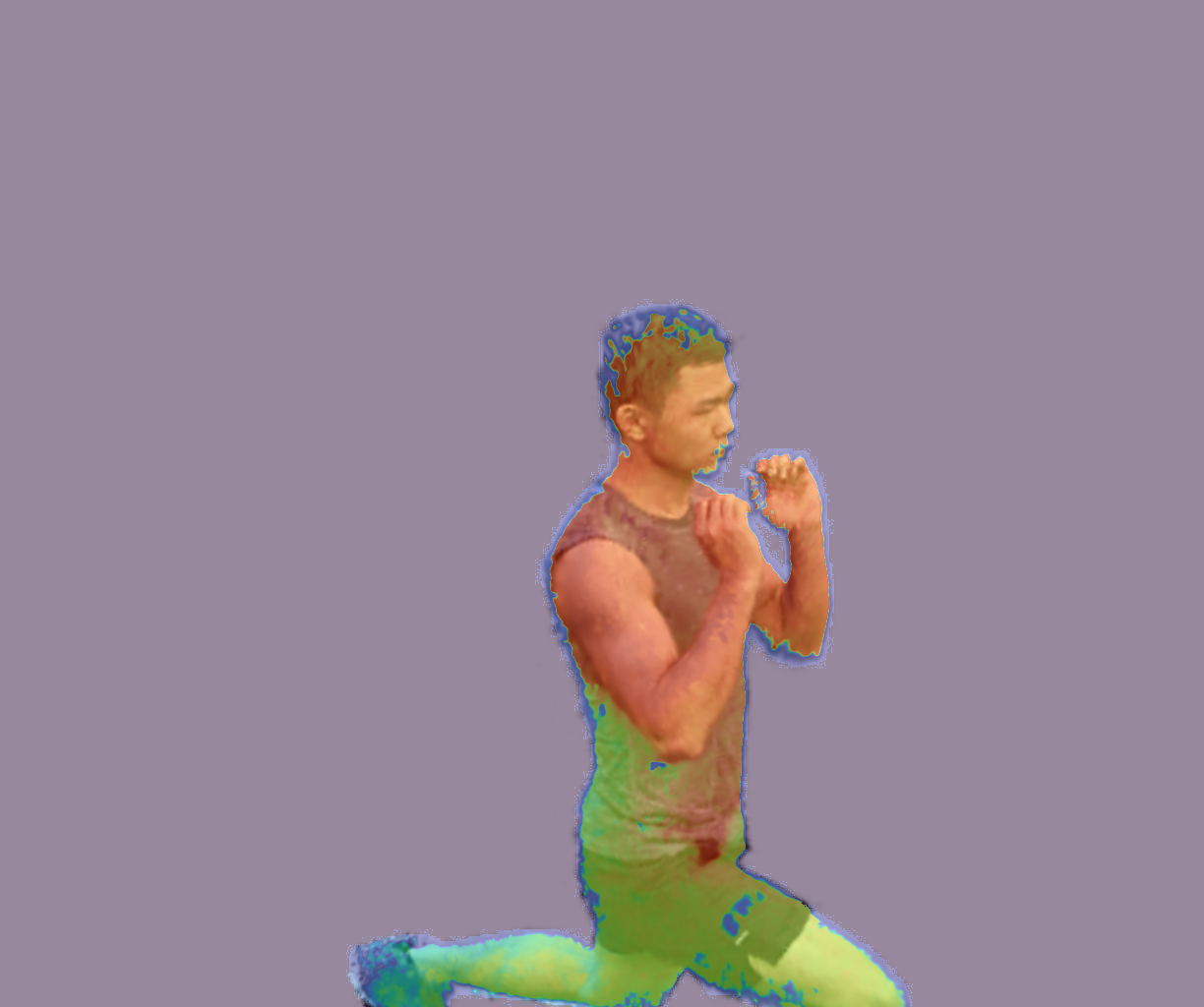}}}{}
\jsubfig{\fcolorbox{red}{red}{\includegraphics[scale=0.063]{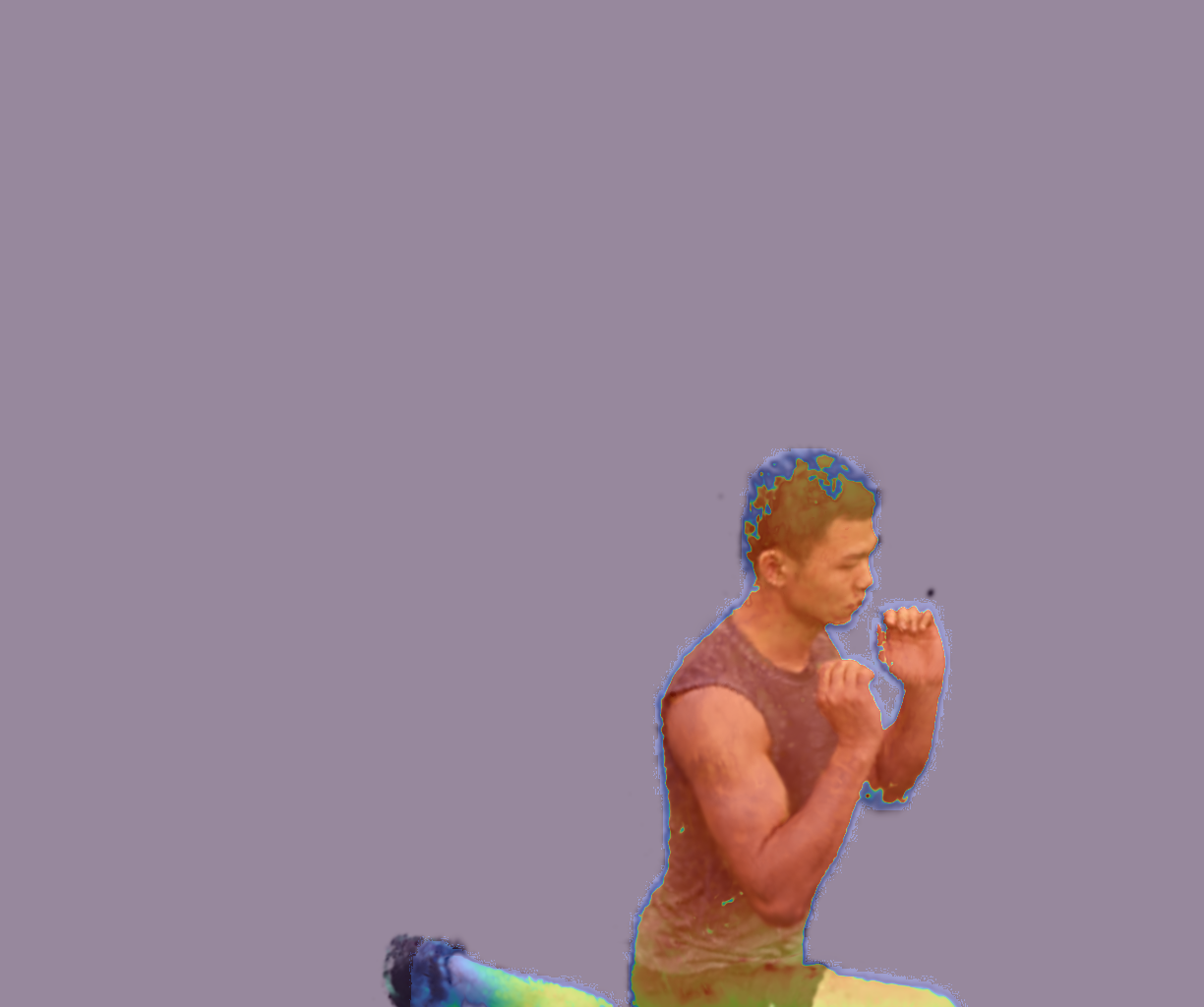}}}{}
\jsubfig{\fcolorbox{red}{red}{\includegraphics[scale=0.063]{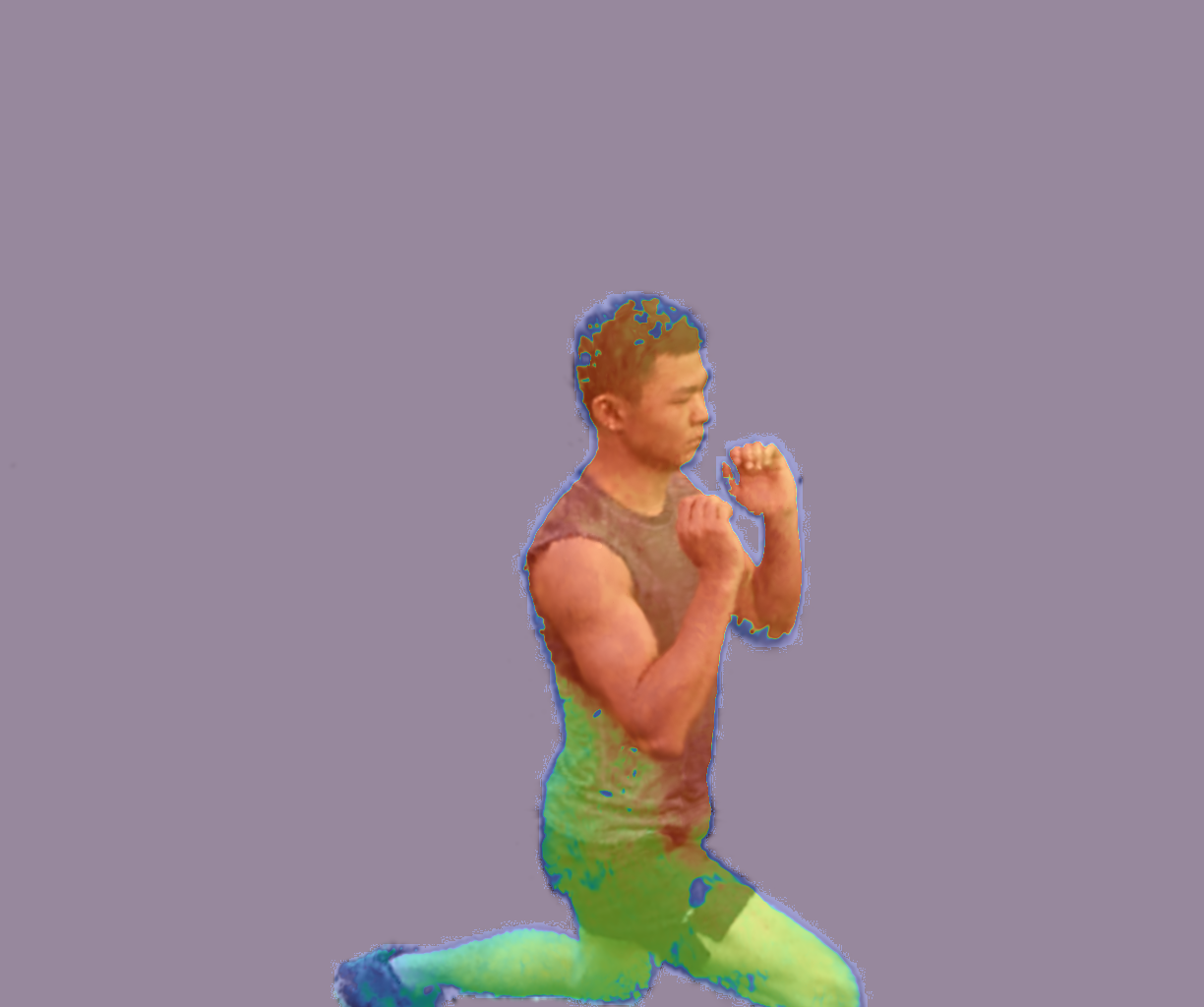}}}{}
\jsubfig{\includegraphics[scale=0.063]{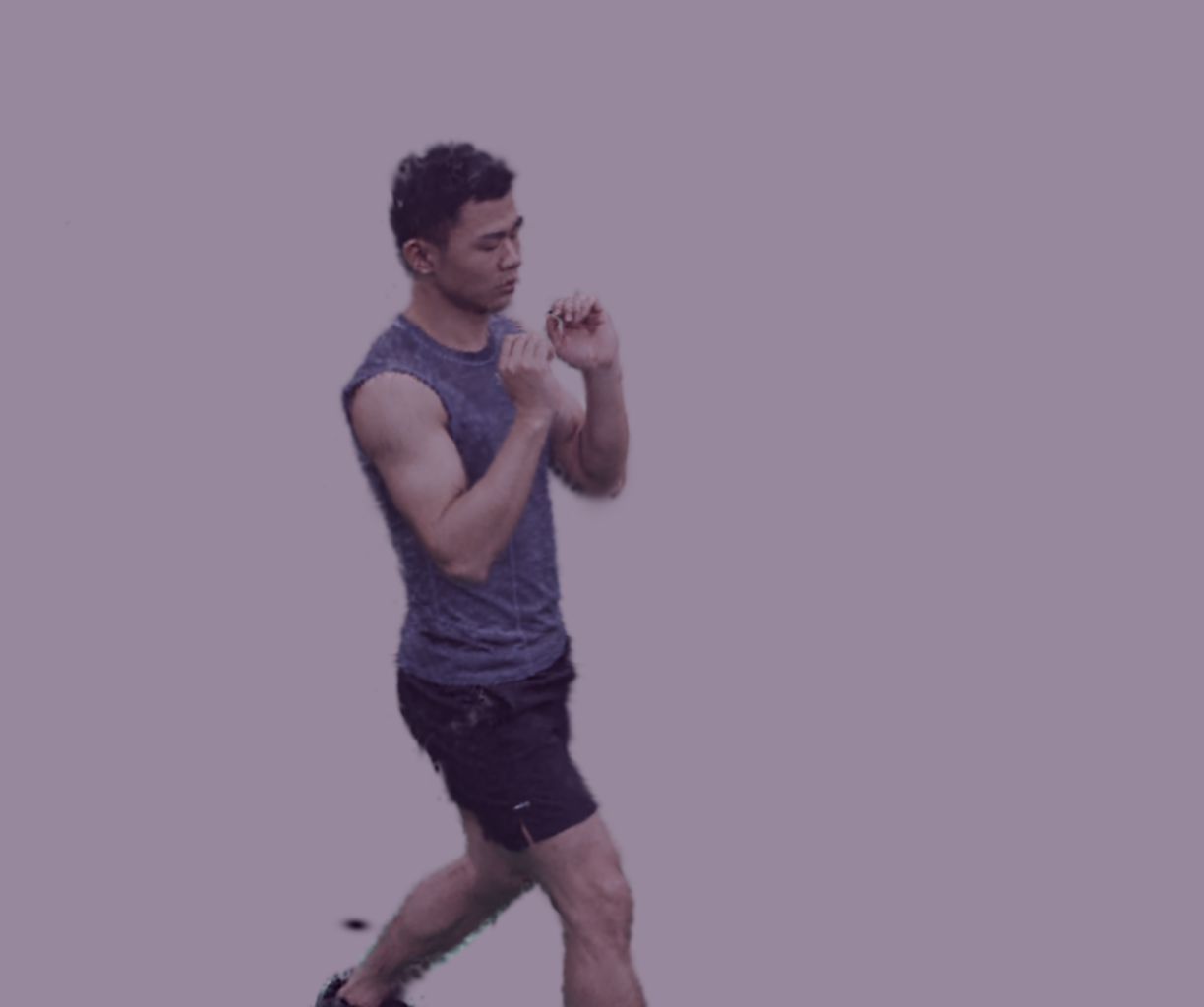}}{}%
\\
\vspace{4pt}
{ Input query: \emph{A person stretching}}
\vspace{4pt}
\\
\jsubfig{\includegraphics[scale=0.063]{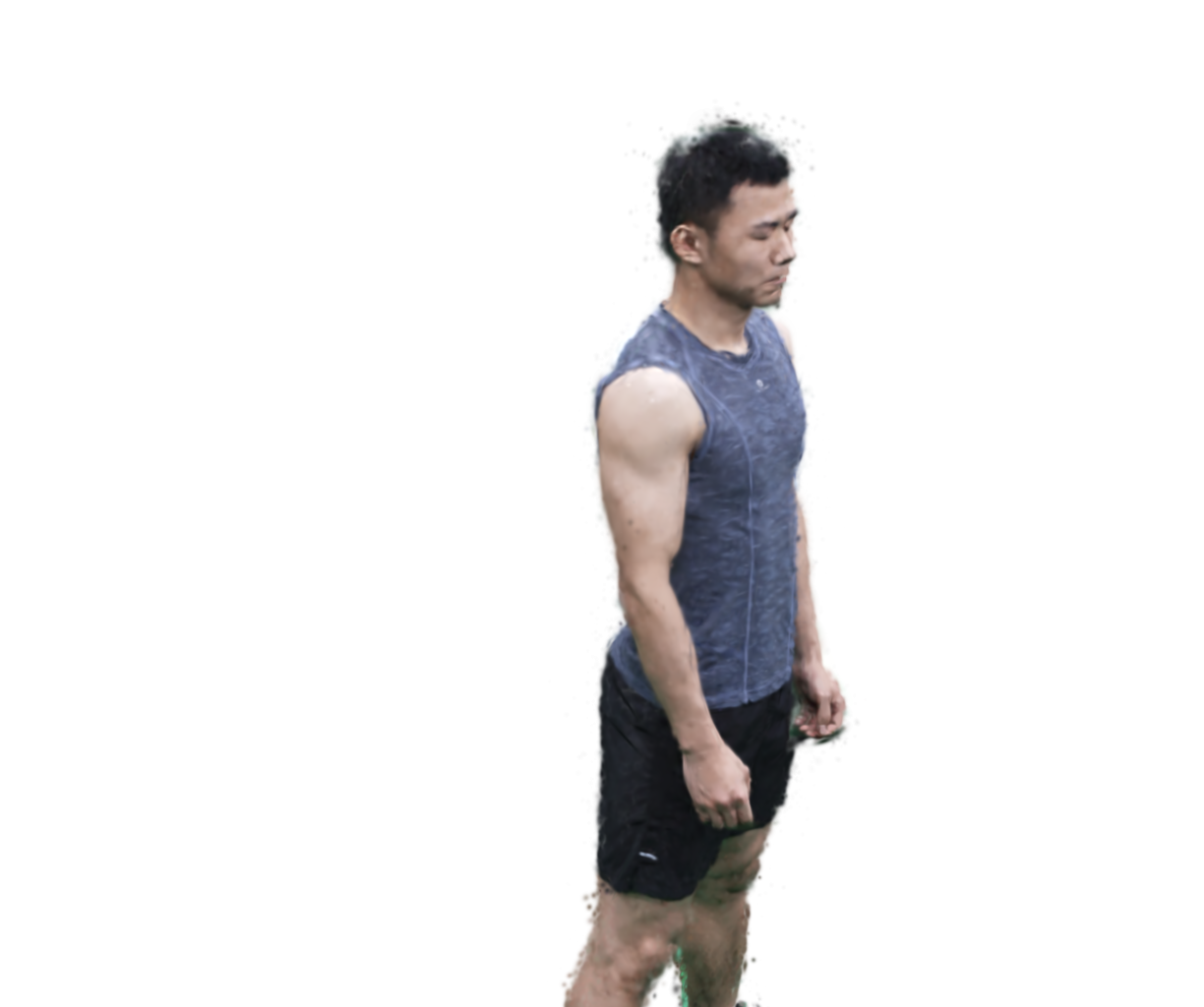}}{}
\jsubfig{{\includegraphics[scale=0.063]{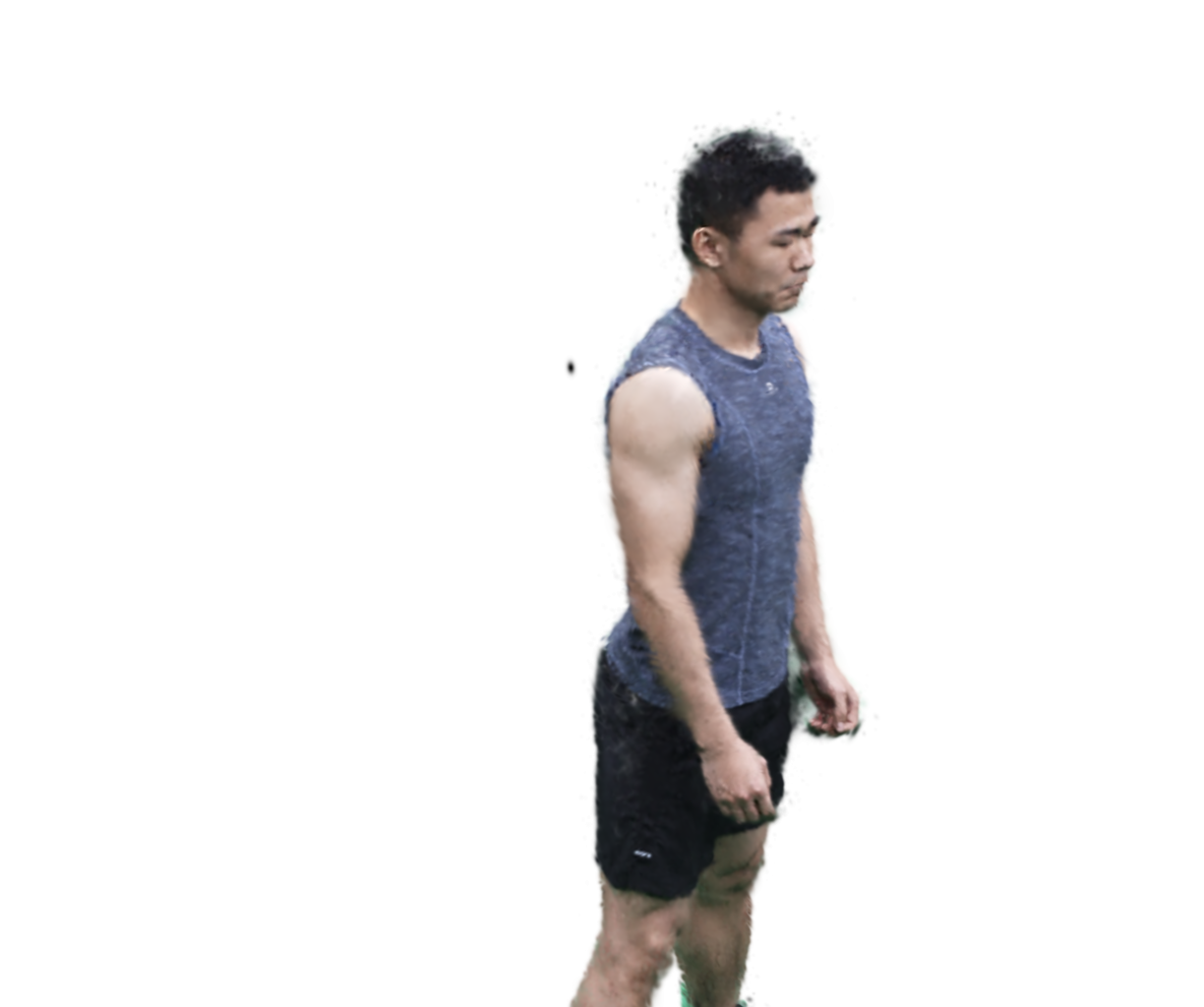}}}{}
\jsubfig{\fcolorbox{white}{white}{\includegraphics[scale=0.063]{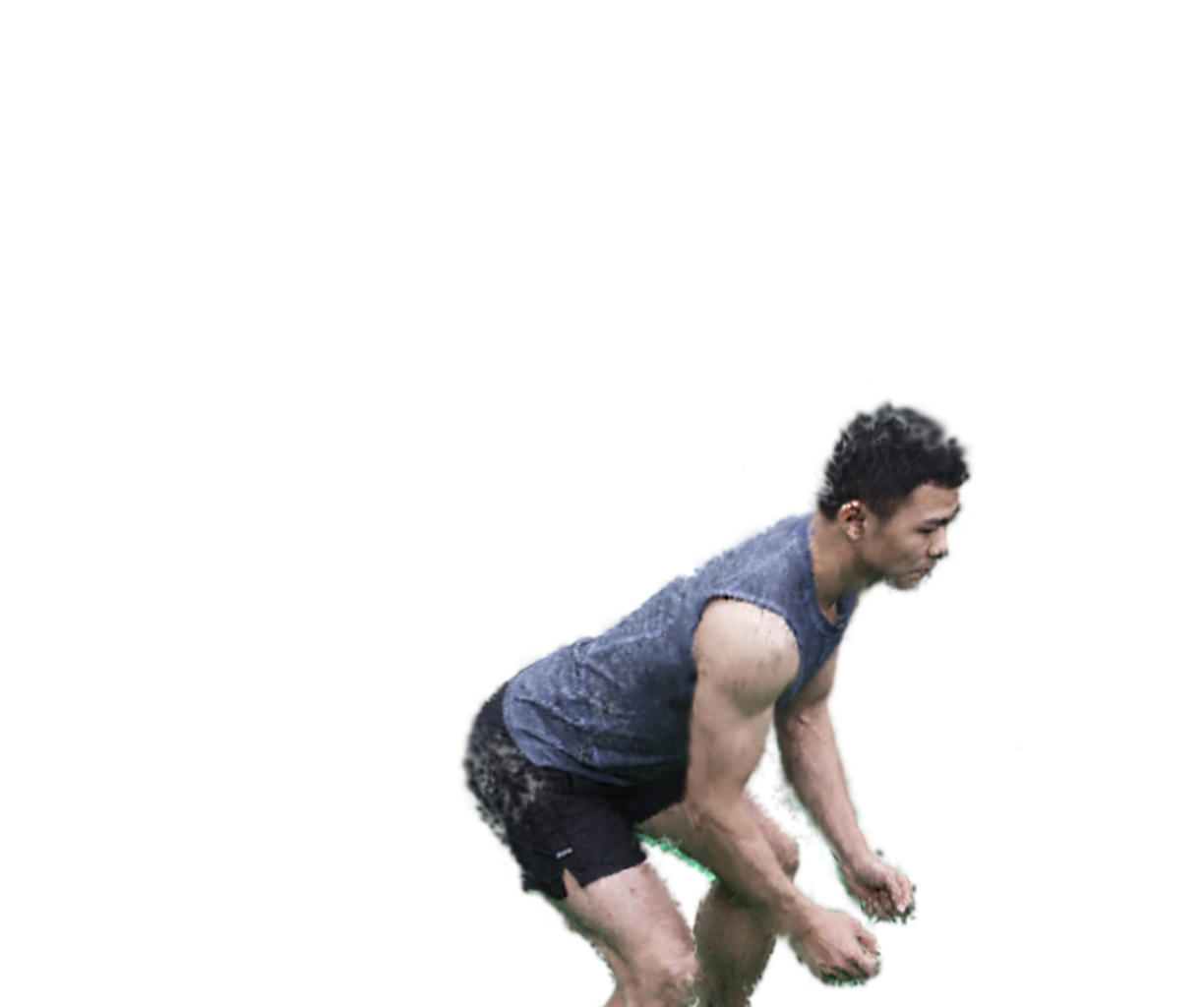}}}{}
\jsubfig{\fcolorbox{white}{white}{\includegraphics[scale=0.063]{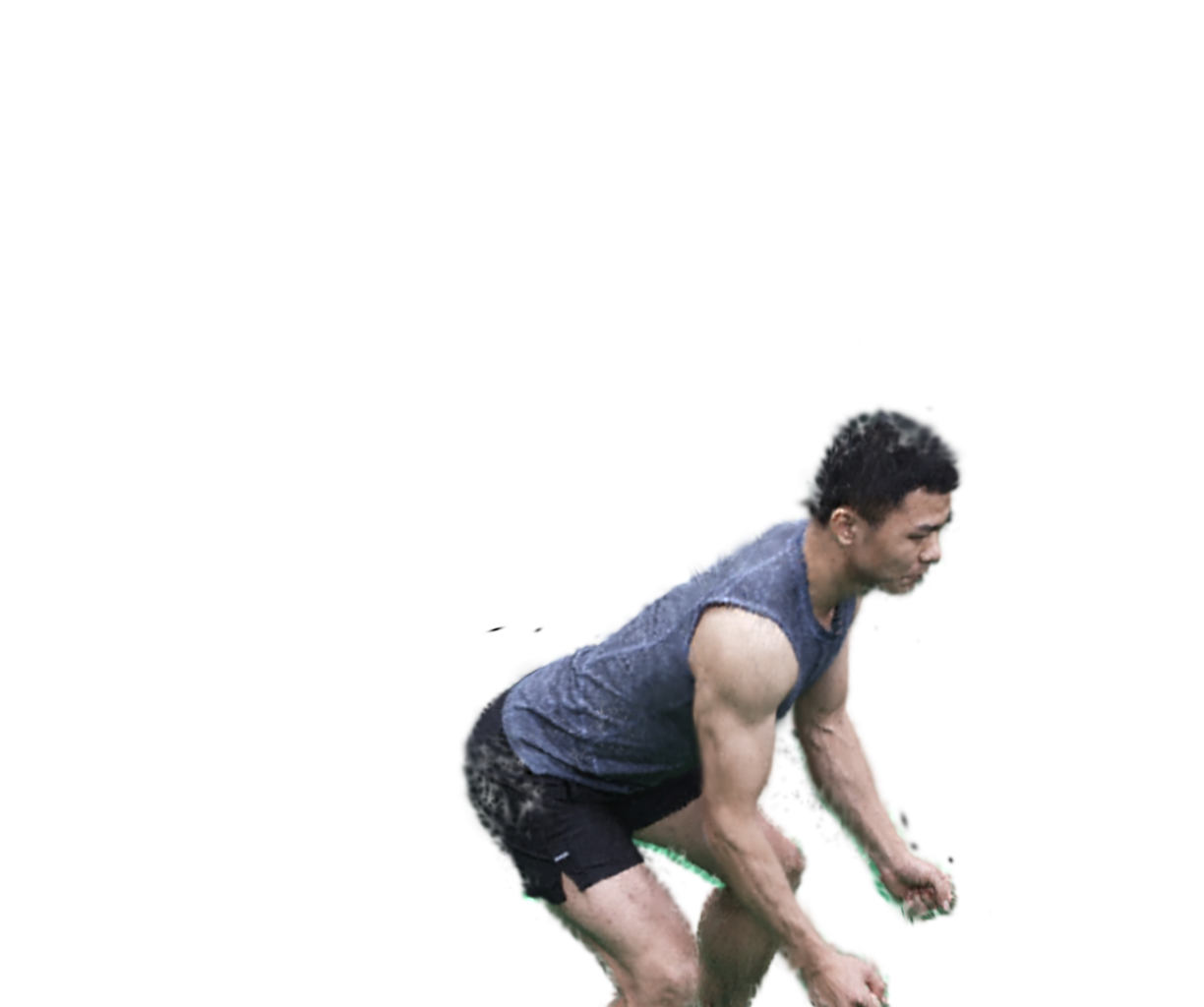}}}{}
\jsubfig{\fcolorbox{white}{white}{\includegraphics[scale=0.063]{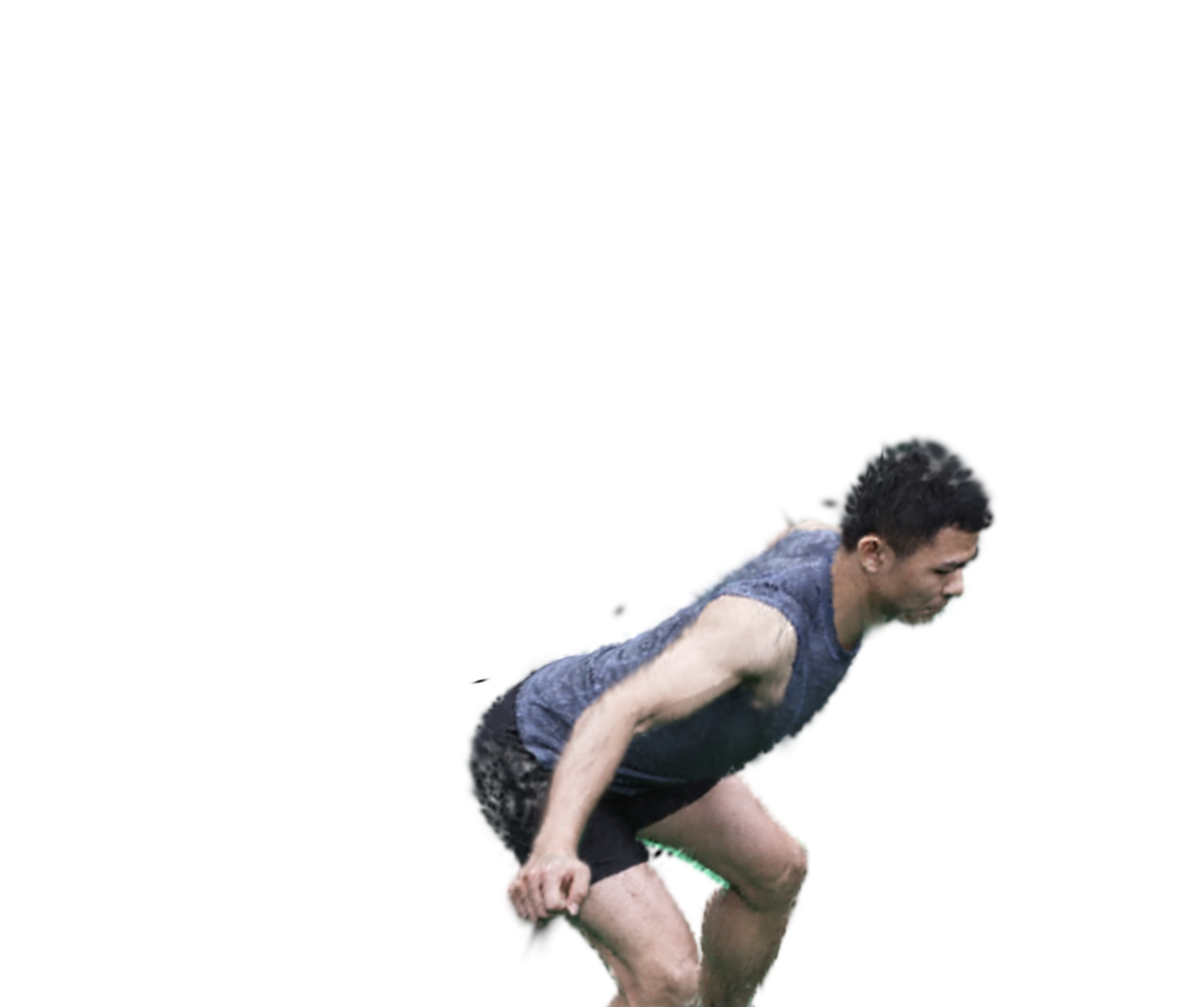}}}{}
\jsubfig{\fcolorbox{white}{white}{\includegraphics[scale=0.063]{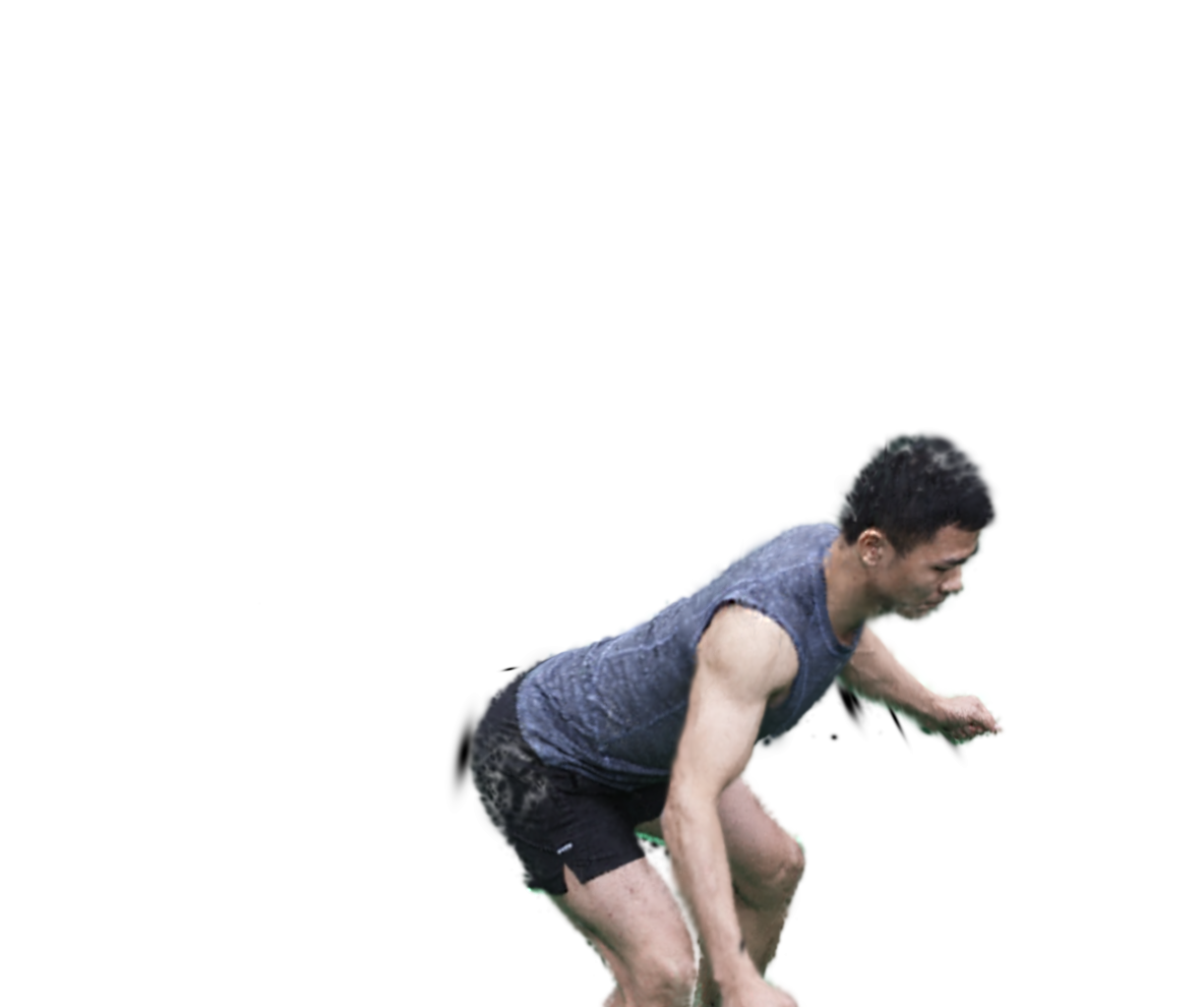}}}{}%
\\
\jsubfig{\includegraphics[scale=0.063]{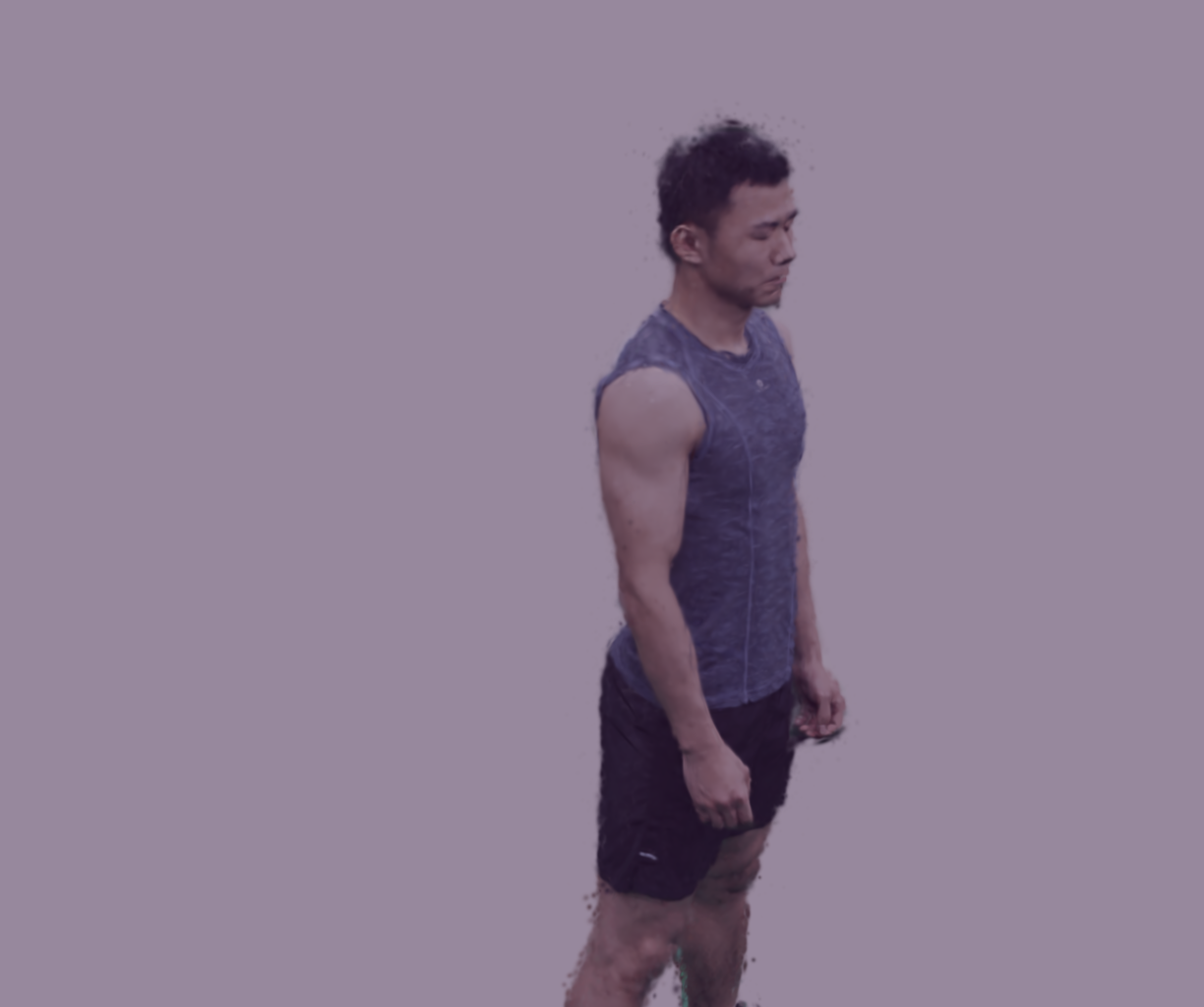}}{}
\jsubfig{{\includegraphics[scale=0.063]{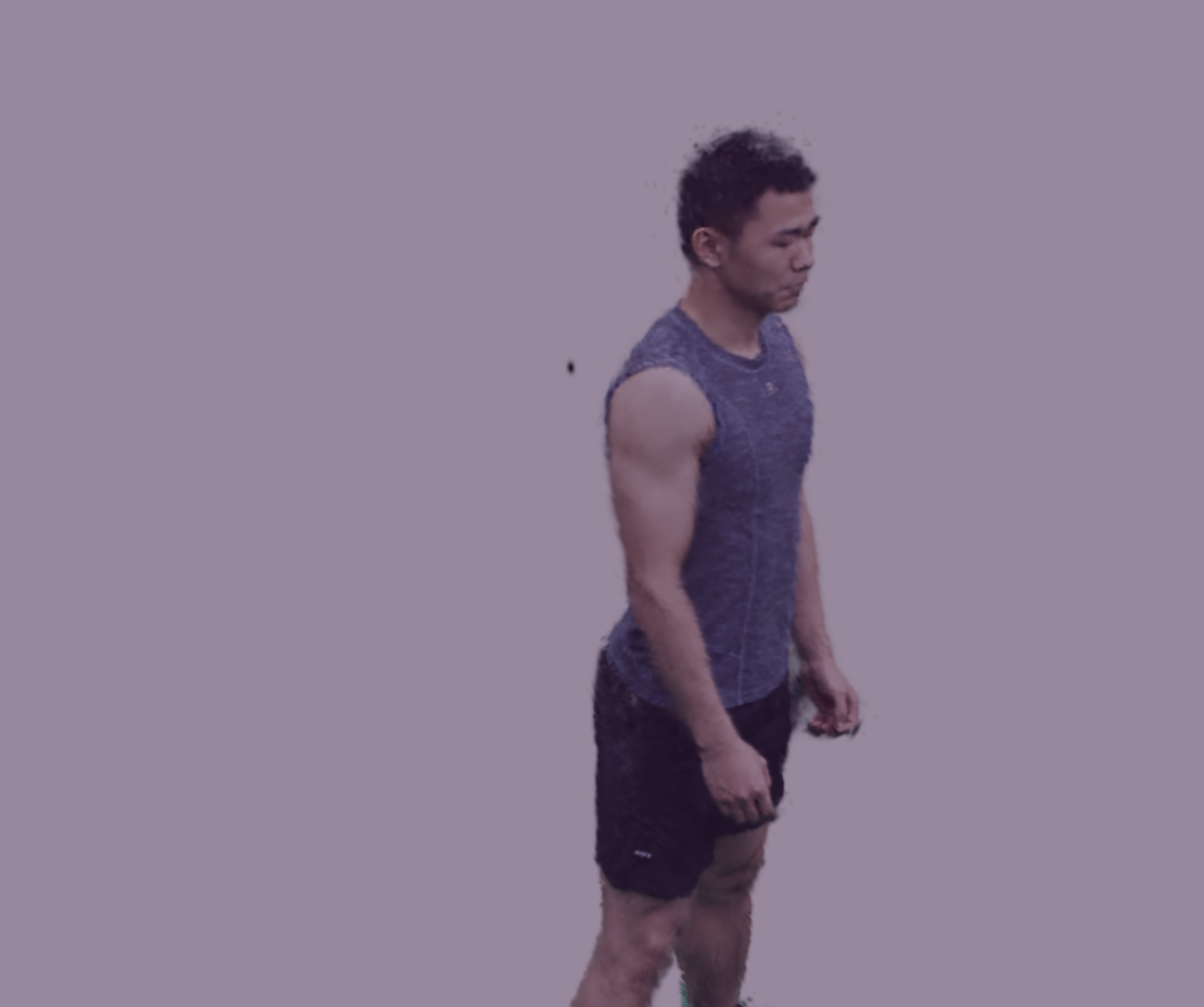}}}{}
\jsubfig{\fcolorbox{red}{red}{\includegraphics[scale=0.063]{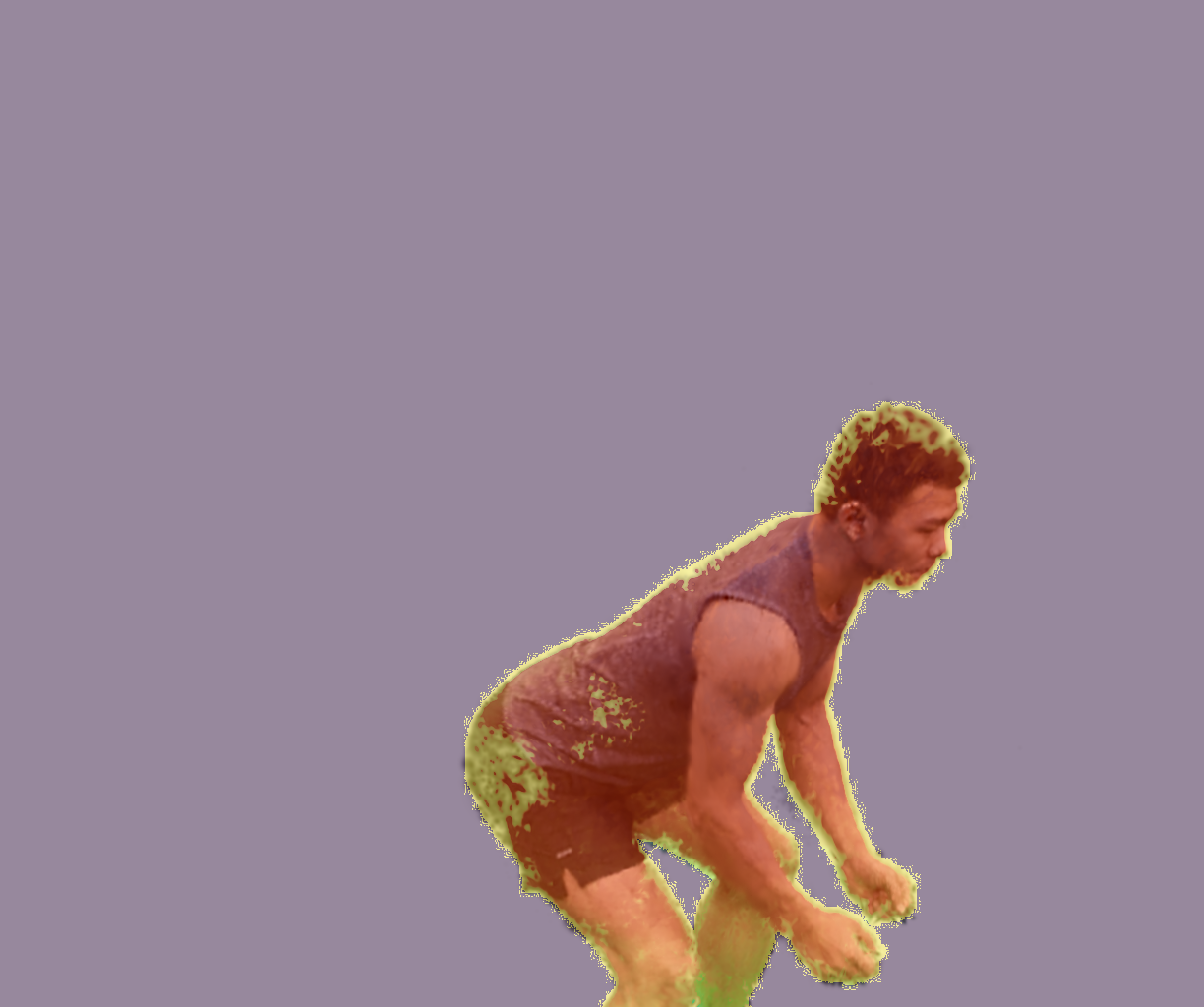}}}{}
\jsubfig{\fcolorbox{red}{red}{\includegraphics[scale=0.063]{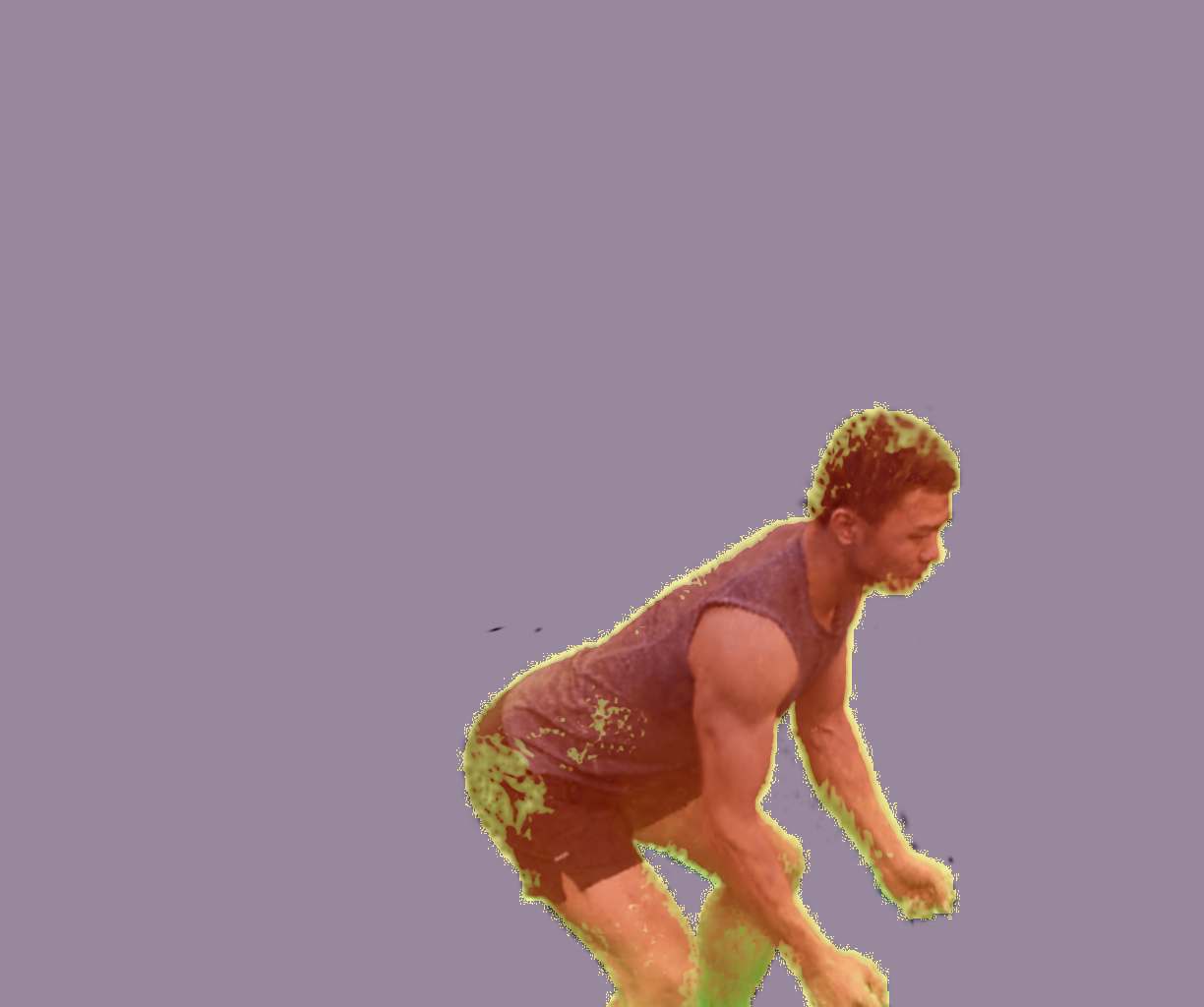}}}{}
\jsubfig{\fcolorbox{red}{red}{\includegraphics[scale=0.063]{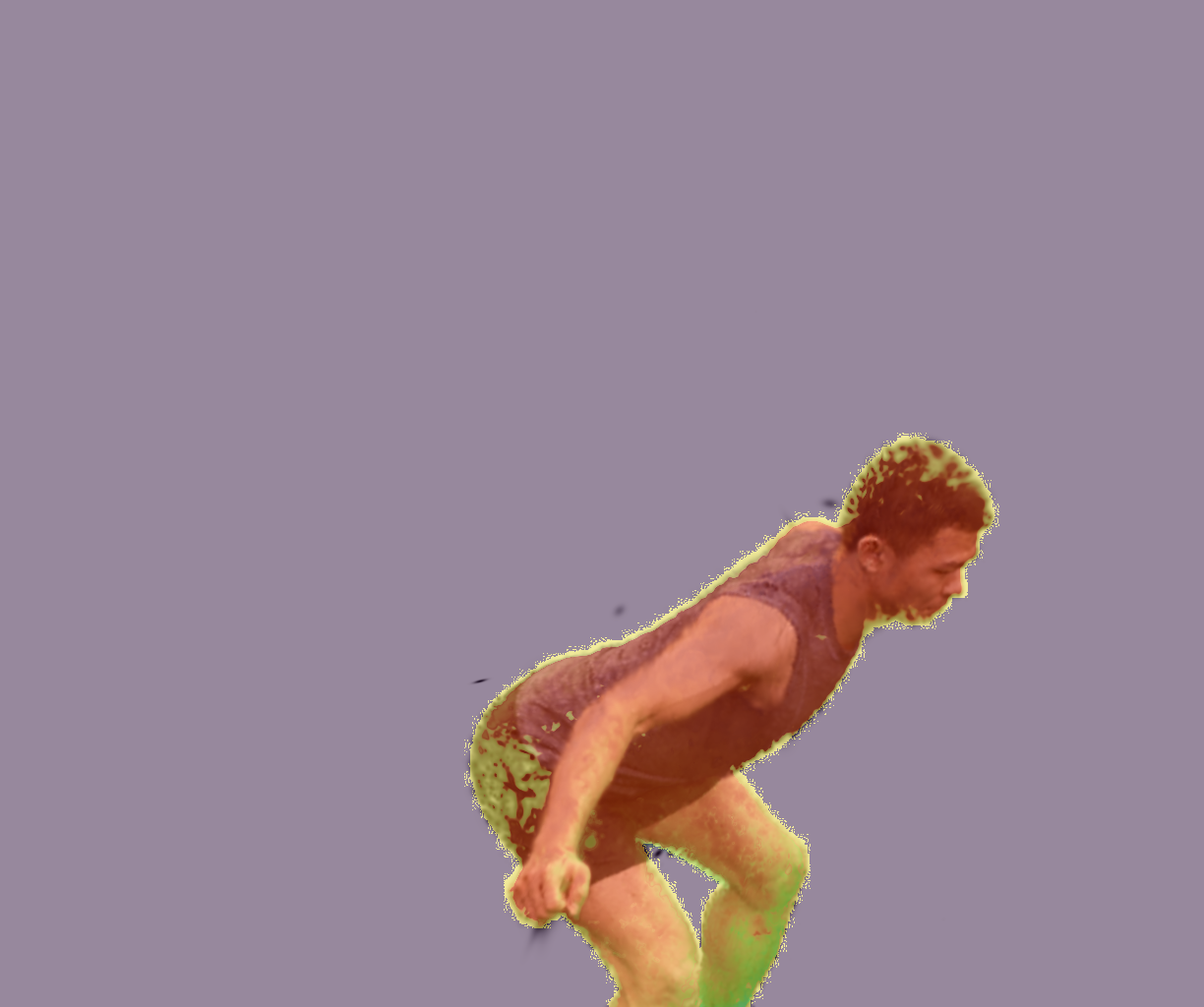}}}{}
\jsubfig{\fcolorbox{red}{red}{\includegraphics[scale=0.063]{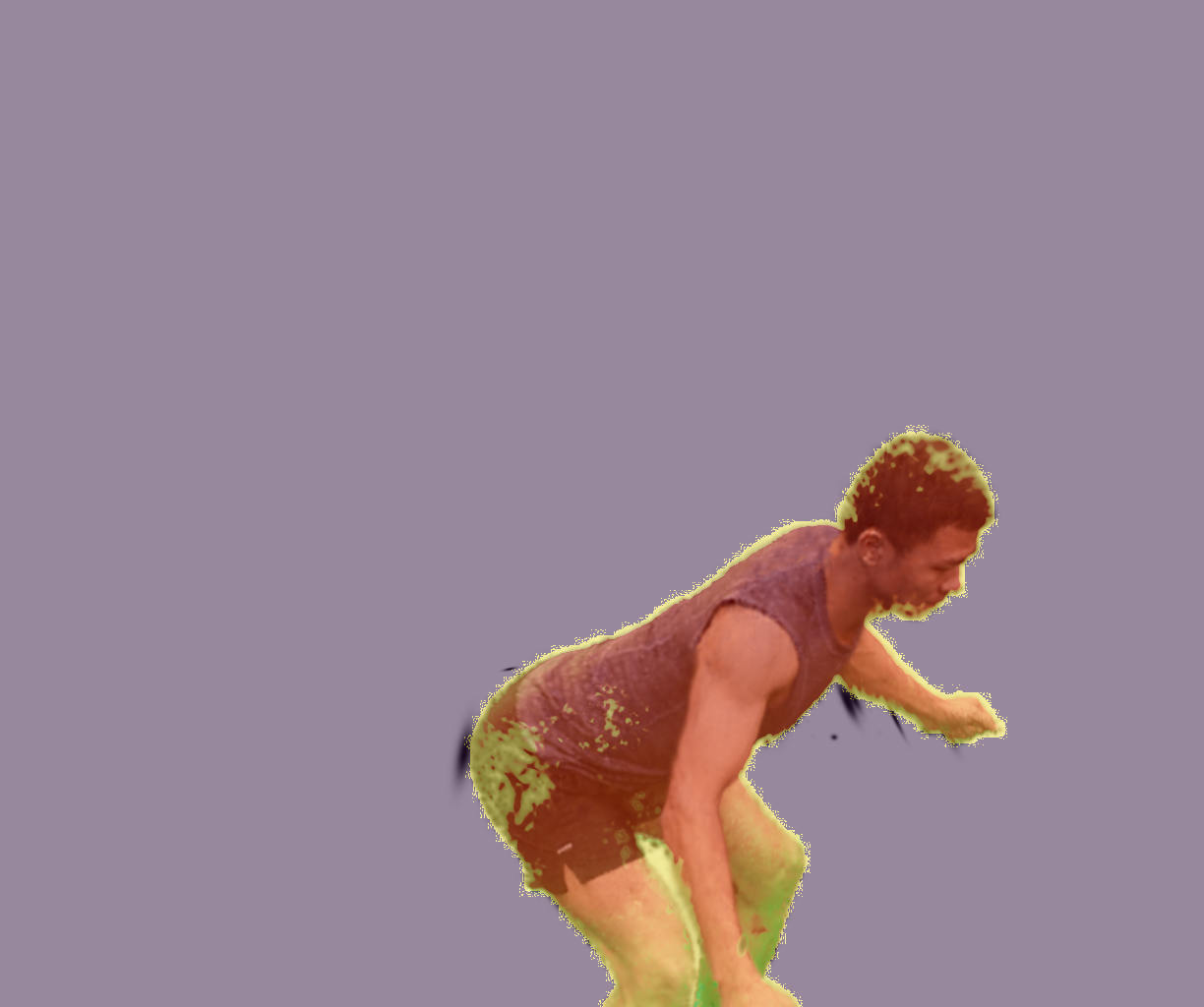}}}{}%
\\
\vspace{4pt}
{ Input query: \emph{A person bending}}
\caption{\textbf{Qualitative results on the Neural Human Rendering \cite{wu2020multi} dataset}, with the input frames depicted on top and the rendered probabilities directly below (frames localized temporally are in red). As illustrated above, our approach succeeds in temporally localizing human actions. We show results over additional scenes in the supplementary material.}  %

\label{fig:results_nhr}
\end{figure*}

\subsection{Spatio-Temporal Highlights}
\label{sec:highlights}
Given a textual query provided by the user, our approach enables to automatically generate videos that highlight the action described by the text. In particular, we demonstrate spatio-temporal highlights such as zooming-in to the action, blurring the background during the detected temporal segment, and visualizing a bullet-time display at the detected temporal \emph{peak}. Several of these highlights are illustrated in Figure \ref{fig:apps} (full temporal sequences are provided in the supplementary material).

To generate these, we first find the viewpoint best depicting the action by locating the viewpoint with the highest average relevance to the prompt. The center of the action in the image is identified using the center of mass of the 2D relevancy map. The 3D location of the action is then calculated using the action’s location  
in the images and the rendered depth of the pixel. The best viewpoint is selected such that the center of the action will be in the middle of the frame.
To detect the temporal segment, we apply dilation over the binary array 
$[s_t > k| \forall t]$ and select the longest positive segment. The temporal peak is selected as the center of the segment. %

\subsection{Spatio-temporal Localization within Multiple Scenes}
\label{sec:selection}
We demonstrate that the spatio-temporal highlights can be extended to a setting where multiple dynamic 3D environments are provided as input. As shown in Figure \ref{fig:scene-selection}, given all 4-LEGS language embedded fields representing the six scenes in the Panoptic Sports dataset, we find the 3D scene that best matches the given text query, localizing the action within this representation in both space and time.

To do so, we define the relevance of each scene to the prompt using
$\max_{t} s_{t}$. In practice, we calculate the maximum over a sampled Gaussian sequence (sampling the temporal domain by 10) to reduce running time. We then select the scene with the highest relevance value. Once selected, we can perform per-scene highlights, as described in the previous section.

\section{Conclusion}

We presented a technique for linking dynamic volumetric representations with textual descriptions, taking a first step towards performing text-driven volumetric video editing.
Generally speaking, in addition to output videos, our framework can potentially support editing for immersive applications, such as augmented and virtual reality platforms, all from captured 2D data. These new possibilities can democratize video editing---advancing the rapidly emerging field of AI generated content from static image generation to dynamic generation that considers temporal and spatial behaviors.

Finally, the ability to link between textual queries and regions within dynamic volumetric representations is important not only for editing videos. We believe it could spark research into many new problems involving dynamic neural representations, such as automatically captioning them or performing volumetric visual question answering, where a model can reason over the full dynamic 3D scene in order to answer a user’s question.

\medskip
\noindent \textbf{Acknowledgments.} This work was partially funded by Google through a TAU-Google grant.

\clearpage

\bibliographystyle{eg-alpha-doi} 
\bibliography{main}

\newcommand{\etalchar}[1]{$^{#1}$}
\begin{thebibliography}{\uppercase{UVDSGS13}}

\bibitem[AHR{\etalchar{*}}23]{attal2023hyperreel}
\textsc{Attal B., Huang J.-B., Richardt C., Zollhoefer M., Kopf J., O'Toole M., Kim C.}:
\newblock {HyperReel}: High-fidelity {6-DoF} video with ray-conditioned sampling.
\newblock In \emph{Conference on Computer Vision and Pattern Recognition (CVPR)} (2023).

\bibitem[AHWS{\etalchar{*}}17]{anne2017localizing}
\textsc{Anne~Hendricks L., Wang O., Shechtman E., Sivic J., Darrell T., Russell B.}:
\newblock Localizing moments in video with natural language.
\newblock In \emph{Proceedings of the IEEE international conference on computer vision} (2017), pp.~5803--5812.

\bibitem[AMKK22]{azuma2022scanqa}
\textsc{Azuma D., Miyanishi T., Kurita S., Kawanabe M.}:
\newblock Scanqa: 3d question answering for spatial scene understanding.
\newblock In \emph{proceedings of the IEEE/CVF conference on computer vision and pattern recognition} (2022), pp.~19129--19139.

\bibitem[APS{\etalchar{*}}14]{arev2014automatic}
\textsc{Arev I., Park H.~S., Sheikh Y., Hodgins J., Shamir A.}:
\newblock Automatic editing of footage from multiple social cameras.
\newblock \emph{ACM Transactions on Graphics (TOG) 33}, 4 (2014), 1--11.

\bibitem[BFO{\etalchar{*}}20]{broxton20}
\textsc{Broxton M., Flynn J., Overbeck R., Erickson D., Hedman P., DuVall M., Dourgarian J., Busch J., Whalen M., Debevec P.}:
\newblock Immersive light field video with a layered mesh representation.
\newblock \emph{SIGGRAPH} (2020).

\bibitem[BTOAF{\etalchar{*}}22]{bar2022text2live}
\textsc{Bar-Tal O., Ofri-Amar D., Fridman R., Kasten Y., Dekel T.}:
\newblock Text2live: Text-driven layered image and video editing.
\newblock In \emph{ECCV} (2022).

\bibitem[CBW{\etalchar{*}}23]{cherti2023reproducible}
\textsc{Cherti M., Beaumont R., Wightman R., Wortsman M., Ilharco G., Gordon C., Schuhmann C., Schmidt L., Jitsev J.}:
\newblock Reproducible scaling laws for contrastive language-image learning.
\newblock In \emph{Proceedings of the IEEE/CVF Conference on Computer Vision and Pattern Recognition} (2023), pp.~2818--2829.

\bibitem[CBWW{\etalchar{*}}22]{cascante2022simvqa}
\textsc{Cascante-Bonilla P., Wu H., Wang L., Feris R.~S., Ordonez V.}:
\newblock Simvqa: Exploring simulated environments for visual question answering.
\newblock In \emph{Proceedings of the IEEE/CVF Conference on Computer Vision and Pattern Recognition} (2022), pp.~5056--5066.

\bibitem[CCN20]{chen2020scanrefer}
\textsc{Chen D.~Z., Chang A.~X., Nie{\ss}ner M.}:
\newblock {ScanRefer: 3D Object Localization in RGB-D Scans using Natural Language}.
\newblock In \emph{Proceedings of the European Conference on Computer Vision (ECCV)} (2020), pp.~202--221.

\bibitem[CFY{\etalchar{*}}23]{cen2023segment}
\textsc{Cen J., Fang J., Yang C., Xie L., Zhang X., Shen W., Tian Q.}:
\newblock Segment any 3d gaussians.
\newblock \emph{arXiv preprint arXiv:2312.00860} (2023).

\bibitem[CGWL23]{chai2023stablevideo}
\textsc{Chai W., Guo X., Wang G., Lu Y.}:
\newblock Stablevideo: Text-driven consistency-aware diffusion video editing.
\newblock In \emph{Proceedings of the IEEE/CVF International Conference on Computer Vision} (2023), pp.~23040--23050.

\bibitem[CHM23]{ceylan2023pix2video}
\textsc{Ceylan D., Huang C.-H.~P., Mitra N.~J.}:
\newblock Pix2video: Video editing using image diffusion.
\newblock In \emph{Proceedings of the IEEE/CVF International Conference on Computer Vision} (2023), pp.~23206--23217.

\bibitem[CKA{\etalchar{*}}21]{cui2021s}
\textsc{Cui Y., Khandelwal A., Artzi Y., Snavely N., Averbuch-Elor H.}:
\newblock Who's waldo? linking people across text and images.
\newblock In \emph{Proceedings of the IEEE/CVF International Conference on Computer Vision} (2021), pp.~1374--1384.

\bibitem[CLW{\etalchar{*}}22]{chen2022ham}
\textsc{Chen J., Luo W., Wei X., Ma L., Zhang W.}:
\newblock Ham: Hierarchical attention model with high performance for 3d visual grounding.
\newblock \emph{arXiv preprint arXiv:2210.12513} (2022).

\bibitem[CMLW19]{chen2019weakly}
\textsc{Chen Z., Ma L., Luo W., Wong K.-Y.~K.}:
\newblock Weakly-supervised spatio-temporally grounding natural sentence in video.
\newblock \emph{arXiv preprint arXiv:1906.02549} (2019).

\bibitem[CWTW17]{cui2017time}
\textsc{Cui Z., Wang O., Tan P., Wang J.}:
\newblock Time slice video synthesis by robust video alignment.
\newblock \emph{ACM Transactions on Graphics (TOG) 36}, 4 (2017), 1--10.

\bibitem[CXI{\etalchar{*}}23]{chen2023open}
\textsc{Chen B., Xia F., Ichter B., Rao K., Gopalakrishnan K., Ryoo M.~S., Stone A., Kappler D.}:
\newblock Open-vocabulary queryable scene representations for real world planning.
\newblock In \emph{2023 IEEE International Conference on Robotics and Automation (ICRA)} (2023), IEEE, pp.~11509--11522.

\bibitem[CZKD22]{corona2022voxel}
\textsc{Corona R., Zhu S., Klein D., Darrell T.}:
\newblock Voxel-informed language grounding.
\newblock \emph{arXiv preprint arXiv:2205.09710} (2022).

\bibitem[DAB{\etalchar{*}}24]{dudai2024halo}
\textsc{Dudai C., Alper M., Bezalel H., Hanocka R., Lang I., Averbuch-Elor H.}:
\newblock Halo-nerf: Learning geometry-guided semantics for exploring unconstrained photo collections.
\newblock In \emph{Computer Graphics Forum} (2024), Wiley Online Library, p.~e15006.

\bibitem[DFBD19]{duvall2019compositing}
\textsc{DuVall M., Flynn J., Broxton M., Debevec P.}:
\newblock Compositing light field video using multiplane images.
\newblock In \emph{ACM SIGGRAPH 2019 Posters} (2019).

\bibitem[GBTBD23]{geyer2023tokenflow}
\textsc{Geyer M., Bar-Tal O., Bagon S., Dekel T.}:
\newblock Tokenflow: Consistent diffusion features for consistent video editing.
\newblock \emph{arXiv preprint arXiv:2307.10373} (2023).

\bibitem[GFH{\etalchar{*}}24]{gu2024context}
\textsc{Gu X., Fan H., Huang Y., Luo T., Zhang L.}:
\newblock Context-guided spatio-temporal video grounding.
\newblock In \emph{Proceedings of the IEEE/CVF Conference on Computer Vision and Pattern Recognition} (2024), pp.~18330--18339.

\bibitem[GKR{\etalchar{*}}18]{gordon2018iqa}
\textsc{Gordon D., Kembhavi A., Rastegari M., Redmon J., Fox D., Farhadi A.}:
\newblock Iqa: Visual question answering in interactive environments.
\newblock In \emph{Proceedings of the IEEE conference on computer vision and pattern recognition} (2018), pp.~4089--4098.

\bibitem[GPZH23]{ge2023expressive}
\textsc{Ge S., Park T., Zhu J.-Y., Huang J.-B.}:
\newblock Expressive text-to-image generation with rich text.
\newblock In \emph{Proceedings of the IEEE/CVF International Conference on Computer Vision} (2023), pp.~7545--7556.

\bibitem[GSKH21]{gao2021Dynamic}
\textsc{Gao C., Saraf A., Kopf J., Huang J.-B.}:
\newblock Dynamic view synthesis from dynamic monocular video.
\newblock In \emph{Proceedings of the IEEE/CVF International Conference on Computer Vision} (2021), pp.~5712--5721.

\bibitem[GSYN17]{gao2017tall}
\textsc{Gao J., Sun C., Yang Z., Nevatia R.}:
\newblock Tall: Temporal activity localization via language query.
\newblock In \emph{Proceedings of the IEEE international conference on computer vision} (2017), pp.~5267--5275.

\bibitem[HCJW22]{huang2022multi}
\textsc{Huang S., Chen Y., Jia J., Wang L.}:
\newblock Multi-view transformer for 3d visual grounding.
\newblock In \emph{CVPR} (2022).

\bibitem[HJA20]{ho2020denoising}
\textsc{Ho J., Jain A., Abbeel P.}:
\newblock Denoising diffusion probabilistic models.
\newblock \emph{Advances in neural information processing systems 33} (2020), 6840--6851.

\bibitem[HMT{\etalchar{*}}22]{hertz2022prompt}
\textsc{Hertz A., Mokady R., Tenenbaum J., Aberman K., Pritch Y., Cohen-Or D.}:
\newblock Prompt-to-prompt image editing with cross attention control.
\newblock \emph{arXiv preprint arXiv:2208.01626} (2022).

\bibitem[HRA{\etalchar{*}}17]{hu2017modeling}
\textsc{Hu R., Rohrbach M., Andreas J., Darrell T., Saenko K.}:
\newblock Modeling relationships in referential expressions with compositional modular networks.
\newblock In \emph{Proceedings of the IEEE conference on computer vision and pattern recognition} (2017), pp.~1115--1124.

\bibitem[HWF{\etalchar{*}}24]{hu2024semantic}
\textsc{Hu X., Wang Y., Fan L., Fan J., Peng J., Lei Z., Li Q., Zhang Z.}:
\newblock Semantic anything in 3d gaussians.
\newblock \emph{arXiv preprint arXiv:2401.17857} (2024).

\bibitem[JLF{\etalchar{*}}16]{jie2016scale}
\textsc{Jie Z., Liang X., Feng J., Lu W.~F., Tay E. H.~F., Yan S.}:
\newblock Scale-aware pixelwise object proposal networks.
\newblock \emph{IEEE Transactions on Image Processing 25}, 10 (2016), 4525--4539.

\bibitem[JLT{\etalchar{*}}15]{joo2015panoptic}
\textsc{Joo H., Liu H., Tan L., Gui L., Nabbe B., Matthews I., Kanade T., Nobuhara S., Sheikh Y.}:
\newblock Panoptic studio: A massively multiview system for social motion capture.
\newblock In \emph{Proceedings of the IEEE international conference on computer vision} (2015), pp.~3334--3342.

\bibitem[KKG{\etalchar{*}}23]{kerr2023lerf}
\textsc{Kerr J., Kim C.~M., Goldberg K., Kanazawa A., Tancik M.}:
\newblock Lerf: Language embedded radiance fields.
\newblock \emph{arXiv preprint arXiv:2303.09553} (2023).

\bibitem[KKLD23]{kerbl20233d}
\textsc{Kerbl B., Kopanas G., Leimk{\"u}hler T., Drettakis G.}:
\newblock 3d gaussian splatting for real-time radiance field rendering.
\newblock \emph{ACM Transactions on Graphics 42}, 4 (2023), 1--14.

\bibitem[KMR{\etalchar{*}}23]{kirillov2023segany}
\textsc{Kirillov A., Mintun E., Ravi N., Mao H., Rolland C., Gustafson L., Xiao T., Whitehead S., Berg A.~C., Lo W.-Y., Doll{\'a}r P., Girshick R.}:
\newblock Segment anything.
\newblock \emph{arXiv:2304.02643} (2023).

\bibitem[KOWD21]{kasten2021layered}
\textsc{Kasten Y., Ofri D., Wang O., Dekel T.}:
\newblock Layered neural atlases for consistent video editing.
\newblock \emph{ACM Transactions on Graphics (TOG) 40}, 6 (2021), 1--12.

\bibitem[KWK{\etalchar{*}}24]{kim2024garfield}
\textsc{Kim C.~M., Wu M., Kerr J., Goldberg K., Tancik M., Kanazawa A.}:
\newblock Garfield: Group anything with radiance fields.
\newblock \emph{arXiv preprint arXiv:2401.09419} (2024).

\bibitem[LCD{\etalchar{*}}20]{lu2020layered}
\textsc{Lu E., Cole F., Dekel T., Xie W., Zisserman A., Salesin D., Freeman W., Rubinstein M.}:
\newblock Layered neural rendering for retiming people in video.
\newblock \emph{ACM Transactions on Graphics(TOG) 39}, 6 (2020).

\bibitem[LCLX23]{li2023spacetime}
\textsc{Li Z., Chen Z., Li Z., Xu Y.}:
\newblock Spacetime gaussian feature splatting for real-time dynamic view synthesis.
\newblock \emph{arXiv preprint arXiv:2312.16812} (2023).

\bibitem[LDTA17]{leake2017computational}
\textsc{Leake M., Davis A., Truong A., Agrawala M.}:
\newblock Computational video editing for dialogue-driven scenes.
\newblock \emph{ACM Transactions on Graphics (TOG) 36}, 4 (2017), 130--1.

\bibitem[LDZY24]{lin2024gaussian}
\textsc{Lin Y., Dai Z., Zhu S., Yao Y.}:
\newblock Gaussian-flow: 4d reconstruction with dynamic 3d gaussian particle.
\newblock In \emph{Proceedings of the IEEE/CVF Conference on Computer Vision and Pattern Recognition} (2024), pp.~21136--21145.

\bibitem[LGH{\etalchar{*}}24]{lu20243d}
\textsc{Lu Z., Guo X., Hui L., Chen T., Yang M., Tang X., Zhu F., Dai Y.}:
\newblock 3d geometry-aware deformable gaussian splatting for dynamic view synthesis.
\newblock \emph{arXiv preprint arXiv:2404.06270} (2024).

\bibitem[LILB24]{labe2024dgd}
\textsc{Labe I., Issachar N., Lang I., Benaim S.}:
\newblock Dgd: Dynamic 3d gaussians distillation.
\newblock In \emph{European Conference on Computer Vision} (2024), Springer, pp.~361--378.

\bibitem[LKLR23]{luiten2023dynamic}
\textsc{Luiten J., Kopanas G., Leibe B., Ramanan D.}:
\newblock Dynamic 3d gaussians: Tracking by persistent dynamic view synthesis.
\newblock \emph{arXiv preprint arXiv:2308.09713} (2023).

\bibitem[LSZ{\etalchar{*}}22]{li2022neural}
\textsc{Li T., Slavcheva M., Zollhoefer M., Green S., Lassner C., Kim C., Schmidt T., Lovegrove S., Goesele M., Newcombe R., et~al.}:
\newblock Neural 3d video synthesis from multi-view video.
\newblock In \emph{Proceedings of the IEEE/CVF Conference on Computer Vision and Pattern Recognition} (2022), pp.~5521--5531.

\bibitem[LXJ{\etalchar{*}}22]{luo2022artemis}
\textsc{Luo H., Xu T., Jiang Y., Zhou C., Qiu Q., Zhang Y., Yang W., Xu L., Yu J.}:
\newblock Artemis: Articulated neural pets with appearance and motion synthesis.
\newblock \emph{arXiv preprint arXiv:2202.05628} (2022).

\bibitem[LZL{\etalchar{*}}23]{liu2023video}
\textsc{Liu S., Zhang Y., Li W., Lin Z., Jia J.}:
\newblock Video-p2p: Video editing with cross-attention control.
\newblock \emph{arXiv preprint arXiv:2303.04761} (2023).

\bibitem[MST{\etalchar{*}}21]{mildenhall2021nerf}
\textsc{Mildenhall B., Srinivasan P.~P., Tancik M., Barron J.~T., Ramamoorthi R., Ng R.}:
\newblock Nerf: Representing scenes as neural radiance fields for view synthesis.
\newblock \emph{Communications of the ACM 65}, 1 (2021), 99--106.

\bibitem[NDR{\etalchar{*}}21]{nichol2021glide}
\textsc{Nichol A., Dhariwal P., Ramesh A., Shyam P., Mishkin P., McGrew B., Sutskever I., Chen M.}:
\newblock Glide: Towards photorealistic image generation and editing with text-guided diffusion models.
\newblock \emph{arXiv preprint arXiv:2112.10741} (2021).

\bibitem[PGA{\etalchar{*}}23]{patashnik2023localizing}
\textsc{Patashnik O., Garibi D., Azuri I., Averbuch-Elor H., Cohen-Or D.}:
\newblock Localizing object-level shape variations with text-to-image diffusion models.
\newblock In \emph{Proceedings of the IEEE/CVF International Conference on Computer Vision} (2023), pp.~23051--23061.

\bibitem[QLZ{\etalchar{*}}23]{qin2023langsplat}
\textsc{Qin M., Li W., Zhou J., Wang H., Pfister H.}:
\newblock Langsplat: 3d language gaussian splatting.
\newblock \emph{arXiv preprint arXiv:2312.16084} (2023).

\bibitem[RHGS15]{ren2015faster}
\textsc{Ren S., He K., Girshick R., Sun J.}:
\newblock Faster r-cnn: Towards real-time object detection with region proposal networks.
\newblock \emph{Advances in neural information processing systems 28} (2015).

\bibitem[RKH{\etalchar{*}}21]{radford2021learning}
\textsc{Radford A., Kim J.~W., Hallacy C., Ramesh A., Goh G., Agarwal S., Sastry G., Askell A., Mishkin P., Clark J., et~al.}:
\newblock Learning transferable visual models from natural language supervision.
\newblock In \emph{International conference on machine learning} (2021), PMLR, pp.~8748--8763.

\bibitem[RLD22]{rozenberszki2022language}
\textsc{Rozenberszki D., Litany O., Dai A.}:
\newblock {Language-Grounded Indoor 3D Semantic Segmentation in the Wild}.
\newblock In \emph{Proceedings of the European Conference on Computer Vision (ECCV)} (2022), pp.~125--141.

\bibitem[SJL{\etalchar{*}}24]{sun20243dgstream}
\textsc{Sun J., Jiao H., Li G., Zhang Z., Zhao L., Xing W.}:
\newblock 3dgstream: On-the-fly training of 3d gaussians for efficient streaming of photo-realistic free-viewpoint videos.
\newblock In \emph{Proceedings of the IEEE/CVF Conference on Computer Vision and Pattern Recognition} (2024), pp.~20675--20685.

\bibitem[SPP{\etalchar{*}}22]{shafiullah2022clip}
\textsc{Shafiullah N. M.~M., Paxton C., Pinto L., Chintala S., Szlam A.}:
\newblock Clip-fields: Weakly supervised semantic fields for robotic memory.
\newblock \emph{arXiv preprint arXiv:2210.05663} (2022).

\bibitem[SWDG23]{shi2023language}
\textsc{Shi J.-C., Wang M., Duan H.-B., Guan S.-H.}:
\newblock Language embedded 3d gaussians for open-vocabulary scene understanding.
\newblock \emph{arXiv preprint arXiv:2311.18482} (2023).

\bibitem[TLL{\etalchar{*}}21]{tang2021human}
\textsc{Tang Z., Liao Y., Liu S., Li G., Jin X., Jiang H., Yu Q., Xu D.}:
\newblock Human-centric spatio-temporal video grounding with visual transformers.
\newblock \emph{IEEE Transactions on Circuits and Systems for Video Technology 32}, 12 (2021), 8238--8249.

\bibitem[TSB{\etalchar{*}}22]{thomason2022language}
\textsc{Thomason J., Shridhar M., Bisk Y., Paxton C., Zettlemoyer L.}:
\newblock Language grounding with 3d objects.
\newblock In \emph{Conference on Robot Learning} (2022), PMLR, pp.~1691--1701.

\bibitem[UVDSGS13]{uijlings2013selective}
\textsc{Uijlings J.~R., Van De~Sande K.~E., Gevers T., Smeulders A.~W.}:
\newblock Selective search for object recognition.
\newblock \emph{International journal of computer vision 104} (2013), 154--171.

\bibitem[VDVG18]{vasudevan2018object}
\textsc{Vasudevan A.~B., Dai D., Van~Gool L.}:
\newblock Object referring in videos with language and human gaze.
\newblock In \emph{Proceedings of the IEEE Conference on Computer Vision and Pattern Recognition} (2018), pp.~4129--4138.

\bibitem[WCC{\etalchar{*}}23]{wang2023tracking}
\textsc{Wang Q., Chang Y.-Y., Cai R., Li Z., Hariharan B., Holynski A., Snavely N.}:
\newblock Tracking everything everywhere all at once.
\newblock In \emph{Proceedings of the IEEE/CVF International Conference on Computer Vision} (2023), pp.~19795--19806.

\bibitem[WGW{\etalchar{*}}23]{wu2023tune}
\textsc{Wu J.~Z., Ge Y., Wang X., Lei S.~W., Gu Y., Shi Y., Hsu W., Shan Y., Qie X., Shou M.~Z.}:
\newblock Tune-a-video: One-shot tuning of image diffusion models for text-to-video generation.
\newblock In \emph{Proceedings of the IEEE/CVF International Conference on Computer Vision} (2023), pp.~7623--7633.

\bibitem[WHL{\etalchar{*}}23]{wang2023internvid}
\textsc{Wang Y., He Y., Li Y., Li K., Yu J., Ma X., Li X., Chen G., Chen X., Wang Y., et~al.}:
\newblock Internvid: A large-scale video-text dataset for multimodal understanding and generation.
\newblock \emph{arXiv preprint arXiv:2307.06942} (2023).

\bibitem[WWHY20]{wu2020multi}
\textsc{Wu M., Wang Y., Hu Q., Yu J.}:
\newblock Multi-view neural human rendering.
\newblock In \emph{Proceedings of the IEEE/CVF Conference on Computer Vision and Pattern Recognition} (2020), pp.~1682--1691.

\bibitem[WYF{\etalchar{*}}23]{wu20234d}
\textsc{Wu G., Yi T., Fang J., Xie L., Zhang X., Wei W., Liu W., Tian Q., Wang X.}:
\newblock 4d gaussian splatting for real-time dynamic scene rendering.
\newblock \emph{arXiv preprint arXiv:2310.08528} (2023).

\bibitem[YDYK23]{ye2023gaussian}
\textsc{Ye M., Danelljan M., Yu F., Ke L.}:
\newblock Gaussian grouping: Segment and edit anything in 3d scenes.
\newblock \emph{arXiv preprint arXiv:2312.00732} (2023).

\bibitem[YGL{\etalchar{*}}23]{yang2023track}
\textsc{Yang J., Gao M., Li Z., Gao S., Wang F., Zheng F.}:
\newblock Track anything: Segment anything meets videos, 2023.
\newblock \href {http://arxiv.org/abs/2304.11968} {\path{arXiv:2304.11968}}.

\bibitem[YGZ{\etalchar{*}}23]{yang2023deformable}
\textsc{Yang Z., Gao X., Zhou W., Jiao S., Zhang Y., Jin X.}:
\newblock Deformable 3d gaussians for high-fidelity monocular dynamic scene reconstruction.
\newblock \emph{arXiv preprint arXiv:2309.13101} (2023).

\bibitem[YMS{\etalchar{*}}22]{yang2022tubedetr}
\textsc{Yang A., Miech A., Sivic J., Laptev I., Schmid C.}:
\newblock Tubedetr: Spatio-temporal video grounding with transformers.
\newblock In \emph{CVPR} (2022).

\bibitem[YSUH17]{yamaguchi2017spatio}
\textsc{Yamaguchi M., Saito K., Ushiku Y., Harada T.}:
\newblock Spatio-temporal person retrieval via natural language queries.
\newblock In \emph{Proceedings of the IEEE international conference on computer vision} (2017), pp.~1453--1462.

\bibitem[ZCJ{\etalchar{*}}23]{zhou2023feature}
\textsc{Zhou S., Chang H., Jiang S., Fan Z., Zhu Z., Xu D., Chari P., You S., Wang Z., Kadambi A.}:
\newblock Feature 3dgs: Supercharging 3d gaussian splatting to enable distilled feature fields.
\newblock \emph{arXiv preprint arXiv:2312.03203} (2023).

\bibitem[ZD14]{zitnick2014edge}
\textsc{Zitnick C.~L., Doll{\'a}r P.}:
\newblock Edge boxes: Locating object proposals from edges.
\newblock In \emph{Computer Vision--ECCV 2014: 13th European Conference, Zurich, Switzerland, September 6-12, 2014, Proceedings, Part V 13} (2014), Springer, pp.~391--405.

\bibitem[ZLC18]{zhou2018weakly}
\textsc{Zhou L., Louis N., Corso J.~J.}:
\newblock Weakly-supervised video object grounding from text by loss weighting and object interaction.
\newblock \emph{arXiv preprint arXiv:1805.02834} (2018).

\bibitem[ZZZ{\etalchar{*}}20a]{zhang2020does}
\textsc{Zhang Z., Zhao Z., Zhao Y., Wang Q., Liu H., Gao L.}:
\newblock Where does it exist: Spatio-temporal video grounding for multi-form sentences.
\newblock In \emph{Proceedings of the IEEE/CVF Conference on Computer Vision and Pattern Recognition} (2020), pp.~10668--10677.

\bibitem[ZZZ{\etalchar{*}}20b]{Zhang_2020_CVPR}
\textsc{Zhang Z., Zhao Z., Zhao Y., Wang Q., Liu H., Gao L.}:
\newblock Where does it exist: Spatio-temporal video grounding for multi-form sentences.
\newblock In \emph{Proceedings of the IEEE/CVF Conference on Computer Vision and Pattern Recognition (CVPR)} (June 2020).

\end{thebibliography}

\clearpage
\medskip \noindent \textbf{\methodname{} Supplementary Material
}

\setcounter{section}{0}

\section{Additional Details}
\label{sec:details}

\subsection{\methodname{} Implementation Details}
\label{sec:imp_details}

For each scene we first use Dynamic 3D Gaussians \cite{luiten2023dynamic} to pretrain our dynamic 3D represntation, using the default hyper parameters that can be found \href{https://github.com/JonathonLuiten/Dynamic3DGaussians/blob/main/train.py}{here}. For our spatio-temporal feature extraction we use the ViCLIP-L-14 \cite{wang2023internvid} model. For the Panoptic Sports scenes we build an image pyramid with scales between 0.15 and 0.6 with 5 steps. For the DFA scenes we build a pyramid with scales between 0.4 and 0.95 with 5 steps. For all scenes we use a temporal window of 8 frames centered around our input image. 

Our autoencoder is implemented by an MLP, which compresses the 768-dimensional ViCLIP features into 128-dimensional latent features. We use an Adam optimizer with a learning rate of 0.0007 and train with a batch size of 1  over all input training views and timesteps ($\sim5$ hrs for a Panoptic Sports scene).

We optimize our \methodname{} while fixing all parameters of the pretrained Dynamic 3D Gaussians. For each Gaussian, we compute attention with its 20 closest neighbors. We train each timestep for 2000 iterations which takes $\sim 5$ minutes per timestep. Our inference includes rendering the language feature, decoding and computing the relevancy score, all this is done at 10 FPS.
All experiments and training are conducted on a single RTX A5000 GPU (24GB VRAM).

\subsection{Temporal Localization Score Details}\label{sec:temp_loc_details}

As described in the main paper, given a decoded spatio-temporal feature $\features{} = \Psi(l)$ extracted from our representation, and an encoded text query $\phi_{\text{Q}}$, we compute a relevancy score $\text{R}(\features{}, \phi_{\text{Q}})$.

To temporally localize this text query, we calculate the relevancy score at each Gaussian $G$ and time step $t$, yielding the following set of all Gaussians: 
\[
\textit{All} = \left\{ \text{R}(\phi^{t}_{g}, \phi_{\text{Q}}) \mid g \in G, t \in T \right\}.
\]
From this set, we extract all scores greater than 0.5, indicating that the text query is more likely to appear at this Gaussian than any canonical phrase. This produces the set of ``relevant" Gaussians:
\[
\textit{Rel} = \left\{ \text{R}(\phi^{t}_{g}, \phi_{\text{Q}}) > 0.5 \mid g \in G, t \in T \right\}.
\]

Next, we compute the average relevancy score of $\textit{Rel}$, denoted as $\textit{Rel}_{\text{avg}}$. We use this average score for extracting:
\[
\textit{Rel}_t = \left\{ \text{R}(\phi^{t}_{g}, \phi_{\text{Q}}) > \textit{Rel}_{\text{avg}} \mid g \in G \right\}.
\]
Finally, for a given timestep $t$, we calculate the localization score $s_t$ using the formula
\[
s_t = \frac{|\textit{Rel}_t|}{\sum_{t' \in T} |\textit{Rel}_{t'}|}.
\]
This provides the fraction of Gaussians at each timestep with an \emph{above-average} relevancy score.

\subsection{Comparisons and Ablations}
\label{sec:prior}
Below we provide all the details needed to reproduce the comparisons and ablations shown in the paper. 

\subsubsection*{TubeDeTR}
We use the \href{https://github.com/antoyang/TubeDETR/}{code provided by the authors} with the 
original videos and corresponding text prompts.
We use the model pretrained on HC-STVG2 with k=2 and res 224.

\subsubsection*{CGSTVG}
We use the \href{https://github.com/HengLan/CGSTVG}{code provided by the authors} with the 
original videos and corresponding text prompts.
We use the model pretrained on HC-STVG2.

\subsubsection*{LangSplat}
We use the \href{https://github.com/minghanqin/LangSplat}{code provided by the authors} and the input parameters used are the default parameters. We train a LangSplat for 20 uniformally sampled timesteps in all 6 scenes. We provide their method with all input views at that timestep.

\subsubsection*{CLIP based ablations}
For our CLIP based ablations, $\text{Static}_\text{CLIP}$ and $\text{AVG}_\text{CLIP}$, we use the same training parameters as for our method. For the feature extraction we use use the OpenClip \cite{cherti2023reproducible} ViTB/16 model.
For the $\text{AVG}_\text{CLIP}$ ablation we use a temporal window of 8 frames centered around our input image, which is also the temporal window we use with ViCLIP.

\subsection{The \datasetname{} Benchmark}
\label{sec:dataset}

To produce ground-truth segmentation maps we used manual labelling combined with pretrained models to preform semi-automated segmentation and tracking through the video. For each camera view and text query, we first generate a masked image for each starting frame of a dynamic annotated interval using Segment-Anything  (SAM)~\cite{kirillov2023segany}. Then,  we expanded the segmentation masks to additional frames up to the whole annotated interval in time with Track-Anything ~\cite{yang2023track}, occasionally using additional manual segmentation with SAM to improve the spatial grounding annotation. 
We note that the Panoptic Sports dataset includes segmentation maps, however these are not specific to an action and are not sufficient for our task.
We chose 5 different views to annotate through all  scenes (cameras 0,6,13,19,23). The temporal ground truth was manually identified using multiple view points to allow consistency in spite of occlusions.

We provide a list of the queries annotated per scene in Table \ref{tab:comparisons-supp} (see the \emph{Query} column).

\section{Additional Ablations and Comparisons}
\subsection{Evaluation Breakdown}
\label{sec:ab}

We report performance per scene and query in Table \ref{tab:comparisons-supp} for our method, as well as for CGSTVG\cite{gu2024context} (which outperforms TubeDETR, as shown in the main paper). As illustrated in the table, our method outperforms CGSTVG over almost all scenes and metrics. We can also observe that the variance amongst different views for CGTSVG is somewhat scene-dependent. For example, there's high variance for the query \emph{A person juggling balls}, because the action is less visible from some viewpoints.

\subsection{Bounding Box Comparison to 2D methods}
\label{sec:bbox-comp}

\begin{figure*}
  \rotatebox{90}{\hspace{-23pt}TubeDeTR \whitetxt{(}}
    \jsubfig{\includegraphics[height=1.21cm]{images/comparisons/tubedetr/6/a_person_playing_tennis_cam6_frame0_no.jpg} \hfill \includegraphics[height=1.21cm]{images/comparisons/tubedetr/6/a_person_playing_tennis_cam6_frame29.jpg}
  \includegraphics[height=1.21cm]{images/comparisons/tubedetr/6/a_person_playing_tennis_cam6_frame89_no.jpg}
    \includegraphics[height=1.21cm]{images/comparisons/tubedetr/6/a_person_playing_tennis_cam6_frame149_no.jpg}
  }{ }
  \hfill
  \jsubfig{\includegraphics[height=1.21cm]{images/comparisons/tubedetr/0/a_ball_flying_cam0_frame0_no.jpg} \hfill \includegraphics[height=1.21cm]{images/comparisons/tubedetr/0/a_ball_flying_cam0_frame14.jpg}
  \includegraphics[height=1.21cm]{images/comparisons/tubedetr/0/a_ball_flying_cam0_frame89_no.jpg}
    \includegraphics[height=1.21cm]{images/comparisons/tubedetr/0/a_ball_flying_cam0_frame149_no.jpg}
  }{ } \\
  \rotatebox{90}{\whitetxt{()}}
    \jsubfig{\includegraphics[height=1.21cm]{images/comparisons/tubedetr/19/a_person_playing_tennis_cam19_frame0_no.jpg} \hfill \includegraphics[height=1.21cm]{images/comparisons/tubedetr/19/a_person_playing_tennis_cam19_frame29.jpg}
  \includegraphics[height=1.21cm]{images/comparisons/tubedetr/19/a_person_playing_tennis_cam19_frame89_no.jpg}
    \includegraphics[height=1.21cm]{images/comparisons/tubedetr/19/a_person_playing_tennis_cam19_frame149_no.jpg}
  }{ }
  \hfill
  \jsubfig{\includegraphics[height=1.21cm]{images/comparisons/tubedetr/23/a_ball_flying_cam23_frame0_no.jpg} \hfill \includegraphics[height=1.21cm]{images/comparisons/tubedetr/23/a_ball_flying_cam23_frame14_no.jpg}
  \includegraphics[height=1.21cm]{images/comparisons/tubedetr/23/a_ball_flying_cam23_frame89.jpg}
    \includegraphics[height=1.21cm]{images/comparisons/tubedetr/23/a_ball_flying_cam23_frame149_no.jpg}
  }{ }\\
   \rotatebox{90}{\hspace{-12pt}CGSTVG\whitetxt{g}}  \jsubfig{\includegraphics[height=1.21cm]{images/comparisons/tubedetr/6/a_person_playing_tennis_cam6_frame0_no.jpg} \hfill \includegraphics[height=1.182cm]{images/comparisons/cgstvg/6/tennis/29_y.png}
  \includegraphics[height=1.21cm]{images/comparisons/tubedetr/6/a_person_playing_tennis_cam6_frame89_no.jpg}
    \includegraphics[height=1.21cm]{images/comparisons/tubedetr/6/a_person_playing_tennis_cam6_frame149_no.jpg}
  }{ }
  \hfill
   \jsubfig{\includegraphics[height=1.182cm]{images/comparisons/cgstvg/0/bball/000000.jpg} \hfill \includegraphics[height=1.182cm]{images/comparisons/cgstvg/0/bball/000014.jpg}
  \includegraphics[height=1.182cm]{images/comparisons/cgstvg/0/bball/000089.jpg}
    \includegraphics[height=1.182cm]{images/comparisons/cgstvg/0/bball/149_y.png}
  }{ } \\
   \rotatebox{90}{\whitetxt{Tubeg}}
   \jsubfig{\includegraphics[height=1.21cm]{images/comparisons/tubedetr/19/a_person_playing_tennis_cam19_frame0_no.jpg} \hfill \includegraphics[height=1.182cm]{images/comparisons/cgstvg/19/tennis/29_y.png}
  \includegraphics[height=1.21cm]{images/comparisons/tubedetr/19/a_person_playing_tennis_cam19_frame89_no.jpg}
    \includegraphics[height=1.21cm]{images/comparisons/tubedetr/19/a_person_playing_tennis_cam19_frame149_no.jpg}
  }{ }
  \hfill
  \jsubfig{\includegraphics[height=1.182cm]{images/comparisons/cgstvg/23/bball/000000.jpg} \hfill \includegraphics[height=1.182cm]{images/comparisons/cgstvg/23/bball/000014.jpg}
  \includegraphics[height=1.182cm]{images/comparisons/cgstvg/23/bball/89_y.png}
    \includegraphics[height=1.182cm]{images/comparisons/cgstvg/23/bball/000149.jpg}
  }{ }\\
    \rotatebox{90}{\hspace{-20pt}Ours (Bbox)}
    \jsubfig{\includegraphics[height=1.182cm]{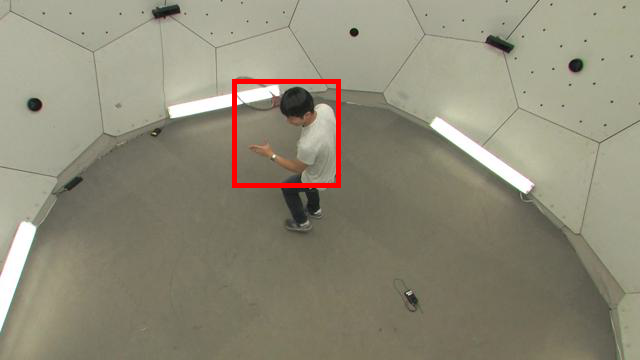} \hfill \includegraphics[height=1.182cm]{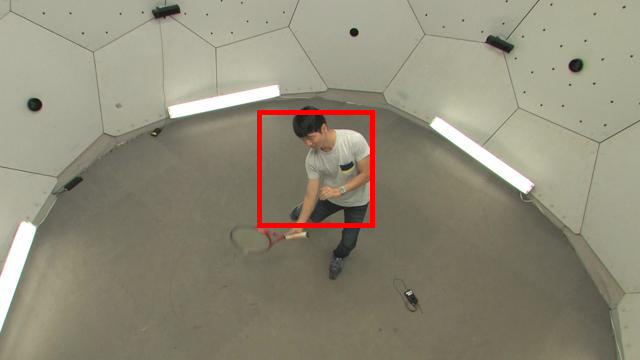}
  \includegraphics[height=1.182cm]{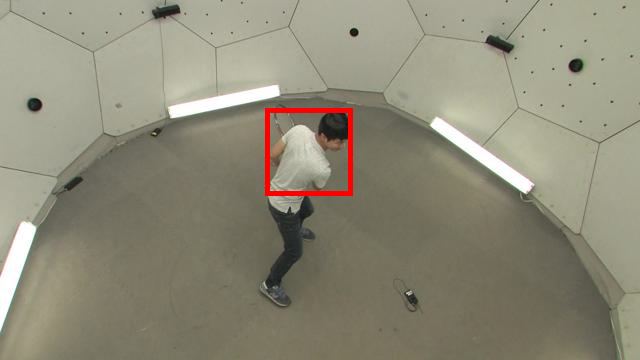}
    \includegraphics[height=1.182cm]{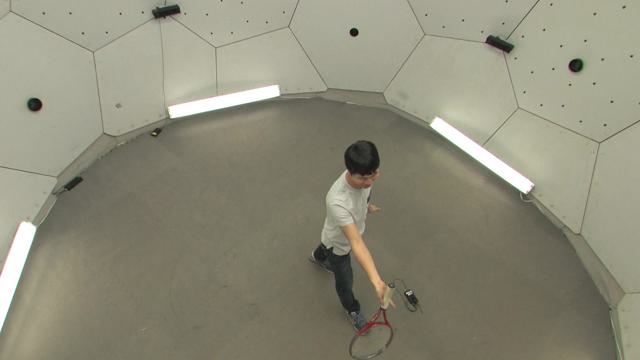}
  }{ }
  \hfill
  \jsubfig{\includegraphics[height=1.182cm]{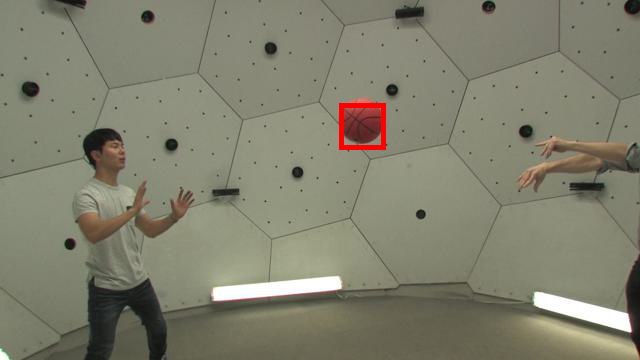} \hfill \includegraphics[height=1.182cm]{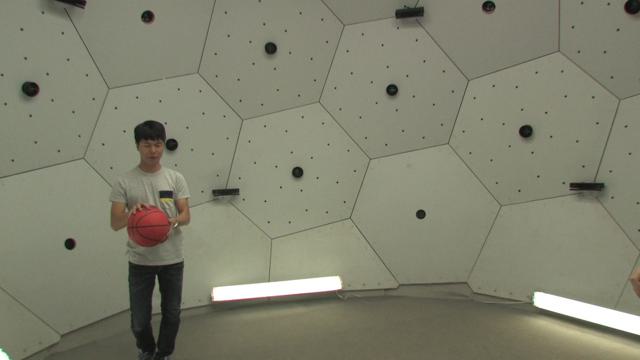}
  \includegraphics[height=1.182cm]{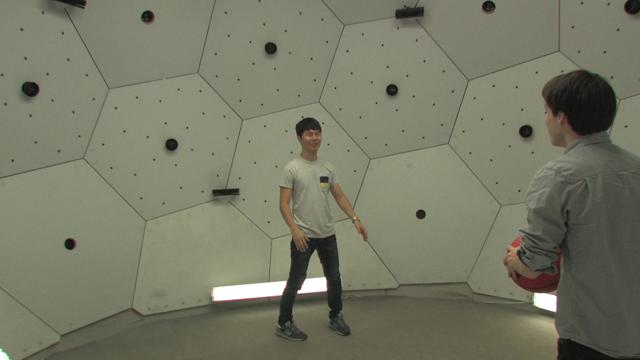}
    \includegraphics[height=1.182cm]{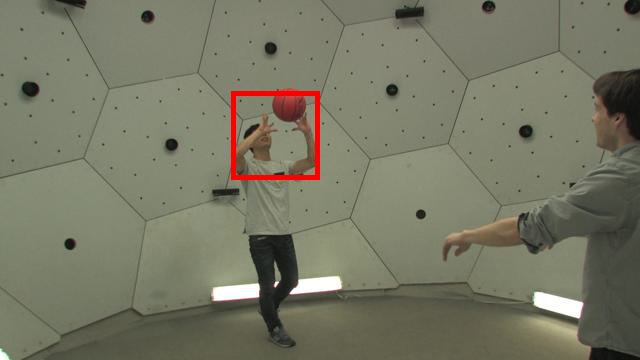}
  }{ } \\
  \rotatebox{90}{\whitetxt{(}}
    \jsubfig{\includegraphics[height=1.182cm]{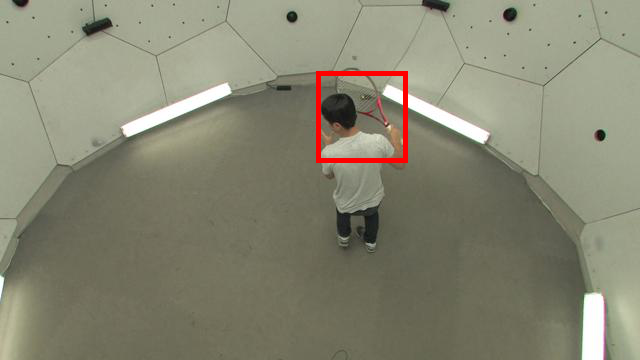} \hfill \includegraphics[height=1.182cm]{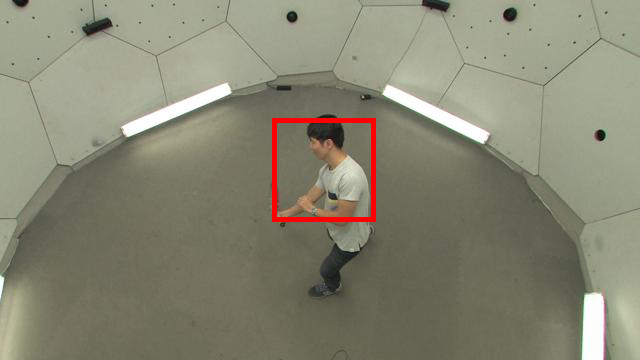}
  \includegraphics[height=1.182cm]{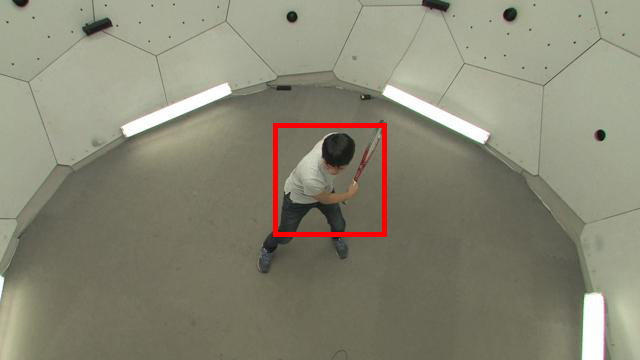}
    \includegraphics[height=1.182cm]{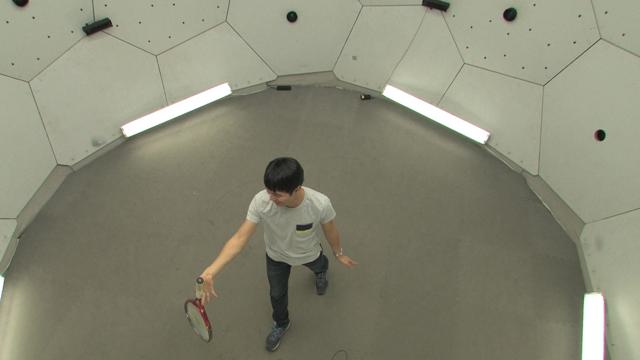}
  }{ }
  \hfill
  \jsubfig{\includegraphics[height=1.182cm]{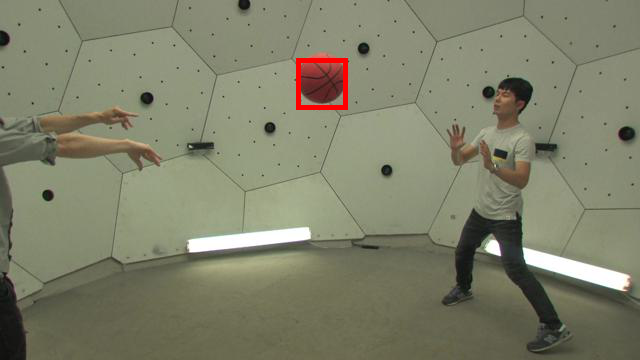} \hfill \includegraphics[height=1.182cm]{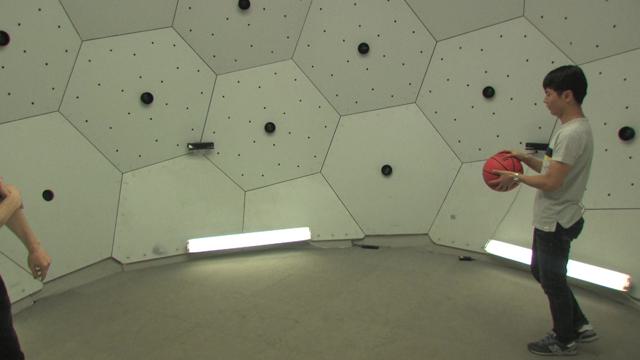}
  \includegraphics[height=1.182cm]{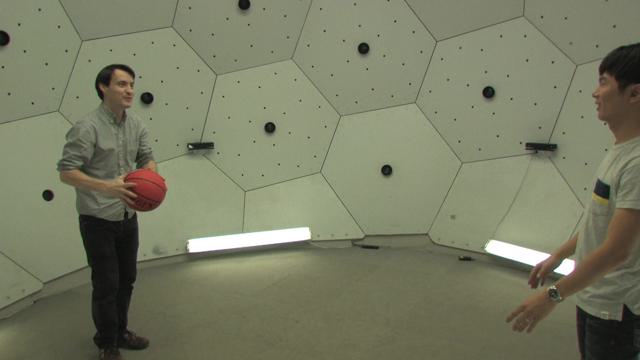}
    \includegraphics[height=1.182cm]{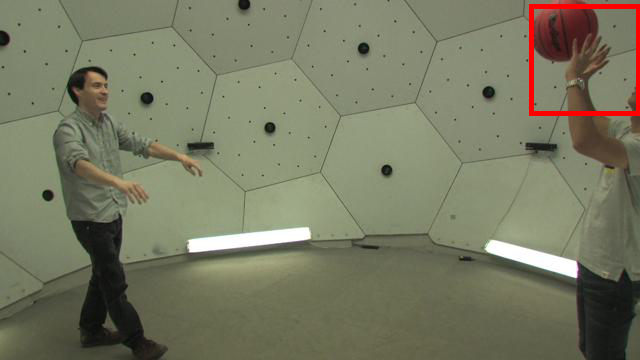}
  }{ } \\
    \rotatebox{90}{\hspace{-6pt}GT\whitetxt{(}}
    \jsubfig{\includegraphics[height=1.182cm]{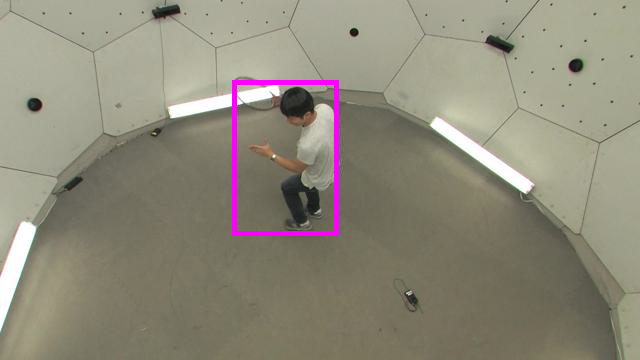} \hfill \includegraphics[height=1.182cm]{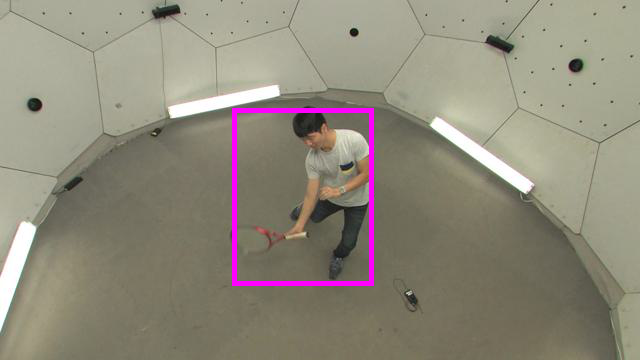}
  \includegraphics[height=1.182cm]{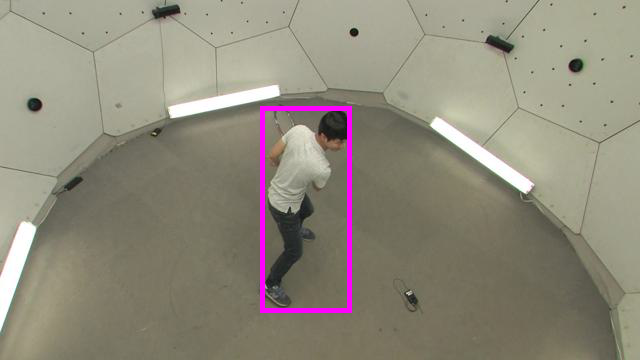}
    \includegraphics[height=1.182cm]{images/comparisons/gt/6/a_person_playing_tennis_cam6_frame149.jpg}
  }{}
  \hfill
  \jsubfig{\includegraphics[height=1.182cm]{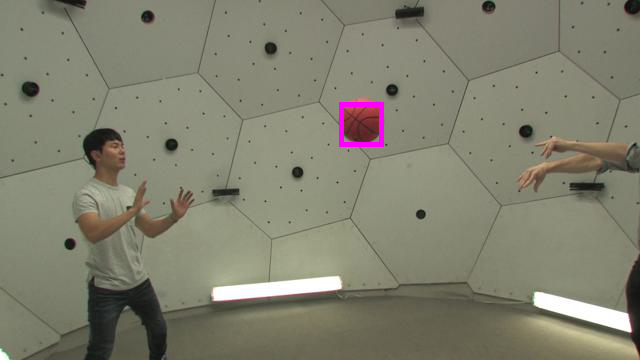} \hfill \includegraphics[height=1.182cm]{images/comparisons/gt/0/a_ball_flying_in_the_air_cam0_frame14.png}
  \includegraphics[height=1.182cm]{images/comparisons/gt/0/a_ball_flying_in_the_air_cam0_frame89.png}
    \includegraphics[height=1.182cm]{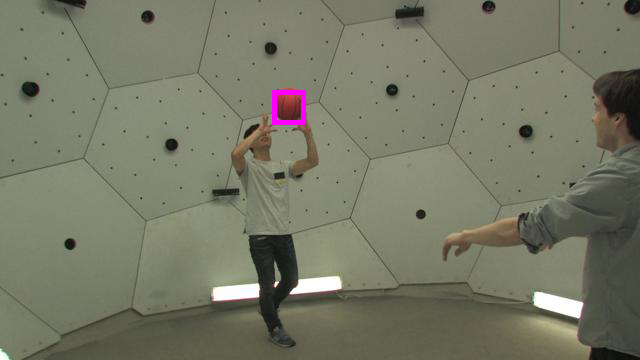}
  }{} \\
   \rotatebox{90}{\whitetxt{())}}
    \jsubfig{\includegraphics[height=1.182cm]{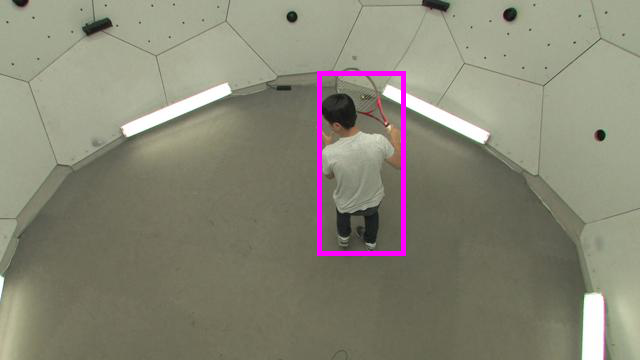} \hfill \includegraphics[height=1.182cm]{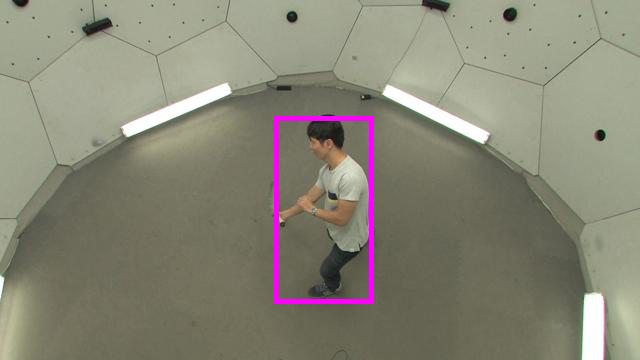}
  \includegraphics[height=1.182cm]{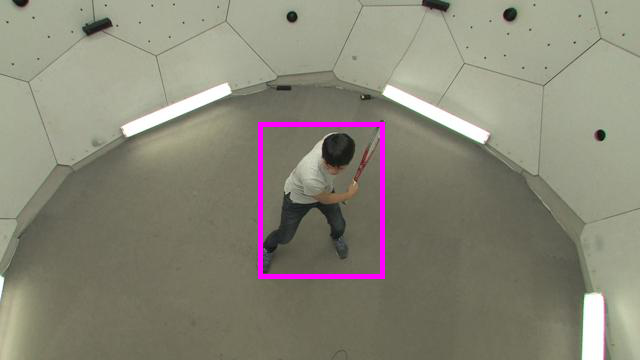}
    \includegraphics[height=1.182cm]{images/comparisons/gt/19/a_person_playing_tennis_cam19_frame149.jpg}
  }{Input query: \emph{A person playing tennis }}
  \hfill
  \jsubfig{\includegraphics[height=1.182cm]{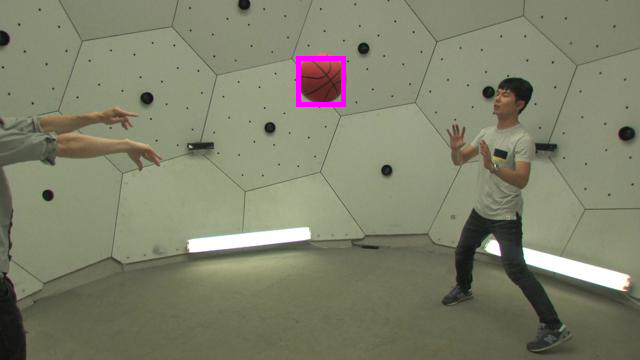} \hfill \includegraphics[height=1.182cm]{images/comparisons/gt/23/a_ball_flying_cam23_frame14.jpg}
  \includegraphics[height=1.182cm]{images/comparisons/gt/23/a_ball_flying_cam23_frame89.jpg}
    \includegraphics[height=1.182cm]{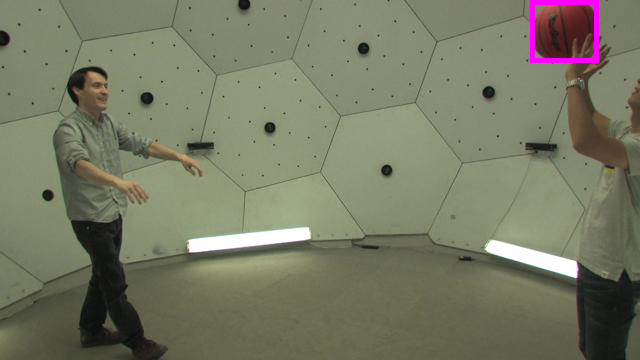}
  }{ Input query: \emph{A ball flying in the air }} \\
    \vspace{-5pt}
    \caption{
  \textbf{Bounding box comparison to 2D spatio-temporal grounding.} We show spatio-temporal localization results for 2D baseline methods (TubeDETR~\cite{yang2022tubedetr} and CGSTVG \cite{gu2024context}), along with our results converted to bounding boxes. The textual queries and ground-truth segmentation maps are taken from our \datasetname{} benchmark. Results are illustrated over two different camera viewpoints (shown on different rows) and four different timestamps (shown on different columns). As illustrated above, albeit our naive conversion method from segmentation maps to bounding boxes, the predicted bounding boxes of our method succeed in spatially localizing the actions.}
  \label{fig:comparison-bbox}
\end{figure*}

As 2D spatio-temporal grounding methods predict bounding boxes rather than segmentation maps, we show a comparison to our method, predicting bounding boxes rather than segmentation maps. For predicting a bounding box with our method, we take the connected component with the maximum score. As shown in 
Table \ref{tab:comparisons-supp-bbox}, our method (denoted as Ours (Bbox) outperforms 2D methods 
, showing strong spatial localization capabilities for both bounding box and segmentation outputs. In Figure \ref{fig:comparison-bbox}, we show a qualitative comparison over bounding box predictions to 2D methods. As illustrated, even though our method outputs segmentation maps and we convert these maps to bounding boxes using a naive method, we outperform the 2D methods in spatially localizing the input queries.

\begin{table}[t]
\centering
\resizebox{1.0\linewidth}{!}{
\begin{tabular}{lcccccc}
\toprule
 Method & $\text{vAP}\uparrow$ &  $\text{vIOU}\uparrow$ & $\text{tIOU}\uparrow$ & $\text{tRec}\uparrow$ & $\text{tPrec}\uparrow$ & $\text{tAP}\uparrow$\\
\midrule
TubeDETR & $21.3\pm3.7$ & $19.9\pm7.3$ & $29.3\pm9.7$ & $44.6\pm16.4$ & $63.7\pm16.3$ & $41.4\pm13.7$ \\
CGSTVG & $25.1\pm8.3$ & $24.3\pm7.7$ & $34.9\pm10.8$ & $56.7\pm17.3$ & $51.6\pm11.3$ & $48.1\pm14.4$ \\
Ours (Bbox) & $\mathbf{29.1\pm0.4}$ & $\mathbf{27.0\pm4.2}$ & $\mathbf{60.8\pm0.0}$ & $\mathbf{83.0\pm0.0}$ & $\mathbf{68.5\pm0.0}$ & $\mathbf{72.6\pm0.0}$\\  
\bottomrule
\end{tabular}
}
\caption{\textbf{Quantitative Evaluation of \methodname{} bounding box variant}. We report performance for our bounding box variant (Ours (Bbox)) as described in Section \ref{sec:bbox-comp} over the spatio-temporal vAP and vIOU metrics, and over four additional metrics quantifying the quality of the temporal localization (tIOU,tRec, tPrec, tAP).} %

\label{tab:comparisons-supp-bbox}
\end{table}

\subsection{Self-Attention Ablations}
\label{sec:ab-att}
\begin{figure*} %
\centering
\rotatebox{90}{\whitetxt{$_{pt}$}$\text{Local}_{\text{Blur}}$}
\jsubfig{\includegraphics[scale=0.12]{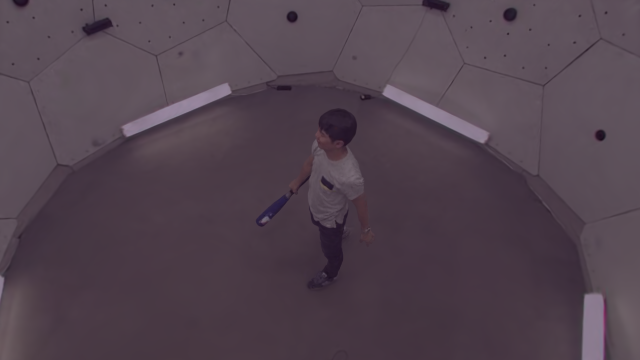}
\includegraphics[scale=0.12]{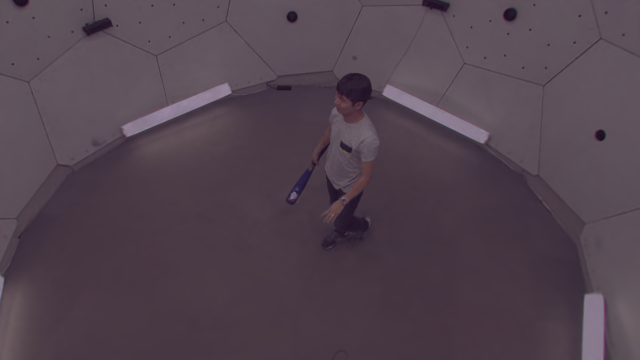}
\includegraphics[scale=0.12]{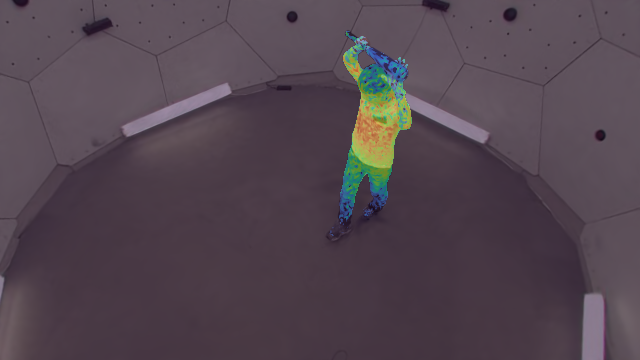}
\includegraphics[scale=0.12]{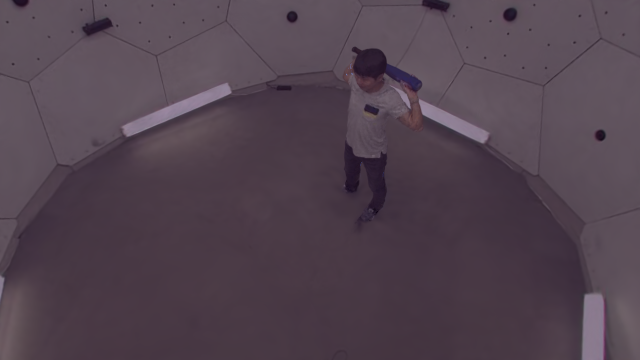}
\includegraphics[scale=0.12]{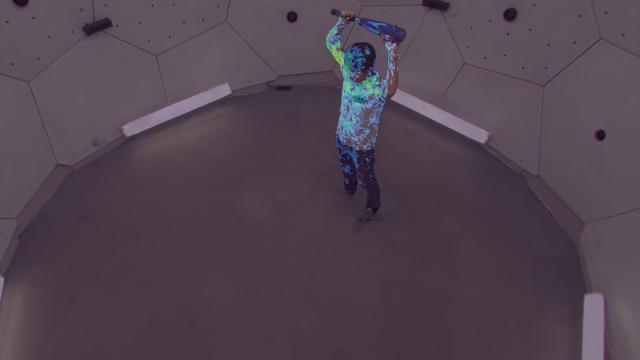}
\includegraphics[scale=0.12]{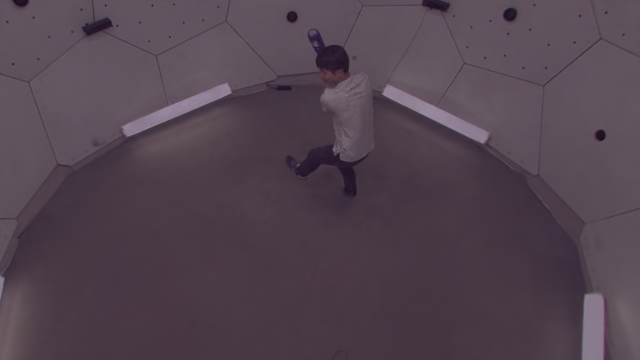}}{}
\\
\rotatebox{90}{\whitetxt{$_{pt}$}$\text{Local}_{\text{Reg}}$}
\jsubfig{\includegraphics[scale=0.12]{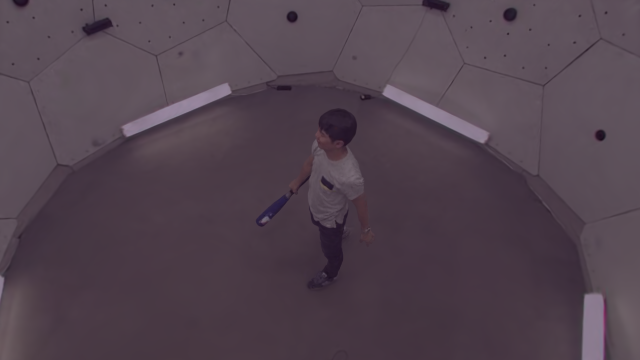}
\includegraphics[scale=0.12]{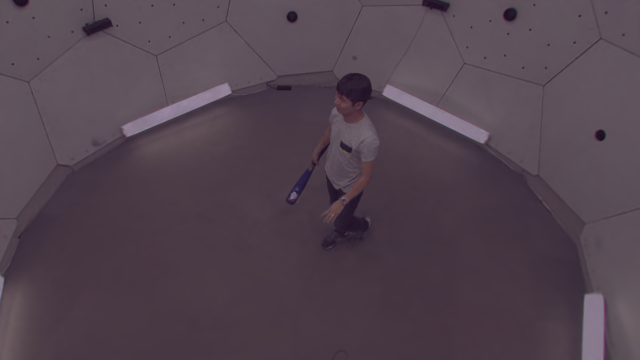}
\includegraphics[scale=0.12]{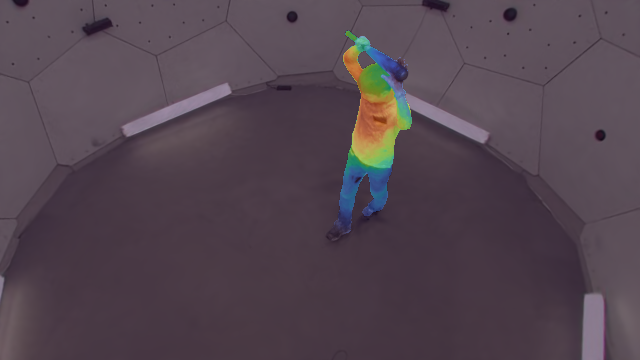}
\includegraphics[scale=0.12]{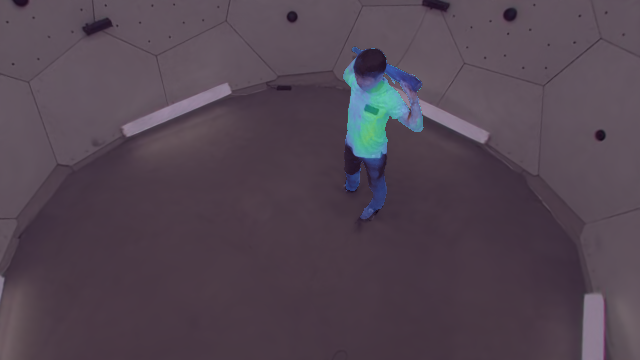}
\includegraphics[scale=0.12]{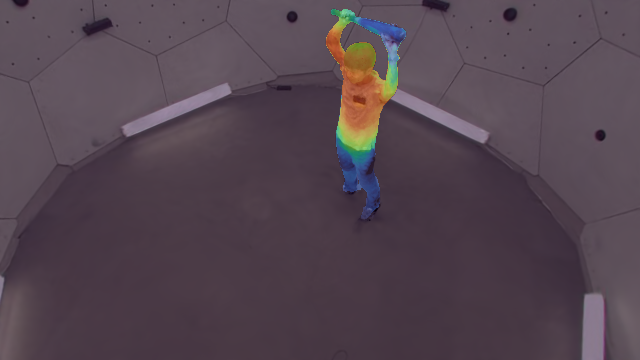}
\includegraphics[scale=0.12]{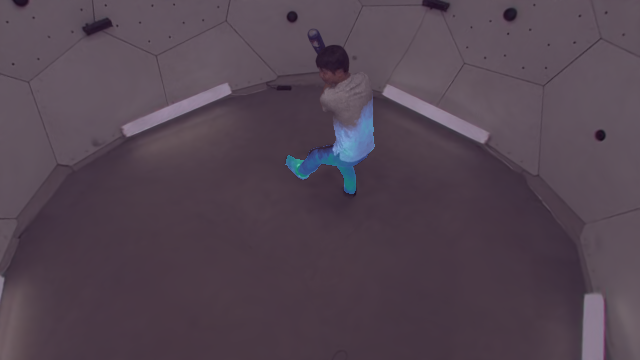}}{}
\\
\rotatebox{90}{\whitetxt{$_{pt}$}$\text{Local}_{\text{MLP}}$}
\jsubfig{\includegraphics[scale=0.12]{images/att_ablations/mlp/comp_0.png}
\includegraphics[scale=0.12]{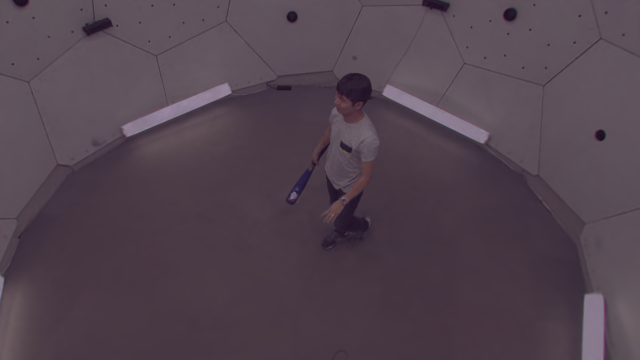}
\includegraphics[scale=0.12]{images/att_ablations/mlp/comp_39.png}
\includegraphics[scale=0.12]{images/att_ablations/mlp/comp_71.png}
\includegraphics[scale=0.12]{images/att_ablations/mlp/comp_79.png}
\includegraphics[scale=0.12]{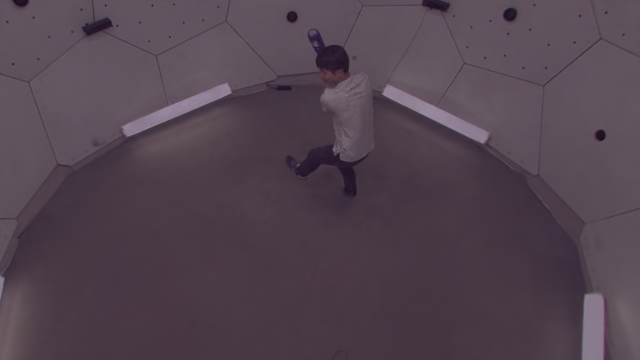}}{}
\\
\rotatebox{90}{\whitetxt{$_{t}$}$\text{w/o Attn}$}
\jsubfig{\includegraphics[scale=0.12]{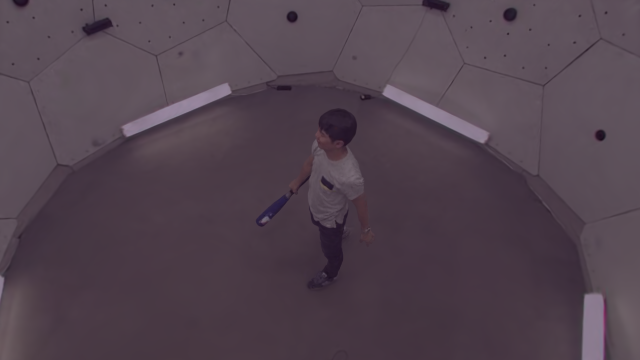}
\includegraphics[scale=0.12]{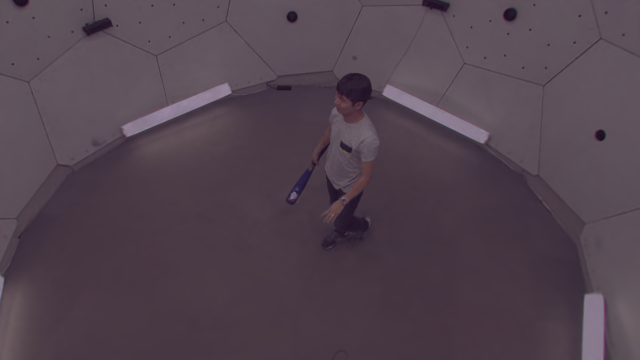}
\includegraphics[scale=0.12]{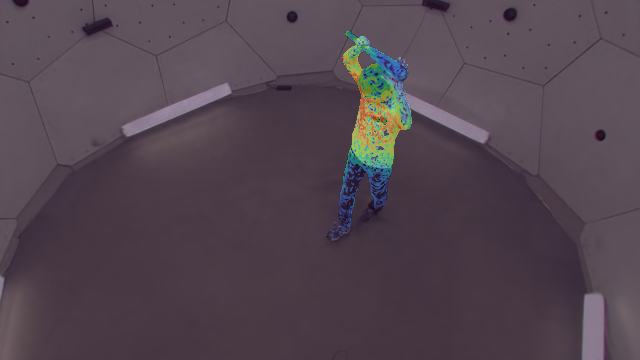}
\includegraphics[scale=0.12]{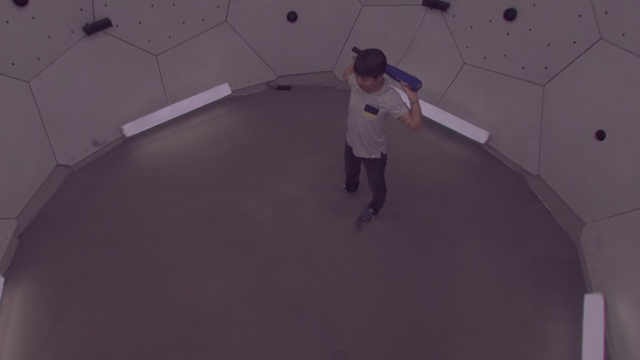}
\includegraphics[scale=0.12]{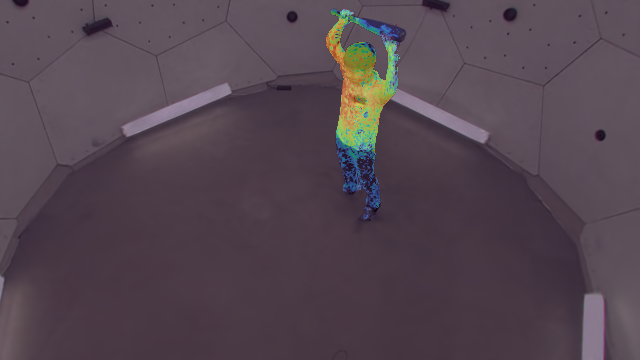}
\includegraphics[scale=0.12]{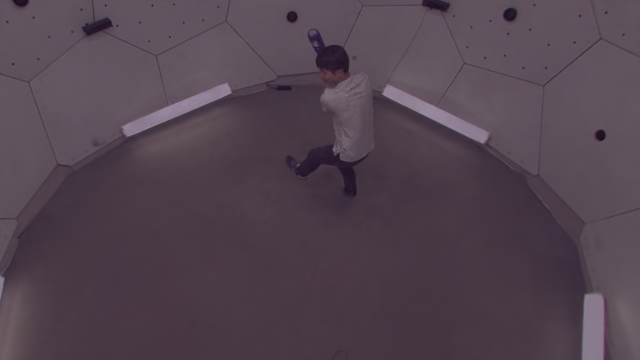}}{}%
\\
\rotatebox{90}{\whitetxt{$_{5pt}$}$\text{Ours}$}
\jsubfig{\includegraphics[scale=0.12]{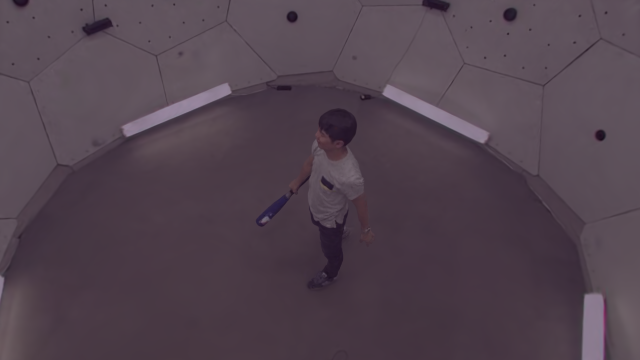}
\includegraphics[scale=0.12]{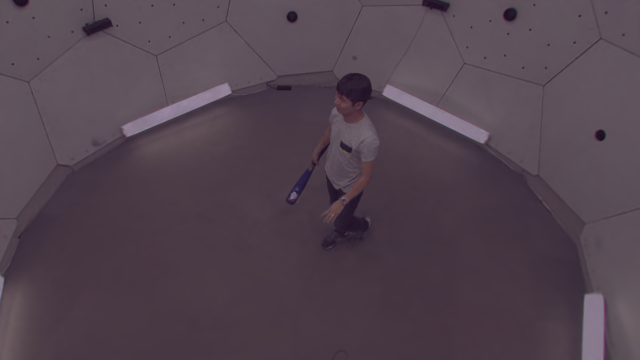}
\includegraphics[scale=0.12]{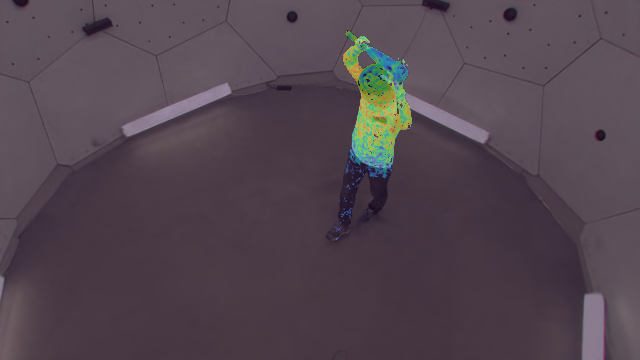}
\includegraphics[scale=0.12]{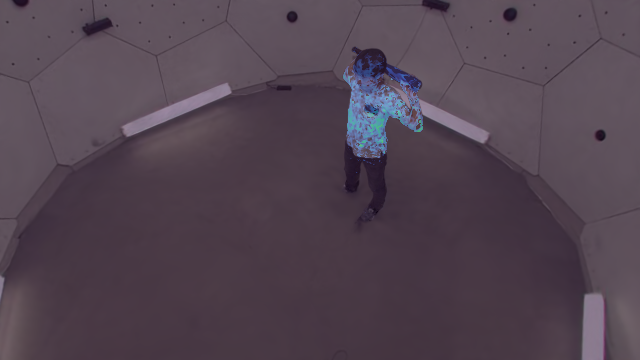}
\includegraphics[scale=0.12]{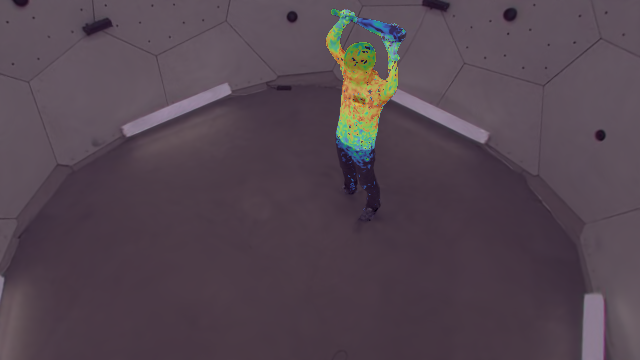}
\includegraphics[scale=0.12]{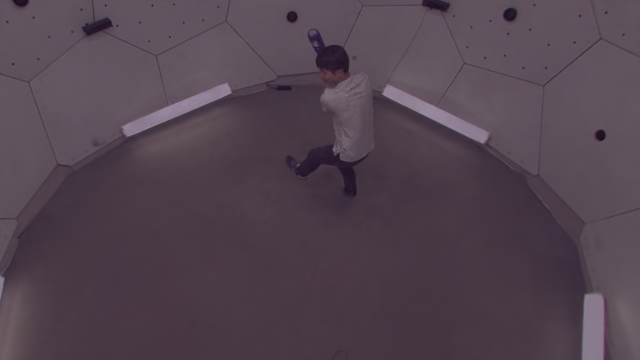}}{}%
\\
\rotatebox{90}{\whitetxt{$_{10pt}$}$\text{GT}$}
\jsubfig{\includegraphics[scale=0.12]{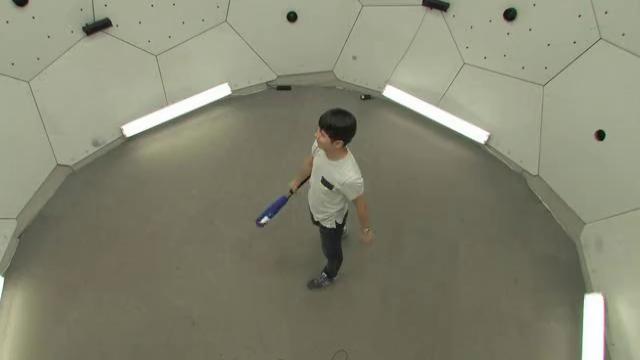}}{}
\jsubfig{\includegraphics[scale=0.12]{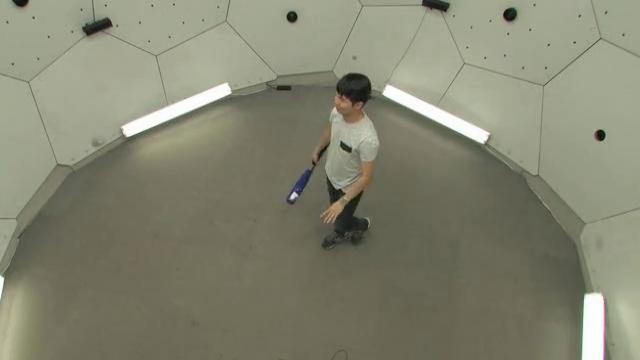}}{}
\jsubfig{{\includegraphics[scale=0.12]{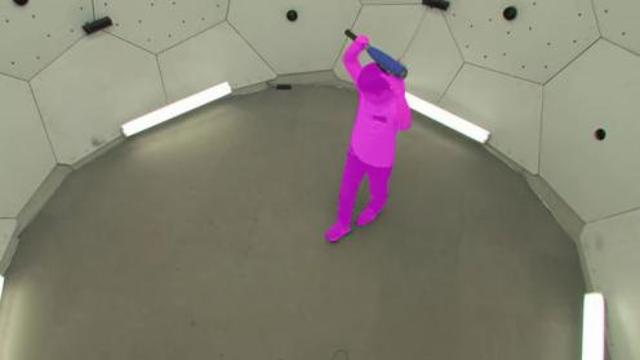}}}{}
\jsubfig{{\includegraphics[scale=0.12]{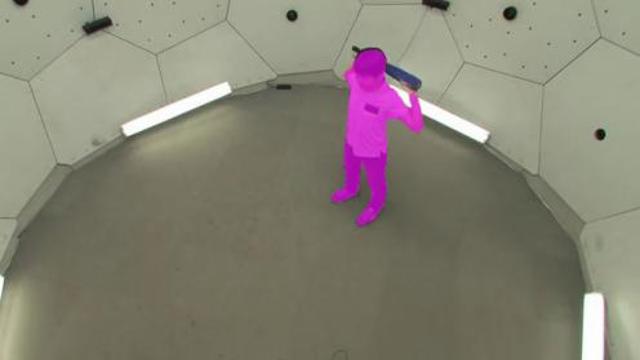}}}{}
\jsubfig{{\includegraphics[scale=0.12]{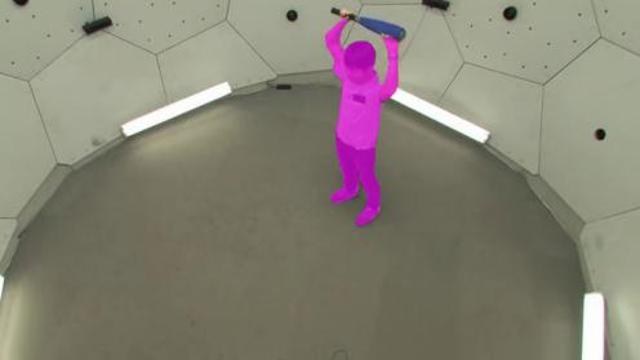}}}{}
\jsubfig{\includegraphics[scale=0.12]{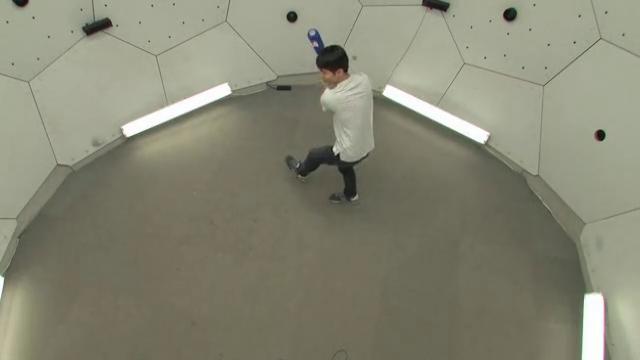}}{ }\\%\hspace{0.001pt}
{Input query: \emph{A person stretching}}
 \vspace{2pt}
\caption{\textbf{Self-Attention Ablations}, We ablate the use of self-attention (w/o Attn) and the effectiveness of self-attention in enforcing local smoothness in the 3D space ($\text{Local}_{\text{Blur}}$, $\text{Local}_{\text{Reg}}$ and $\text{Local}_{\text{MLP}}$). As illustrated above, our approach outperforms these ablations -- both spatially and temporally.}

\label{fig:ablations-att}
\end{figure*}

We perform ablations to motivate the use of self-attention and the self-attention mechanism's ability to enable local spatial smoothness between Gaussians' features. To ablate the use of self-attention we compare against a 4D language feature field distilled without attention (denoted as w/o Attn). We ablate the ability of the self-attention mechanism to enforce  local smoothness in 3D space against three methods of local smoothing, (i)a neighborhood blur (denoted as $\text{Local}_{\text{Blur}}$), (ii) a neighborhood smoothness regularization loss (denoted as $\text{Local}_{\text{Reg}}$) and (iii) a MLP for the local neighborhood (denoted as $\text{Local}_{\text{MLP}}$. For each one of these methods we perform $\text{kNN}$ on Gaussian locations and then smooth the features of each Gaussian, $g$, with the features of its $k$ spatially nearest Gaussians, $\text{kNN}(g)$, using each one of the smoothness methods: 
\begin{itemize}
    \item[(i)] For every Gaussian over all timesteps we average the features of $kNN(x^{t}_i)$, where $x^{t}_i$ is the mean location of the $i$-th Gaussian at timestep $t$ and $\text{kNN}$ returns the features of the $k$ spatially nearest Gaussians.
    \item[(ii)]\hspace{0.02cm}   We add a regularization loss for training, thus given a time $t$, we compute the loss over all Gaussians' features, $L^{t}_i$ denoting the spatio-temporal feature of the $i$-th Gaussian at timestep $t$:
        \begin{equation}
            \label{formula: reg_loss}
                L^{t}_{\text{smooth\_reg}} = \sum_{g^{t}_i \in G} \sum_{g^{t}_k \in \text{kNN}(x^{t}_i)} \left\|L^{t}_i - L^{t}_k\right\|_2
        \end{equation} 
    \item[(iii)]\hspace{0.05cm} Instead of performing self-attention, $\text{SelfAttention}(\text{kNN}(x^{t}_i))$, between each Gaussian and its local neighborhood we learn a MLP, $\text{MLP}(\text{kNN}(x^{t}_i))$, on each Gaussian's local neighborhood.
\end{itemize}
A quantitative evaluation over these ablations is reported in Table 3 in the main paper and a qualitative comparison is shown in Figure \ref{fig:ablations-att} . As illustrated in the table, using self-attention improves performance and outperforms all other local smoothing methods.

\subsection{Additional Ablations}
\label{sec:ab-add}

We conduct additional ablations to motivate our feature extraction design choice. We perform the following modifications to our aggregation method for feature extraction, (i) concatenation and (ii) single scale.

\smallskip \noindent \textbf{Feature Concatenation} 
For this ablation, after extracting the multi-scale features, for each pixel $\textbf{p}$ and timestamp $t$, we extract spatio-temporal features of multiple scales $s$, $\tilde{\phi}^{t}_{s}(\textbf{p})$.
To obtain the final spatio-temporal features used for optimization, we concatenate the features across all scales:
    \begin{equation}
    \label{formula: feature_concat}
        \phi^{t}(\textbf{p}) = \text{Concat}(\tilde{\phi}^{t}_{s}(\textbf{p}), s\in S.
    \end{equation} 
We train an autoencoder per scene to compress the 3840 dimensional concatenated ViCLIP features into 128 dimensional latent features. Since the the text encoder feature dimension must match the decoded features to obtain a direct calculation of the relevancy score, we calculate the score at each scale of the concatenated vector and average the scores of each scale for our final relevancy score.

\smallskip \noindent \textbf{Single Scale Features}
For this ablation we extract the features from a single scale of image crops. We ablate over the scales we used for creating our image pyramid: $\text{Scale}_{0}$ - 0.15, $\text{Scale}_{1}$ - 0.2625, $\text{Scale}_{2}$ - 0.375, $\text{Scale}_{3}$ - 0.4875, $\text{Scale}_{4}$ - 0.6. We note that as ViCLIP is not pixel-aligned, using a single scale alignment effectively determines the granularity of the features.

\begin{figure*} %
\centering
\rotatebox{90}{\whitetxt{$_{5pt}$}$\text{Scale}_{0}$}
\jsubfig{\includegraphics[scale=0.12]{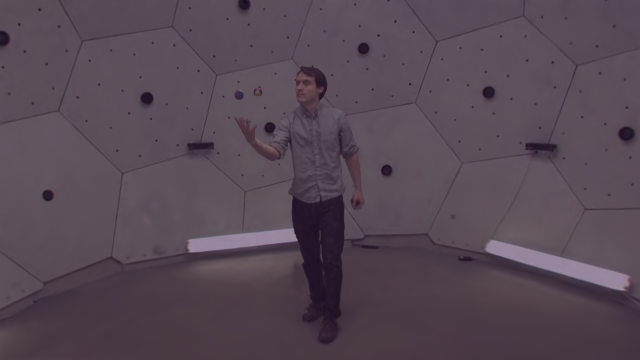}}{}
\jsubfig{{\includegraphics[scale=0.12]{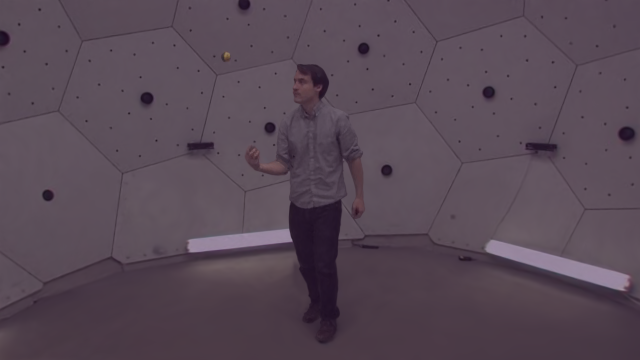}}}{}
\jsubfig{{\includegraphics[scale=0.12]{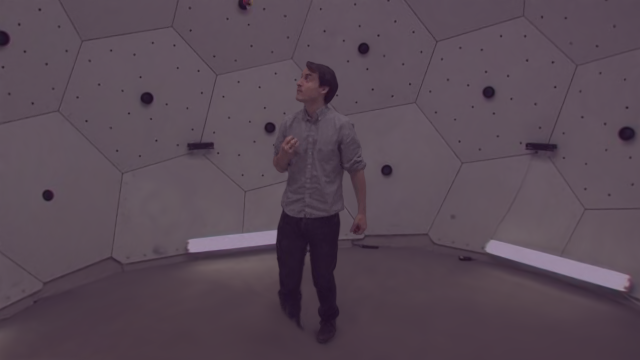}}}{}
\jsubfig{\includegraphics[scale=0.12]{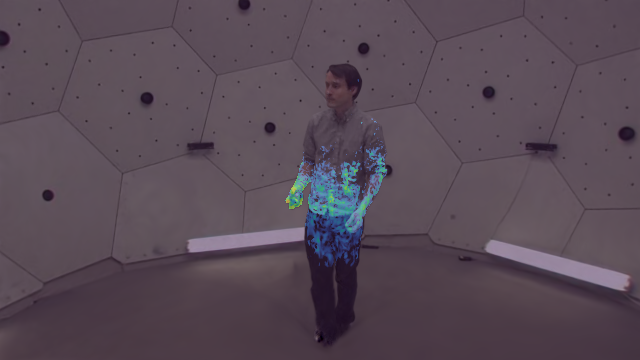}}{}
\jsubfig{{\includegraphics[scale=0.12]{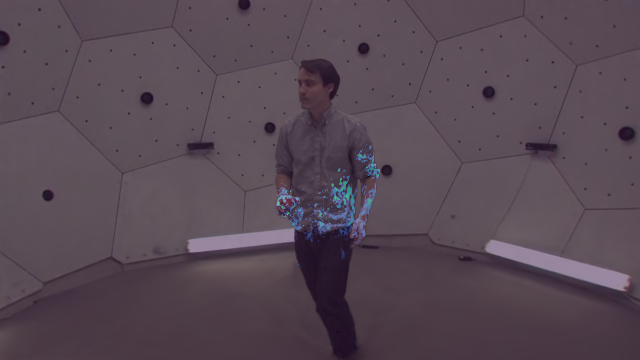}}}{}
\jsubfig{\includegraphics[scale=0.12]{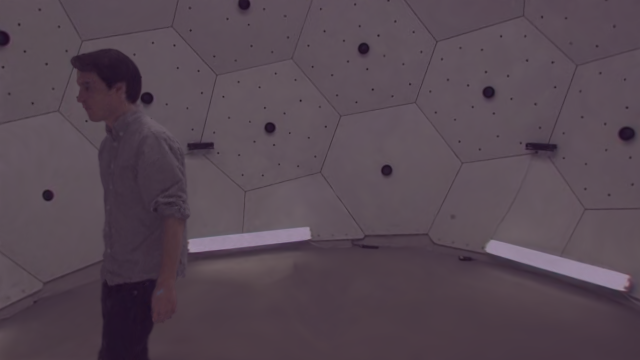}}{}
\\
\rotatebox{90}{\whitetxt{$_{5pt}$}$\text{Scale}_{1}$}
\jsubfig{\includegraphics[scale=0.12]{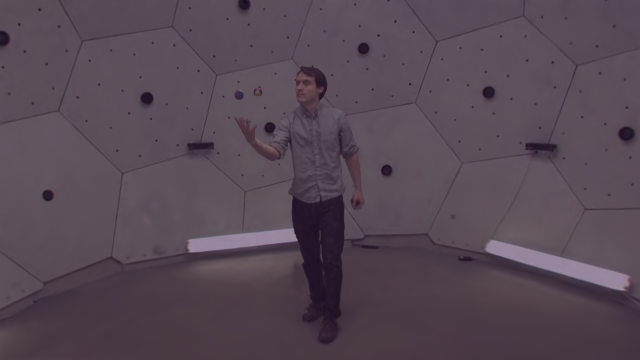}}{}
\jsubfig{{\includegraphics[scale=0.12]{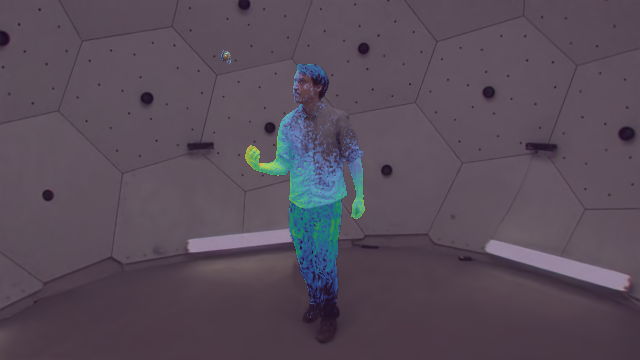}}}{}
\jsubfig{{\includegraphics[scale=0.12]{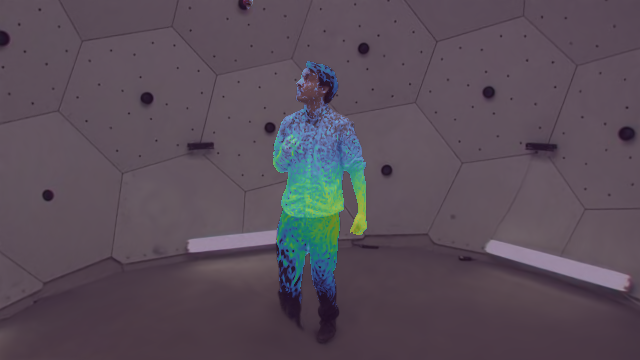}}}{}
\jsubfig{\includegraphics[scale=0.12]{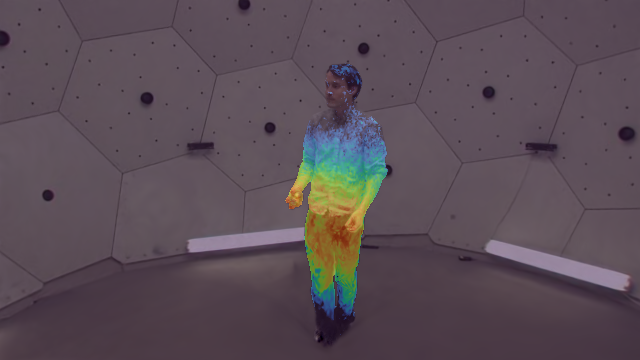}}{}
\jsubfig{{\includegraphics[scale=0.12]{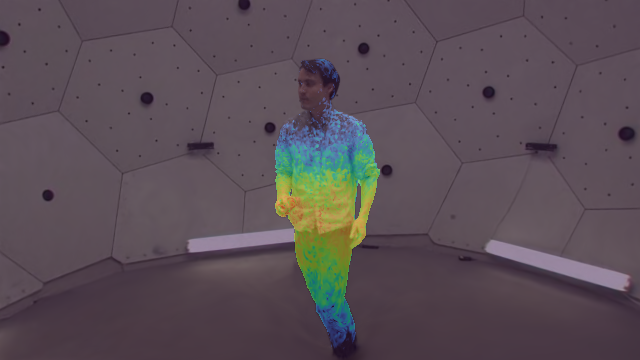}}}{}
\jsubfig{\includegraphics[scale=0.12]{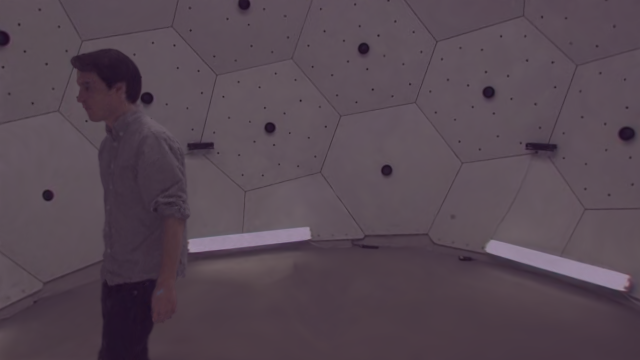}}{}
\\
\rotatebox{90}{\whitetxt{$_{5pt}$}$\text{Scale}_{2}$}
\jsubfig{\includegraphics[scale=0.12]{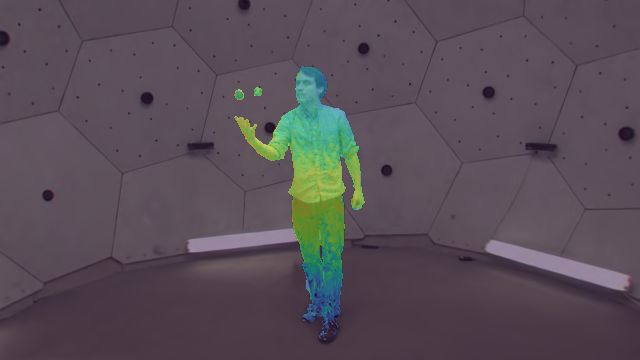}}{}
\jsubfig{{\includegraphics[scale=0.12]{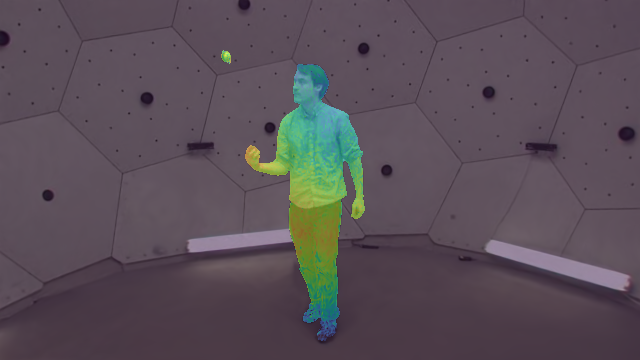}}}{}
\jsubfig{{\includegraphics[scale=0.12]{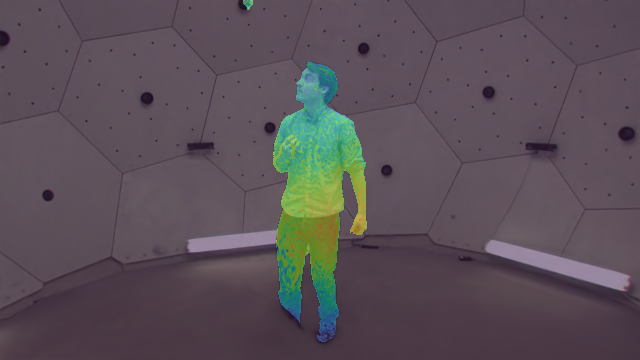}}}{}
\jsubfig{\includegraphics[scale=0.12]{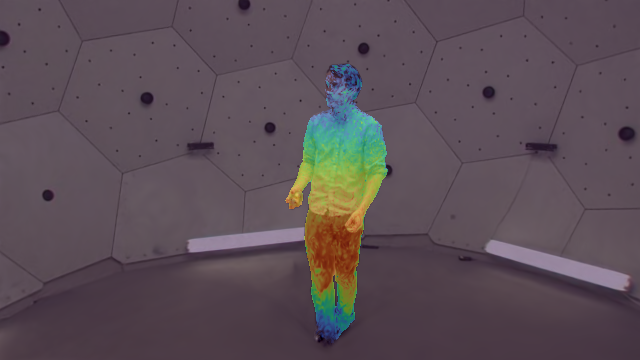}}{}
\jsubfig{{\includegraphics[scale=0.12]{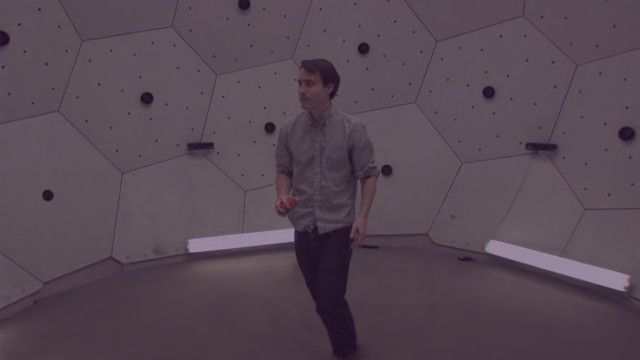}}}{}
\jsubfig{\includegraphics[scale=0.12]{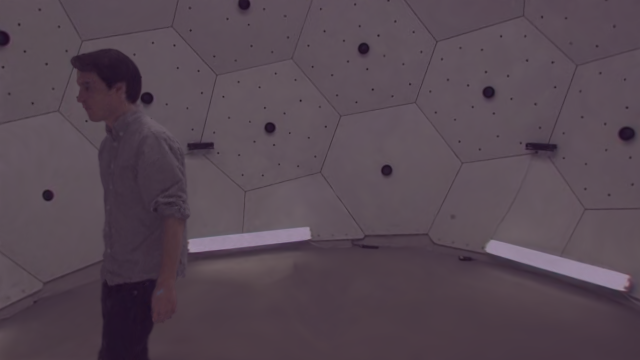}}{}
\\
\rotatebox{90}{\whitetxt{$_{5pt}$}$\text{Scale}_{3}$}
\jsubfig{\includegraphics[scale=0.12]{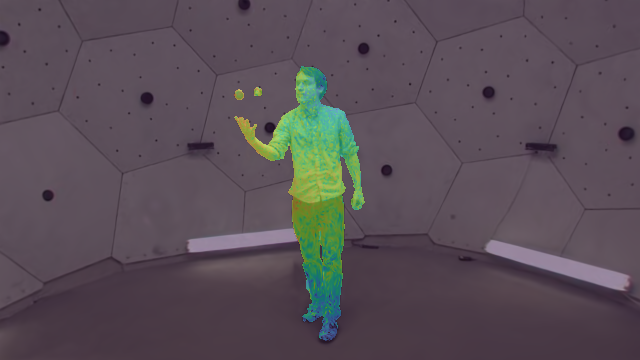}}{}
\jsubfig{{\includegraphics[scale=0.12]{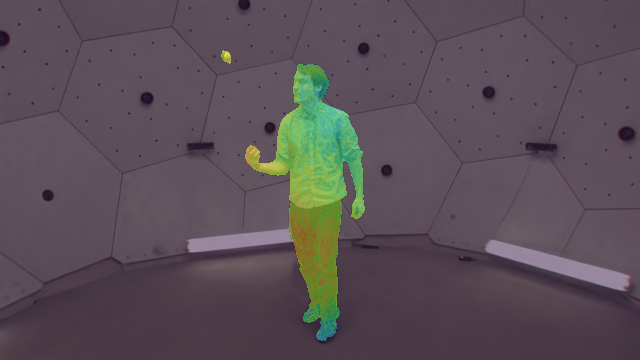}}}{}
\jsubfig{{\includegraphics[scale=0.12]{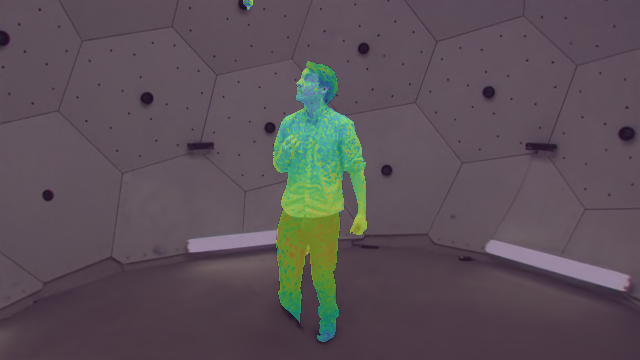}}}{}
\jsubfig{\includegraphics[scale=0.12]{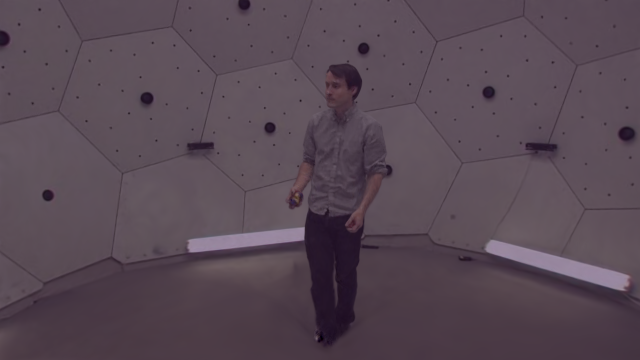}}{}
\jsubfig{{\includegraphics[scale=0.12]{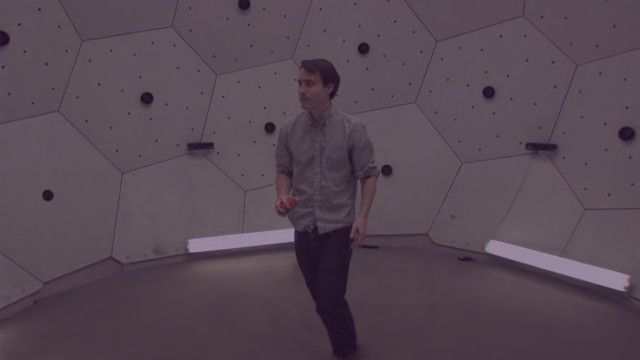}}}{}
\jsubfig{\includegraphics[scale=0.12]{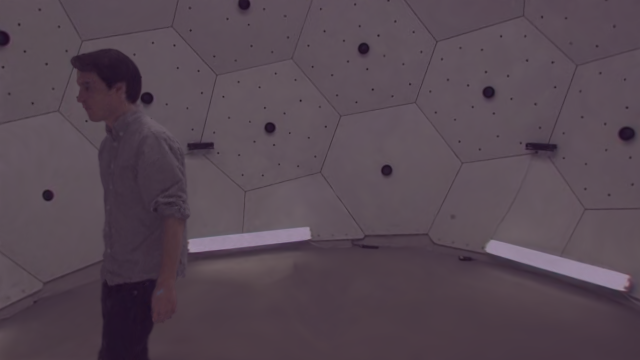}}{}
\\
\rotatebox{90}{\whitetxt{$_{5pt}$}$\text{Scale}_{4}$}
\jsubfig{\includegraphics[scale=0.12]{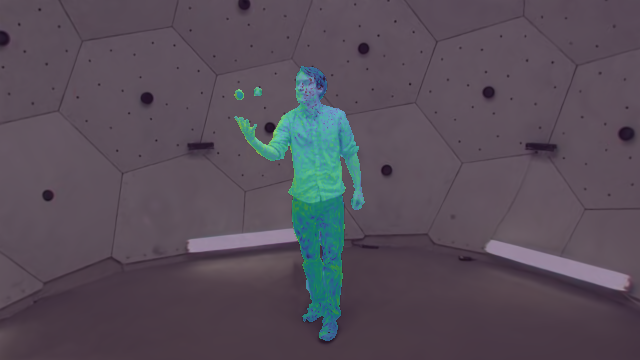}}{}
\jsubfig{{\includegraphics[scale=0.12]{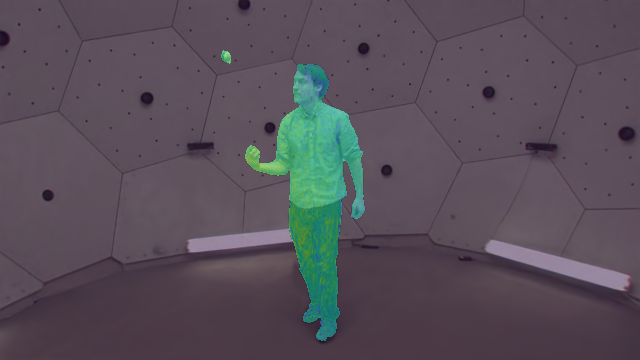}}}{}
\jsubfig{{\includegraphics[scale=0.12]{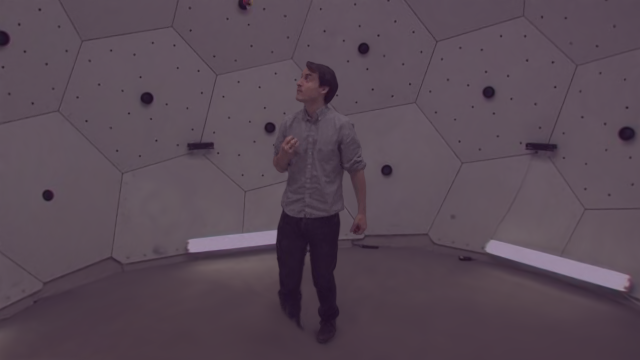}}}{}
\jsubfig{\includegraphics[scale=0.12]{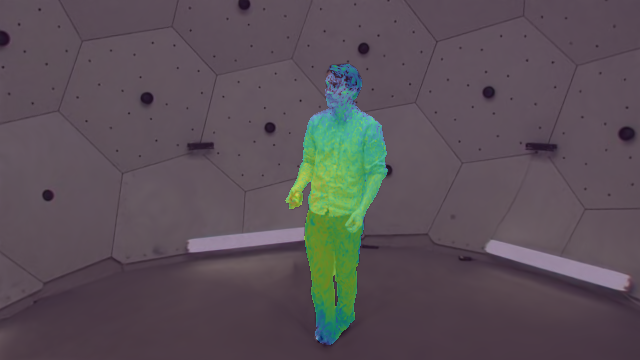}}{}
\jsubfig{{\includegraphics[scale=0.12]{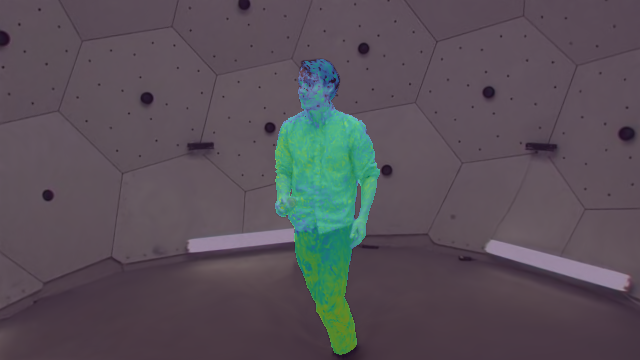}}}{}
\jsubfig{\includegraphics[scale=0.12]{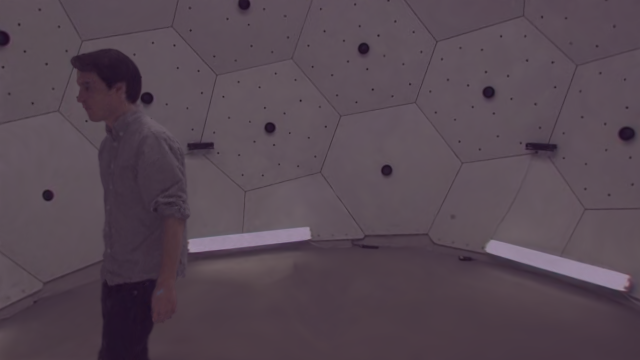}}{}
\\
\rotatebox{90}{\whitetxt{$_{pt}$}$\text{Concat}$}
\jsubfig{\includegraphics[scale=0.12]{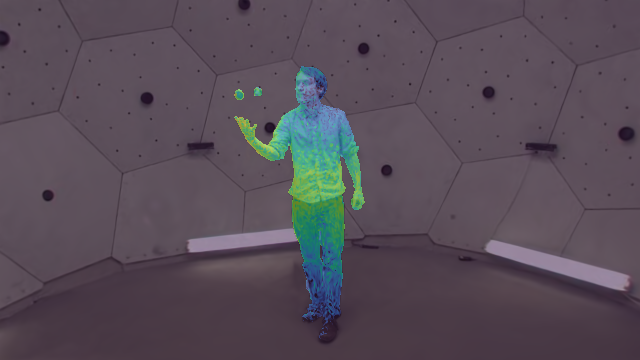}
\includegraphics[scale=0.12]{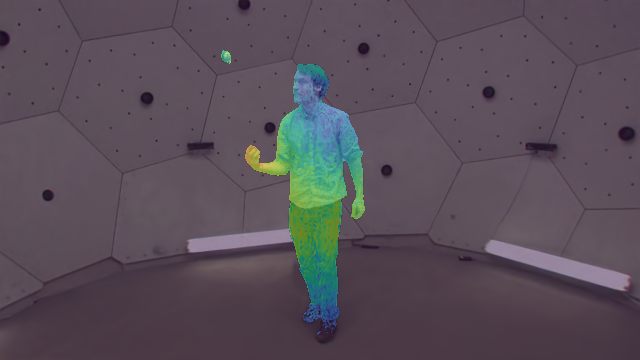}
\includegraphics[scale=0.12]{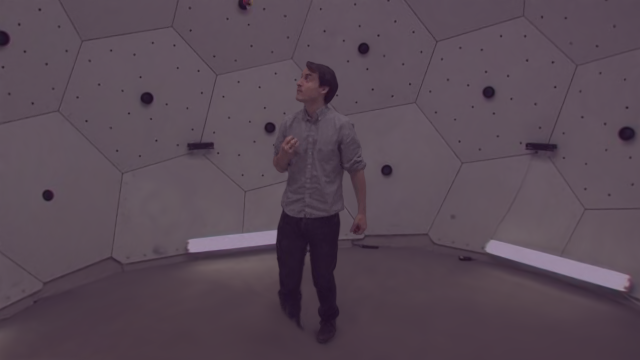}
\includegraphics[scale=0.12]{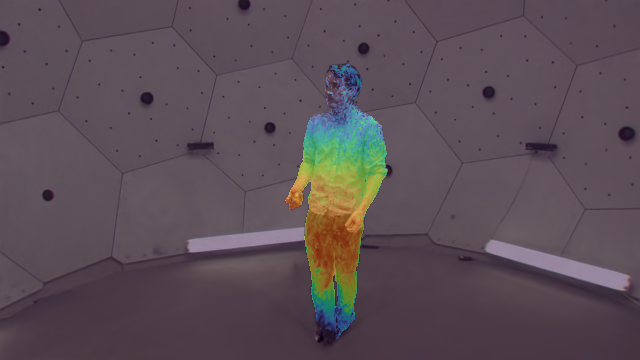}
\includegraphics[scale=0.12]{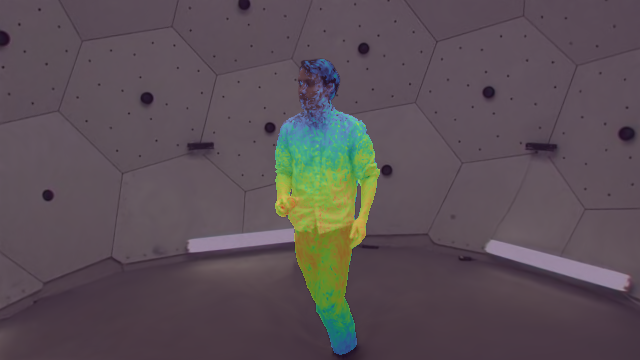}
\includegraphics[scale=0.12]{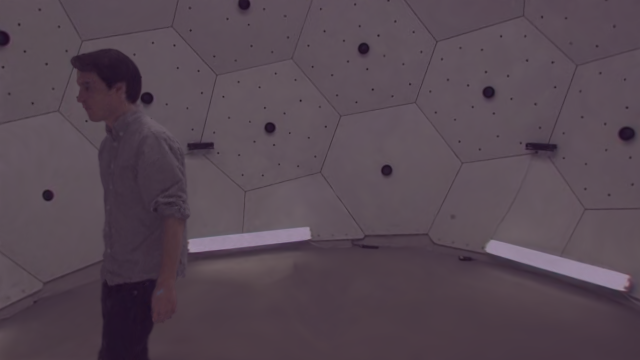}}{}%
\\
\rotatebox{90}{\whitetxt{$_{t}$}$\text{Weight}_{\text{Size}}$}
\jsubfig{\includegraphics[scale=0.12]{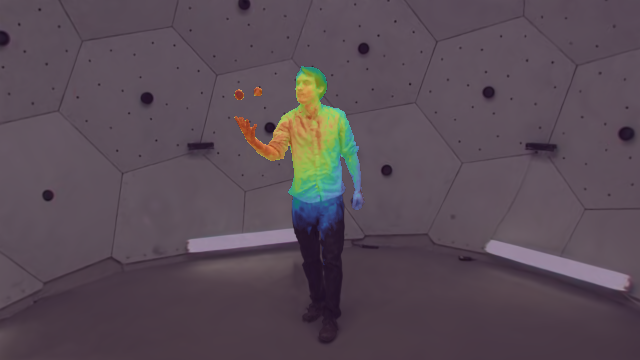}
\includegraphics[scale=0.12]{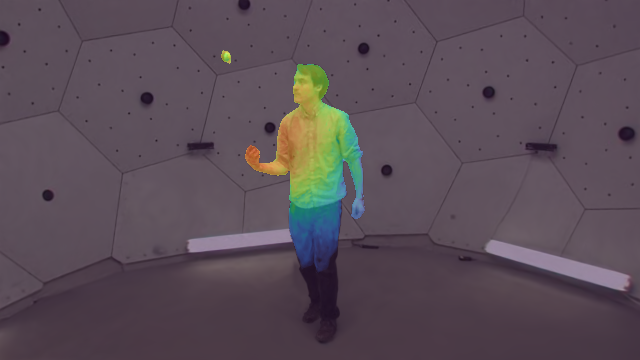}
\includegraphics[scale=0.12]{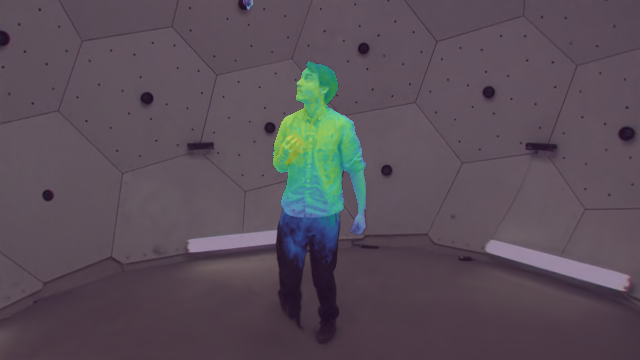}
\includegraphics[scale=0.12]{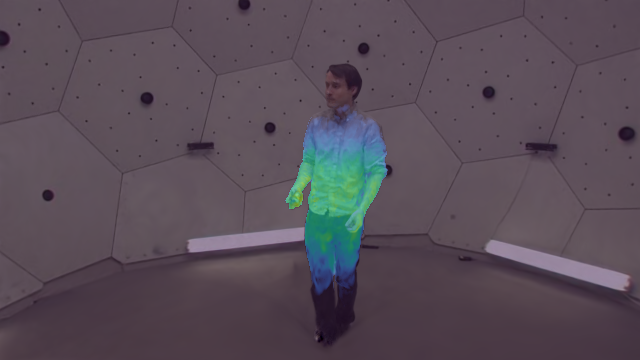}
\includegraphics[scale=0.12]{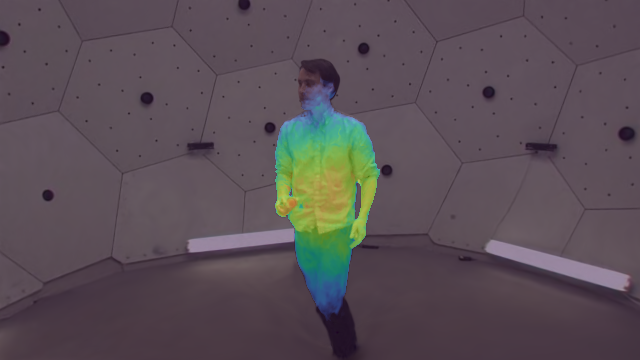}
\includegraphics[scale=0.12]{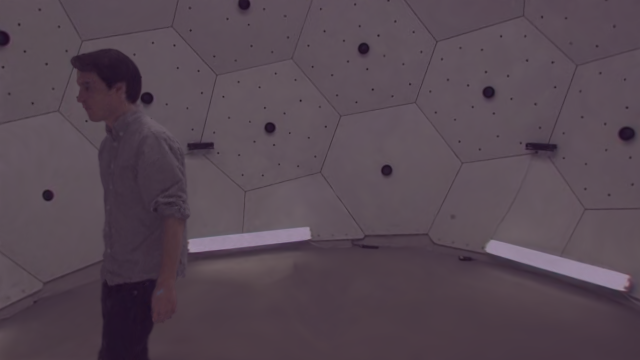}}{}%
\\
\rotatebox{90}{$\text{Latent}_{\text{Query}}$}
\jsubfig{\includegraphics[scale=0.12]{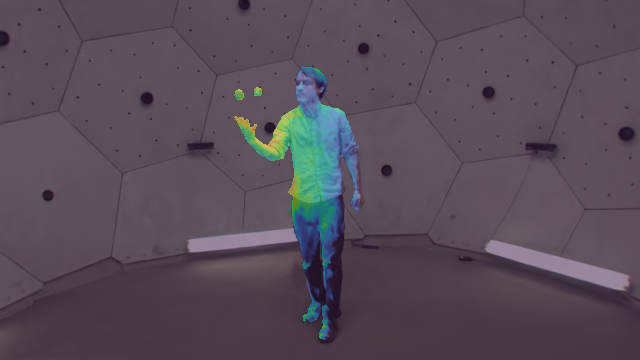}
\includegraphics[scale=0.12]{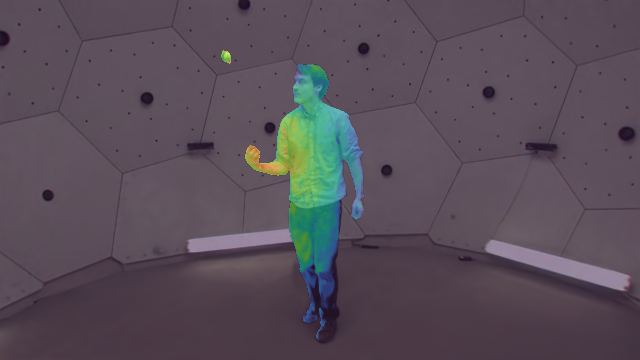}
\includegraphics[scale=0.12]{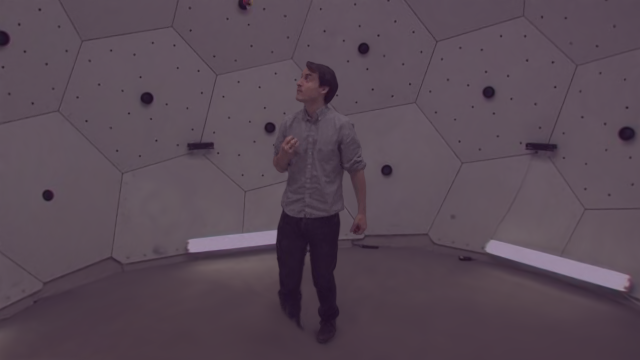}
\includegraphics[scale=0.12]{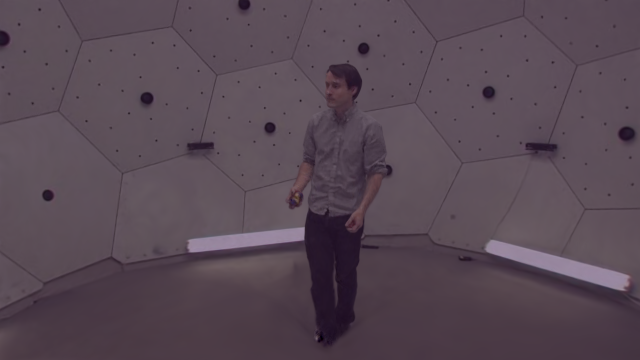}
\includegraphics[scale=0.12]{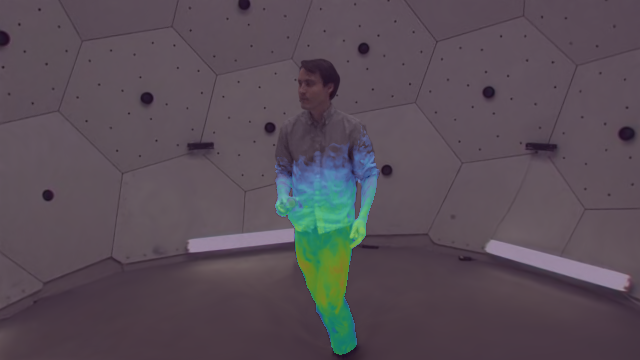}
\includegraphics[scale=0.12]{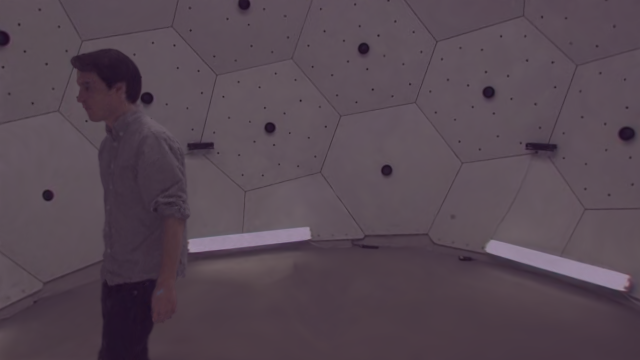}}{}%
\\
\rotatebox{90}{\whitetxt{$_{5pt}$}$\text{Ours}$}
\jsubfig{\includegraphics[scale=0.12]{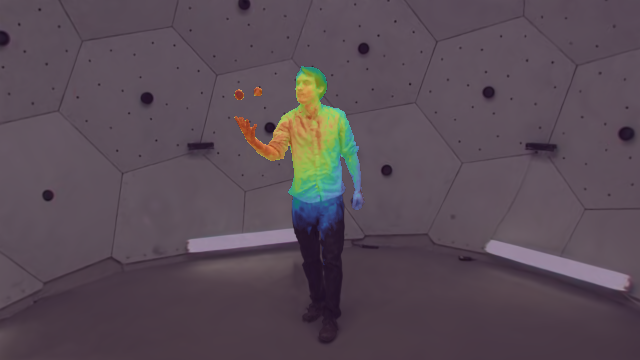}
\includegraphics[scale=0.12]{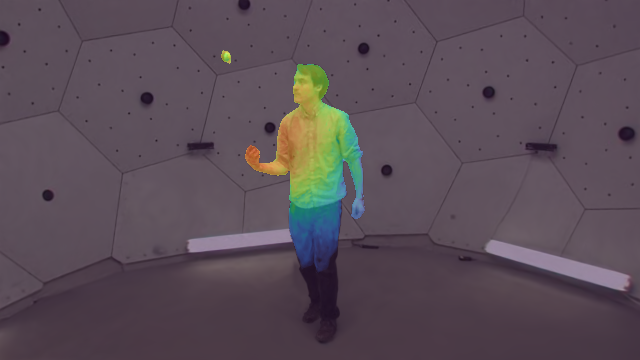}
\includegraphics[scale=0.12]{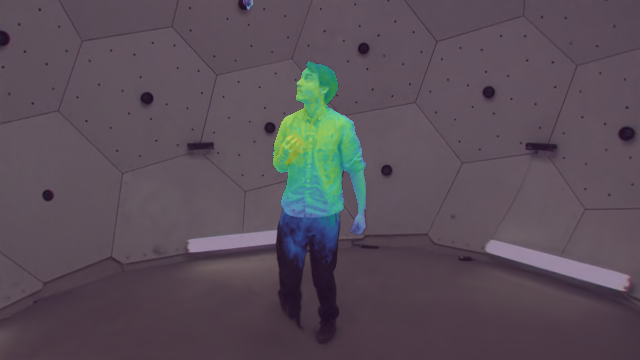}
\includegraphics[scale=0.12]{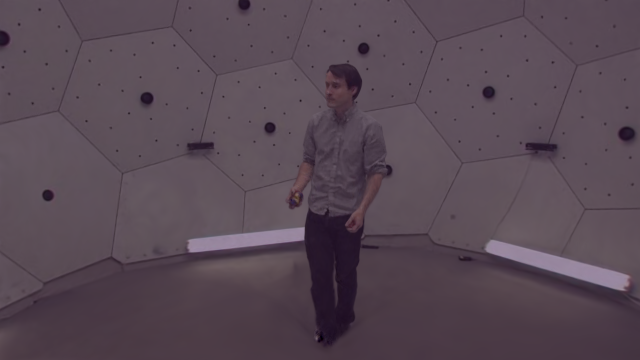}
\includegraphics[scale=0.12]{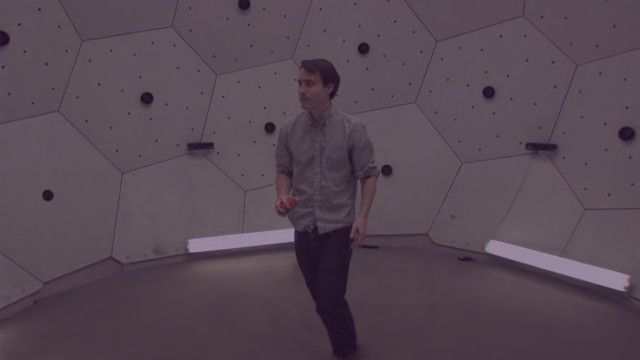}
\includegraphics[scale=0.12]{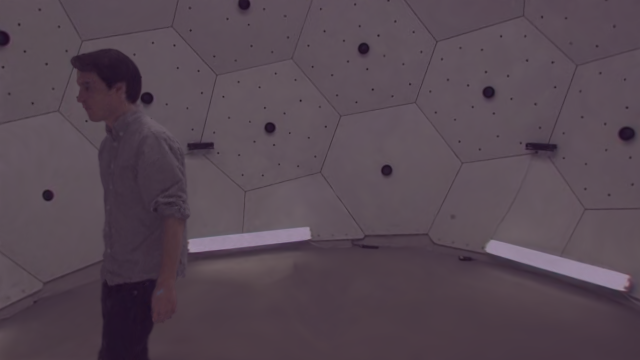}}{}%
\\
\rotatebox{90}{\whitetxt{$_{10pt}$}$\text{GT}$}
\jsubfig{\includegraphics[scale=0.12]{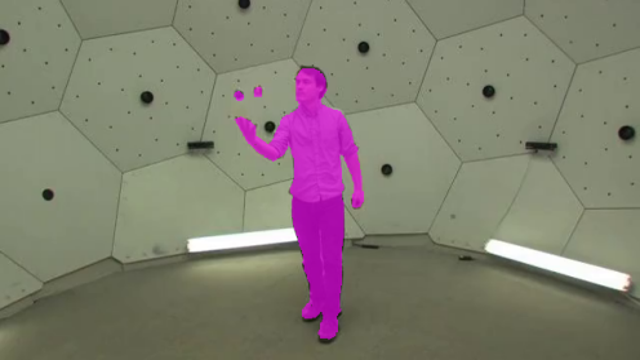}}{}
\jsubfig{\includegraphics[scale=0.12]{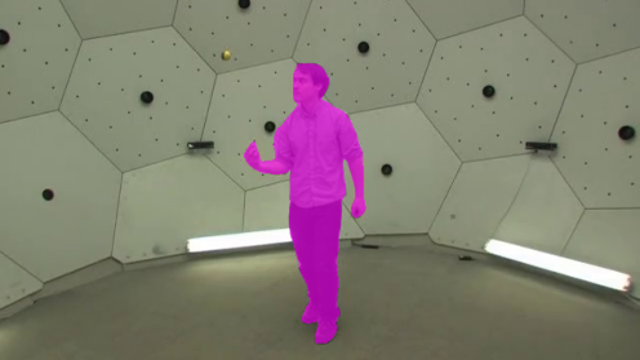}}{}
\jsubfig{{\includegraphics[scale=0.12]{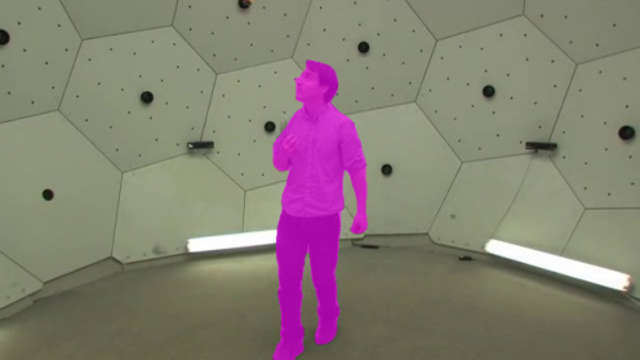}}}{}
\jsubfig{{\includegraphics[scale=0.12]{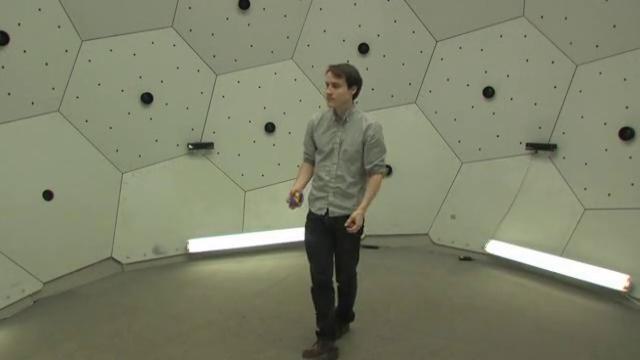}}}{}
\jsubfig{{\includegraphics[scale=0.12]{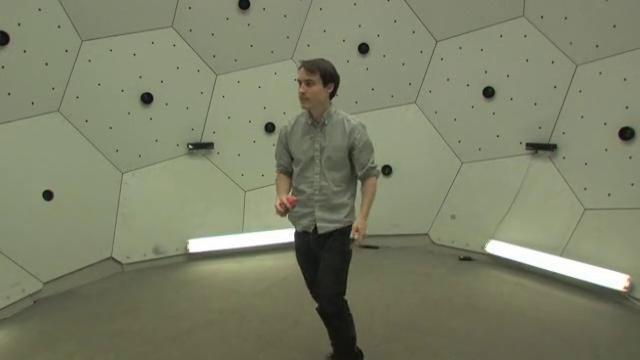}}}{}
\jsubfig{\includegraphics[scale=0.12]{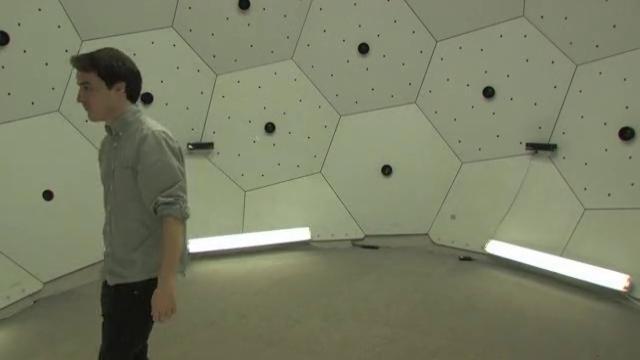}}{ }\\%\hspace{0.001pt}
{Input query: \emph{A person juggling balls}}
 \vspace{2pt}
\caption{\textbf{Additional Ablations.} We ablate the feature extraction design choice by concatenating the multi-scale features (Concat) and by extracting the features from a single resolution scale ($\text{Scale}_{0}$, $\text{Scale}_{1}$, $\text{Scale}_{2}$, $\text{Scale}_{3}$, $\text{Scale}_{4}$). We also ablate the weighting method of our temporal localization score ($\text{Weight}_{\text{Size}}$) and the querying space of open-vocabulary querying ($\text{Latent}_{\text{Query}}$). As illustrated above, our approach outperforms these ablations -- both spatially and temporally.}
\label{fig:ablations_agg}
\end{figure*}

\smallskip \noindent \textbf{Temporal Localization Score Weighting}
For this ablation we compute the size of each Gaussian and weight our relevancy score by each Gaussian's size. As illustrated in Table \ref{tab:add-ablations}  and in Figure \ref{fig:ablations_agg} using the cardinality of the relevant Gaussians outperforms weighting by the Gaussians' size.

\smallskip \noindent \textbf{Open-Vocabulary Querying Space}
For this ablation we perform open-vocabulary querying in the latent feature space. We use our pre-trained scene specific encoder to encode the input text query to latent space. As illustrated in Table \ref{tab:add-ablations} and in Figure \ref{fig:ablations_agg} querying in the original feature space outperforms latent space querying.

A quantitative evaluation over these additional ablations are reported in Table \ref{tab:add-ablations} and a qualitative comparison is shown in Figure \ref{fig:ablations_agg}. As illustrated in the table, our averaging of the features 
, weighting the tempral localization scores by cardinality and open-vocabulary querying in the feature space improves temporal and spatial localization.  

\begin{table}[!t]
\vspace{-8pt}
\centering
\resizebox{\linewidth}{!}{
\begin{tabular}{lcccccc}
\toprule
 Method & $\text{vAP}\uparrow$ & $\text{vIOU}\uparrow$&  $\text{tIOU}\uparrow$ & $\text{tRec}\uparrow$ & $\text{tPrec}\uparrow$ & $\text{tAP}\uparrow$\\
\midrule
$\text{Scale}_{0}$ & $44.0\pm1.0$ &$14.1\pm3.2$ & $46.0\pm0.0$ & $78.1\pm0.0$ & $52.2\pm0.0$& $59.2\pm0.0$ \\
$\text{Scale}_{1}$ & $46.1\pm1.1$ &$15.8\pm3.7$&  $48.2\pm0.0$ & $80.7\pm0.0$ &  $54.7\pm0.0$ & $62.4\pm0.0$\\ 
$\text{Scale}_{2}$ & $47.9\pm1.1$ &$16.7\pm3.9$& $50.0\pm0.0$ &$80.0\pm0.0$ & $56.2\pm0.0$  & $63.3\pm0.0$ \\
$\text{Scale}_{3}$ & $46.8\pm1.1$ &$16.4\pm4.1$& $48.9\pm0.0$ & $78.7\pm0.0$ & $55.9\pm0.0$ & $62.6\pm0.0$\\
$\text{Scale}_{4}$ & $44.9\pm1.0$ &$15.8\pm5.1$&  $46.9\pm0.0$ & $78.0\pm0.0$ &  $53.5\pm0.0$ & $61.0\pm0.0$\\ 
Concat & $47.2\pm1.1$ &$16.1\pm4.2$& $49.4\pm0.0$ &$80.0\pm0.0$ & $56.8\pm0.0$ & $61.0\pm0.0$ \\
$\text{Weight}_{\text{Size}}$ & $50.7\pm1.2$ & $18.8\pm2.7$& $53.1\pm0.0$ & $\mathbf{87.1\pm0.0}$ & $58.2\pm0.0$&$66.3\pm0.0$ \\
$\text{Latent}_{\text{Query}}$ & $50.3\pm0.9$ & $15.9\pm3.7$& $52.7\pm0.0$ & $86.9\pm0.0$ & $57.6\pm0.0$&$65.9\pm0.0$ \\
Ours & $\mathbf{58.7\pm0.6}$ &$\mathbf{25.6\pm4.6}$ & $\mathbf{60.8\pm0.0}$ & $83.0\pm0.0$ & $\mathbf{68.5\pm0.0}$ & $\mathbf{72.6\pm0.0}$ \\   
\bottomrule
\end{tabular}
}
\caption{\textbf{Additional Ablations}, we ablate our feature extraction mechanism , our temporal localization score weighting method and the querying space of open-vocabulary querying. We compare against a concatenation of multi-scale features (Concat), five different resolutions of single scale features , weighting the temporal localization scores by Gaussian sizes ($\text{Weight}_{\text{Size}}$) and open-vocabulary querying in the latent space ($\text{Latent}_{\text{Query}}$). See Section \ref{sec:ab-add} for additional details. Best results are highlighted in bold.}
\label{tab:add-ablations}
\end{table}

\section{Free-View Results}
\begin{figure*} %
\centering
\jsubfig{\includegraphics[scale=0.12]{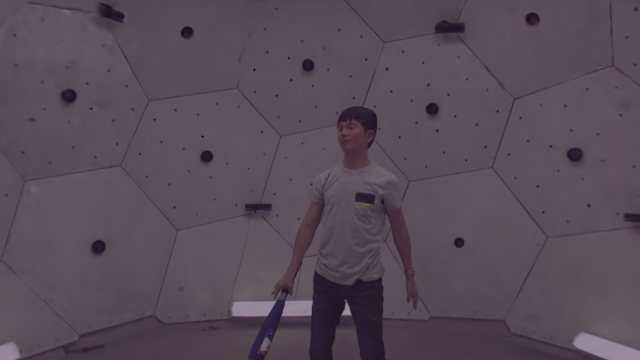}}{}
\jsubfig{\includegraphics[scale=0.12]{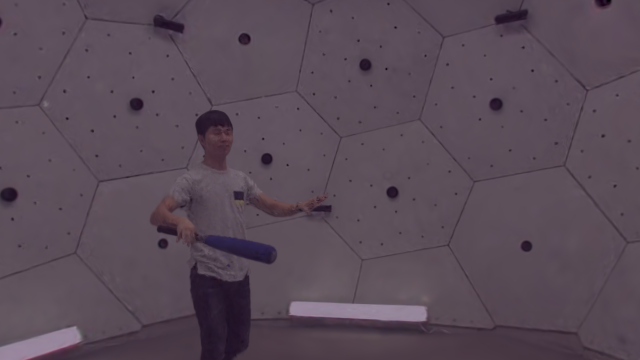}}{}
\jsubfig{\fcolorbox{red}{red}{\includegraphics[scale=0.12]{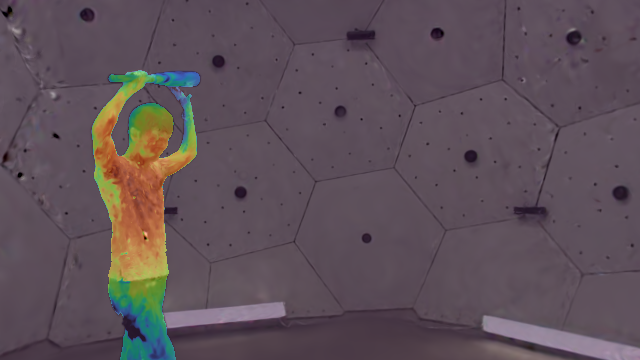}}}{}
\jsubfig{\fcolorbox{red}{red}{\includegraphics[scale=0.12]{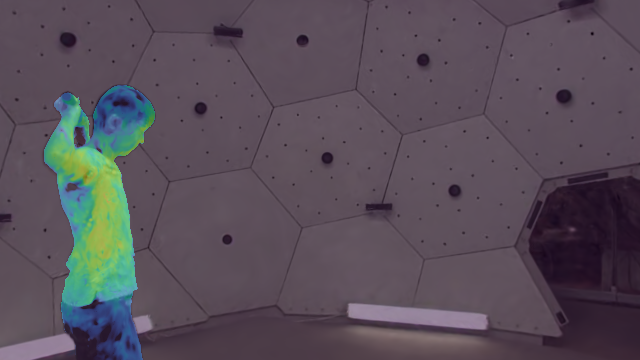}}}{}
\jsubfig{\fcolorbox{red}{red}{\includegraphics[scale=0.12]{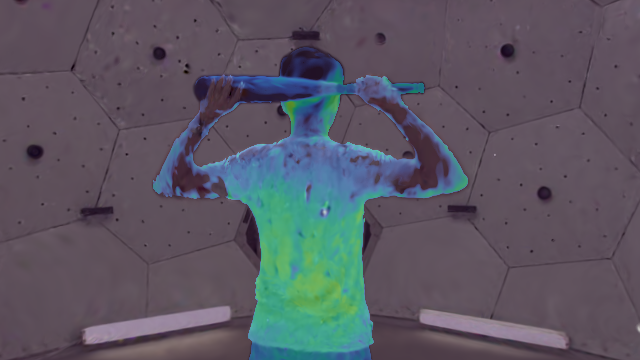}}}{}
\jsubfig{\includegraphics[scale=0.12]{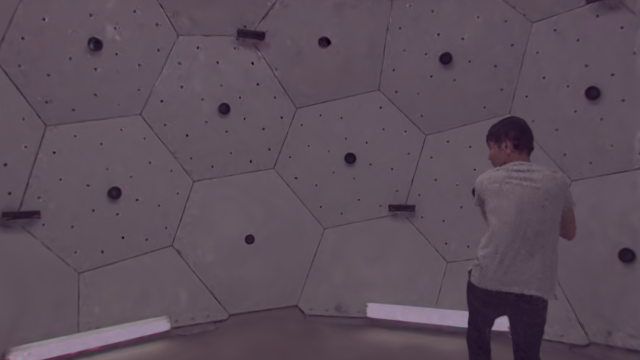}}{}
\\\vspace{2pt}
{ Input query: \emph{A person stretching}}
\\
\jsubfig{\fcolorbox{red}{red}{\includegraphics[scale=0.12]{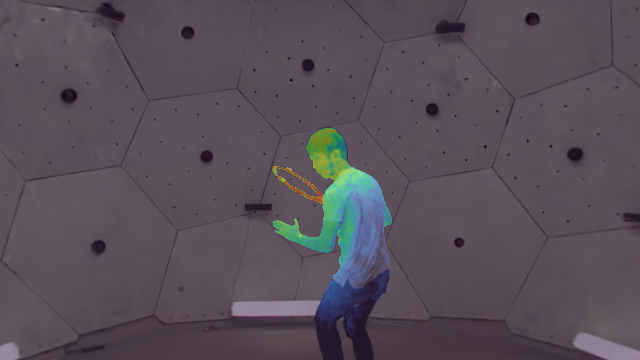}}}{}
\jsubfig{\fcolorbox{red}{red}{\includegraphics[scale=0.12]{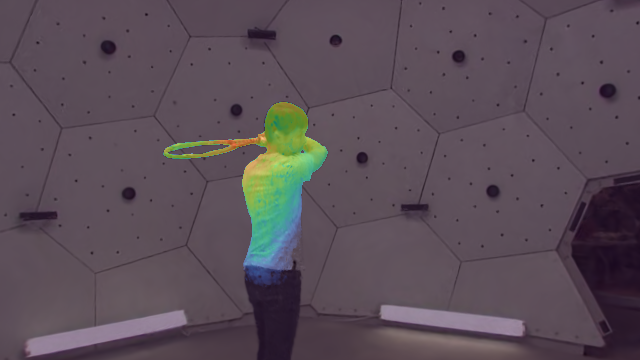}}}{}
\jsubfig{\fcolorbox{red}{red}{\includegraphics[scale=0.12]{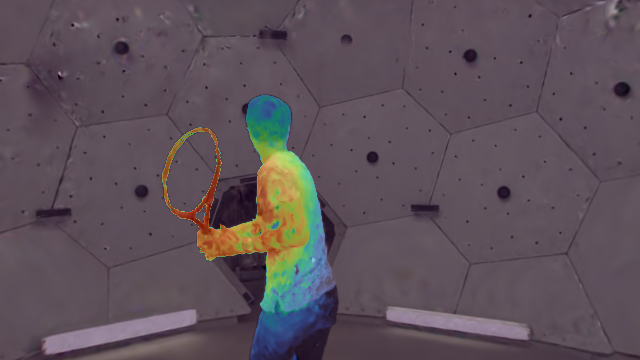}}}{}
\jsubfig{\fcolorbox{red}{red}{\includegraphics[scale=0.12]{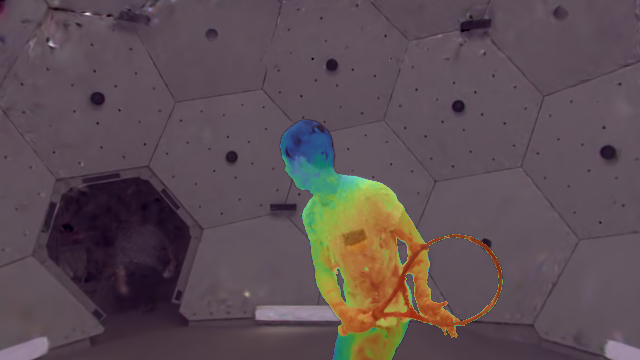}}}{}
\jsubfig{\includegraphics[scale=0.12]{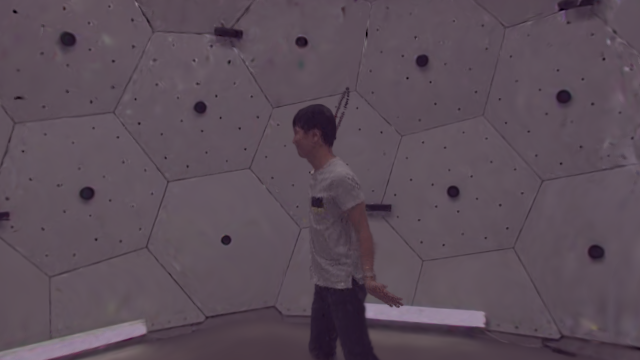}}{}
\jsubfig{{\includegraphics[scale=0.12]{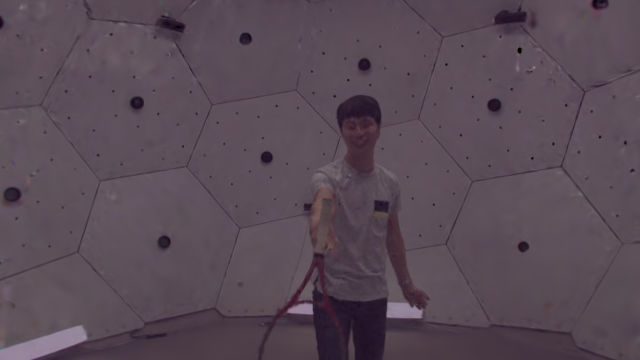}}}{}%
\\
\vspace{2pt}
{ Input query: \emph{A person playing tennis}}
\vspace{-5pt}
\caption{\textbf{Free-View Results.} We show novel camera view relevancy maps with the frames that are localized temporally in red. Please refer to the accompanying interactive visualizations for additional free-view results (e.g., depicting a bullet-time effect).}%

\label{fig:free_view}
\end{figure*}

We optimize a volumetric representation of a scene, thus allowing the user to render novel camera views. As we distill our spatio-temporal features onto this volumetric representation, our method can produce novel view spatio-temporal relevancy maps for input text queries. This is illustrated in Figure \ref{fig:free_view}.

\begin{table*}[t]
\centering
\resizebox{\linewidth}{!}{
\begin{tabular}{llcccccccccccc}
\toprule
  & & \multicolumn{5}{c}{$\text{CGSTVG}$} & \multicolumn{5}{c}{$\text{Ours}$} \\
 Scene & Query & $\text{vAP}$ &  $\text{vIOU}$ & $\text{tIOU}$ & $\text{tRec}$ & $\text{tPrec}$ & $\text{tAP}$ & $\text{vAP}$ &  $\text{vIOU}$ & $\text{tIOU}$ & $\text{tRec}$ & $\text{tPrec}$ & $\text{tAP}$  \\
\midrule
\multirow{4}{*}{Basketball} & A ball flying in the air & $3.1 \pm 0.2$& $2.3 \pm 2.2$& $23.5\pm5.5$&$57.5\pm19.0$&$29.8\pm8.6$ & $37.8\pm 7.5$ &$\mathbf{76.8 \pm 1.1}$& $\mathbf{7.4 \pm 2.5}$ & $\mathbf{81.0 \pm 0.0}$ & $\mathbf{95.9 \pm 0.0}$ & $\mathbf{84.0\pm 0.0}$ & $\mathbf{89.5\pm 0.0}$ \\
 & A person catching & $39.0 \pm 6.2$&$3.6 \pm 5.2$&$8.0\pm10.6$&$24.7\pm31.2$&$9.1\pm11.8$ & $13.2\pm 5.2$ & $\mathbf{52.9 \pm 0.4}$& $\mathbf{7.7 \pm 2.4}$&$\mathbf{55.7 \pm 0.0}$ & $\mathbf{78.9 \pm 0.0}$ & $\mathbf{65.5\pm 0.0}$ & $\mathbf{71.6\pm 0.0}$\\
 & A person holding the ball & $42.6 \pm 7.8$&$\mathbf{41.4 \pm 8.0}$&$54.9\pm12.2$&$66.3\pm20.7$&$80.1\pm9.5$ & $70.1\pm 11.3$ & $\mathbf{87.3 \pm 1.2}$&$40.0 \pm 6.2$&$\mathbf{90.9 \pm 0.0}$ & $\mathbf{99.0 \pm 0.0}$ & $\mathbf{91.7 \pm 0.0}$ & $\mathbf{95.2\pm 0.0}$ \\
 & A person throwing & $10.0 \pm 4.5$&$9.5 \pm 3.7$ &$15.6\pm7.3$&$\mathbf{59.2\pm35.1}$&$17.2\pm6.0$ & $26.2\pm 5.6$ & $\mathbf{30.3 \pm 0.6}$& $\mathbf{11.0 \pm 2.6}$ &$\mathbf{32.5 \pm 0.0}$ & $56.0 \pm 0.0$ & $\mathbf{43.7 \pm 0.0}$ & $\mathbf{49.2\pm 0.0}$ \vspace{3pt} \\
 \multirow{3}{*}{Boxes} & A person moving boxes &  $28.8 \pm 4.2$ & $27.9 \pm 1.9$& $36.6\pm3.2$ & $36.6\pm3.2$ & $\mathbf{100.0\pm0.0}$ & $53.6\pm 4.3$ & $\mathbf{62.3 \pm 0.6}$& $\mathbf{29.2\pm4.5}$ &$\mathbf{62.2 \pm 0.0}$ & $\mathbf{70.2 \pm 0.0}$ & $88.6 \pm 0.0$ & $\mathbf{78.3\pm 0.0}$\\
 &A person picking up a box & $18.3 \pm3.1$ &$17.8 \pm 7.0$ & $22.7\pm7.6$ & $34.8\pm12.6$ & $39.4\pm11.8$  & $36.3\pm 7.9$ & $\mathbf{58.3 \pm 0.5}$& $\mathbf{27.9 \pm 6.7}$& $\mathbf{60.2 \pm 0.0}$ & $\mathbf{98.1 \pm 0.0}$ & $\mathbf{60.9 \pm 0.0}$ & $\mathbf{75.1\pm 0.0}$\\
 & A person putting down a box & $\mathbf{29.8 \pm 8.2}$ &$\mathbf{28.9 \pm 9.0}$& $\mathbf{37.9\pm11.4}$ & $\mathbf{94.4\pm11.7}$ & $\mathbf{39.2\pm11.2}$ & $\mathbf{54.7\pm 8.3}$ &$7.1 \pm 0.1$& $3.6 \pm 0.7$&$8.3 \pm 0.0$ & $32.0 \pm 0.0$ & $10 \pm 0.0$ & $15.2\pm 0.0$
 \vspace{3pt} \\
 \multirow{2}{*}{Football} &  A ball flying in the air & $6.2 \pm1.1$ & $5.3 \pm 3.5$&$29.5\pm9.6$ & $65.5\pm2.4$ & $35.6\pm13.5$ & $44.7\pm 6.9$ & $\mathbf{63.6 \pm 1.2}$& $\mathbf{5.4 \pm 2.8}$&$\mathbf{66.0 \pm 0.0}$ & $\mathbf{100.0 \pm 0.0}$ & $\mathbf{66.0 \pm 0.0}$ & $\mathbf{79.5\pm 0.0}$\\
 & A person catching & $4.8 \pm2.2$& $4.5 \pm 2.4$& $8.2\pm4.9$ & $29.5\pm17.9$ & $9.9\pm5.5$ & $14.8\pm 4.1$ & $\mathbf{22.6 \pm 0.3}$&$\mathbf{8.1 \pm 0.9}$&$\mathbf{23.4 \pm 0.0}$ & $\mathbf{100.0 \pm 0.0}$ & $\mathbf{23.4 \pm 0.0}$ & $\mathbf{38.0\pm 0.0}$
  \vspace{3pt} \\
 \multirow{2}{*}{Juggle} & A person walking & $\mathbf{50.2 \pm17.3}$ & $\mathbf{49.2 \pm 19.1}$& $\mathbf{57.3\pm22.2}$ & $82.9\pm5.9$ & $\mathbf{66.9\pm28.6}$ & $\mathbf{70.2\pm 10.3}$ & $46.5 \pm 0.4$& $25.1 \pm 4.3$&$47.8 \pm 0.0$ & $\mathbf{83.1 \pm 0.0}$ & $53.5 \pm 0.0$ & $64.7\pm 0.0$\\
 & A person juggling balls & $54.0 \pm9.7$ & $\mathbf{52.7 \pm 10.9}$& $63.0\pm12.4$ & $69.0\pm9.4$ & $87.8\pm15.6$ & $76.6\pm 11.4$ & $\mathbf{88.2 \pm 0.9}$& $46.4 \pm 4.9$&$\mathbf{90.7 \pm 0.0}$ & $\mathbf{94.5 \pm 0.0}$ & $\mathbf{95.8 \pm 0.0}$& $\mathbf{95.2\pm 0.0}$
  \vspace{3pt} \\
 \multirow{2}{*}{Softball} & A person swinging the bat & $22.6 \pm8.1$ & $21.9 \pm 8.4$& $28.0\pm11.1$ & $64.6\pm26.2$ & $31.9\pm9.5$ & $42.6\pm 9.3$ & $\mathbf{67.8 \pm 0.4}$&$\mathbf{40.4 \pm 6.3}$&$\mathbf{69.2 \pm 0.0}$ & $\mathbf{100.0 \pm 0.0}$ & $\mathbf{69.1 \pm 0.0}$ & $\mathbf{81.8\pm 0.0}$\\
 & A person stretching  & $41.7 \pm20.6$ & $\mathbf{40.1 \pm 22.8}$& $48.8\pm26.9$ & $66.0\pm38.1$ & $55.5\pm28.4$ & $59.3\pm 26.9$ & $\mathbf{77.8 \pm 0.5}$& $36.9 \pm 8.9$&$\mathbf{80.0\pm0.0}$ & $\mathbf{86.7 \pm 0.0}$ & $\mathbf{90.9 \pm 0.0}$ & $\mathbf{88.9\pm 0.0}$
  \vspace{3pt} \\
 \multirow{2}{*}{Tennis} & A person playing tennis & $27.3 \pm8.4$ &$26.3 \pm 6.6$& $40.6\pm11.8$ & $50.4\pm20.4$ & $71.5\pm7.9$ & $56.7\pm 16.9$ & $\mathbf{64.0 \pm 0.5}$& $\mathbf{36.9 \pm 6.6}$&$\mathbf{65.7 \pm 0.0}$ & $\mathbf{76.0 \pm 0.0}$ & $\mathbf{82.9 \pm 0.0}$ & $\mathbf{79.3\pm 0.0}$ \\
 & tennis shot & $34.3 \pm3.1$ & $33.0 \pm 4.9$& $48.9\pm5.2$ & $48.9\pm5.2$ & $\mathbf{100.0 \pm 0.0}$ & $66.1\pm 15.6$ &$\mathbf{71.7 \pm 2.3}$& $\mathbf{36.9 \pm 5.8}$&$\mathbf{73.8 \pm 0.0}$ & $\mathbf{73.8 \pm 0.0}$ & $\mathbf{100.0 \pm 0.0}$ & $\mathbf{85.0\pm 0.0}$ \\
\bottomrule
\end{tabular}
}
\caption{\textbf{Performance Breakdown over the \datasetname{} Benchmark}. We report performance over the spatio-temporal vAP metric, and over three additional metrics quantifying the quality of the temporal localization (tIOU,tRec, tPrec) over each scene and query in our benchmark.  }

\label{tab:comparisons-supp}
\end{table*}

\begin{figure*} %
\centering
\jsubfig{\includegraphics[scale=0.063]{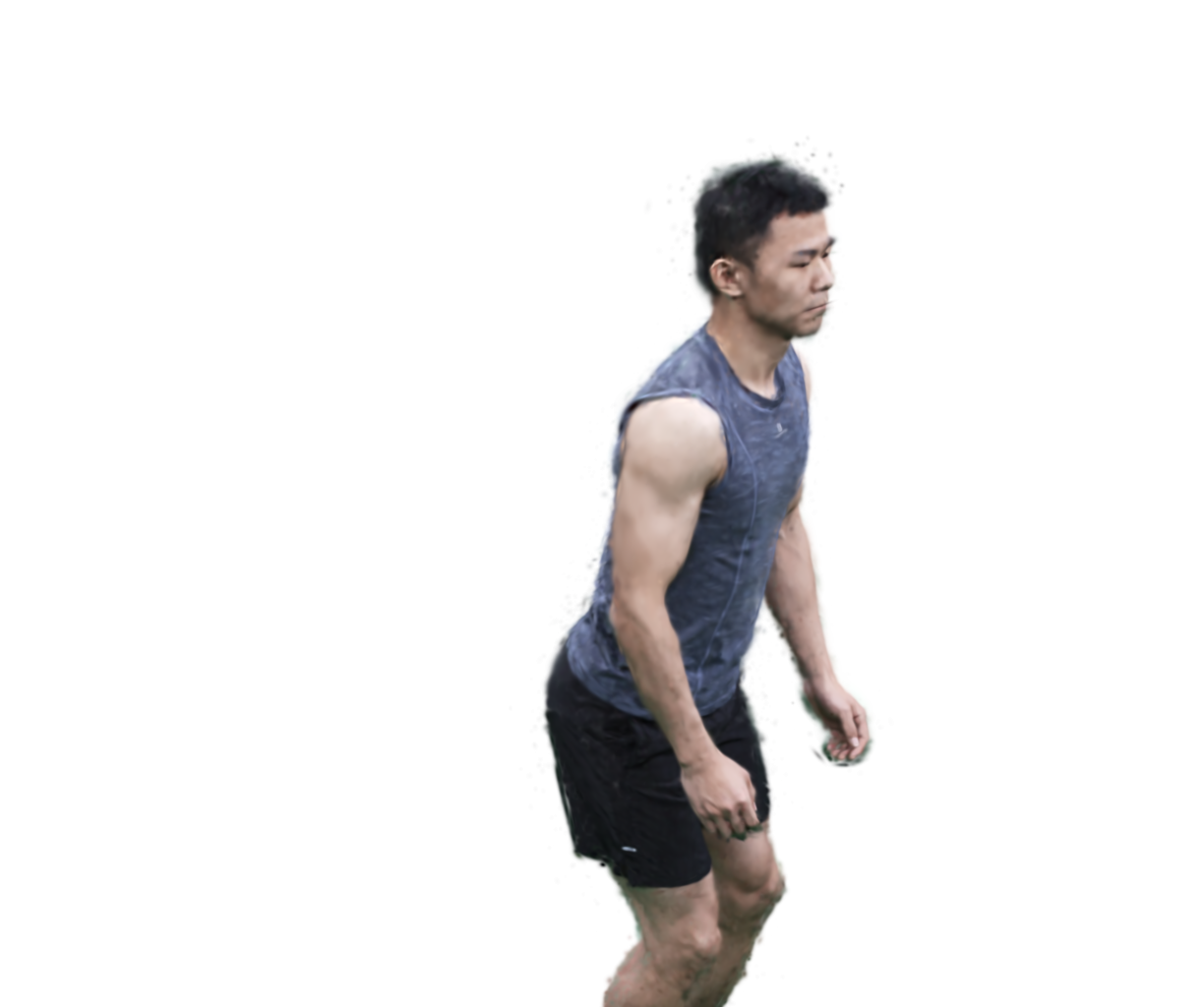}}{}
\jsubfig{\fcolorbox{white}{white}{\includegraphics[scale=0.063]{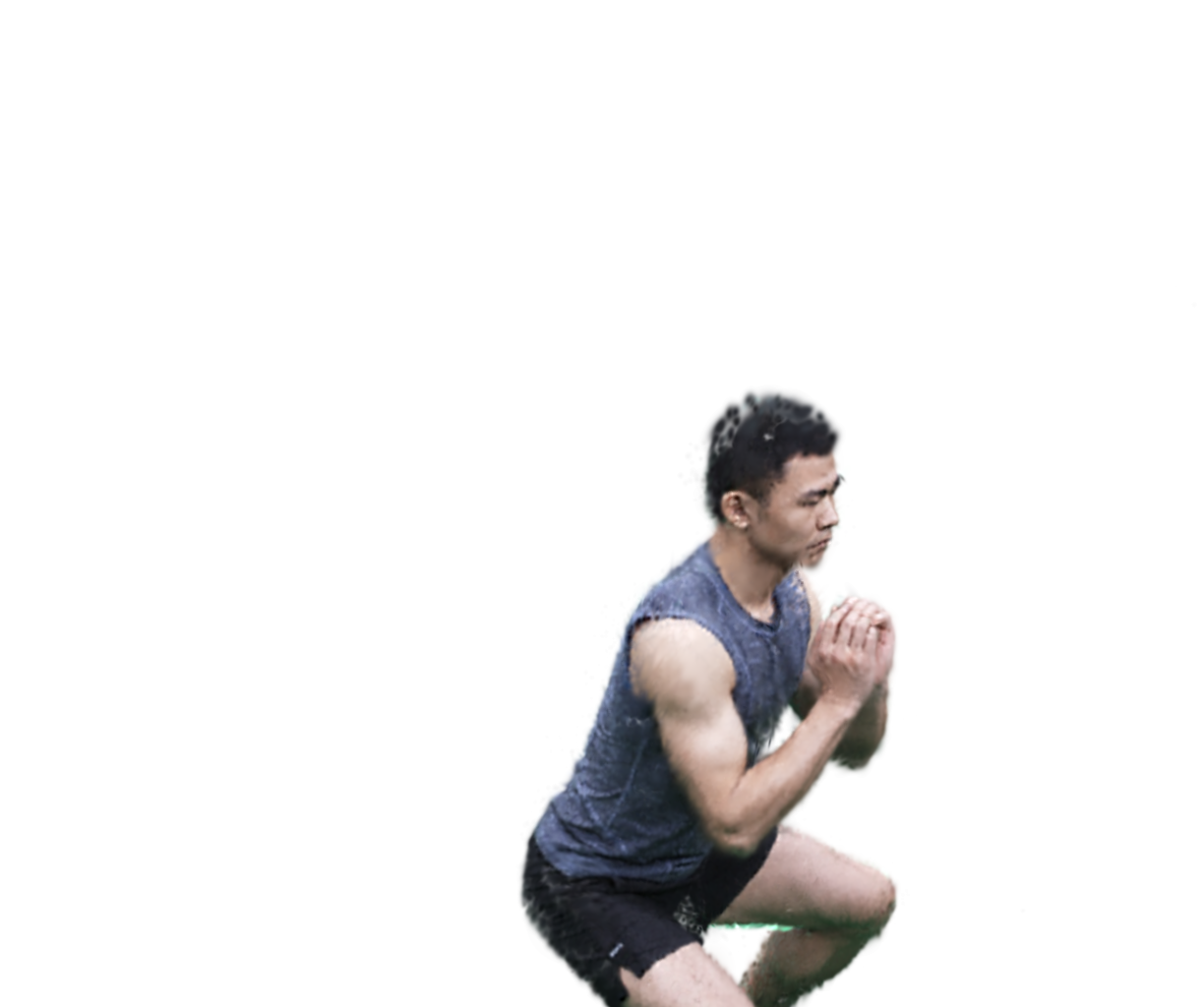}}}{}
\jsubfig{\fcolorbox{white}{white}{\includegraphics[scale=0.063]{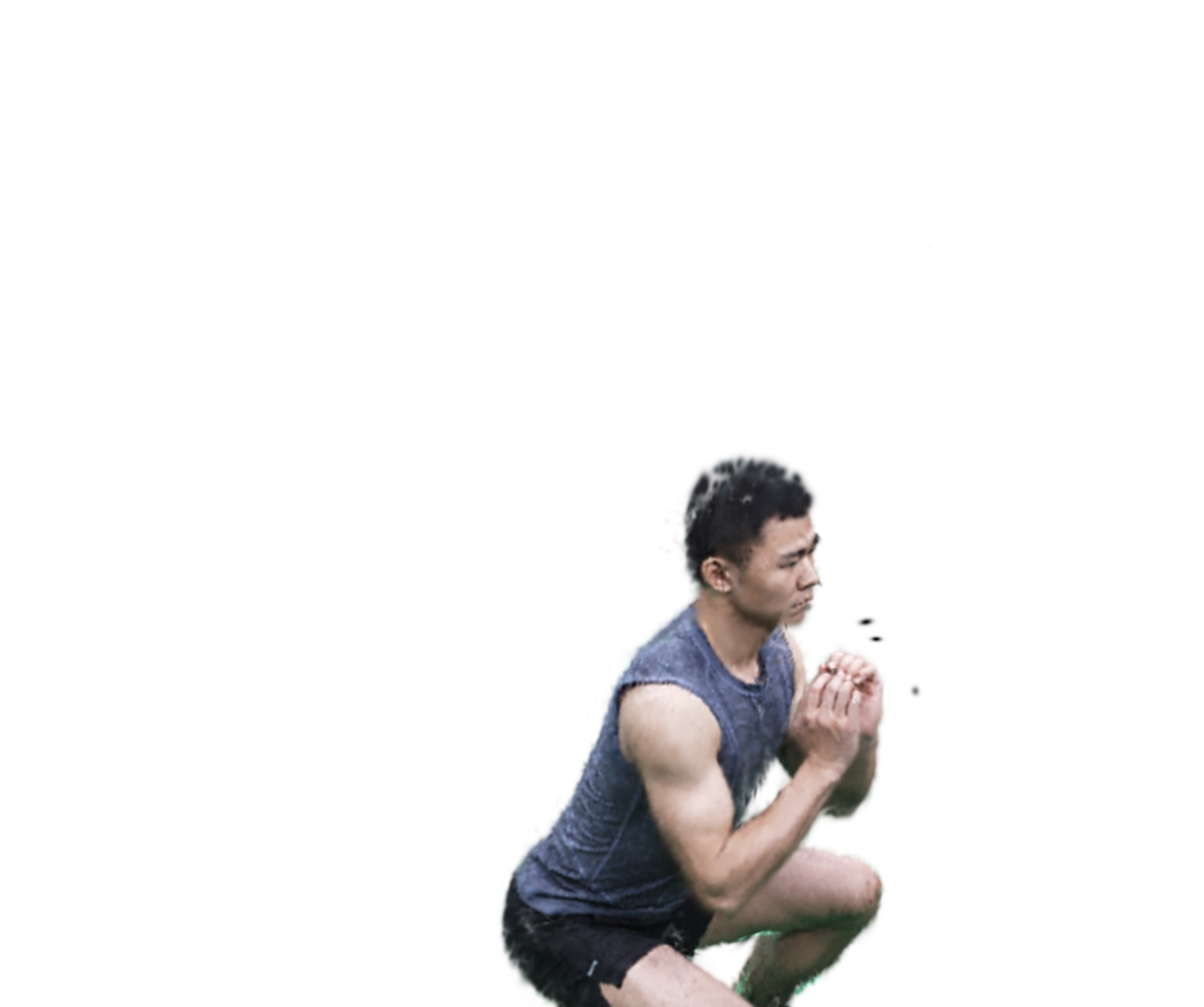}}}{}
\jsubfig{\fcolorbox{white}{white}{\includegraphics[scale=0.063]{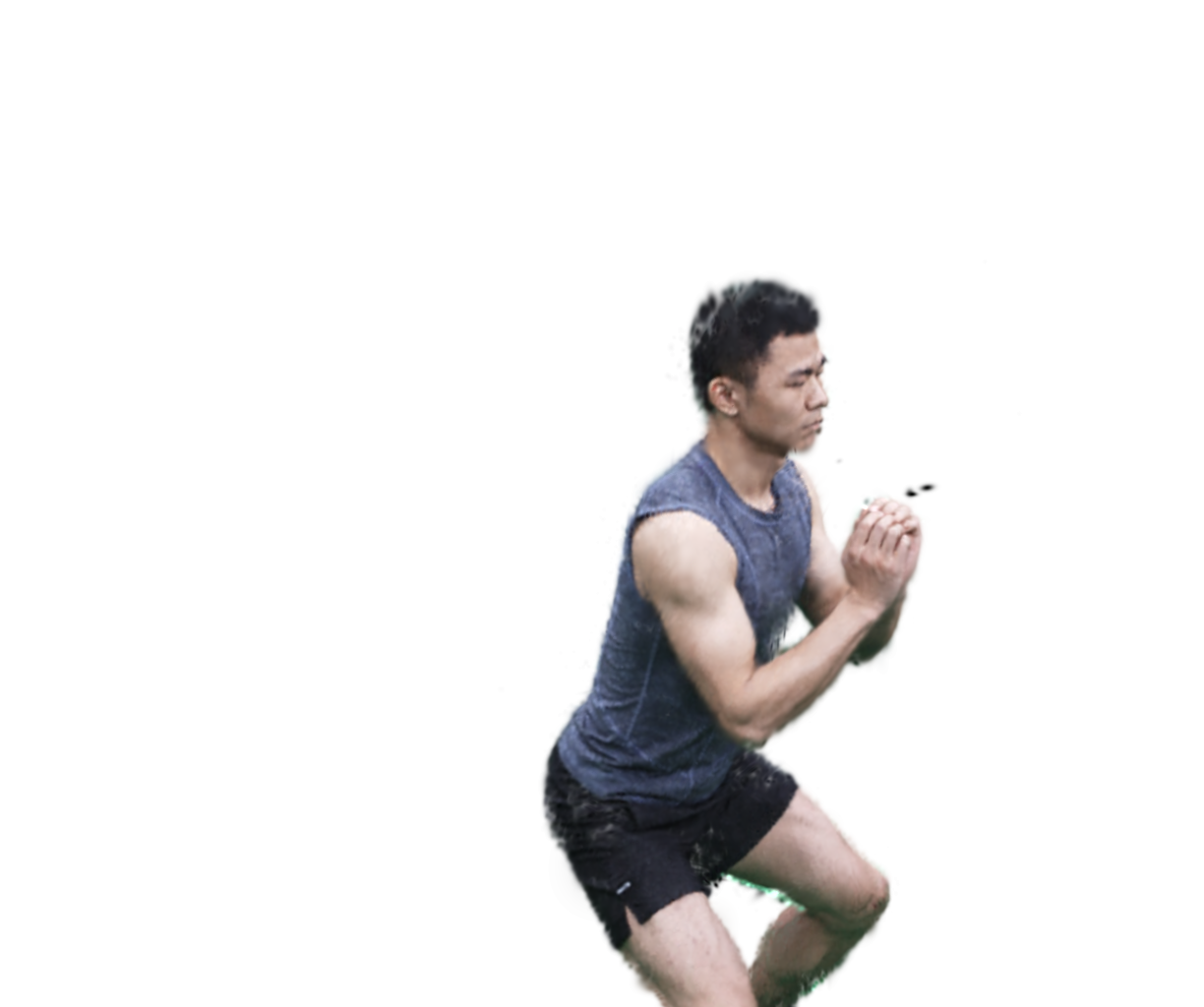}}}{}
\jsubfig{\includegraphics[scale=0.063]{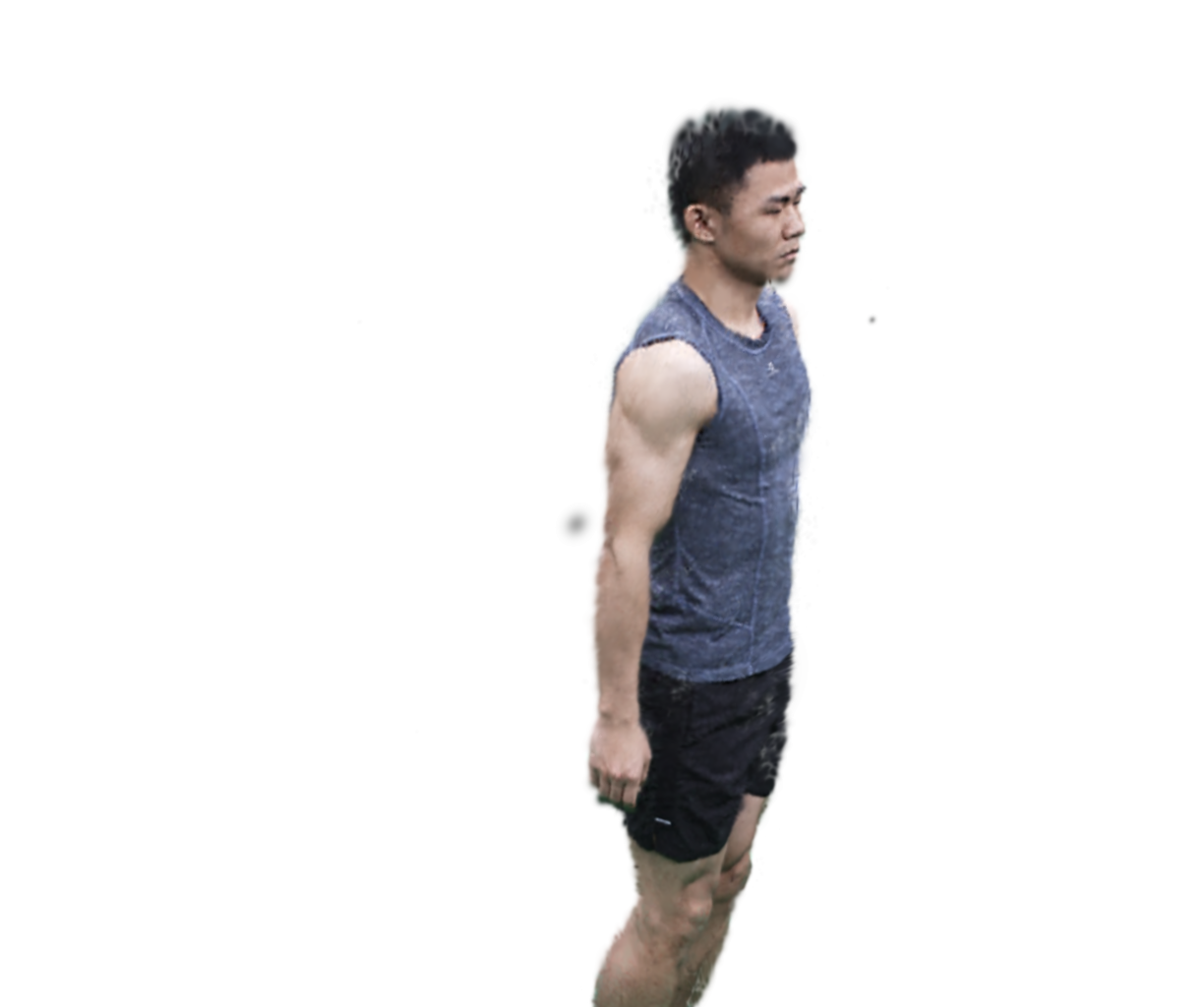}}{}
\jsubfig{\includegraphics[scale=0.063]{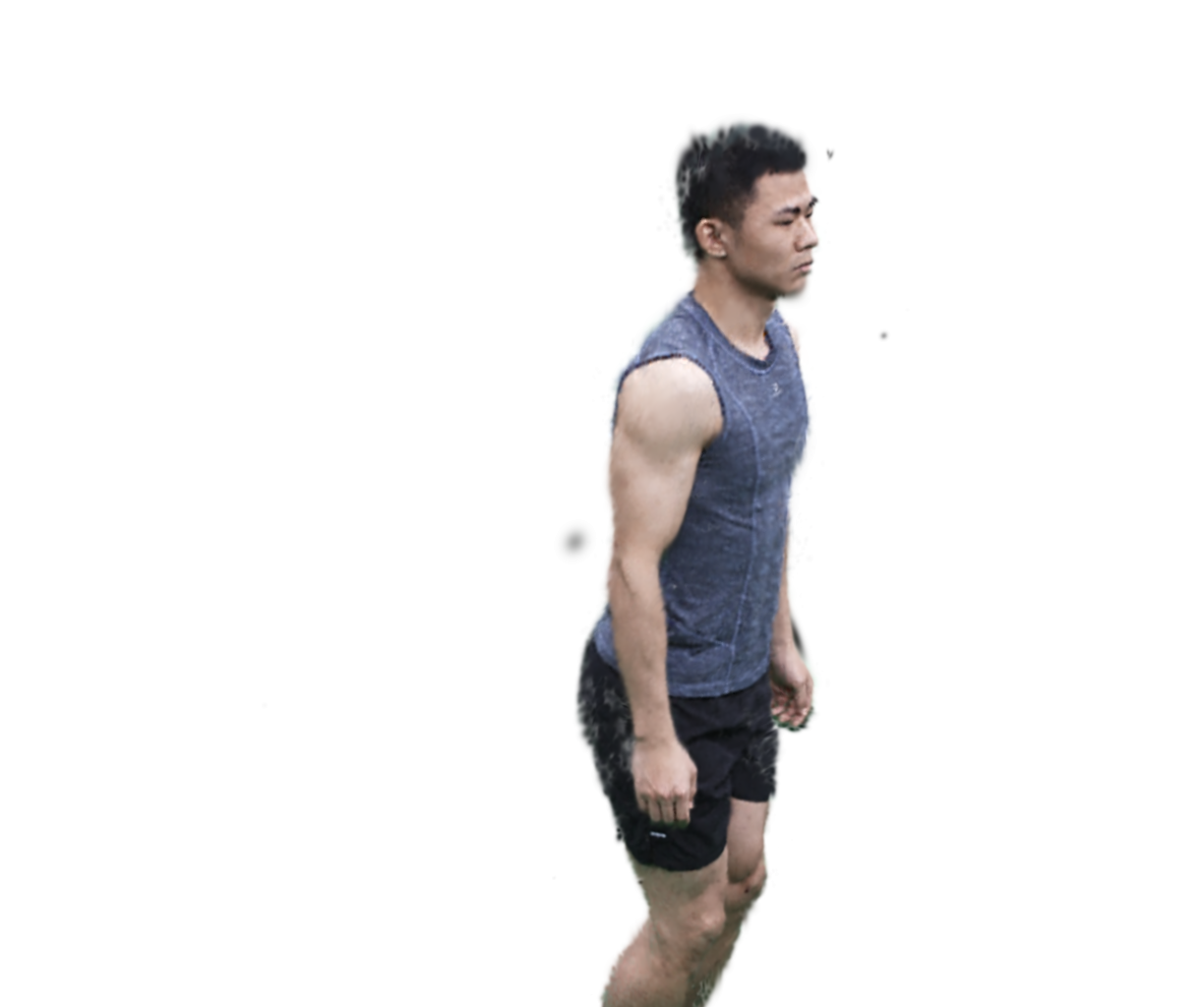}}{}%
\\
\jsubfig{\includegraphics[scale=0.063]{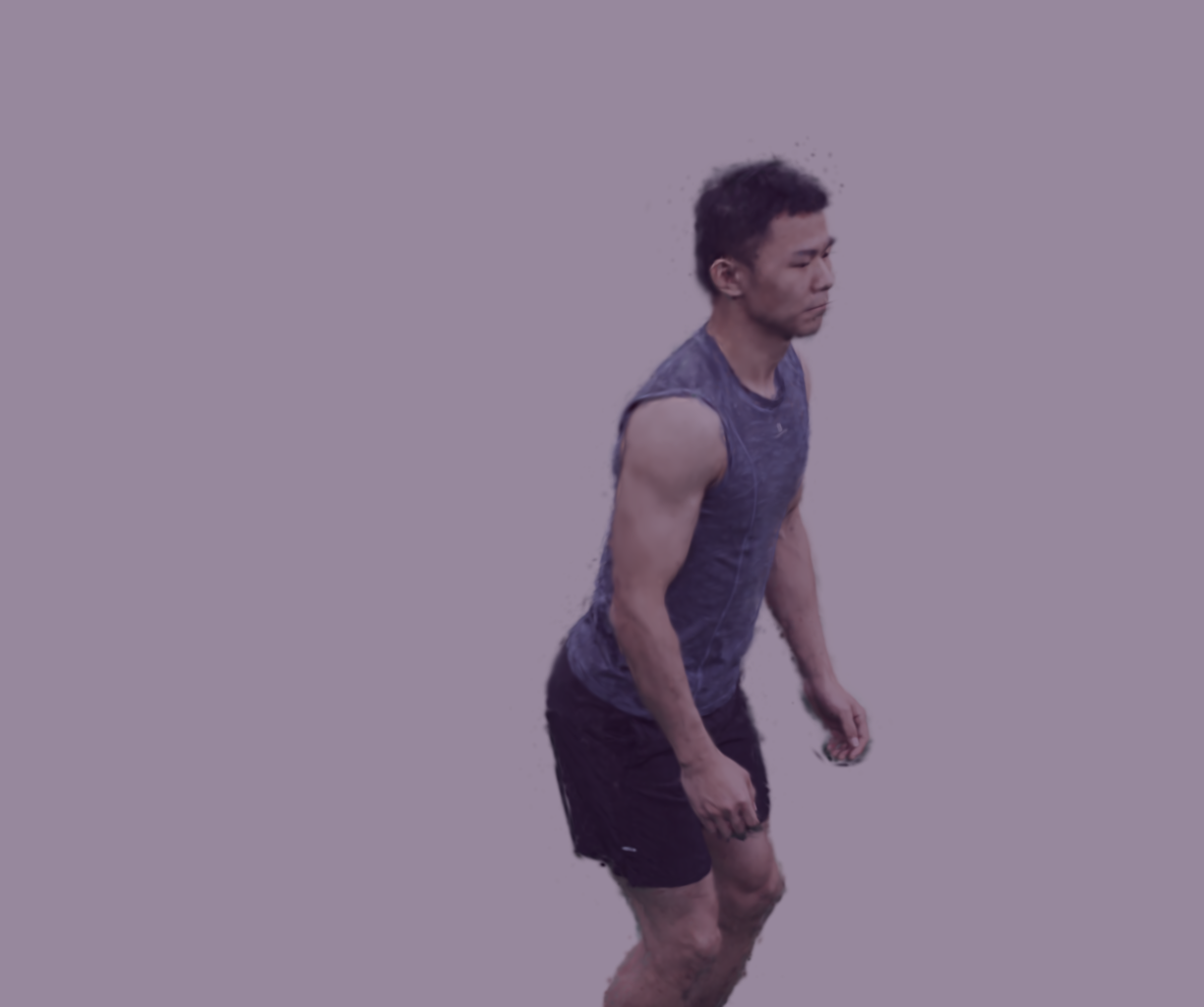}}{}
\jsubfig{\fcolorbox{red}{red}{\includegraphics[scale=0.063]{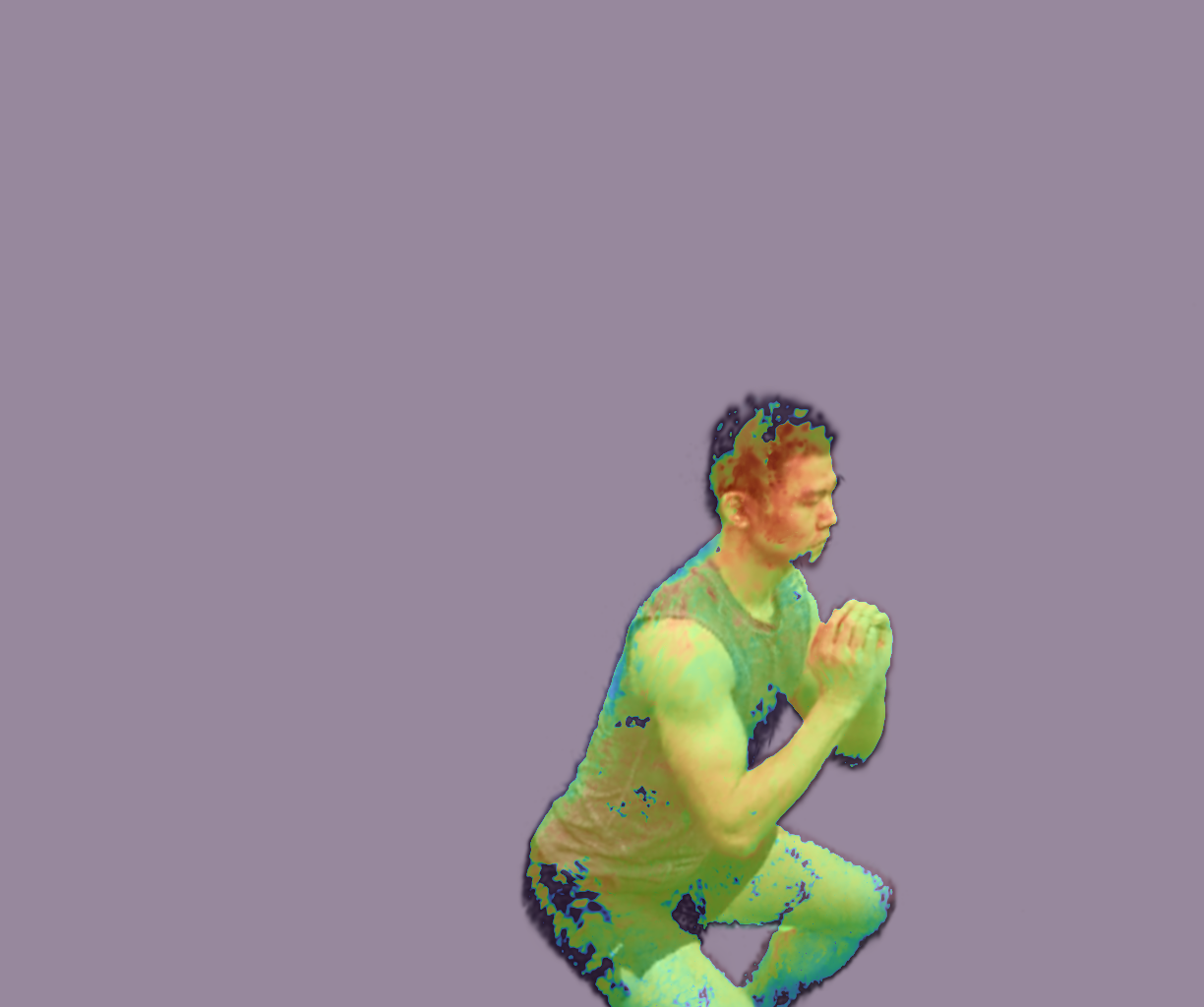}}}{}
\jsubfig{\fcolorbox{red}{red}{\includegraphics[scale=0.063]{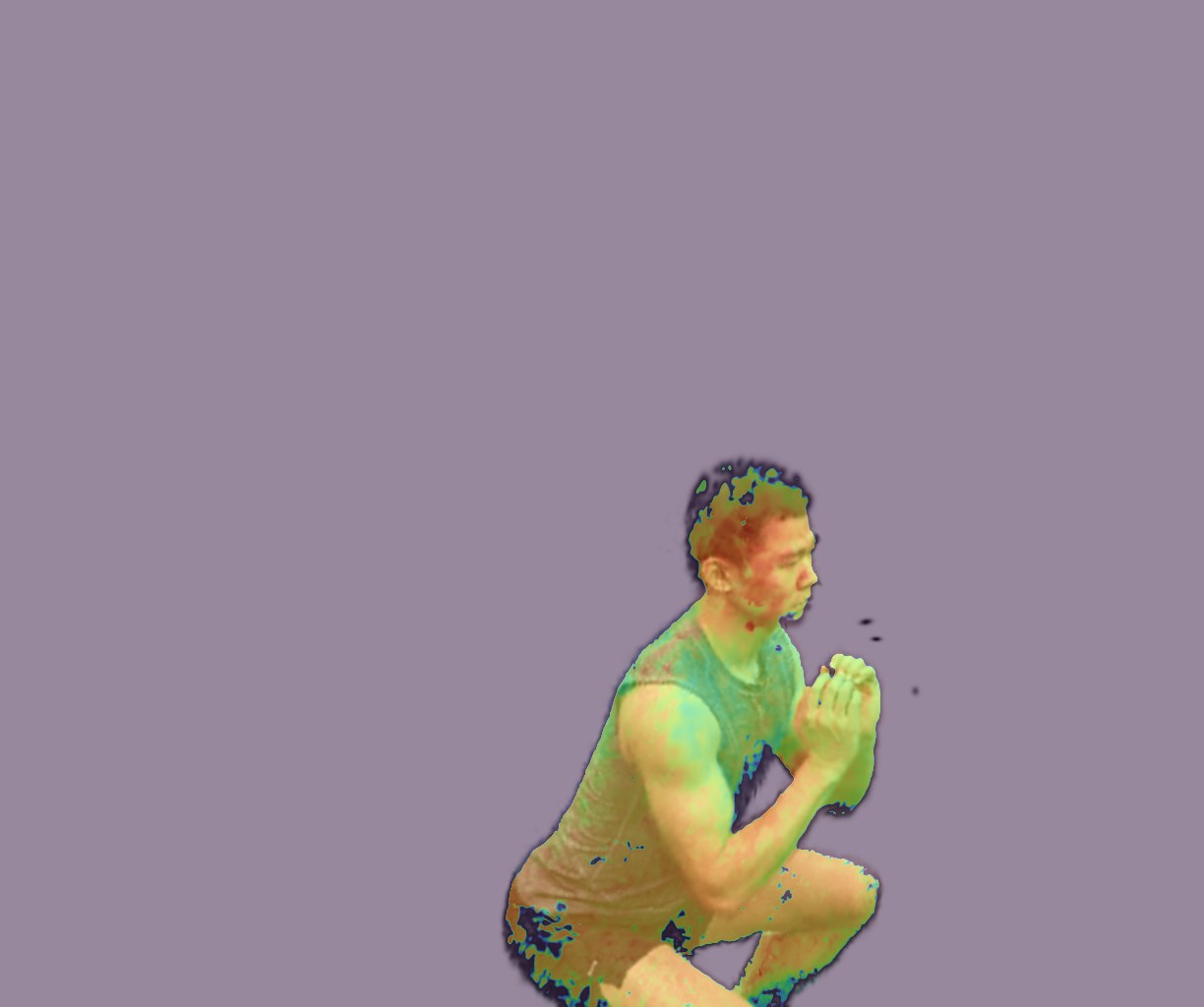}}}{}
\jsubfig{\fcolorbox{red}{red}{\includegraphics[scale=0.063]{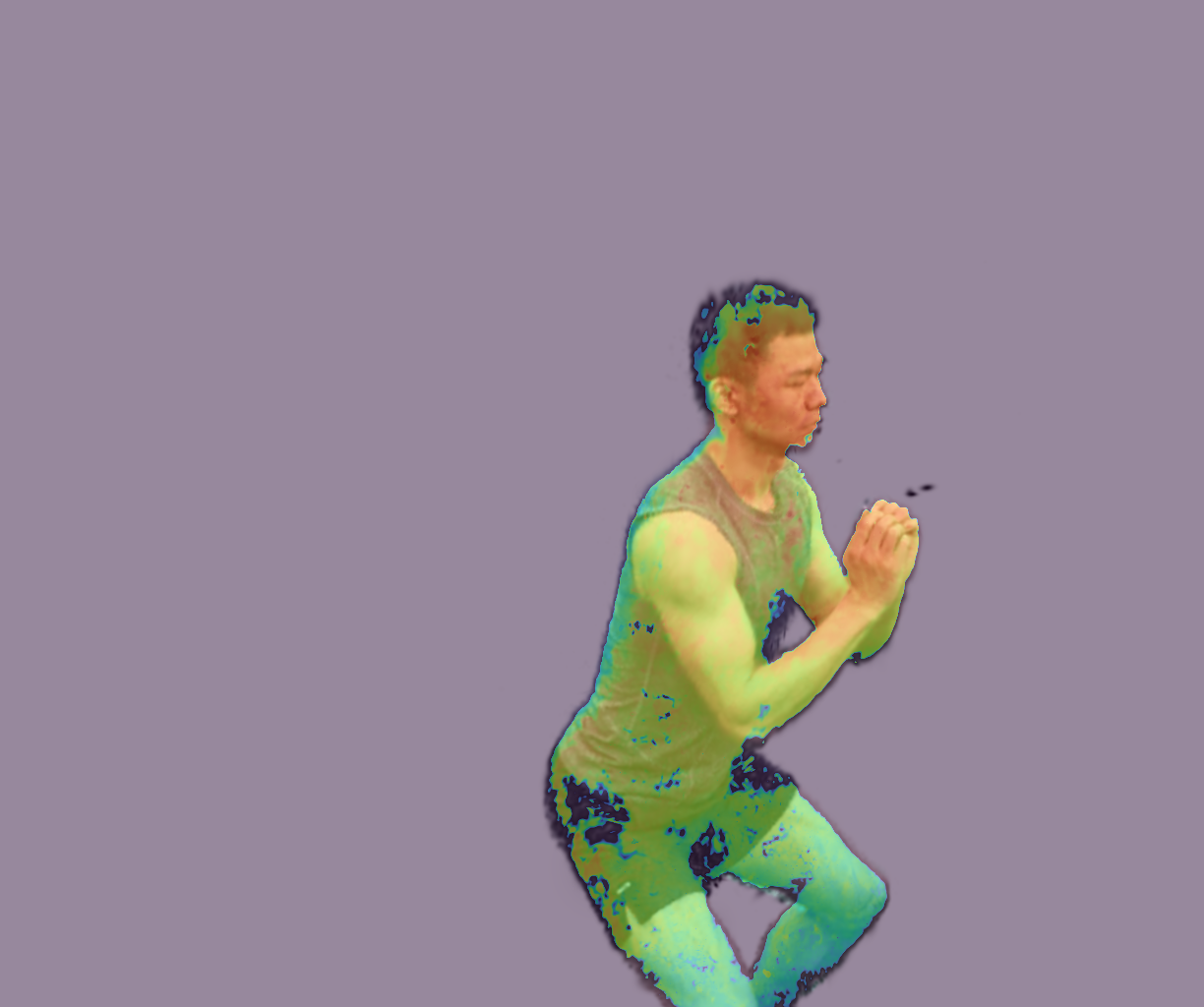}}}{}
\jsubfig{\includegraphics[scale=0.063]{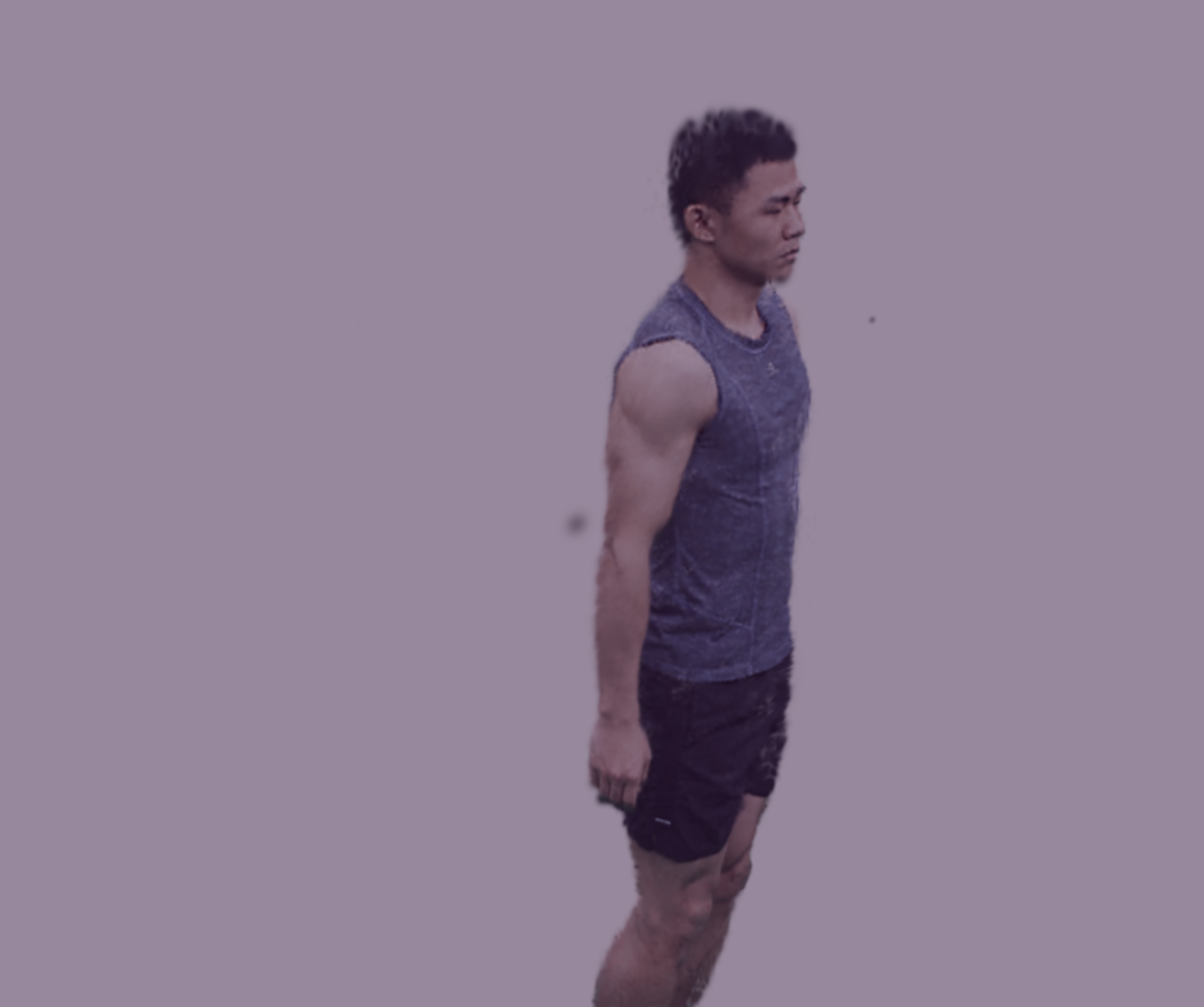}}{}
\jsubfig{\includegraphics[scale=0.063]{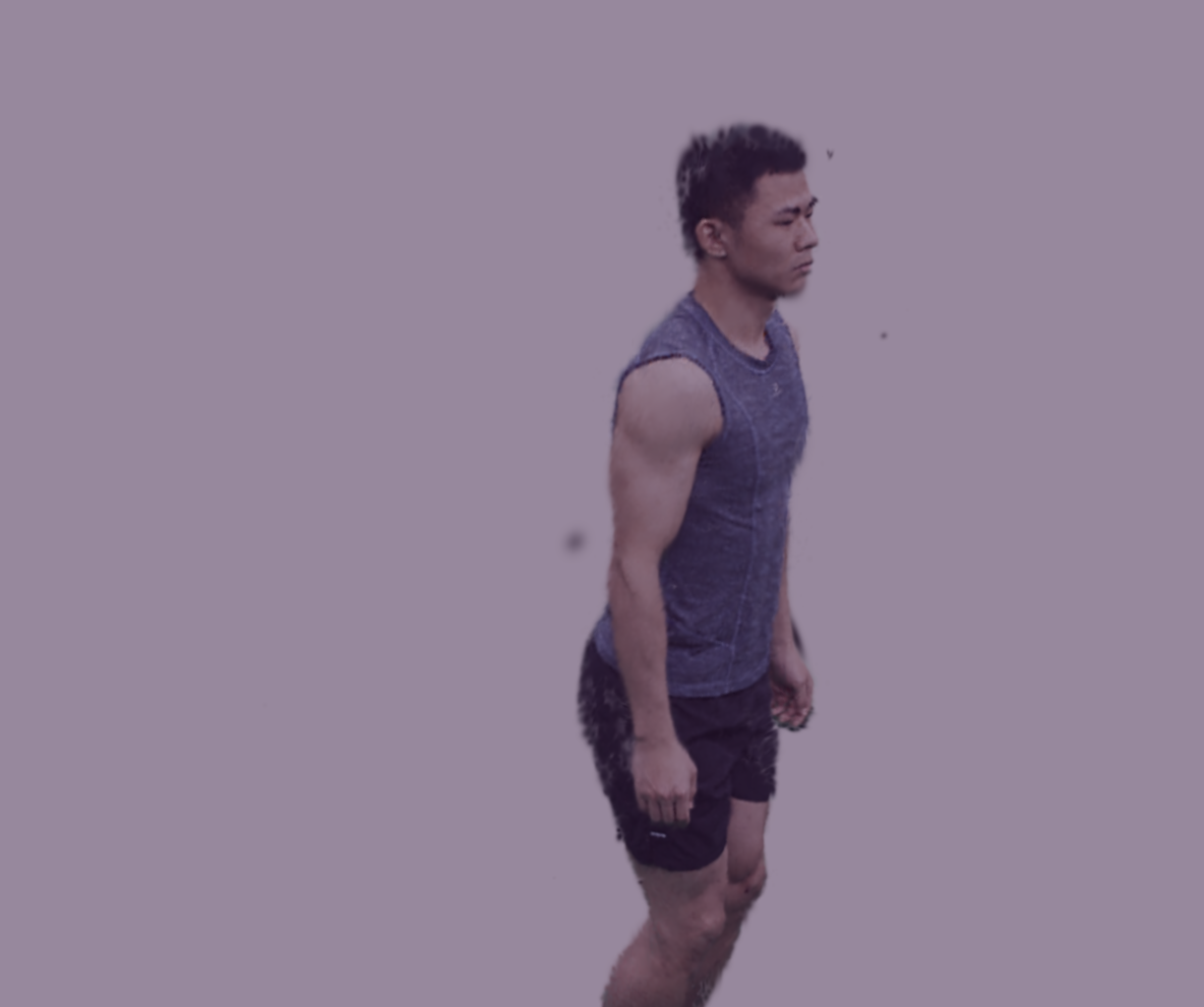}}{}%
\\
\vspace{4pt}
{ Input query: \emph{A person crouching}}
\vspace{4pt}
\\
\jsubfig{\includegraphics[scale=0.063]{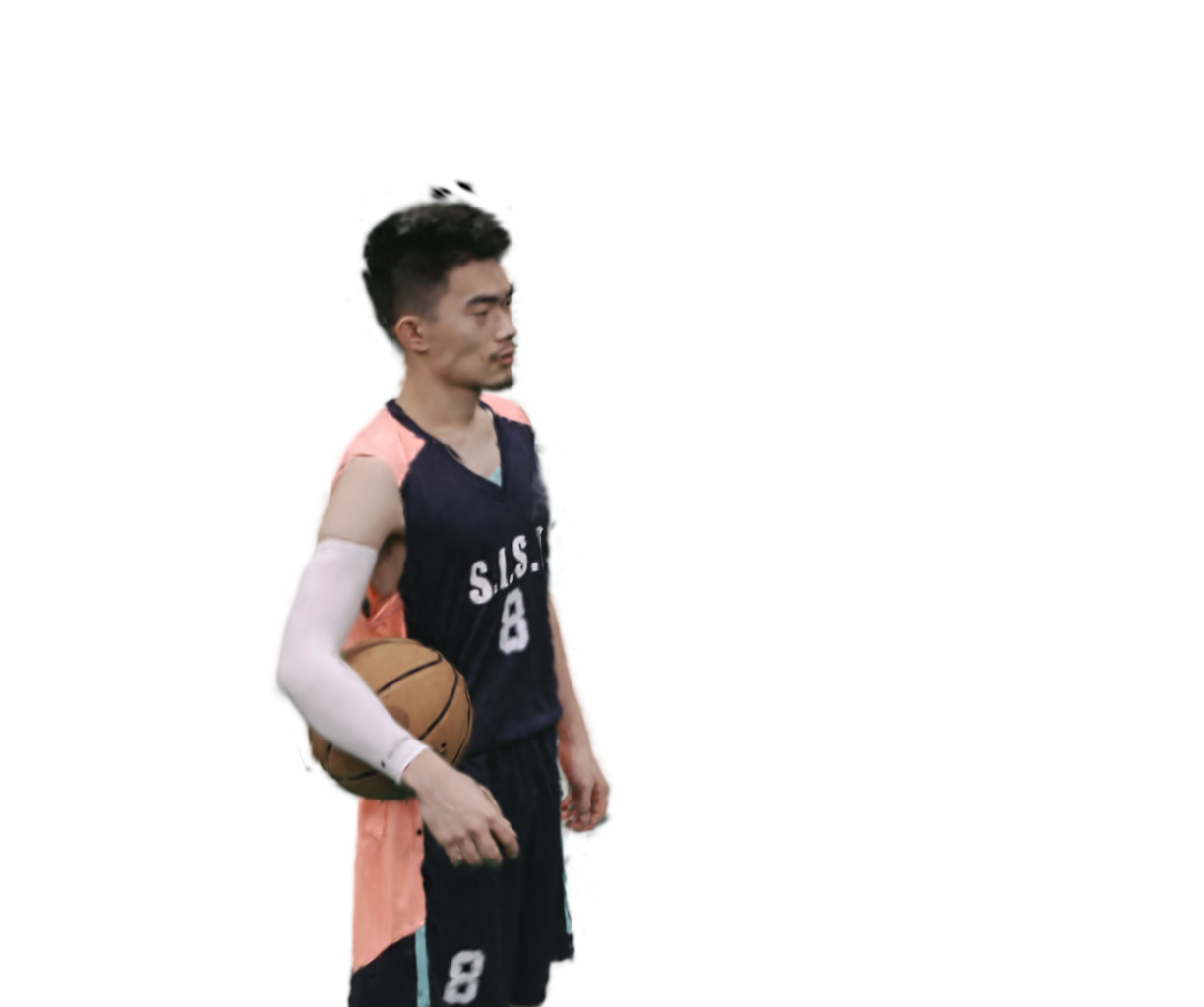}}{}
\jsubfig{{\includegraphics[scale=0.063]{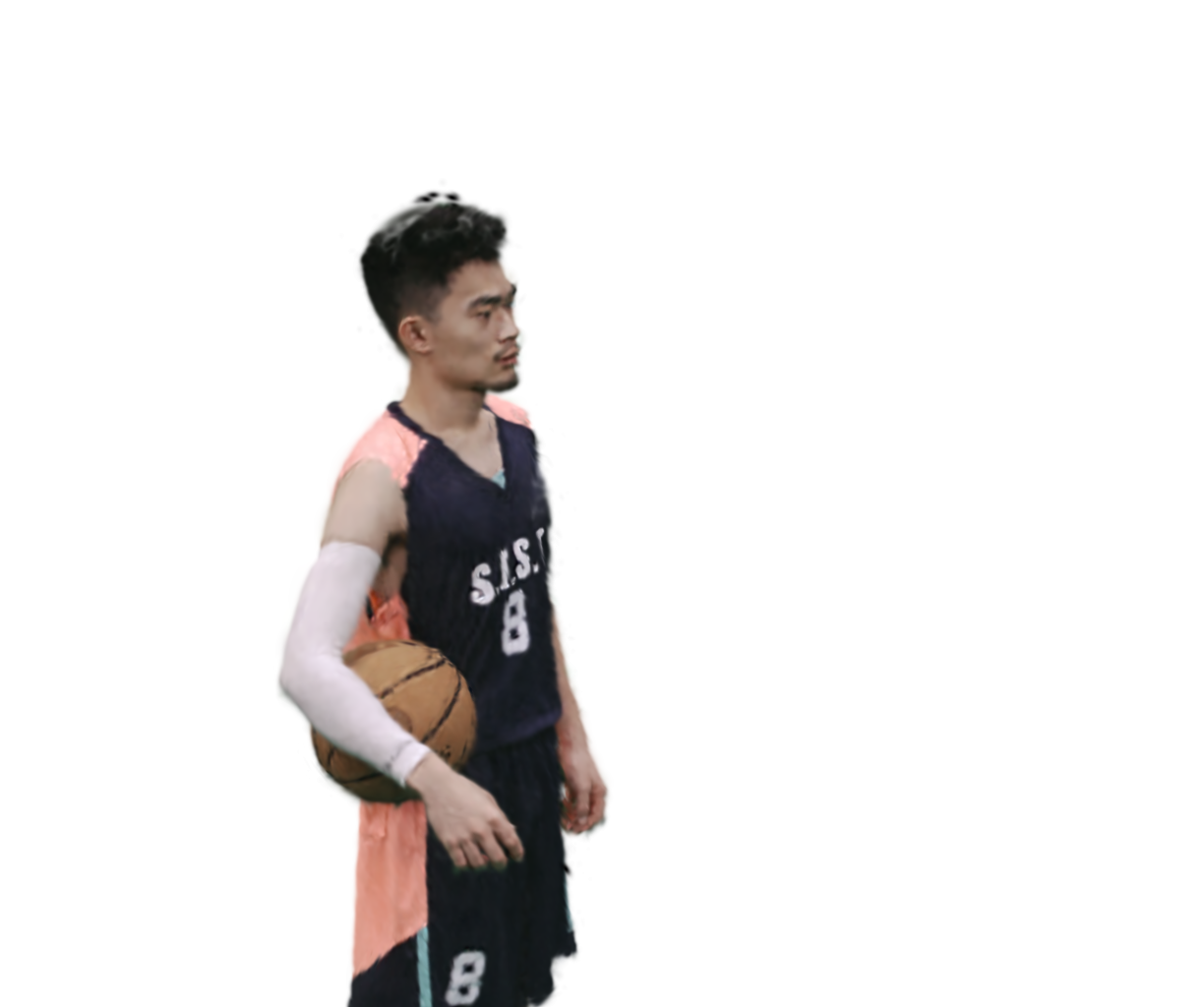}}}{}
\jsubfig{\fcolorbox{white}{white}{\includegraphics[scale=0.063]{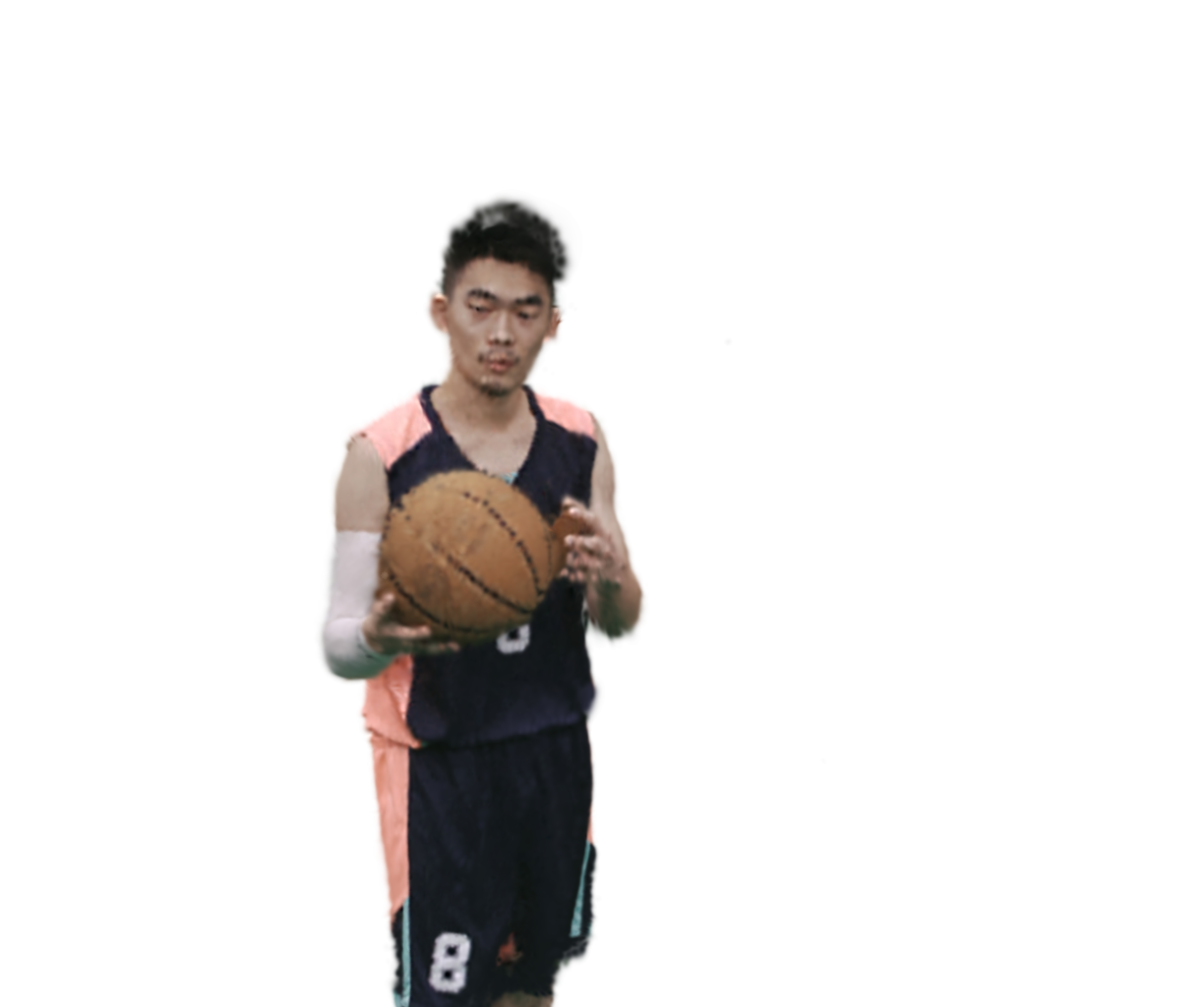}}}{}
\jsubfig{\fcolorbox{white}{white}{\includegraphics[scale=0.063]{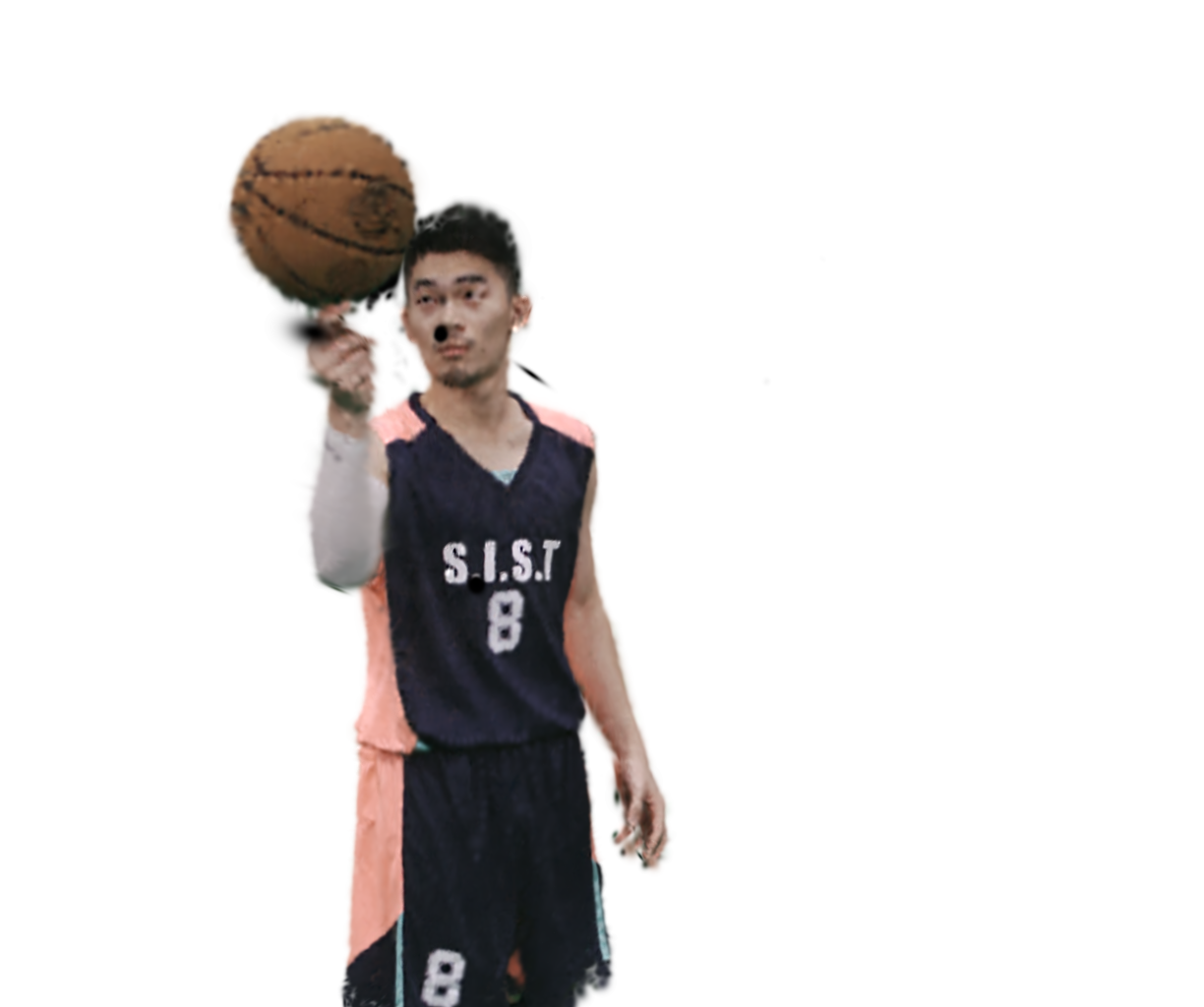}}}{}
\jsubfig{\fcolorbox{white}{white}{\includegraphics[scale=0.063]{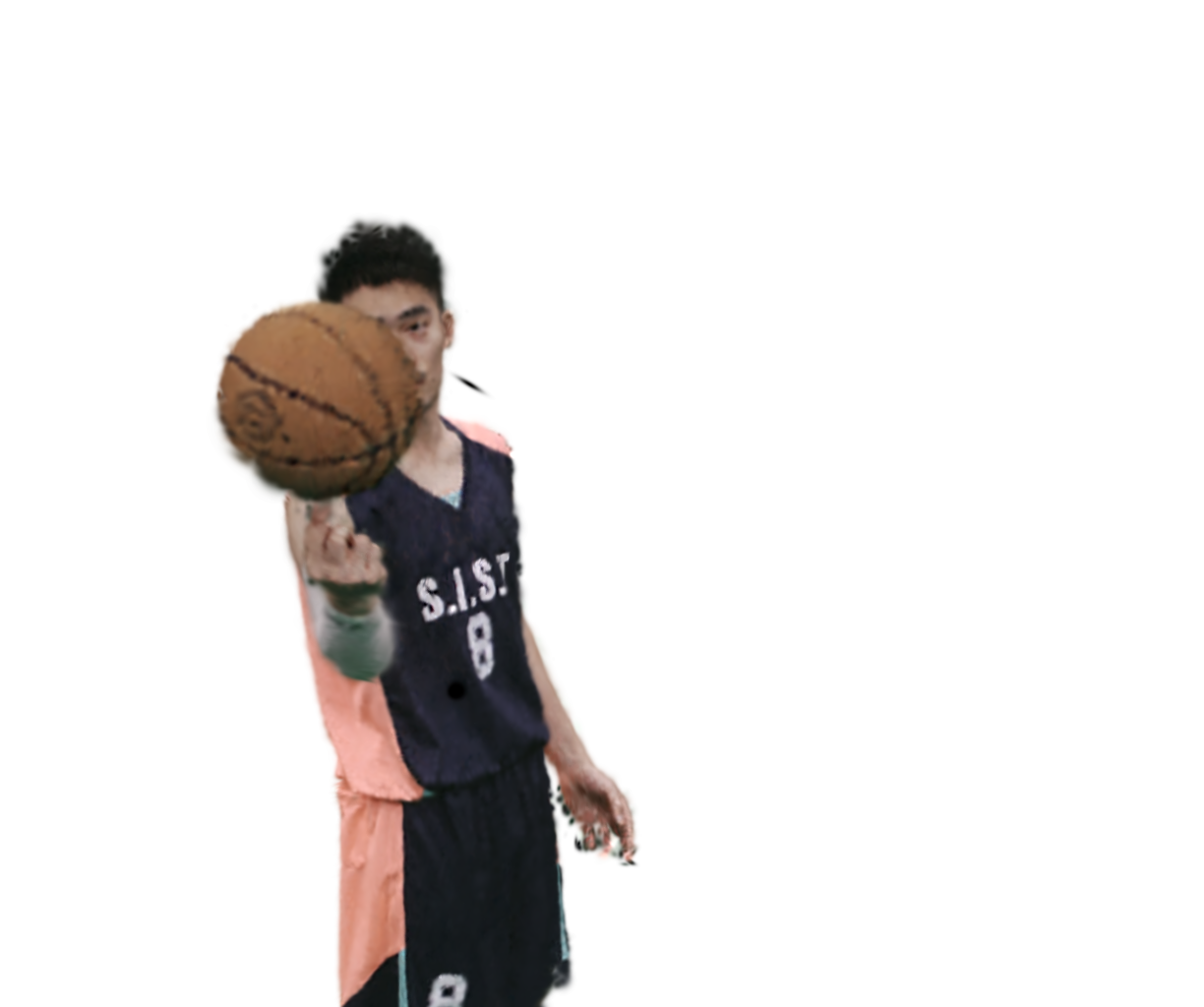}}}{}
\jsubfig{\fcolorbox{white}{white}{\includegraphics[scale=0.063]{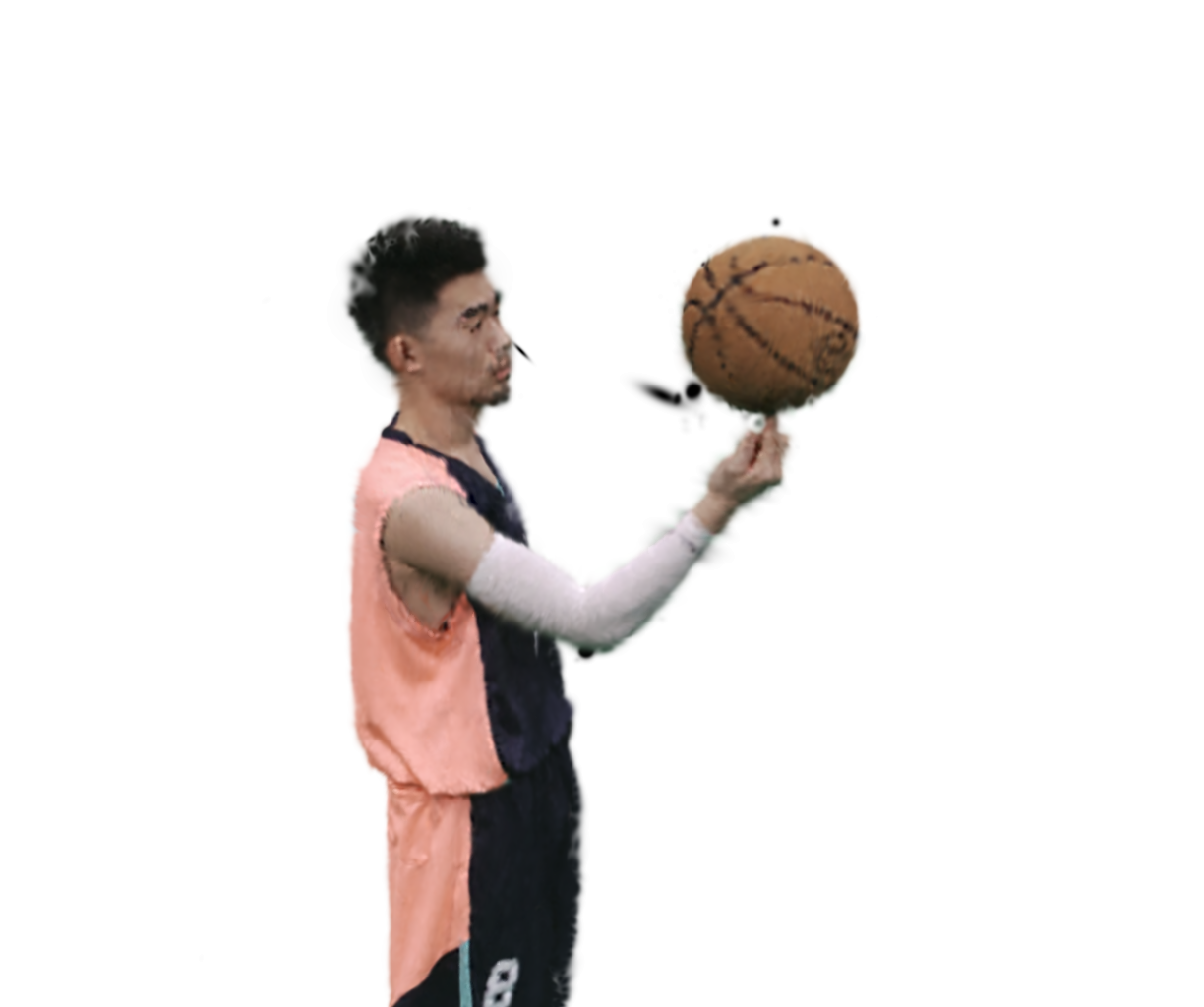}}}{}%
\\
\jsubfig{\includegraphics[scale=0.063]{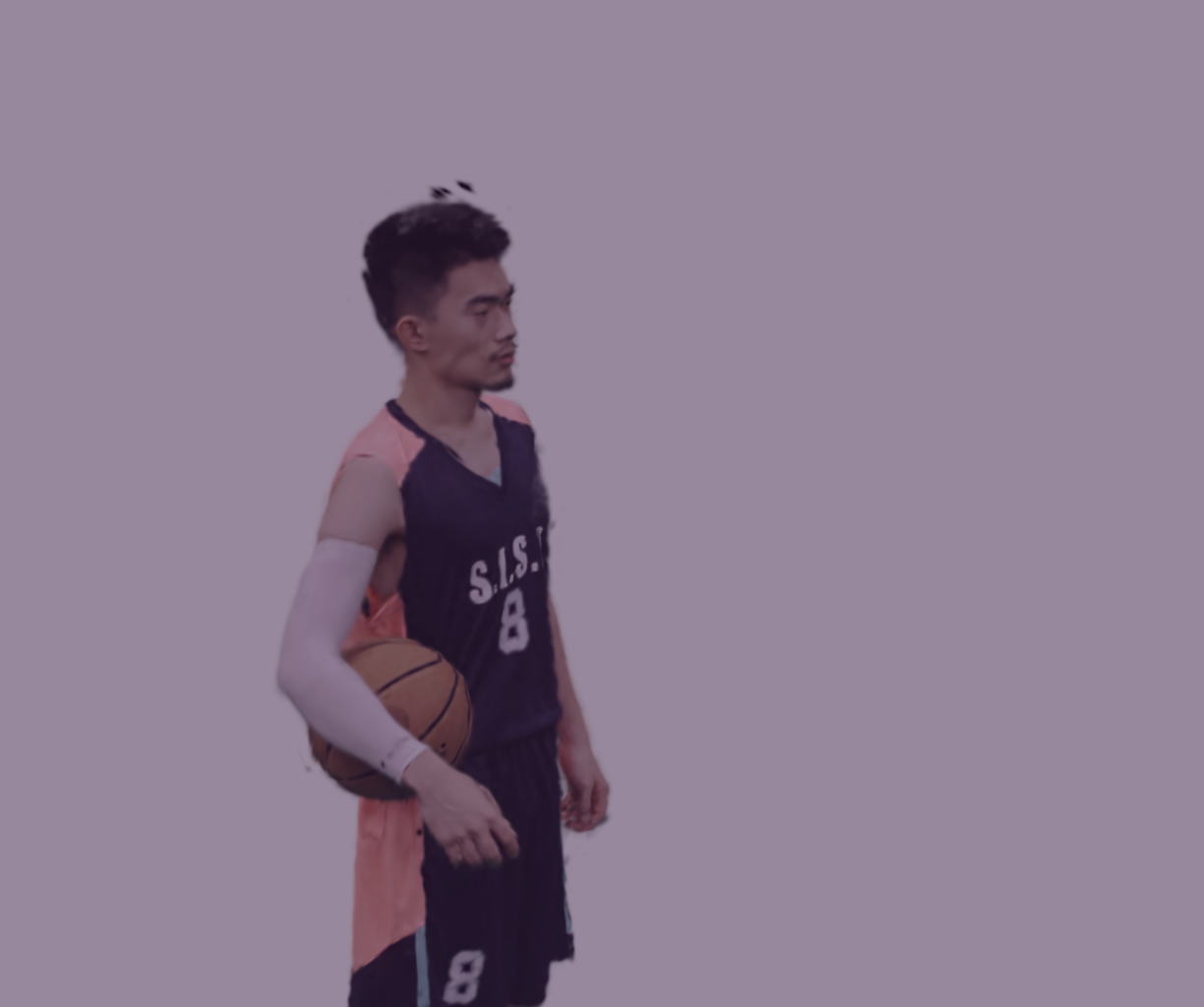}}{}
\jsubfig{{\includegraphics[scale=0.063]{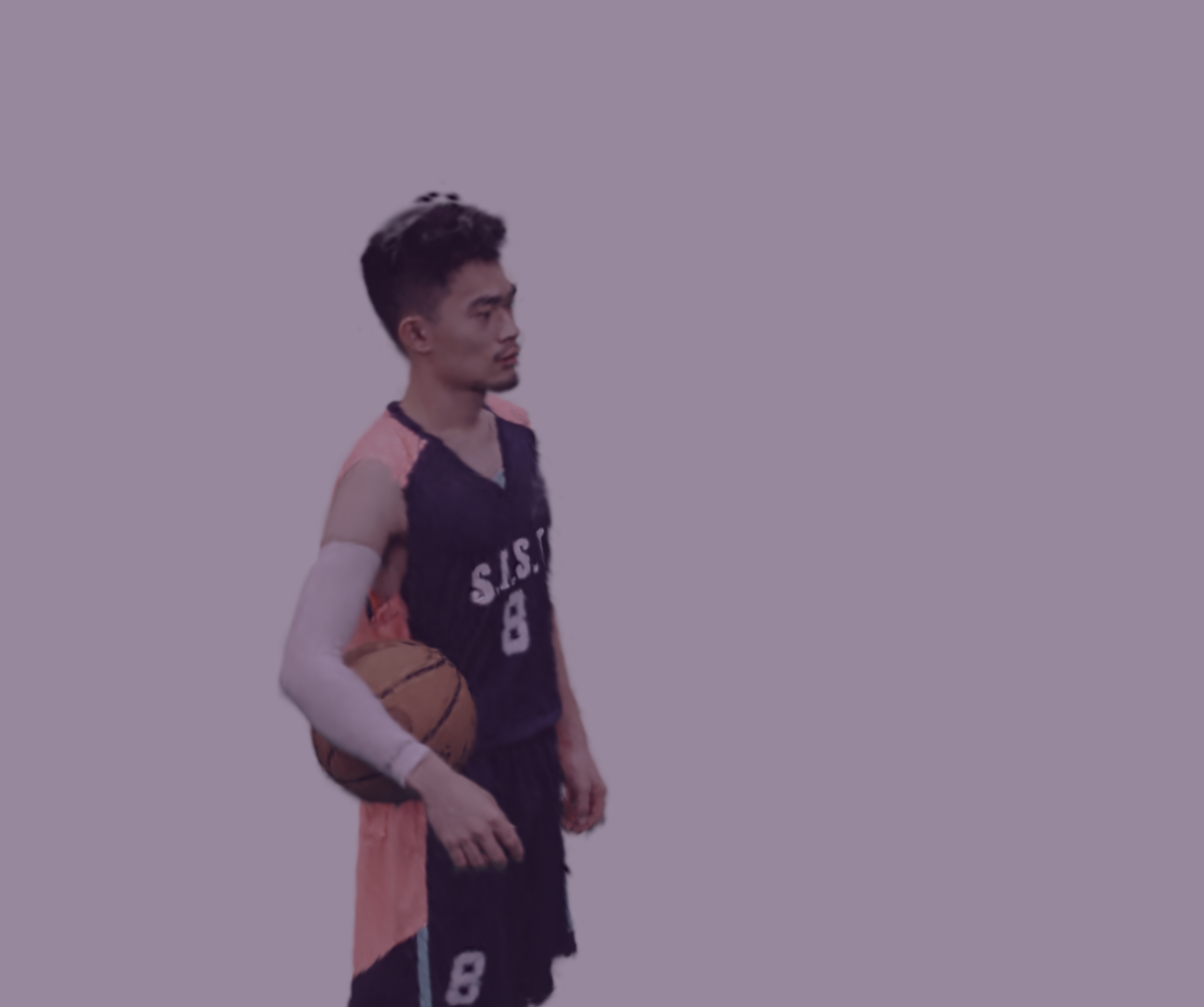}}}{}
\jsubfig{\fcolorbox{red}{red}{\includegraphics[scale=0.063]{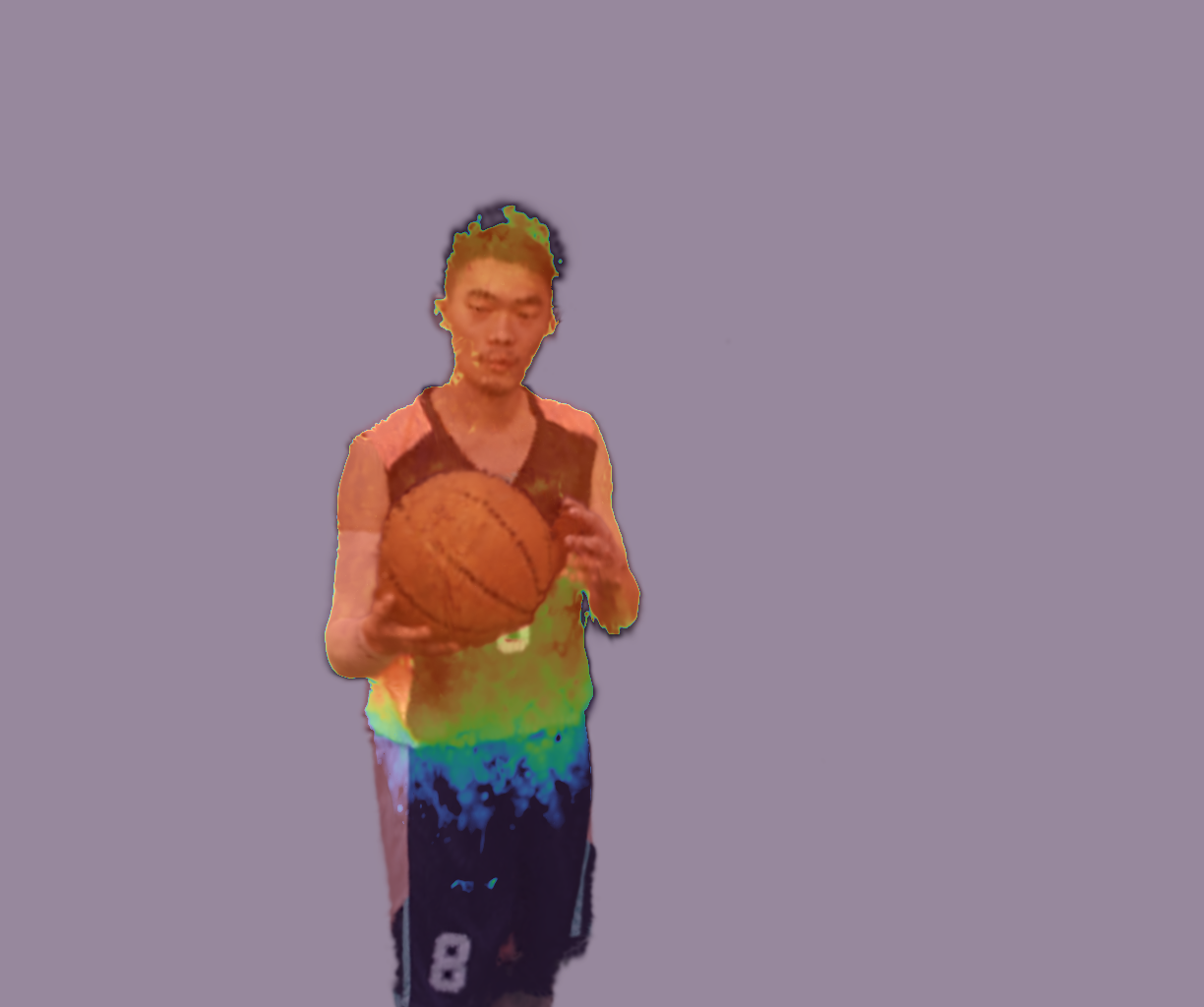}}}{}
\jsubfig{\fcolorbox{red}{red}{\includegraphics[scale=0.063]{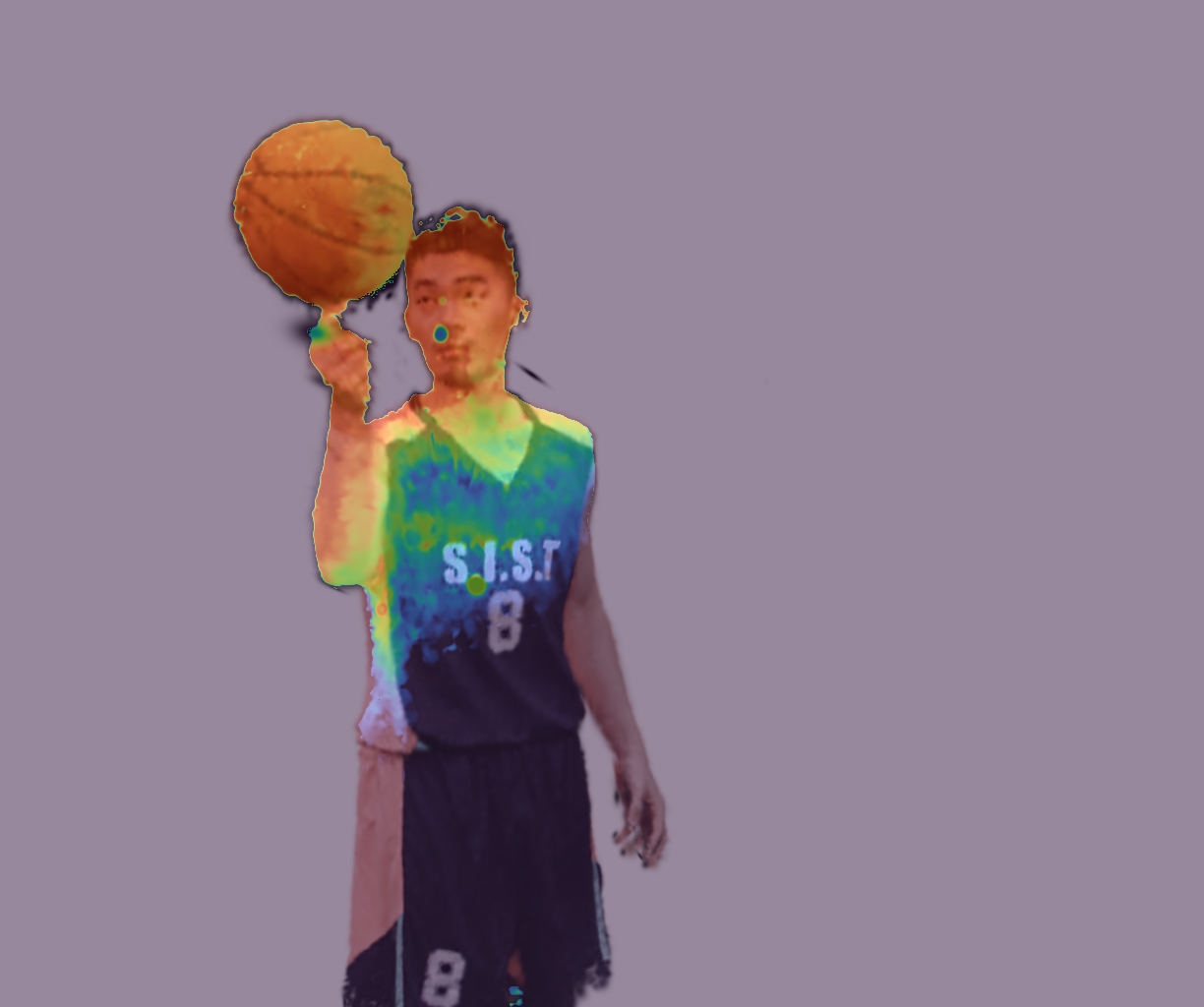}}}{}
\jsubfig{\fcolorbox{red}{red}{\includegraphics[scale=0.063]{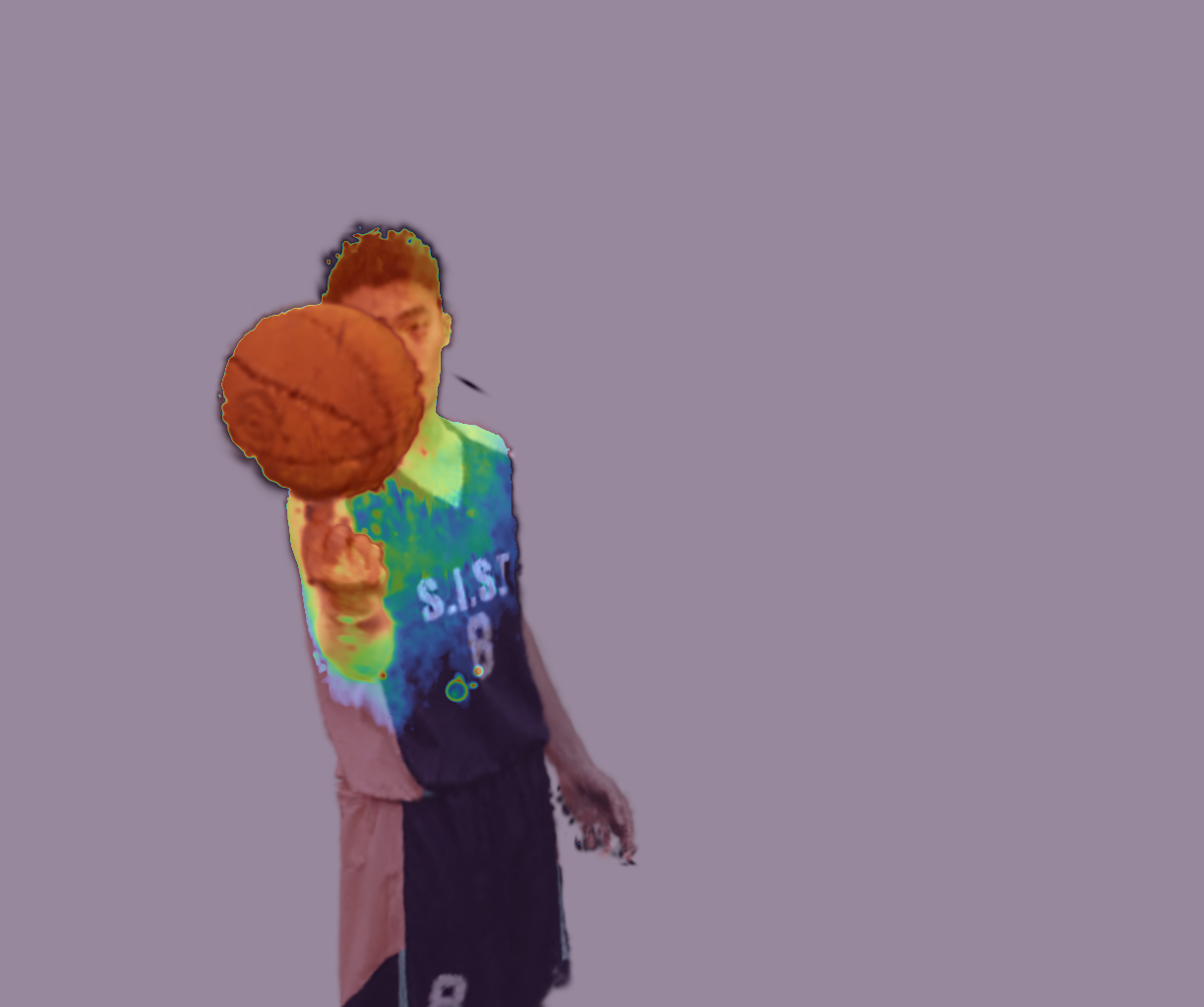}}}{}
\jsubfig{\fcolorbox{red}{red}{\includegraphics[scale=0.063]{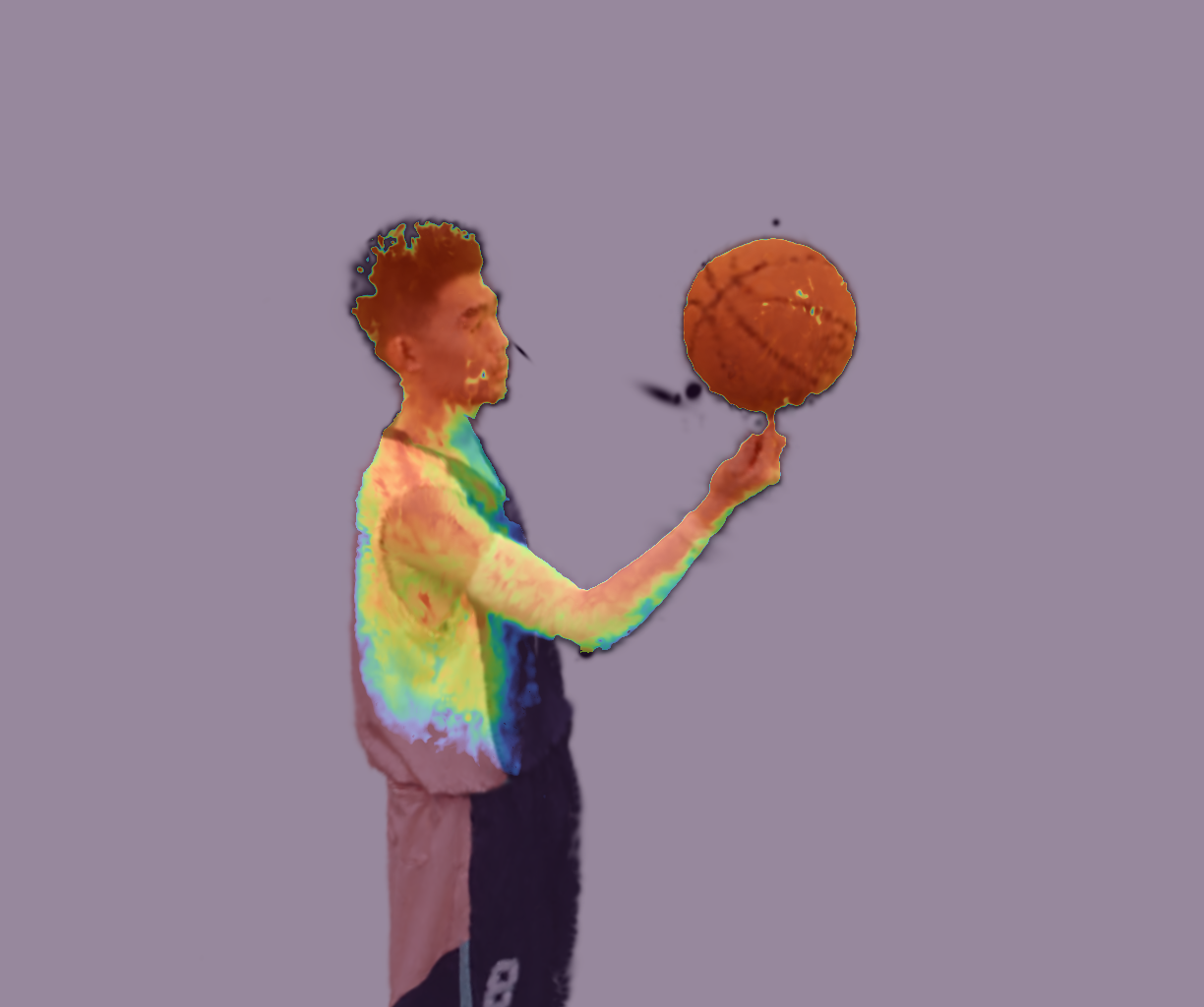}}}{}%
\\
\vspace{4pt}
{ Input query: \emph{A person playing with basketball}}
\caption{\textbf{Additional qualitative results on the Neural Human Rendering \cite{wu2020multi} dataset}, with the input frames depicted on top and the rendered probabilities directly below (frames localized temporally are in red).}

\label{fig:supp_results}
\end{figure*}

\end{document}